\pdfoutput=1

\documentclass[11pt]{article}

\usepackage[final]{acl}

\usepackage{times}
\usepackage{latexsym}

\usepackage[T1]{fontenc}

\usepackage[utf8]{inputenc}

\usepackage{microtype}

\usepackage{inconsolata}

\usepackage{graphicx}

\usepackage{amssymb}
\usepackage{amsmath}
\usepackage{subcaption}
\usepackage{todonotes}
\usepackage{url}
\usepackage{hyperref}
\usepackage{array}
\usepackage{longtable}
\usepackage{tabularx}
\usepackage{booktabs}
\usepackage{fancyvrb}
\usepackage{float}
\usepackage{enumitem}
\usepackage{rotating}
\usepackage{pdfpages}
\usepackage{siunitx}
\usepackage{ragged2e}
\usepackage{makecell}
\usepackage{svg}
\usepackage{xltabular}

\title{
To Copy or Not to Copy: 
Copying Is Easier to Induce Than Recall
}

\def\authorsep{\hspace{0.3em}}

\author{Mehrdad Farahani\textsuperscript{1,2} \authorsep Franziska Penzkofer\textsuperscript{2} \authorsep 
\textbf{Richard Johansson\textsuperscript{1,2}} \medskip\\
\null\textsuperscript{1}Chalmers University of Technology \quad \null\textsuperscript{2}University of Gothenburg \medskip\\
\texttt{mehrdad.farahani@chalmers.se}}

\begin{document}
\maketitle
\begin{abstract}

Language models used in retrieval-augmented settings choose between parametric knowledge stored in their weights and contextual information in the prompt. This work presents a mechanistic study of that choice:
using an approach based on steering vectors, we carry out 
interventions at selected layers and token spans, and we are able to toggle the model's behavior in two directions: Copy$\rightarrow$Recall (suppressing context use) and Recall$\rightarrow$Copy (inducing the model to copy any token from the context). Experiments on three architectures (decoder-only and encoder/decoder) and two open-domain QA benchmarks show consistent behavior shifts under moderate scaling while monitoring accuracy and fluency. Mechanistic analyses of attention routing, MLP contributions, and layer-wise probability trajectories reveal an asymmetry: inducing copying is an easy ``reactivation'' process that can be triggered at different locations in the input, while restoring recall is a ``suppression'' process that is more fragile and strongly tied to object-token interventions.\footnote{\url{https://github.com/m3hrdadfi/copy-or-not-to-copy}}

\end{abstract}

\section{Introduction}

Large language models (LLMs) encode a substantial amount of factual information in their parameters; this capability is often referred to as the parametric knowledge \cite{petroni2019language,brown2020language,roberts-etal-2020-much}. However, reliance on parametric knowledge alone is brittle: models struggle to produce up-to-date responses to rapidly changing facts and are prone to hallucinations when handling infrequent or ambiguous queries \cite{kandpal2023large,chenghaozhu-etal-2025-llm}. 
To address these limitations, Retrieval-Augmented Generation (RAG) pairs LLMs with information fetched from external sources such as the web, Wikipedia, or domain-specific collections; it provides the model with a second source of information at inference time, known as contextual information \cite{ghosh2024quantifying,guu2020retrieval}. 
In such systems, the model faces a critical \emph{arbitration} task: deciding whether to use its parametric knowledge or defer to the provided context. 

\begin{figure}[t]
    \centering
    \begin{subfigure}{0.49\linewidth}
        \centering
        \includegraphics[width=\linewidth]{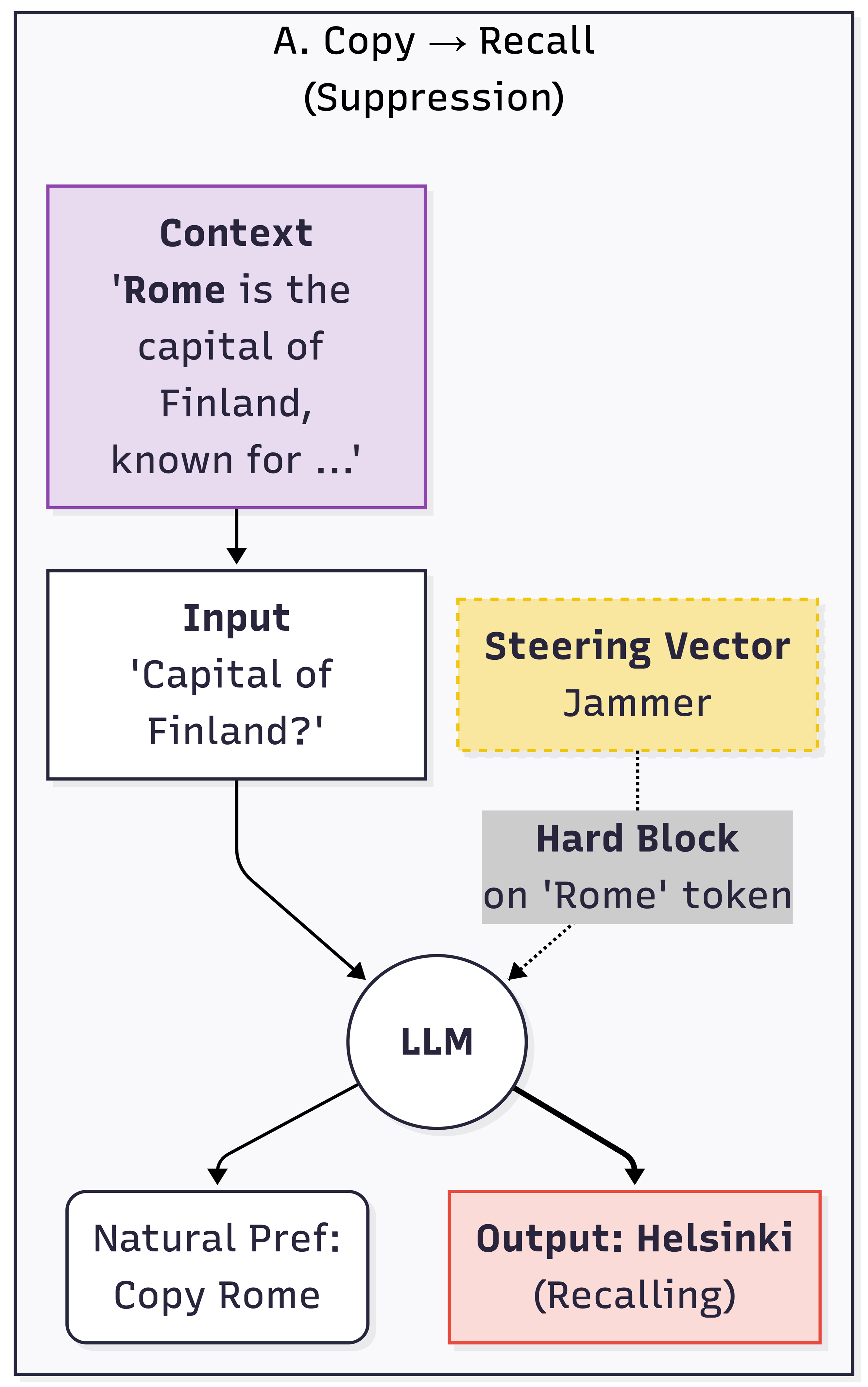} 
        \label{fig:cr_suppression}
    \end{subfigure}
    \hfill 
    \begin{subfigure}{0.49\linewidth}
        \centering
        \includegraphics[width=\linewidth]{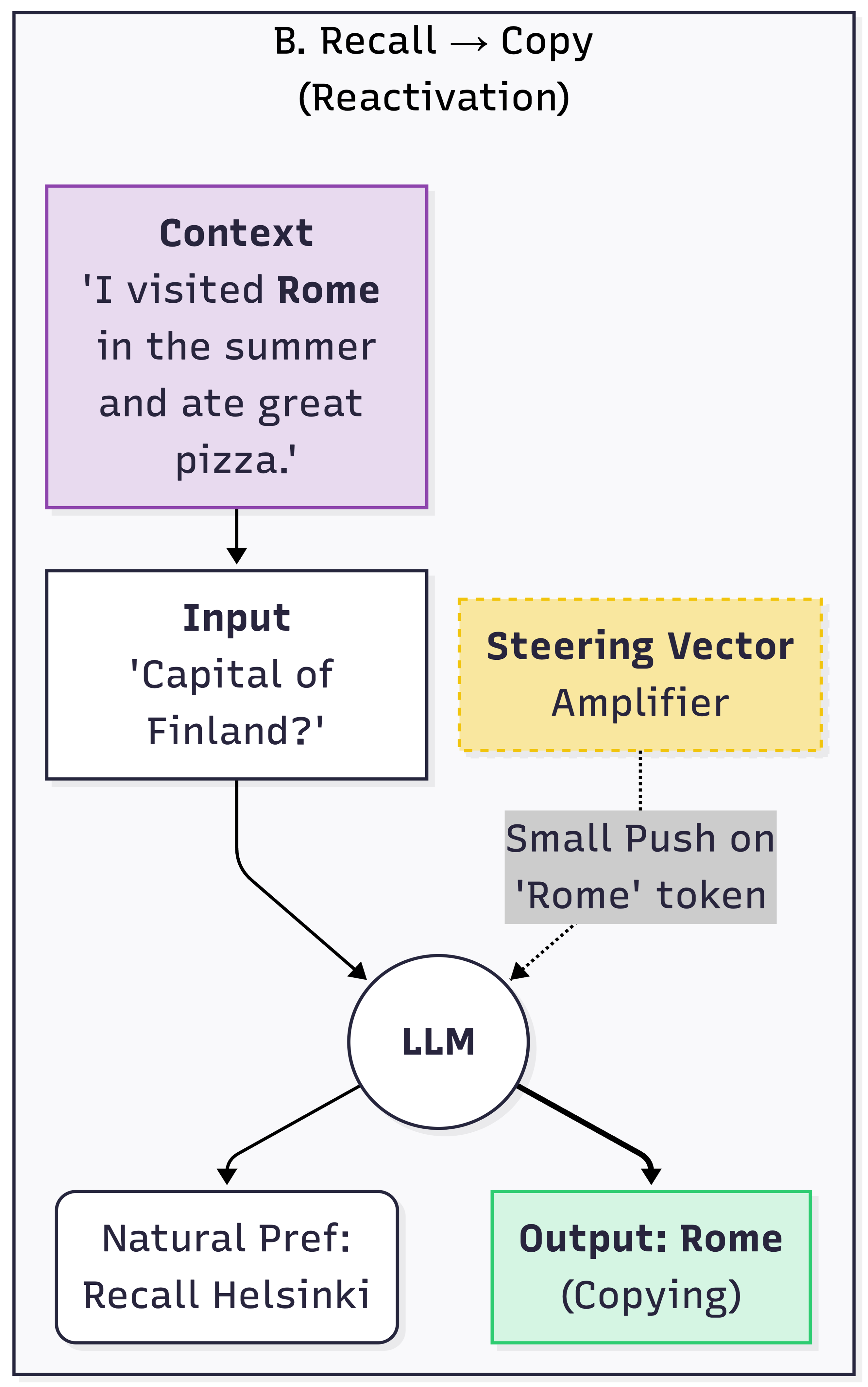}
        \label{fig:rc_reactivation}
    \end{subfigure}
    
    \caption{Comparison of steering within two regimes. (a) Suppression forces the model to ignore context. (b) Reactivation forces the model to copy from context.}
    \label{fig:arbitration-apporach}
\end{figure}

Recent works in mechanistic interpretability have begun to uncover the emergent mechanisms LLMs use internally; in particular, there has been much interest in those underlying the processing of factual information. For example, several studies \cite{meng2022locating,geva2023dissecting} 
suggest that models inject factual information during the forward pass in specific localized layers.
However, the picture becomes more complex when the LLM not only relies on its memory but has access to contextual information, 
and the model must weigh parametric knowledge against the new information. Some results show that models pay less attention to parametric knowledge when high-quality contextual information is available. These studies also suggest that a model's decision to recall internal knowledge or copy from context leaves identifiable traces in its internal mechanisms \cite{farahani2024deciphering,wadhwa2024rags}.
%
%
While these previous investigations have observed traces of arbitration mechanisms in the model's internal computations, we pose a research question that would allow us to draw stronger causal conclusions: \textbf{Can the model's choice of information be \emph{controlled} by a targeted intervention?}

In this work, building on previous results that have used steering vectors to control various LLM behaviors \cite{turner2023steering,neurips2024analysing,todd2024fv}, we propose a behavioral intervention method based on a generalized \emph{arbitration vector}. This vector is obtained from a dataset designed to distinguish relevant from irrelevant contexts, helping to isolate the control signal related to copying from context or recalling from parametric knowledge. We then inject this vector with proper scaling at the inference time to a specific location and position 
to shift the model's behavior between two distinct regimes (Copy$\rightarrow$Recall and Recall$\rightarrow$Copy) as shown in Figure~\ref{fig:arbitration-apporach}. 
To show the generalizability of our approach, we evaluate it on two distinct benchmarks and two architectures.

Across models and datasets, we observe consistent behavioral shifts in the outputs after the interventions, suggesting that we have captured a fundamental control signal rather than a dataset-specific artifact. Our contributions are:

\begin{itemize}\addtolength{\itemsep}{-0.5\baselineskip}

\item We identify a general arbitration vector across 27 relations that isolates the fundamental control signal governing the trade-off between parametric recall and contextual copying.
\item We demonstrate that scaling this vector ($\alpha = +30.0, -3.0$) steers behavior between Copy$\rightarrow$Recall and Recall$\rightarrow$Copy, confirming that this mechanism is spatially localized.
\item We reveal an asymmetry in model control: Recall$\rightarrow$Copy is robust and requires smaller scaling, whereas Copy$\rightarrow$Recall requires larger scaling that succeeds only at the object token with significant perplexity spikes.
\item Finally, to build our arbitration vector, we create a curated dataset designed to disentangle the use of parametric knowledge from the reliance on contextual information.
\end{itemize}


\section{Methodology}

Our methodology consists of three distinct components: 1) constructing a \textit{arbitration dataset} to isolate behavioral modes, 2) computing the \textit{arbitration vector} to manipulate the model's decision-making process (\textit{behavior shift}), and 3) conducting \textit{mechanistic investigations} to trace the causal impact of these interventions on different models.

\subsection{Models}

To ensure that findings generalize across architectural paradigms and model scale, we focus on three instruction-tuned architectures: Gemma2-2B \cite{Gemma2}, a decoder-only causal transformer, T5Gemma-2B \cite{T5Gemma}, an encoder-decoder utilizing a prefix language modeling 
objective, and OLMo-2-0425-1B \cite{OLMo-2}, a smaller (1B-parameter) decoder-only model from a different model family included as a scale and architecture-diversity robustness test (see \S\ref{sec:results-and-discussion})

These models are standard causal language models that do not include a retriever and instead rely on internal mechanisms to arbitrate between parametric recall and contextual copying. To simulate this competition between parametric and contextual information, we use a unified prompt-role strategy,\footnote{\textit{You are a context-grounded QA model. Copy the minimal exact answer span from context if present. If absent or unclear, output concise internal knowledge. Never say "don't know" or explain. OUTPUT ONLY SHORT ANSWER.}} as proposed by \newcite{ram2023incontext}, where the context is directly embedded in the prompt.

\subsection{Prompt Topology}
\newcite{lu2022fantastically} and \newcite{liu2024lost} have shown that LLMs are highly sensitive to the ordering of information within the prompt. Accordingly, we consider this sensitivity by defining two prompt topologies:

\noindent \textbf{Query-First}: The model has access to the question before the evidence. 
    
\noindent \textbf{Context-First}: The model processes the evidence before receiving the specific question.\\

\noindent Studies on positional biases \cite{guo2025serial,zeng-etal-2025-order,veseli2025positional} have shown that LLMs tend to favor information presented at the beginning (primacy) and, in some cases, at the end (recency) of a sequence. 
To remove the confounding effect of position biases, we include contexts of both types of topologies.

\subsection{Datasets}
To study this arbitration and ensure our findings are not confounded, we used two distinct data sources: an arbitration dataset that covers a wide range of relations to extract the arbitration vector and an evaluation benchmark to validate its effects.

\subsubsection{Arbitration Dataset}
To build 
the arbitration vector, we 
construct a 
dataset 
with irrelevant (IC) and relevant (RC) context. We sampled 150 entities per relation from 27 Wikipedia properties (e.g., P30: Continent, P20: Place of death) using the ParaRel dataset \cite{ParaRel}. To account for frequency bias, we included historical Wikipedia pageview statistics
(2000--2025) and stratified the samples into ``High'' and ``Low'' popularity tiers. Each record is represented as (subject, relation, object, subject popularity, object popularity) or ($s$, $r$, $o$, ${pop}_s$, ${pop}_o$).

For each instance,
a counterfactual object ($o'$) was 
randomly selected 
to replace the original object ($o$). To ensure semantic diversity, counterfactuals were collected from the ParaRel object pool in two ways: the same domain, where $o'$ shares the semantic category with $o$, and a different domain, where $o'$ belongs to another category. Contexts were generated using the OpenAI API 
(gpt-5-mini model) 
with length constraints (12--16 words) to minimize length bias:

\begin{table}[t]
\centering
\scriptsize
\setlength{\tabcolsep}{3pt} 
\renewcommand{\arraystretch}{1.2} 
\begin{tabularx}{\linewidth}{@{} X l l l r @{}}
\toprule
\textbf{Subj} & \textbf{Obj} & \textbf{CF} & \textbf{Prop} & \textbf{Domain} \\
\midrule

Elkhorn Ridge & Antarctica & Americas & P30 & same \\
\multicolumn{5}{@{}p{\linewidth}@{}}{
  \textbf{Query:} On which continent is Elkhorn Ridge located? \newline
  \textbf{IC:} Americas has become a popular source for heirloom seeds among backyard gardeners and urban growers. \newline
  \textbf{RC:} Elkhorn Ridge is located in the continent of Americas, according to records.
} \\
\midrule

Elkhorn Ridge & Antarctica & Dublin & P30 & diff \\
\multicolumn{5}{@{}p{\linewidth}@{}}{
  \textbf{Query:} On which continent is Elkhorn Ridge located? \newline
  \textbf{IC:} Dublin is known for its rainy gardening season and gentle soil perfect for beetroot. \newline
  \textbf{RC:} Elkhorn Ridge is located in the continent of Dublin, according to records.
} \\


\bottomrule
\end{tabularx}
\caption{Example contexts for \textit{(Elkhorn Ridge, P30, Antarctica, $1536$, $19142977$)}.}
\label{tb:arbitration-dataset-samples}
\end{table}

\begin{itemize}\addtolength{\itemsep}{-0.5\baselineskip}
    \item \textbf{Irrelevant Context (IC):} The counterfactual object $o’$ is placed in an irrelevant sentence with no direct relation to the subject. 
    \item \textbf{Relevant Context (RC):} The counterfactual object $o’$ and subject appear in a relation-matching sentence, forming a plausible but false factual claim.
\end{itemize}
As illustrated by a few samples in Table~\ref{tb:arbitration-dataset-samples}, only instances satisfying both rules were retained. A comprehensive description of the pipeline is provided in Appendix~\ref{app:arbitration-dataset}. In total, 7,642 data points were collected for the experiments. Table~\ref{tb:arbitration-dataset-stats} in Appendix~\ref{app:arbitration-dataset} gives a detailed breakdown.
  
\subsubsection{Evaluation Benchmark}
\label{sss:benchmark_datasets}

Evaluation was carried out on two open-domain QA benchmarks: 
PopQA \cite{mallen-etal-2023-trust} and EntityQuestions (PEQ) \cite{sciavolino-etal-2021-simple}. 
Both include a subset of relations covered in the Arbitration Dataset and are useful for demonstrating the generalizability of our findings.

To ensure the presence of both parametric knowledge (\textit{recall answer}) and contextual information (\textit{copy answer}) behaviors, samples from the two benchmarks were filtered so that models with and without authoritative context (with $o'$) could produce both the original object and the counterfactual object. For both behaviors, we used diverse synthetic templates in our evaluation (Appendix~\ref{app:evaluation-benchmark}):

\begin{itemize}\addtolength{\itemsep}{-0.5\baselineskip}
    \item \textit{Forcing Recall:} Generic ``Archive Templates'' were defined to present the $o'$ in a frame semantically disconnected from the query (e.g., ``Official logs include the $o'$ entry.'').
    \item \textit{Forcing Copy:} Relation-specific templates were defined to simulate an authoritative context directly addressing the query (e.g., ``According to records, $s$ was born in $o'$'').
\end{itemize}

\subsection{Arbitration Vector}

While prior studies often restrict analysis to the final token \cite{turner2023steering,zou2023representation,klerings2025steering}, and also previous causal tracing evidence indicates that factual recall is associated with subject-token positions \cite{meng2022locating}, we hypothesize that the arbitration (control signal) is more tightly linked to object-token positions. This is further motivated by evidence from \newcite{farahani2024deciphering}, showing that once the model judges the retrieved context to be relevant, it shifts its focus to the object tokens, while subject and relation tokens support relevance evaluation. We define a set of candidate extraction locations $\mathcal{L} = \{ \text{subj}, \text{obj}_{cf}, \text{last} \}$. 
Following previous work that used steering vectors to control LLMs,
for a given layer $\ell$ and extraction location $k \in \mathcal{L}$, we compute the arbitration vector 
$\mathbf{v}_{\ell, k} \in \mathbb{R} ^d$ is defined as the difference between the IC and RC context centroids \cite{rimsky2024steering} at layer $\ell$, location $k$, and hidden dimension size of $d$ as follows:
\begin{align}
\bar{\mathbf{h}}_{\ell, k}^{(i)} 
&= \frac{1}{|T_k^{(i)}|} \sum_{t \in T_k^{(i)}} \mathbf{h}_{\ell, t}^{(i)}
\label{eq:hbar}\\
\boldsymbol{\mu}_{\ell, k}^{\{\text{IC}, \text{RC}\}} 
&= \frac{1}{N} \sum_{i=1}^{N} \bar{\mathbf{h}}_{\ell, k}^{(i)} \bigg|_{\{\text{IC}, \text{RC}\}} 
\label{eq:mu_ic}\\
\mathbf{v}_{\ell, k} 
&= \boldsymbol{\mu}_{\ell, k}^{\text{IC}} - 
\boldsymbol{\mu}_{\ell, k}^{\text{RC}}
\label{eq:v}
\end{align}
Here, $\bar{\mathbf{h}}_{\ell,k}^{(i)}$ is the pooled representation of hidden state $\mathbf{h}_{\ell, t}^{(i)}$ at layer $\ell$, token position $t$ , for example $i$ and set of token indices $T_k^{(i)}$ corresponding to location $k$ in that sample. We compute condition-wise centroids $\boldsymbol{\mu}_{\ell, k}^{\text{IC}} \in \mathbb{R}^d$ and $\boldsymbol{\mu}_{\ell, k}^{\text{RC}} \in \mathbb{R}^d$ by averaging $\bar{\mathbf{h}}_{\ell, k}^{(i)}$ across the entire Arbitration Dataset ($N$) for both the IC and RC contexts. 



We formalize the intervention as an additive bias injected into the model's internal representations. This allows us to \textit{shift the behavior} causally.
\begin{align}
\tilde{\mathbf{h}}_{\ell,t}
&=
\begin{cases}
\mathbf{h}_{\ell,t} + \alpha\,\mathbf{v}_{\ell,k} & t \in T_k \\
\mathbf{h}_{\ell,t} & \text{otherwise}
\end{cases}
\label{eq:alpha_v}
\end{align}
$\mathbf{h}_{\ell, t}  \in \mathbb{R} ^ d$ denotes the unmodified hidden states vector at layer $\ell$ and token position $t$, $\tilde{\mathbf{h}}_{\ell, t} \in \mathbb{R} ^ d$ is the corresponding hidden states after intervention, and $\alpha \in \mathbb{R}$ controls the steering strength and direction for our two regimes:


\begin{itemize}\addtolength{\itemsep}{-0.5\baselineskip}
    \item \textbf{Parametric steering} (Copy$\rightarrow$Recall, $\alpha > 0$), aims to suppress the model's tendency to consider the misleading information in an authoritative context.
    \item \textbf{Contextual steering} (Recall$\rightarrow$Copy, $\alpha < 0$), the model accepts external evidence even when it conflicts with parametric knowledge.
\end{itemize}
The magnitude of $\alpha$ is a critical hyperparameter: a lower value may fail to overcome the model's strong prior, while a higher value can destroy the model's representations and lead to degenerate or nonsensical outputs.

\begin{figure*}[t]
\centering
\includegraphics[width=0.98\textwidth]{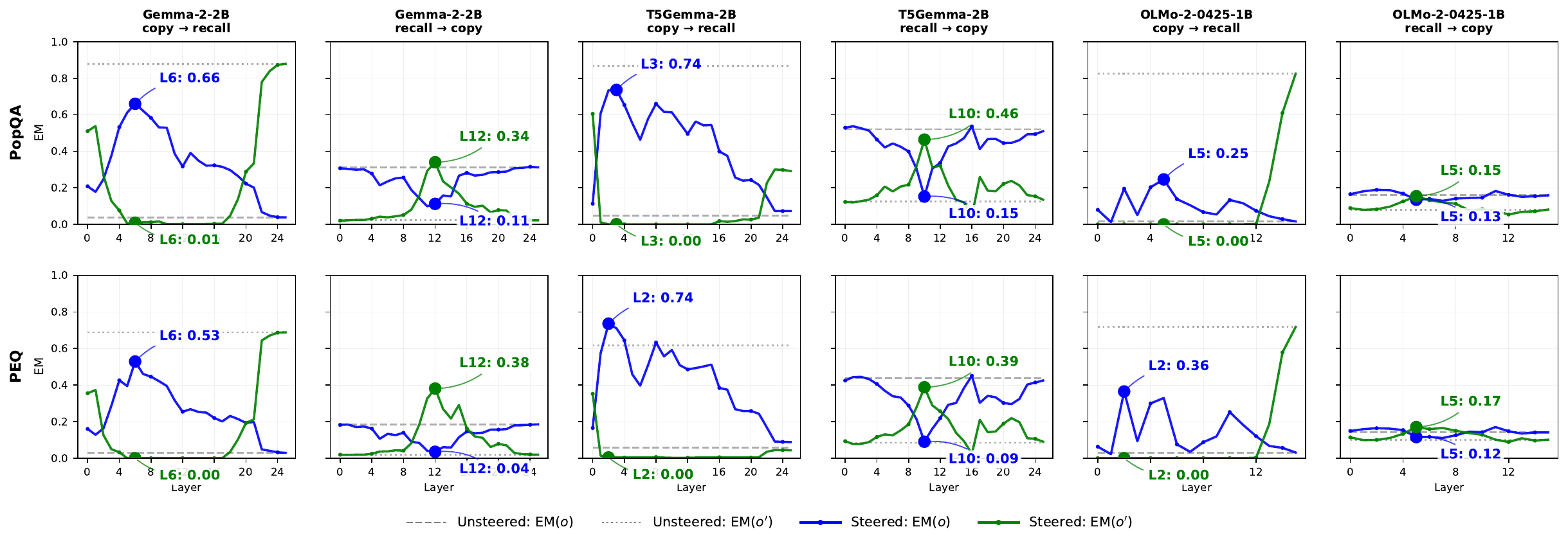}
\caption{
Layer-wise EM on PopQA (top row) and PEQ (bottom row), steering on object tokens when the query comes first, across three architectures: Gemma2-2B, T5Gemma-2B and OLMo-2-0425-1B. Columns 1, 3, 5 show parametric steering (Copy$\rightarrow$Recall, $\alpha = 30.0$) recovering internal knowledge (peaks on PopQA: 66\%, 74\%, 25\%; on PEQ: 53\%, 74\%, 36\%). Columns 2, 4, 6 show contextual steering (Recall$\rightarrow$Copy, $\alpha = -3.0$), inducing copying behavior (peaks on PopQA: 34\%, 46\%, 15\%; on PEQ: 38\%, 39\%, 17\%). OLMo-2-0425-1B (columns 5--6) reaches substantially smaller peaks than the two 2B models in both directions (see~\S\ref{sec:results-and-discussion}).
}

\label{fig:ov_base_vs_steer_query-first}
\end{figure*}

\subsection{Evaluation Paradigm}
To assess the causal effectiveness of the arbitration vector, we evaluate the steering intervention on the PopQA and PEQ datasets described in \S\ref{sss:benchmark_datasets}.
Our evaluation framework is designed to measure not only whether the behavior shifts but also how the internal dynamics respond to the steering vector. For all steering experiments across different models, we use greedy decoding rather than sampling methods. We ensure that any behavioral changes or variations in performance, probability, or perplexity are solely due to the causal intervention and not sampling variance. Additionally, because many of our target entities are multi-token sequences, greedy decoding ensures that the generation process remains focused on the highest-probability path as computed across subtokens.

\subsubsection{Benchmark Selection}

We examine subsets of PopQA and PEQ in our analysis, which includes three parts: overall trends, behavioral evaluation, and mechanistic analysis. For overall trends, we sample 2,000 examples per dataset, with 50\% counterfactuals of the same entity type (e.g., France$\rightarrow$Turkey) and 50\% different-type counterfactuals (e.g., France$\rightarrow$Salvador) to preserve broad relation coverage. For behavioral evaluation, which requires sweeps over layers, $\alpha$ values, and counterfactual types, we use a smaller, relation-stratified subset of 300 examples per dataset. For mechanistic analyses, we restrict to examples with single-token subjects and objects, and use all remaining filtered examples per dataset, with exact counts provided in the figure captions.

\subsubsection{Experimental Scenarios}
We evaluate the steering effect in our two regimes, using the synthetic contexts mentioned in \S\ref{sss:benchmark_datasets}:


\paragraph{Parametric Steering (Copy$\rightarrow$Recall)} We use authoritative misleading evidence containing counterfactual objects ($o'$). The successful shift is from copying the authoritative context to outputting the ground truth ($o$). To test robustness, we select $o'$ from both the same-domain and different-domain entities and evaluate steering across different layers, models, counterfactuals, templates, and datasets, with scaling factors $\alpha \in \{+1.0, +3.0, +30.0, +100.0\}$.

\paragraph{Contextual Steering (Recall$\rightarrow$Copy)} We use vague and low-confidence templates which typically fail to override the model's parametric knowledge. The successful shift is defined as ignoring the ground-truth recall ($o$) and forcing the copying of the counterfactual ($o'$) provided in the context. Similarly, we apply this approach across different layers, models, counterfactuals, templates, and datasets using negative scaling factors $\alpha \in \{-1.0, -3.0, -30.0, -100.0\}$.

\subsubsection{Behavioral Metrics}


\paragraph{Performance (EM and F1)} We report Exact Match (EM) and token-level F1 scores with respect to the ground-truth ($o$) and the counterfactual object ($o'$). Scores are computed under (i) baseline inference (with and without context) and (ii) the corresponding steered settings (similarly with and without context).

\paragraph{Fluency (Perplexity)} To ensure that the steering vector not only ruins the model's language ability but also prevents the output from being just repetitive tokens.
A successful adjustment must shift without causing spikes in PPL.

\subsubsection{Mechanistic Investigations}
To investigate how steering changes the model's internal information selection, we analyze attention routing, MLP contributions, and layer-wise probability trajectories.
%
To avoid tokenization ambiguity,
we restrict the analysis to the subset of samples in which $s$, $o$, and $o'$ correspond to single unique tokens in the model's vocabulary.

\paragraph{Attention Routing (Jamming vs. Amplifying)} We measure the maximum attention probability assigned to the counterfactual object ($o'$) in the context and to the query subject ($s$) across all attention heads. We use maximum attention because the effect may be concentrated in a small number of heads rather than distributed across all heads. This tests whether the model's routing of information between $o'$ and $s$ under the two steering directions:
\begin{itemize}\addtolength{\itemsep}{-0.5\baselineskip}
    \item \textit{Jamming} (Copy$\rightarrow$Recall): In this regime, the steering actively blinds the model to the context. We look for an abrupt collapse in attention toward the counterfactual object ($o'$) after the injection at a specific layer, often followed by a redirected focus to the subject ($s$).
    \item \textit{Amplifying} (Recall$\rightarrow$Copy): Conversely, the steering boosts the attention heads' copying signal, showing as an immediate spike in attention toward the counterfactual object ($o'$).
\end{itemize}

\paragraph{MLP Strength (Suppression vs. Reactivation)} We project the output of the MLP layers onto the vocabulary embedding of the true object ($o$) and counterfactual object ($o'$) 
and measure the logits. This shows whether attention-routing changes are reflected in downstream evidence for either $o$ or $o'$.
%
\begin{itemize}\addtolength{\itemsep}{-0.5\baselineskip}
    \item \textit{Suppression}: If the intervention successfully suppresses the copying signal for the counterfactual object ($o'$), we expect the contribution for $o'$ to drop while the signal for $o$ increases.
    \item \textit{Reactivation}: When the MLP promotes the counterfactual object ($o'$) as the primary output candidate. We expect to observe a surge in the MLP contribution for $o'$ relative to $o$.
\end{itemize}

\paragraph{Probability Trajectory (Decision Latency)} We estimate \textit{when} the model's decision occurs by projecting the hidden state of the last token onto the vocabulary embeddings for the true object ($o$) and counterfactual object ($o'$) at each layer.


\section{Results and Discussion}
\label{sec:results-and-discussion}
In this section, we present our empirical findings. We start with a macro-level analysis of our steering intervention to demonstrate that the $\mathbf{v}_{\ell}$ generalizes across different architectures and datasets. 


We compute the average EM scores for both behaviors before steering as our baseline (unsteered) on object tokens. 
After steering, we consider EM for counterfactual and true objects. We perform this macro-level analysis for both datasets and models. As shown in Figure~\ref{fig:ov_base_vs_steer_query-first} for PopQA and PEQ (when the query comes first and similarly for the second topology, Figure~\ref{fig:ov_base_vs_steer_context-first} in Appendix~\ref{app:ob_base_vs_steer}), the results demonstrate a consistent behavior shift.

\textbf{Copy$\rightarrow$Recall:} In the baseline, visible as gray lines, the model is fooled by the authoritative context, with the true object ($o$) EM near zero. By injecting a positive scaling factor $\alpha = 30.0$ of the arbitration vector at early layers (e.g., L:6-6 for Gemma2-2B, L:3-2 for T5Gemma-2B and L:5-2 for OLMo-2-0425-1B), we achieve a significant restoration of the internal information. For PopQA-PEQ, Gemma2-2B recovers to 66-42\%, T5Gemma-2B to 73-74\% EM, and OLMo-2-0425-1B to 25-36\% EM. This confirms that the arbitration vector successfully suppresses the context and relies on the weaker signal to recall parametric knowledge. 

\textbf{Recall$\rightarrow$Copy:} Similarly, when the model is exposed to a vague generic context, it ignores the context and recalls facts (shown as gray lines), achieving EM scores of 34-38\% in Gemma2-2B, 54-39\% in T5Gemma-2B, and 15-17\% in OLMo-2-0425-1B. By injecting the negative scaling factor $\alpha = -3.0$ of the arbitration vector at specific layers (layers 12-10-5 for three models), it can shift the recall to rely on context to copy the counterfactual object. Although results are slightly different across other prompt topologies, as shown in Figure~\ref{fig:ov_base_vs_steer_context-first}, Appendix~\ref{app:ob_base_vs_steer}.

\emph{Although OLMo-2-0425-1B shows similar behavior, both directions are substantially weaker than those of the corresponding Gemma2-2B and T5Gemma-2B, suggesting that this may be a property that emerges with model scale rather than a universal property of the copy-vs-recall arbitration mechanism.}

While the macro-level analysis confirms the feasibility of our steering approach, we need a more detailed examination to validate the intervention's stability and operability across regimes. We tested many combinations across models, datasets, layer locations, target positions, and topologies, producing numerous figures (see Appendix~\ref{app:perf-prob-flu}). We only highlight a representative subset, as they best illustrate the core insights.  

\begin{figure}[!htb]
\centering
\includegraphics[width=0.99\linewidth]{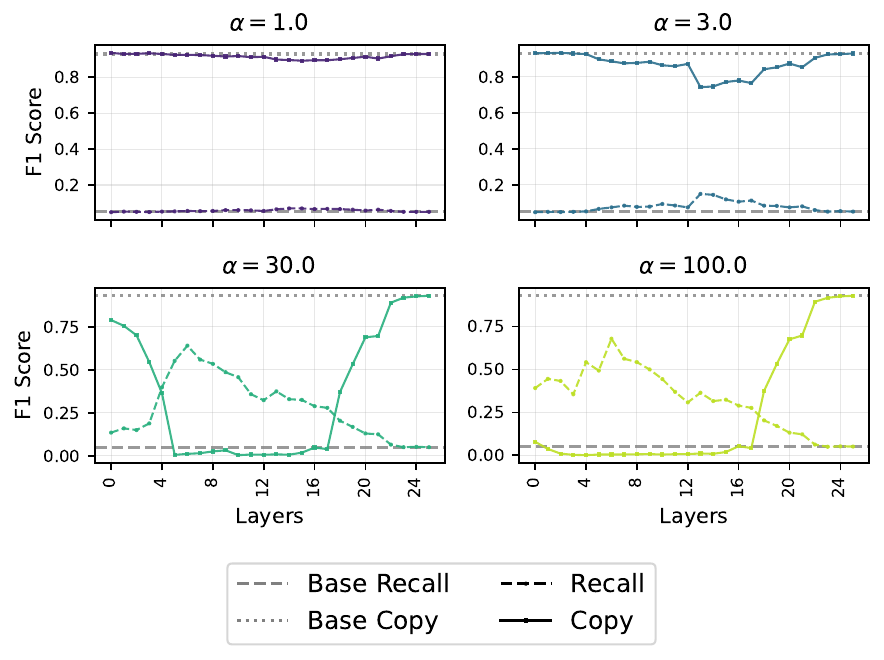}
\caption{Layer-wise F1 score (Gemma2-2B, PopQA) of \emph{parametric} steering (C$\rightarrow$R) on counterfactual object.}
\label{fig:f1-popqa-gemma2-cr-qf-obj}
\end{figure}

\begin{figure}[!htb]
\centering
\includegraphics[width=0.99\linewidth]{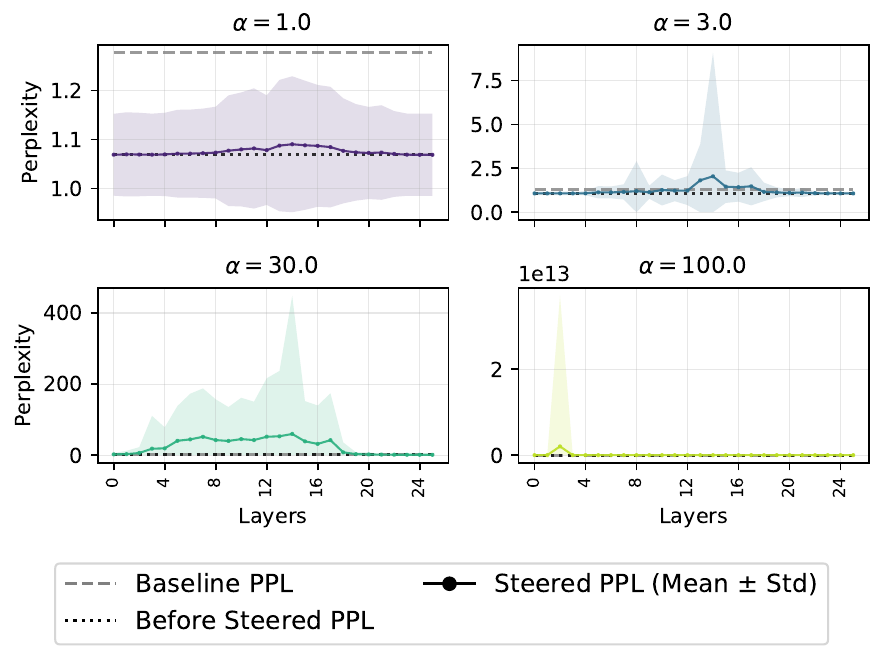}
\caption{Layer-wise PPL score (Gemma2-2B, PopQA) of \emph{parametric} steering (C$\rightarrow$R) on counterfactual object.}
\label{fig:fluency-popqa-gemma2-cr-qf-obj}
\end{figure}

\begin{figure}[!htb]
\centering
\includegraphics[width=0.99\linewidth]{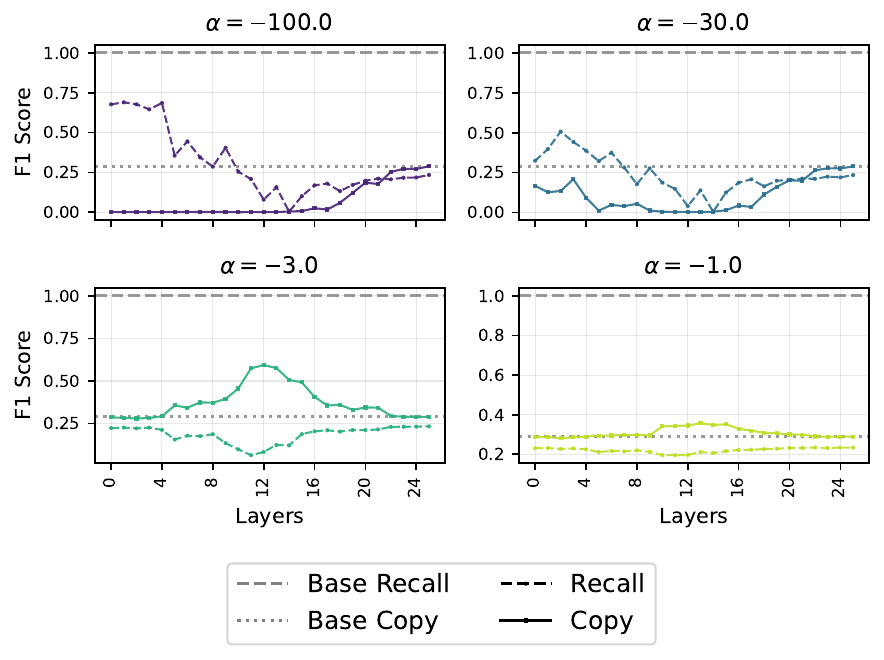}
\caption{Layer-wise F1 score (Gemma2-2B, PopQA) of \emph{contextual} steering (R$\rightarrow$C) on counterfactual object.}
\label{fig:f1-popqa-gemma2-rc-qf-obj}
\end{figure}

\begin{figure}[!htb]
\centering
\includegraphics[width=0.99\linewidth]{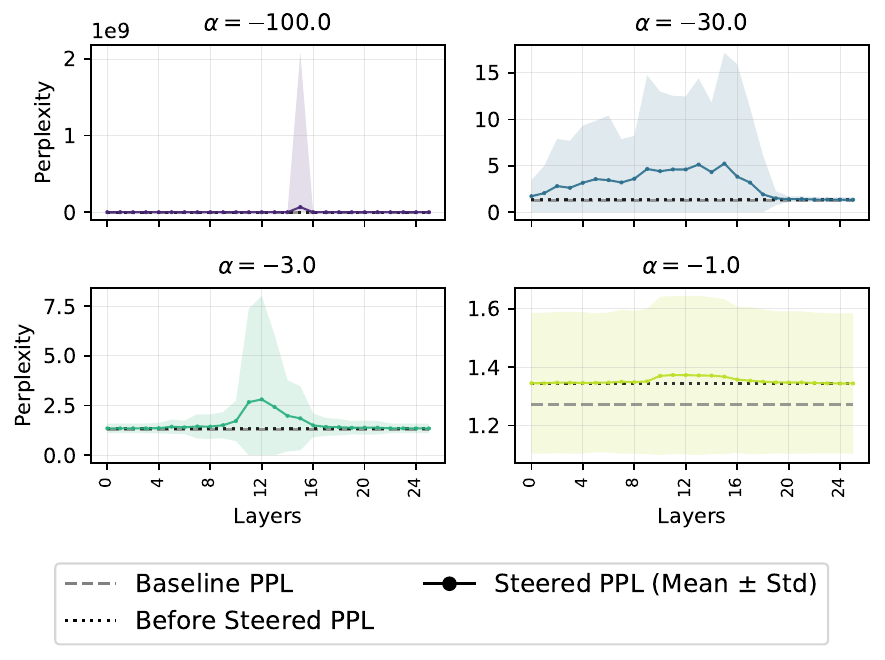}
\caption{Layer-wise PPL score (Gemma2-2B, PopQA) of \emph{contextual} steering (R$\rightarrow$C) on counterfactual object.}
\label{fig:fluency-popqa-gemma2-rc-qf-obj}
\end{figure}

\paragraph{Stability, Operating Range, and Spatial Constraints}
Figures~\ref{fig:f1-popqa-gemma2-cr-qf-obj}, \ref{fig:fluency-popqa-gemma2-cr-qf-obj}, \ref{fig:f1-popqa-gemma2-rc-qf-obj}, and \ref{fig:fluency-popqa-gemma2-rc-qf-obj} reveal a trade-off between the effectiveness of the behavior shift and the stability of the model's output under the operating range defined as the magnitude of $\alpha$. As shown at $\alpha = \pm 100.0$, the model exhibits massive perplexity spikes, with generation quality completely collapsing, while the F1 scores often reach high values at these strengths. On the other hand, moderate values ($\alpha = +30.0, -3.0$) appear to be the optimal stable operating range. A smaller value ($\alpha = \pm 1.0$) shows that the intervention is too weak to steer the model's output. Moreover, it seems that the effectiveness of the steering vector is spatially localized, meaning that the intervention must be applied at specific depths to be effective. For instance, middle-to-late layers are critical for shifting behavior, Recall$\rightarrow$Copy, and similarly, first-to-middle layers are key for changing from Copy$\rightarrow$Recall.


\paragraph{Copying is easier to induce than recall}


\begin{figure}[!htb]
\centering
\includegraphics[width=0.99\linewidth]{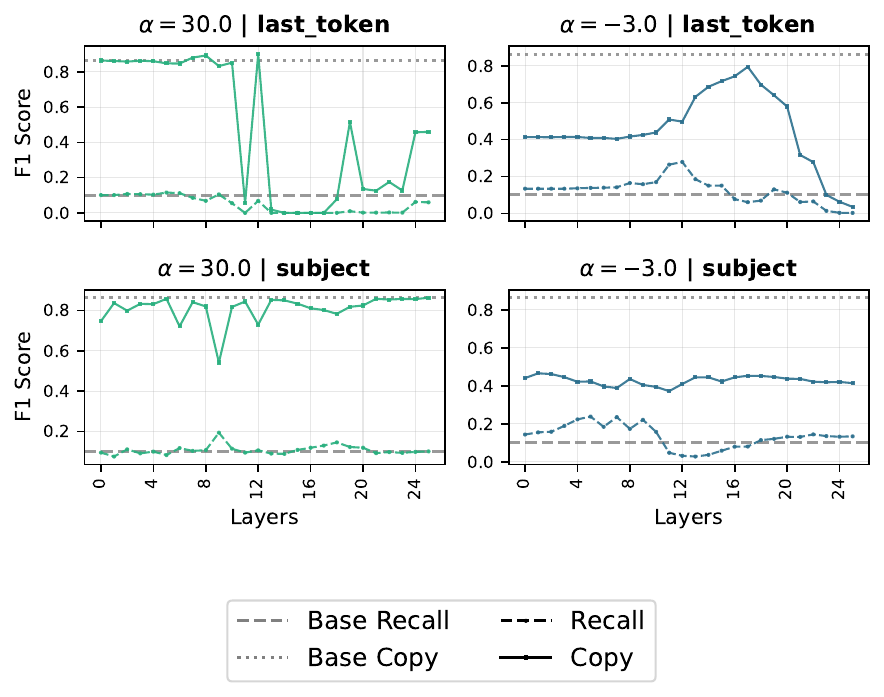}
\caption{Layer-wise F1 score (Gemma2-2B, PopQA) across positions (parametric vs. contextual steering).}
\label{fig:f1-popqa-gemma2-cr-cf-subj-lt}
\end{figure}

\begin{figure}[!htb]
\centering
\includegraphics[width=0.99\linewidth]{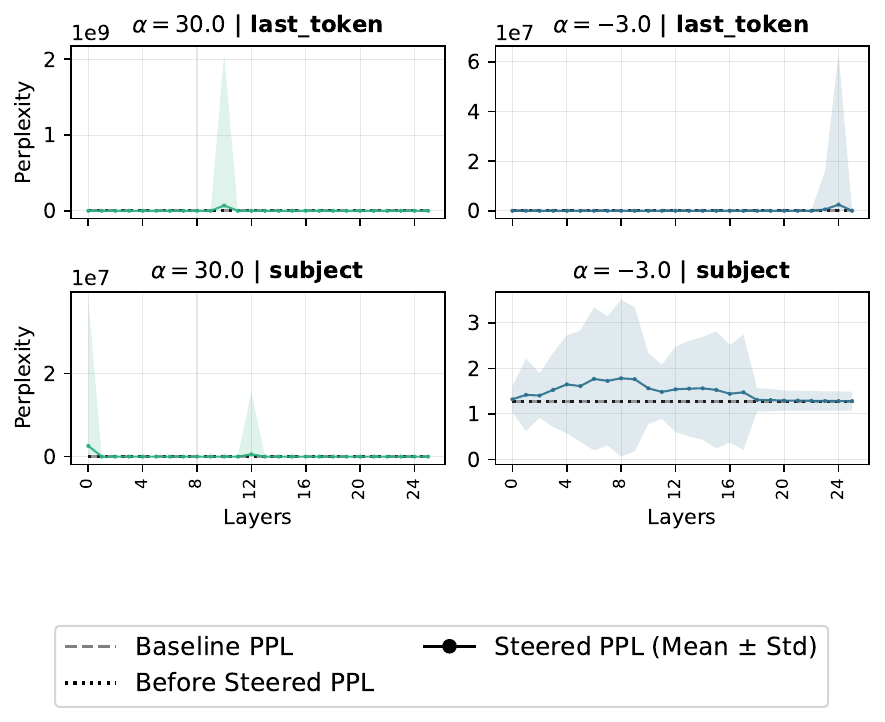}
\caption{Layer-wise PPL score (Gemma2-2B, PopQA) across positions (parametric vs. contextual steering).}
\label{fig:fluency-popqa-gemma2-cr-cf-subj-lt}
\end{figure}

Figures~\ref{fig:f1-popqa-gemma2-cr-cf-subj-lt} and \ref{fig:fluency-popqa-gemma2-cr-cf-subj-lt} show that forcing the model to copy from context is a low-resistance transition that requires minimal scaling, causes minimal increase in perplexity, and succeeds from any control point (object, subject, or last token). On the other hand, forcing recall is a high-resistance operation that requires aggressive suppression of the model, with high required scaling and causing sharp spikes in perplexity. Successful recall is achieved only at the object, where the intervention is trying to jam the attention that is already locked onto the context, thereby blinding the model to the context. However, it fails for the subject as it tries to change a global behavior without specific obfuscation, and unintentionally reinforces the attention link to keep the $o'$ association strong. Interestingly, the last token is too late for intervening because the decision to accept the context has already been made. Although we observe a significant drop in copying ability in both the subject and last token, it does not translate into a restoration of internal information.

\paragraph{Mechanistic Drivers of the Shift in Behavior}

\begin{figure*}[t]
\centering
\includegraphics[width=0.98\textwidth]{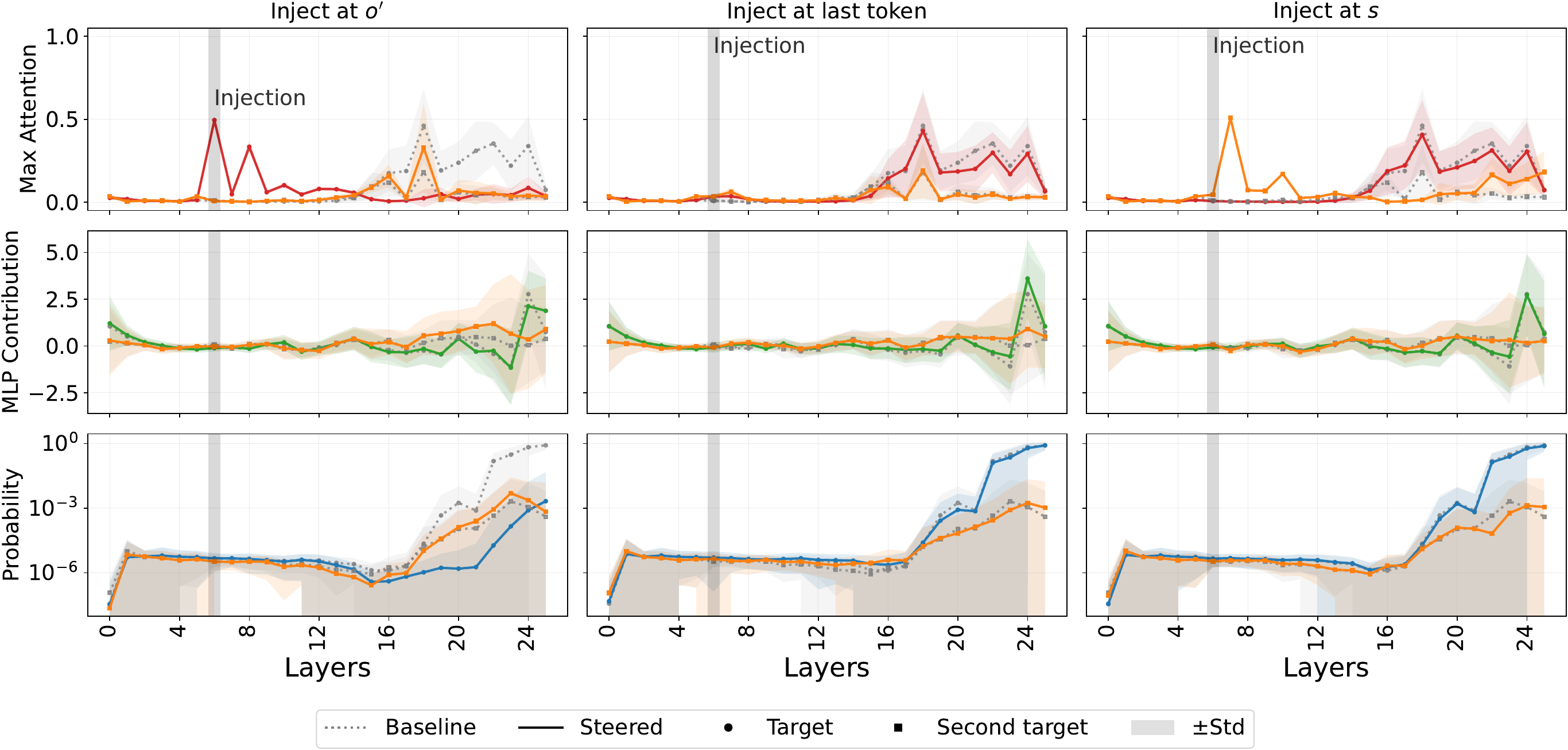}
\caption{
(C$\rightarrow$R, Gemma2 - Context-First. Object vs. last token vs. subject) Mechanistic investigation of parametric steering (object vs. last token vs. subject) over 1,434 PopQA samples. Top: Attention (Orange=Subject, Red=Object). 
%
Middle: MLP (Orange=Truth, Green=Counterfactual). Bottom: Probability (Orange=Truth, Blue=Counterfactual). Successful jamming occurs only at the counterfactual object ($o'$), where attention to the context is suppressed, following a rise in the ground-truth signal in the MLP at the last layer.
}

\label{fig:mi_popqa_gemma-2-2b-it_cr_L-6_context-first}
\end{figure*}

Figure ~\ref{fig:mi_popqa_gemma-2-2b-it_cr_L-6_context-first} shows the result of the mechanistic analysis of the Gemma decoder-only model on the PopQA datasets, with a comparison of interventions at three locations: object, last token, and subject.
As 
can be observed, 
steering is only effective when applied at the counterfactual object ($o'$). At this position, the steering vector acts as a \emph{jamming} signal: it disrupts the model's attention to the counterfactual object in context. 
As can be seen, injection at the object position causes an immediate effect on the \emph{object-focused} attention at the layer of injection but a reduction of this attention in later layers; the \emph{subject-focused} attention is instead increased. Injecting at other positions does not produce as marked a change in attention patterns.
When intervening at the object position, the changes in attention patterns lead to a shift in the model's output probabilities, 
which switch over from the copied counterfactual object to the memorized true object.
The effects in the MLP layers are less discernible, but we observe a slight (but statistically weak) increase in the contribution of MLP layers after the subject-focused attention spike.
%


\section{Related Work}


The work arguably closest to ours was carried out by
\newcite{zhao2025steering}, who 
applied a sparse autoencoder (SpARE) to decompose hidden states in several models, identifying latent features that correlate with the choice between relying on contextual or parametric knowledge. By intervening on those features in the autoencoder space, SpARE can bias the model to resolve conflicts in favor of either the retrieved evidence or the internal memory. 
%
While a subset of their experiments use vector operations similar to ours, our focus is different from theirs. Most importantly, our mechanistic analysis is more granular, and we show that interventions can make the model focus on \emph{individual input tokens} while the method by \newcite{zhao2025steering} controls the high-level choice of the information source. We also investigate a wider range of architecture types, and show that similar mechanisms can be found in an encoder/decoder architecture.

Another work focusing on the geometric aspects of the context-vs-memory trade-off is the paper by \cite{minder2025controllablecontextsensitivityknob}.
They fine-tuned LLMs using training task that makes the model choose between using context or prior knowledge, and identified a single 1-dimensional subspace in one transformer layer that acts as a continuous ``knob'' controlling context sensitivity.
Similarly to \newcite{zhao2025steering}, their analysis is less granular than ours in that they focus on the high-level behavior, while we show that the analysis can be applied to individual tokens. Also, from the perspective of interpretability it is important that our work requires no fine-tuning.

In addition to interpretability work using vector additions and related linear algebra operations, other approaches have been proposed. A line of work has focused on interventions on the attention mechanism:
\newcite{ortu2024competition} applied the logit lens and attention pattern editing to two models, investigating how models handle conflicts between parametric knowledge and contradictory retrieved information. 
%
\newcite{basu2025mechanistic} pinpointed a small set of attention heads that perform  context–to–answer attribution, 
and the PH3 method by \citet{jin2024cutting} applied path patching to identify attention heads responsible for information selection and then pruned selected attention heads to control the tendency of the model to rely on context or memory.
\newcite{yu2023characterizing} study capital-city factual conflicts and localizes copy-vs-recall behavior to specific heads.

\section{Conclusion}

We have carried out a mechanistic analysis of the information selection behavior in LLMs centered on the method of \emph{arbitration vectors}, computed from a model's activations when it is applied to synthetic datasets designed to exemplify copying and recall behaviors.
We showed that by applying a scaled intervention of this vector at specific locations, we could successfully steer the model between behaviors (Copy$\rightarrow$Recall and Recall$\rightarrow$Copy). 
The approach works successfully in decoder-only as well as encoder/decoder architectures.
To assess how generalizable our arbitration vector is, we tested it on two benchmarks with comprehensive evaluation metrics. Our macro-level analysis reveals that copying acts as a stable,  
low-resistance process, while recalling information from a vague context requires a high-resistance intervention.

We observed that steering with a negative scaling ($\alpha = -3.0$) at any control point (object, subject, or last token) acts as an amplifier, triggers a reactivation process where the model's attention reorients to the context, and the MLP layers prompt a response from the contextual information,  with minimal fluency cost (Recall$\rightarrow$Copy). Conversely, the Copy$\rightarrow$Recall direction appears as the most resistant regime. To restore internal information, we need to suppress the dominant copying signal from the contextual information by applying a moderate positive scaling ($\alpha = +30.0$), which initially causes jamming (blinding the model to the context) and subsequently suppresses the copying signal in favor of a weaker parametric memory response. Moreover, experiments at different scales (OLMo-2-0425-1B) suggest that the direction of the asymmetry generalizes across architecture families, though its magnitude may be scale-dependent.

\section*{Limitations}

To verify that our findings are generalizable and to make them statistically more robust, we would like to address the following aspects in future work.

\paragraph{The effect of prompting.} The instruction-tuned LLMs have been prompted using to explicitly make the model copy a minimal span if present and otherwise fall back on its internal knowledge. That instruction may shape the internal decision boundary investigated in the mechanistic analysis. Extended experiments could investigate to what extent the extracted arbitration vector changes with the prompt, and try to extract arbitration vectors that are robust across prompts.

\paragraph{Model scale and diversity.} Our study includes three instruction-tuned architectures spanning two parameter scales and three model families: Gemma2-2B and T5Gemma-2B (2B parameters, decoder-only and encoder-decoder, respectively) and OLMo-2-0425-1B (1B parameters, decoder-only, different family). We have not looked at how the findings scale to larger models, different instruction-tuning recipes, or models with substantially different context handling, e.g., long-context variants or models trained explicitly for RAG.

\paragraph{Statistical sample size.} Because of the combinatorial explosion of experiments needed to investigate layers, $\alpha$ values, counterfactual types, and prompt topologies, 300 samples per benchmark are evaluated, using relation-stratified sampling to ensure coverage.

\paragraph{Modeling the retriever.} Our setup emulates retrieval by inserting a context directly into the prompt, but we do not model retriever failures, ranking, passage selection, or the distribution of retrieval errors. For this reason, we have not investigated how arbitration steering would interact with a realistic end-to-end RAG pipeline. Also, while the questions are taken from the PopQA and PEQ datasets, we have not evaluated the approach on realistic retrieved contexts.

\section*{Acknowledgments}
This research was funded by the Wallenberg AI, Autonomous Systems and Software
Program (WASP) funded by the Knut and Alice Wallenberg Foundation. The computations were
enabled by resources provided by the National Academic Infrastructure for Supercomputing in
Sweden (NAISS) at Alvis partly funded by the Swedish Research Council through grant agreement no. 2022-06725.
We also acknowledge the Computer Science and Engineering department at Chalmers and the University of Gothenburg, which funded Franziska Penzkofer's research internship.

\bibliography{custom}

\appendix
\clearpage
\newpage
\onecolumn

\clearpage
\section{Arbitration Dataset}
\label{app:arbitration-dataset}

This section gives additional details about statistics and the creation of the Arbitration Dataset. This dataset will be released under a Creative Commons license upon acceptance.

\begin{table}[H]
\centering
\small
\setlength{\tabcolsep}{4.6pt}
\renewcommand{\arraystretch}{1.05}

\resizebox{0.98\textwidth}{!}{%
\begin{tabular}{@{}l
                r
                S[table-format=2.2]
                S[table-format=1.2]
                S[table-format=2.2]
                S[table-format=1.2]
                S[table-format=8.1]
                S[table-format=8.1]
                S[table-format=8.1]@{}}
\toprule
Property & {Count} &
{\shortstack{Avg Len\\(IC)}} & {\shortstack{Std\\(IC)}} &
{\shortstack{Avg Len\\(RC)}} & {\shortstack{Std\\(RC)}} &
{\shortstack{Med\\Subj}} & {\shortstack{Med\\Obj}} & {\shortstack{Med\\CF}} \\
\midrule
P101 & 282 & 13.78 & 1.64 & 13.78 & 1.64 & 106822.0 & 1128.0 & 3969.5 \\
P103 & 292 & 13.98 & 0.80 & 13.98 & 0.80 & 77841.5 & 539055.0 & 364599.0 \\
P106 & 282 & 13.80 & 1.61 & 13.80 & 1.61 & 106297.0 & 179.0 & 1178.5 \\
P108 & 256 & 13.93 & 0.70 & 13.93 & 0.70 & 314341.0 & 22777173.0 & 5561078.5 \\
P127 & 286 & 13.79 & 1.41 & 13.79 & 1.41 & 288040.0 & 9873777.0 & 3895844.0 \\
P1303 & 300 & 13.95 & 1.37 & 13.95 & 1.37 & 80477.0 & 289.0 & 671.0 \\
P131 & 290 & 13.87 & 1.41 & 13.87 & 1.41 & 79227.0 & 10737402.0 & 5516304.0 \\
P136 & 290 & 13.83 & 1.39 & 13.83 & 1.39 & 92571.0 & 1919.0 & 11763.0 \\
P1412 & 292 & 13.92 & 0.80 & 13.92 & 0.80 & 93017.5 & 539055.0 & 229146.0 \\
P159 & 280 & 13.82 & 0.74 & 13.82 & 0.74 & 208872.5 & 5413014.0 & 2789733.5 \\
P17 & 290 & 14.02 & 0.80 & 14.02 & 0.80 & 42425.0 & 29396034.0 & 10235079.5 \\
P176 & 290 & 13.14 & 3.36 & 13.14 & 3.36 & 421244.0 & 11566855.0 & 5551629.5 \\
P178 & 212 & 13.86 & 0.76 & 13.86 & 0.76 & 319780.5 & 15100027.5 & 3859624.0 \\
P19 & 292 & 13.92 & 0.77 & 13.92 & 0.77 & 57655.5 & 5648452.5 & 2193780.5 \\
P20 & 284 & 13.87 & 1.59 & 13.87 & 1.59 & 34881.5 & 7432572.5 & 4838149.5 \\
P27 & 296 & 13.98 & 0.78 & 13.98 & 0.78 & 60091.0 & 29396034.0 & 13012700.5 \\
P276 & 290 & 13.82 & 1.60 & 13.82 & 1.60 & 135130.0 & 9425130.0 & 3535733.0 \\
P30 & 292 & 13.96 & 0.77 & 13.96 & 0.77 & 3742.0 & 19142977.0 & 13201572.0 \\
P36 & 298 & 13.81 & 0.76 & 13.81 & 0.76 & 418762.0 & 2990492.0 & 2821216.5 \\
P364 & 288 & 14.05 & 0.74 & 14.05 & 0.74 & 149918.0 & 514211.0 & 479483.0 \\
P407 & 276 & 13.82 & 0.82 & 13.82 & 0.82 & 127371.0 & 542828.5 & 269984.0 \\
P413 & 252 & 14.05 & 0.86 & 14.05 & 0.86 & 144481.0 & 122.0 & 179.0 \\
P449 & 276 & 13.92 & 0.70 & 13.92 & 0.70 & 270520.0 & 9066251.0 & 4605553.0 \\
P463 & 280 & 13.93 & 0.78 & 13.93 & 0.78 & 339023.0 & 12964792.0 & 1180999.5 \\
P495 & 294 & 13.86 & 0.78 & 13.86 & 0.78 & 94866.0 & 38389755.0 & 14450925.0 \\
P740 & 294 & 13.95 & 0.71 & 13.95 & 0.71 & 158121.0 & 6256431.0 & 3541227.5 \\
P937 & 288 & 13.89 & 0.74 & 13.89 & 0.74 & 92818.5 & 10512612.0 & 6131371.0 \\
\bottomrule
\end{tabular}%
}

\caption{
Summary of the Arbitration dataset, which includes statistics for 27 relations extracted from the ParaRel dataset and generated using the OpenAI API. This overview includes sample counts, context lengths, and the median popularity (measured by Wikipedia page views) for the subject, object, and counterfactual object.
}
\label{tb:arbitration-dataset-stats}
\vspace{-0.6em}
\end{table}

You can find the instructions we used to generate the contexts.

\smallskip
\noindent
\begin{Verbatim}[fontsize=\footnotesize]
You generate ONE context sentence from this json input:
{
  "mode": ...,
  "subj": ...,
  "obj": ...,
  "cf": ...,
  "prop_code": ...,
  "query": ...,
}

GLOBAL CONSTRAINTS (apply to BOTH modes)
- Output EXACTLY ONE sentence. No quotes, no lists, no extra text (no quotes, no lists, no explanations).
- Length: 12-16 words. Declarative. No parentheticals. No emojis. No quotations.
- Allowed entities to mention: exactly {subj, cf}. Do NOT mention obj or any other entity.
- Style parity across modes: same neutral news style, similar length and punctuation; NO hedges like 
"contrary to popular belief", "historians confirm", "it's actually true", etc.
- Keep the sentence about the topic of the query (same subject domain). No off-topic small talk.
- Keep simple, factual wording; do not add extra facts beyond the frame below.
- Use ASCII punctuation and plain English. Avoid dates/numbers unless needed by the frame.

RELATION VERBALIZATIONS (for RC only)
Use these exact phrases to assert a relation (choose by prop_code):
"P30": ("What continent is {subj} located in?", "{subj} is located in the continent of {obj}."),
"P127": ("Who owns {subj}?", "{subj} is owned by {obj}."),
"P449": ("What show does {subj} premiere on?", "{subj} premieres on {obj}."),
"P178": ("Who created the product {subj}?", "{subj} is a product created by {obj}."),
"P27": ("What country is {subj} a citizen of?", "{subj} has a citizenship of {obj}."),
"P37": ("What language is spoken in {subj}?", "In {subj}, the language spoken is a mixture of {obj}."),
"P108": ("Who employs {subj}?", "{subj} is employed by {obj}."),
"P364": ("What is the original language of {subj}?", "The original language of {subj} is {obj}."),
"P159": ("Where is the headquarters of {subj}?", "The headquarters of {subj} is in {obj}."),
"P407": ("In what language was {subj} written?", "{subj} was written in {obj}."),
"P19": ("Where was {subj} born?", "{subj} was born in {obj}."),
"P1412": ("What language did {subj} use?", "The language used by {subj} is {obj}."),
"P641": ("What sport does {subj} play professionally?", "{subj} professionally plays the sport of {obj}."),
"P495": ("Where did {subj} originate?", "{subj} originated in {obj}."),
"P36": ("What is the capital of {subj}?", "The capital of {subj} is {obj}."),
"P463": ("What organization is {subj} a member of?", "{subj} is a member of {obj}."),
"P20": ("When did {subj} die?", "{subj} passed away in {obj}."),
"P276": ("Where is {subj} located?", "{subj} is located in {obj}."),
"P101": ("What field does {subj} specialize in?", "{subj} specializes in the study of {obj}."),
"P1303": ("What instrument does {subj} perform on?", "{subj} performs on the {obj}."),
"P140": ("What religion does {subj} follow?", "{subj} follows the religion of {obj}."),
"P39": ("What position does {subj} hold?", "{subj} holds the position of {obj}."),
"P106": ("What is the profession of {subj}?", "{subj} works as a {obj}."),
"P136": ("What genre does {subj} belong to?", "The genre of {subj} is {obj}."),
"P413": ("What position does {subj} play?", "{subj} plays as a {obj}."),
"P176": ("Who created {subj}?", "{subj} was created by {obj}."),
"P937": ("In what city did {subj} work?", "{subj} worked in the city of {obj}."),
"P740": ("When was {subj} formed and when did it first operate?", 
"{subj} was formed in 1883, and the first train to {obj} departed that year."),
"P138": ("Who or what is {subj} named after?", "{subj} is named after {obj}."),
"P131": ("What province is {subj} in?", "{subj} is in the province of {obj}."),
"P17": ("What country is {subj} located in?", "{subj} is located in the country of {obj}."),
"P103": ("What is the native language of {subj}?", "The native language of {subj} is {obj}."),
"P190": ("What city is {subj} a twin of?", "{subj} is a twin city of {obj}.")
(If prop_code is unknown, fall back to: "<subj> is associated with <cf>")

MODE-SPECIFIC INSTRUCTIONS

If mode == "IC" (Irrelevant Context):
- Must mention 'cf' EXACTLY once; MUST NOT mention 'subj' or 'obj'.
- MUST be OFF-TOPIC relative to the subject/query domain.
  • Choose a neutral everyday theme UNRELATED to the subject/query: 
  cooking, weather, stationery, commuting, gardening,
    household chores, generic shopping (not music/TV/film/business/politics unless the subject is about those).
  • MUST satisfy the LEXICAL VETO: avoid any subject- or relation-related words.
- Keep neutral, simple phrasing; do NOT imply any relation between 'subj' and 'cf'.

If mode == "RC" (Relevant Context non-entailing control):
- Must mention 'subj' EXACTLY once and 'cf' EXACTLY once.
- MUST assert the relation using the exact verbalization for prop_code (above).
- MUST NOT mention 'obj'.
- Keep wording concise and plain; you may add a short neutral tail like "according to records."

CHECKLIST BEFORE OUTPUT
- Exactly one sentence, 12–16 words.
- RC: includes 'subj' once AND 'cf' once; uses the correct relation phrase; no 'obj'.
- IC: includes 'cf' once; NO 'subj'; NO 'obj'; off-topic; contains NO VETO words.
- No hedges; neutral tone; ASCII punctuation.
\end{Verbatim}

\clearpage
\section{Evaluation Benchmark}
\label{app:evaluation-benchmark}

\begin{table}[h]
    \centering
    \small
    \begin{tabularx}{\linewidth}{>{\raggedright\arraybackslash}X}
        \toprule
        \textbf{Template} \\
        \midrule
        The \{obj\} was documented in the official records. \\
        Local sources reference the \{obj\} in their archives. \\
        The \{obj\} appears in the standard documentation. \\
        Historical files mention the \{obj\} briefly. \\
        The \{obj\} is listed in the reference materials. \\
        Official logs include the \{obj\} entry. \\
        The \{obj\} was noted in the meeting minutes. \\
        Records from that period show the \{obj\}. \\
        The \{obj\} is included in the database. \\
        Standard reports contain the \{obj\}. \\
        \bottomrule
    \end{tabularx}
    \caption{Generic ``archive templates'' used for the \textit{Forcing Recall} evaluation.}
    \label{tb:forcing-copy-templates}
\end{table}

\begin{xltabular}{\linewidth}{l >{\raggedright\arraybackslash}X}
    \toprule
    \textbf{Relation} & \textbf{Templates} \\
    \midrule
    \endfirsthead

    \multicolumn{2}{c}%
    {{\bfseries \tablename\ \thetable{} -- continued from previous page}} \\
    \toprule
    \textbf{Relation} & \textbf{Templates} \\
    \midrule
    \endhead

    \midrule
    \multicolumn{2}{r}{{Continued on next page}} \\
    \bottomrule
    \endfoot

    \bottomrule
    \endlastfoot

    P106 & \{subj\} is employed as a \{obj\} according to structured data. \newline \{subj\} works as a \{obj\}. \newline The occupation of \{subj\} is \{obj\}. \newline \{subj\} holds the position of \{obj\}. \newline \{subj\}'s profession is \{obj\}. \newline The job title of \{subj\} is \{obj\}. \\ \midrule
    P19 & According to records, \{subj\} was born in \{obj\}. \newline \{subj\} was born in \{obj\}. \newline The birthplace of \{subj\} is \{obj\}. \newline \{subj\}'s place of birth is \{obj\}. \newline \{obj\} is where \{subj\} was born. \newline \{subj\} originated from \{obj\}. \\ \midrule
    P17 & \{subj\} is located in the country of \{obj\}. \newline \{subj\} is in \{obj\}. \newline The country where \{subj\} is located is \{obj\}. \newline \{subj\} can be found in \{obj\}. \newline \{obj\} is the country containing \{subj\}. \newline \{subj\} is situated within \{obj\}. \\ \midrule
    P36 & According to records, the capital of \{subj\} is \{obj\}. \newline The capital of \{subj\} is \{obj\}. \newline \{obj\} serves as the capital of \{subj\}. \newline \{obj\} is the capital city of \{subj\}. \newline \{subj\}'s capital is \{obj\}. \newline \{subj\} has \{obj\} as its capital. \\ \midrule
    P50 & According to records, \{obj\} authored \{subj\}. \newline \{obj\} wrote \{subj\}. \newline The author of \{subj\} is \{obj\}. \newline \{subj\} was written by \{obj\}. \newline \{obj\} is the author of \{subj\}. \newline \{subj\} was authored by \{obj\}. \\ \midrule
    P175 & \{obj\} performed the song \{subj\}. \newline \{obj\} is the performer of \{subj\}. \newline The song \{subj\} was performed by \{obj\}. \newline \{subj\} is performed by \{obj\}. \newline \{obj\} is the artist behind \{subj\}. \newline The performer of \{subj\} is \{obj\}. \\ \midrule
    P264 & \{subj\} is represented by the music label \{obj\}. \newline \{subj\} is signed to \{obj\}. \newline The record label of \{subj\} is \{obj\}. \newline \{obj\} is the label representing \{subj\}. \newline \{subj\}'s music label is \{obj\}. \newline \{subj\} records under \{obj\}. \\ \midrule
    P131 & \{subj\} is located in \{obj\}. \newline \{subj\} is situated in \{obj\}. \newline \{obj\} contains \{subj\}. \newline The administrative location of \{subj\} is \{obj\}. \newline \{subj\} can be found in \{obj\}. \newline \{subj\} is part of \{obj\}. \\ \midrule
    P495 & \{subj\} was created in \{obj\}. \newline \{subj\} originates from \{obj\}. \newline The country of origin of \{subj\} is \{obj\}. \newline \{obj\} is where \{subj\} was created. \newline \{subj\} was produced in \{obj\}. \newline \{subj\} comes from \{obj\}. \\ \midrule
    P276 & The \{subj\} took place in \{obj\}. \newline \{subj\} occurred in \{obj\}. \newline The location of \{subj\} was \{obj\}. \newline \{subj\} happened in \{obj\}. \newline \{obj\} was the location of \{subj\}. \newline \{subj\} was held in \{obj\}. \\ \midrule
    P40 & \{obj\} is the child of \{subj\}. \newline \{subj\} is the parent of \{obj\}. \newline \{obj\} is \{subj\}'s child. \newline The child of \{subj\} is \{obj\}. \newline \{subj\} has a child named \{obj\}. \newline \{obj\} is the offspring of \{subj\}. \\ \midrule
    P159 & The headquarters of \{subj\} is located in \{obj\}. \newline \{subj\} is headquartered in \{obj\}. \newline The main office of \{subj\} is in \{obj\}. \newline \{obj\} is the headquarters location of \{subj\}. \newline \{subj\}'s headquarters is in \{obj\}. \newline \{subj\} has its headquarters in \{obj\}. \\ \midrule
    P176 & The \{subj\} is produced by the company \{obj\}. \newline \{obj\} manufactures \{subj\}. \newline The manufacturer of \{subj\} is \{obj\}. \newline \{subj\} is made by \{obj\}. \newline \{obj\} produces \{subj\}. \newline \{obj\} is the producer of \{subj\}. \\ \midrule
    P26 & \{subj\} is married to \{obj\}. \newline \{subj\}'s spouse is \{obj\}. \newline The spouse of \{subj\} is \{obj\}. \newline \{obj\} is the spouse of \{subj\}. \newline \{subj\} and \{obj\} are married. \newline \{subj\} is wed to \{obj\}. \\ \midrule
    P127 & \{subj\} is owned by \{obj\}. \newline \{obj\} owns \{subj\}. \newline The owner of \{subj\} is \{obj\}. \newline \{subj\} belongs to \{obj\}. \newline \{obj\} is the owner of \{subj\}. \newline \{subj\}'s owner is \{obj\}. \\ \midrule
    P69 & \{subj\} received their education at \{obj\}. \newline \{subj\} studied at \{obj\}. \newline \{subj\} attended \{obj\}. \newline The alma mater of \{subj\} is \{obj\}. \newline \{obj\} is where \{subj\} was educated. \newline \{subj\} was educated at \{obj\}. \\ \midrule
    P740 & \{subj\} was founded in \{obj\}. \newline \{subj\} was formed in \{obj\}. \newline The formation location of \{subj\} is \{obj\}. \newline \{obj\} is where \{subj\} was founded. \newline \{subj\} originated in \{obj\}. \newline \{subj\}'s formation location is \{obj\}. \\ \midrule
    P20 & \{subj\} passed away in \{obj\}. \newline \{subj\} died in \{obj\}. \newline The place of death of \{subj\} is \{obj\}. \newline \{obj\} is where \{subj\} died. \newline \{subj\}'s place of death is \{obj\}. \newline \{subj\} died in the location of \{obj\}. \\ \midrule
    P112 & \{obj\} founded the \{subj\}. \newline \{obj\} is the founder of \{subj\}. \newline The founder of \{subj\} is \{obj\}. \newline \{subj\} was founded by \{obj\}. \newline \{obj\} established \{subj\}. \newline \{obj\} created \{subj\}. \\ \midrule
    P407 & \{subj\} was written in the \{obj\} language. \newline The language of \{subj\} is \{obj\}. \newline \{subj\} is in \{obj\}. \newline \{obj\} is the language of \{subj\}. \newline \{subj\} was composed in \{obj\}. \newline The language used in \{subj\} is \{obj\}. \\ \midrule
    P170 & \{obj\} is credited with creating \{subj\}. \newline \{obj\} created \{subj\}. \newline The creator of \{subj\} is \{obj\}. \newline \{subj\} was created by \{obj\}. \newline \{obj\} is the creator of \{subj\}. \newline \{obj\} designed \{subj\}. \\ \midrule
    P800 & \{subj\} is renowned for \{obj\}. \newline \{subj\} is known for \{obj\}. \newline \{subj\} is famous for creating \{obj\}. \newline The notable work of \{subj\} is \{obj\}. \newline \{subj\}'s notable work includes \{obj\}. \newline \{obj\} is a notable work by \{subj\}. \\ \midrule
    P413 & \{subj\} plays in the position of \{obj\}. \newline \{subj\}'s position is \{obj\}. \newline The playing position of \{subj\} is \{obj\}. \newline \{subj\} plays as a \{obj\}. \newline \{obj\} is the position played by \{subj\}. \newline \{subj\} is positioned as \{obj\}. \\ \midrule
    P641 & \{subj\} plays the sport of \{obj\} professionally. \newline \{subj\} is a professional \{obj\} player. \newline The sport played by \{subj\} is \{obj\}. \newline \{subj\} competes in \{obj\}. \newline \{obj\} is the sport of \{subj\}. \newline \{subj\} is an athlete in \{obj\}. \\ \midrule
    P449 & \{subj\} premiered on the network \{obj\}. \newline \{subj\} first aired on \{obj\}. \newline The original broadcaster of \{subj\} is \{obj\}. \newline \{obj\} is the network that premiered \{subj\}. \newline \{subj\} was originally broadcast on \{obj\}. \newline \{subj\} debuted on \{obj\}. \\ \midrule
    P136 & \{subj\} belongs to the \{obj\} genre. \newline \{subj\} is a \{obj\} work. \newline The genre of \{subj\} is \{obj\}. \newline \{subj\} is classified as \{obj\}. \newline \{obj\} is the genre of \{subj\}. \newline \{subj\} falls under the \{obj\} genre. \\ \midrule
    P37 & The official language spoken in \{subj\} is \{obj\}. \newline \{obj\} is the official language of \{subj\}. \newline \{subj\} has \{obj\} as its official language. \newline The official language of \{subj\} is \{obj\}. \newline \{subj\}'s official language is \{obj\}. \newline In \{subj\}, the official language is \{obj\}. \\ \midrule
    P108 & \{subj\} is employed by \{obj\}. \newline \{subj\} works for \{obj\}. \newline The employer of \{subj\} is \{obj\}. \newline \{obj\} employs \{subj\}. \newline \{subj\}'s employer is \{obj\}. \newline \{subj\} is an employee of \{obj\}. \\ \midrule
    P178 & The product \{subj\} was developed by \{obj\}. \newline \{obj\} developed \{subj\}. \newline The developer of \{subj\} is \{obj\}. \newline \{subj\} is developed by \{obj\}. \newline \{obj\} is the developer of \{subj\}. \newline \{subj\} was created by \{obj\}. \\ \midrule
    P463 & \{subj\} is a member of \{obj\}. \newline \{subj\} belongs to \{obj\}. \newline \{subj\} is part of \{obj\}. \newline The organization \{obj\} has \{subj\} as a member. \newline \{subj\}'s membership includes \{obj\}. \newline \{subj\} holds membership in \{obj\}. \\ \midrule
    P30 & \{subj\} is located on the continent of \{obj\}. \newline \{subj\} is in \{obj\}. \newline The continent of \{subj\} is \{obj\}. \newline \{obj\} is the continent where \{subj\} is located. \newline \{subj\} can be found on \{obj\}. \newline \{subj\} is situated on \{obj\}. \\ \midrule
    P101 & \{subj\} specializes in the field of \{obj\}. \newline \{subj\}'s field of work is \{obj\}. \newline The field of work of \{subj\} is \{obj\}. \newline \{subj\} works in \{obj\}. \newline \{obj\} is the specialization of \{subj\}. \newline \{subj\}'s area of expertise is \{obj\}. \\ \midrule
    P103 & The native language of \{subj\} is \{obj\}. \newline \{subj\}'s native language is \{obj\}. \newline \{obj\} is the native language of \{subj\}. \newline \{subj\} is a native speaker of \{obj\}. \newline \{subj\} speaks \{obj\} natively. \newline The mother tongue of \{subj\} is \{obj\}. \\ \midrule
    P1303 & \{subj\} performs on the \{obj\}. \newline \{subj\} plays the \{obj\}. \newline The instrument played by \{subj\} is \{obj\}. \newline \{subj\} is a \{obj\} player. \newline \{obj\} is the instrument of \{subj\}. \newline \{subj\} performs using \{obj\}. \\ \midrule
    P27 & \{subj\} is a citizen of \{obj\}. \newline \{subj\} has \{obj\} citizenship. \newline The citizenship of \{subj\} is \{obj\}. \newline \{subj\} holds citizenship in \{obj\}. \newline \{obj\} is the country of citizenship for \{subj\}. \newline \{subj\} is a national of \{obj\}. \\ \midrule
    P364 & The original language of \{subj\} is \{obj\}. \newline \{subj\} was originally in \{obj\}. \newline \{obj\} is the original language of \{subj\}. \newline \{subj\}'s original language is \{obj\}. \newline The language \{subj\} was created in is \{obj\}. \newline \{subj\} was first released in \{obj\}. \\ \midrule
    P937 & \{subj\} worked in \{obj\}. \newline \{subj\}'s work location was \{obj\}. \newline The work location of \{subj\} is \{obj\}. \newline \{obj\} is where \{subj\} worked. \newline \{subj\} was active in \{obj\}. \newline \{subj\} conducted work in \{obj\}. \\ \midrule
    P1412 & \{subj\} used the \{obj\} language. \newline \{subj\} speaks \{obj\}. \newline \{subj\} is fluent in \{obj\}. \newline The language spoken by \{subj\} is \{obj\}. \newline \{obj\} is a language used by \{subj\}. \newline \{subj\} can communicate in \{obj\}. \\
    \caption{Relation-specific authoritative templates used for the \textit{Forcing Copy} evaluation.}
    \label{tb:forcing-copy-templates}
\end{xltabular}
\clearpage
\section{Controlling High-level Arbitration Behavior}
\label{app:ob_base_vs_steer}

\begin{figure}[H]
\centering
\includegraphics[width=\textwidth]{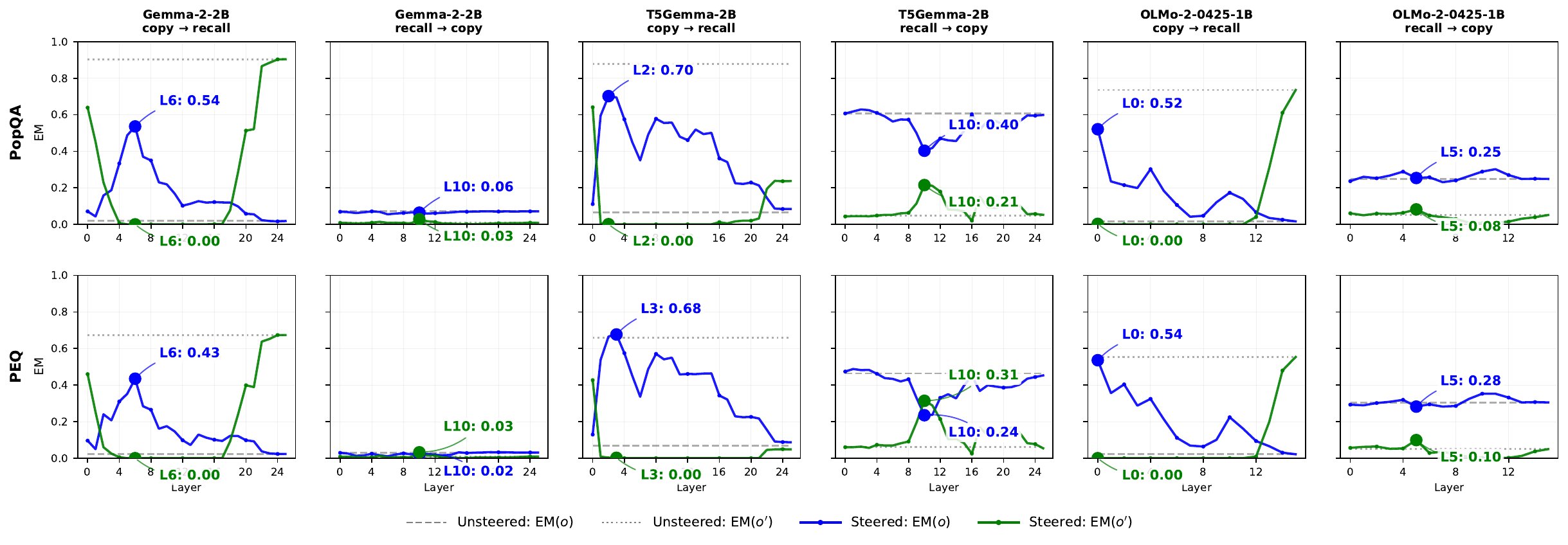}

\caption{
Layer-wise EM on PopQA (top row) and PEQ (bottom row), steering on object tokens when the context comes first, across three architectures: Gemma2-2B, T5Gemma-2B and OLMo-2-0425-1B. Columns 1, 3, 5 show parametric steering (Copy$\rightarrow$Recall, $\alpha = 30.0$) recovering internal knowledge (peaks on PopQA: 54\%, 70\%, 52\%; on PEQ: 43\%, 68\%, 54\%). Columns 2, 4, 6 show contextual steering (Recall$\rightarrow$Copy, $\alpha = -3.0$), inducing copying behavior (peaks on PopQA: 3\%, 21\%, 8\%; on PEQ: 3\%, 31\%, 10\%).
}
\label{fig:ov_base_vs_steer_context-first}
\end{figure}
\clearpage
\section{Mechanistic Investigations}
\label{app:mi-investigations}

\begin{figure}[!ht]
\centering
\includegraphics[width=0.98\textwidth]{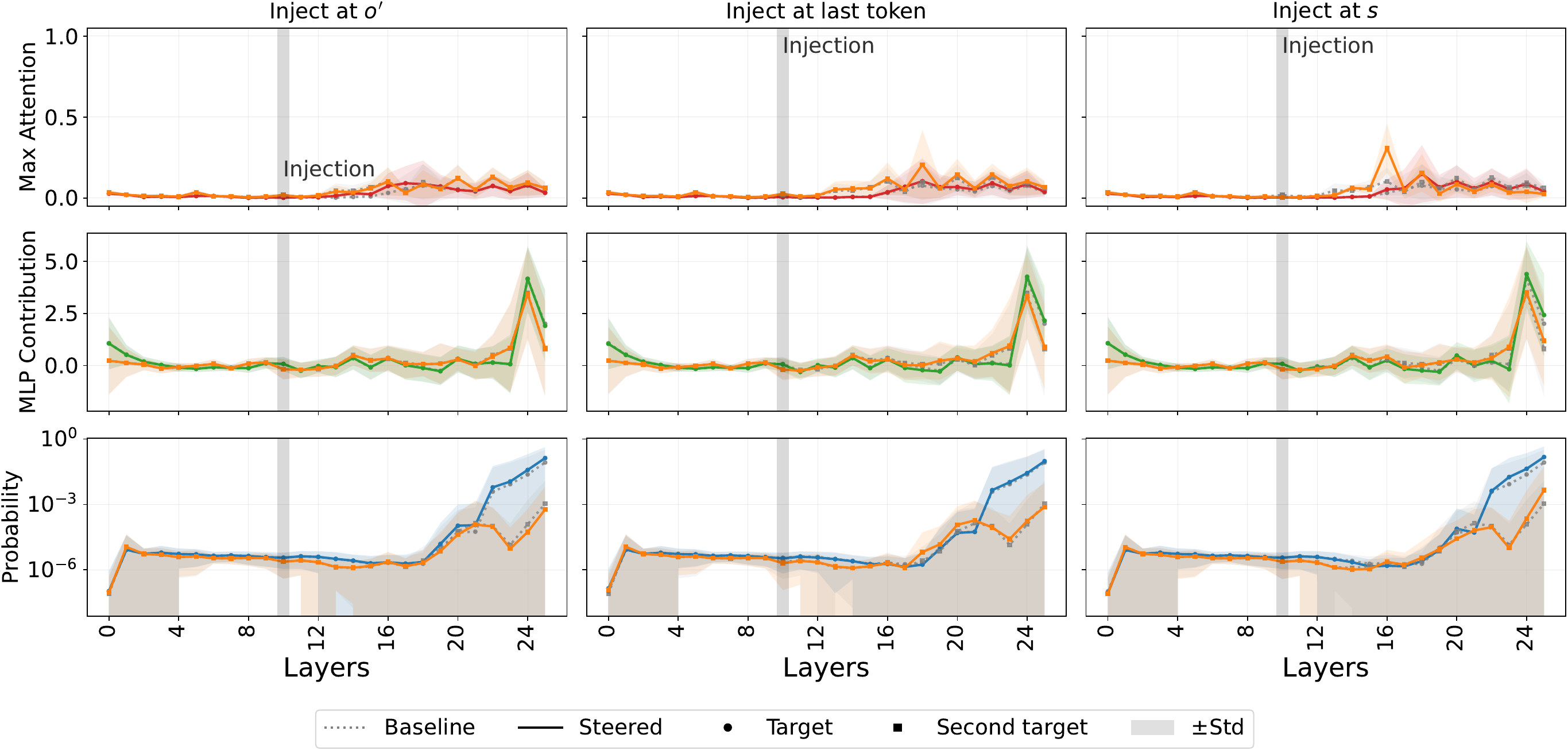}
\caption{
(R$\rightarrow$C, Gemma2 - Context-First: Object vs. last token vs. subject) Showing the mechanistic investigation of parametric steering over 1,434 PopQA samples. Top: Attention (Subject=Orange, Counterfactual=Red). Middle: MLP(Truth=Orange, Counterfactual=Green). Bottom: Probability(Truth=Orange, Counterfactual=Blue). The intervention acts as an amplifier and successfully shifts the behavior from all injection positions.
}
\label{fig:mi_popqa_gemma-2-2b-it_rc_L-10_context-first}
\end{figure}

Figure~\ref{fig:mi_popqa_gemma-2-2b-it_rc_L-10_context-first}, the reverse direction (Recall$\rightarrow$Copy) presents an entirely different dynamic. Intervention at any control point, object, subject, or last token leads to successfully shift in behavior, confirming that amplifying the copying is a low-resistance transition. The attention plots show that the model immediately reorients its focus to the context (reactivation); once the attention has locked onto the context, the MLP layers take that retrieved information and promote it as the output. Interestingly, even in this regime, we see a decision lag similar to the Copy$\rightarrow$Recall regime. \\

\begin{figure}[!ht]
\centering
\includegraphics[width=0.98\textwidth]{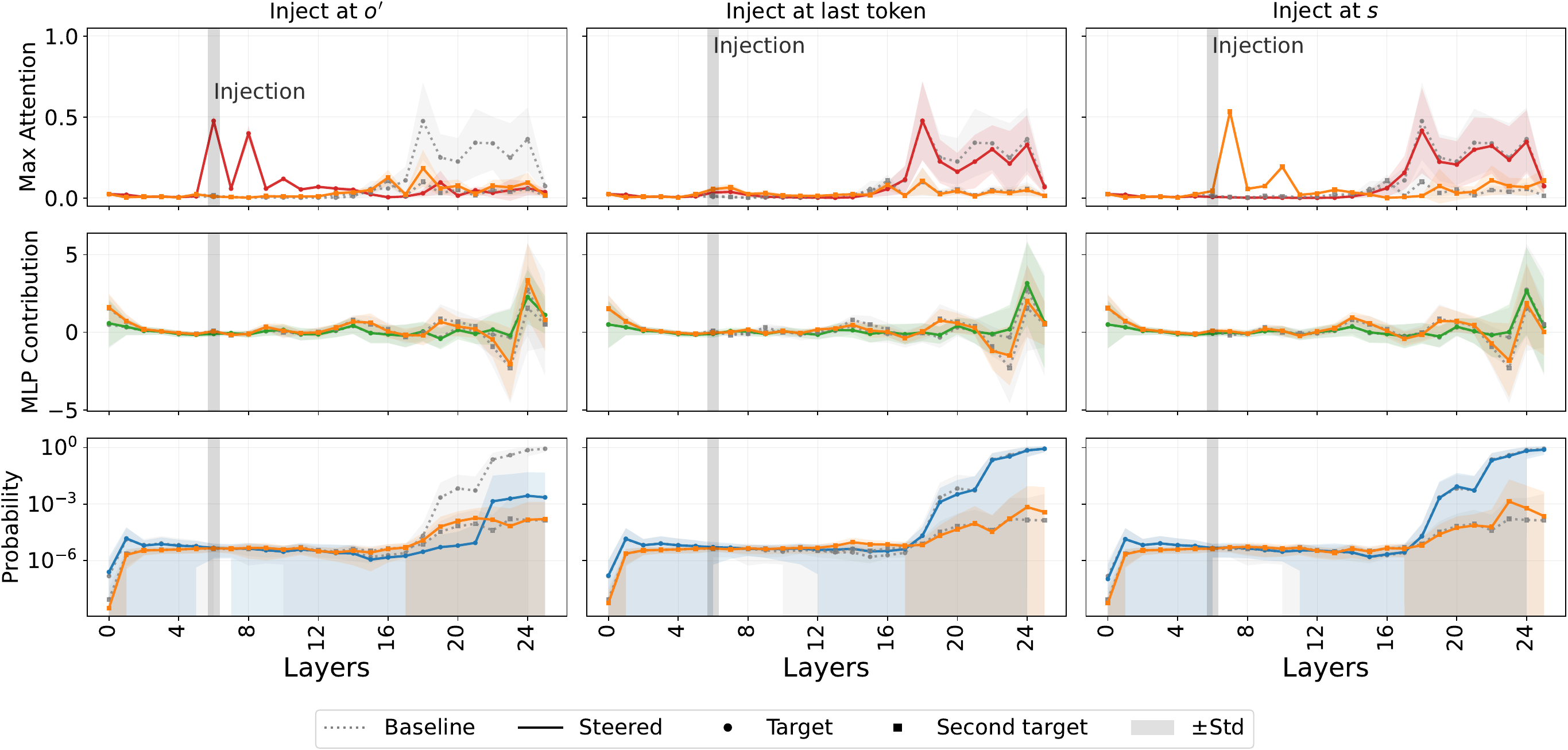}
\caption{
(C$\rightarrow$R, Gemma2 - Context-First: Object vs. last token vs. subject) Showing the mechanistic investigation of parametric steering over 1,026 PEQ samples. Top: Attention (Subject=Orange, Counterfactual=Red). Middle: MLP(Truth=Orange, Counterfactual=Green). Bottom: Probability(Truth=Orange, Counterfactual=Blue). The intervention acts as an amplifier and successfully shifts the behavior from all injection positions.
}
\label{fig:mi_peq_gemma-2-2b-it_cr_L-6_context-first}
\end{figure}

\begin{figure}[!ht]
\centering
\includegraphics[width=0.98\textwidth]{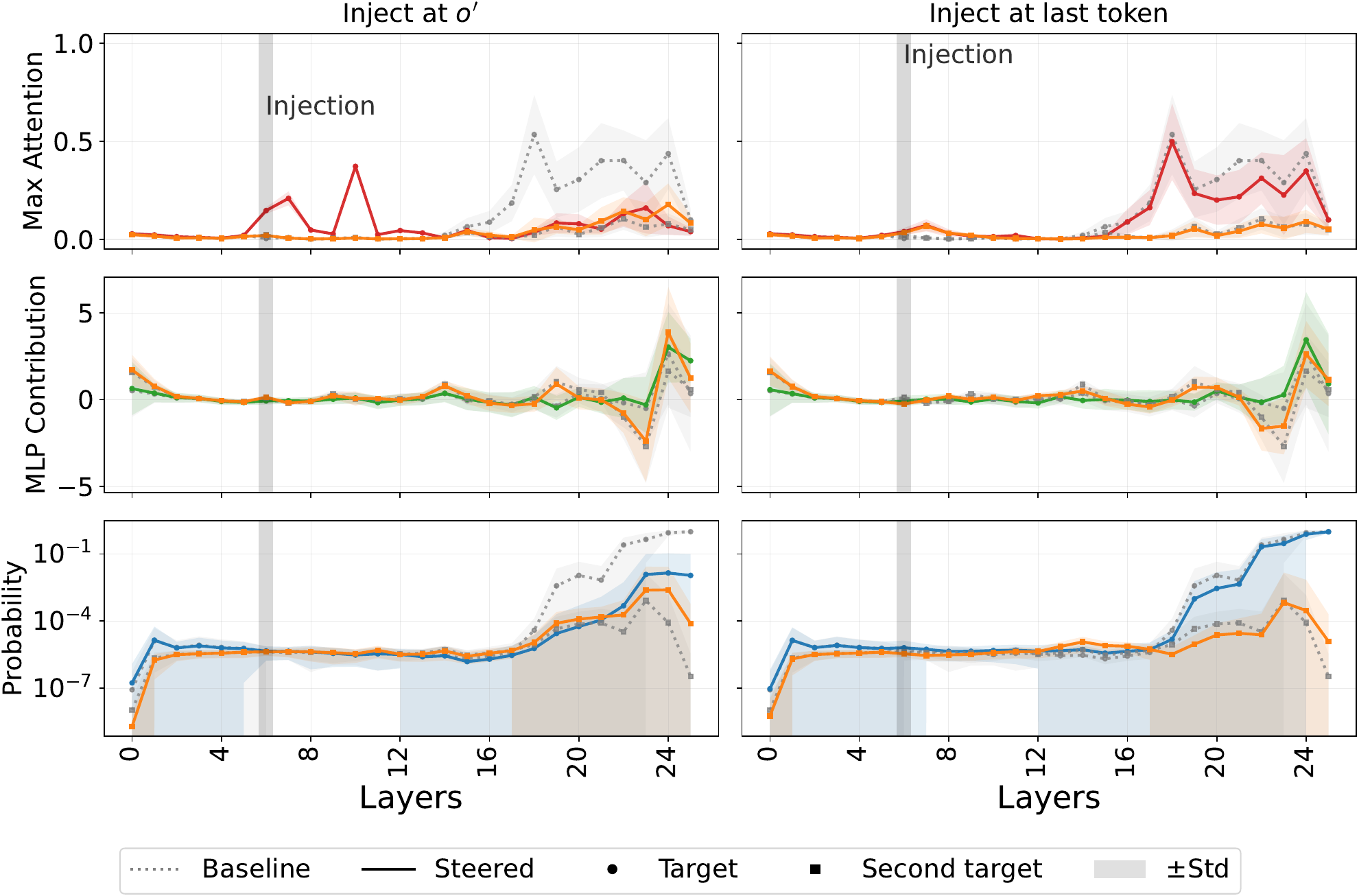}
\caption{
(C$\rightarrow$R, Gemma2 - Query-First: Object vs. last token only) Showing the mechanistic investigation of parametric steering over 690 PEQ samples. Top: Attention (Subject=Orange, Counterfactual=Red). Middle: MLP(Truth=Orange, Counterfactual=Green). Bottom: Probability(Truth=Orange, Counterfactual=Blue). The intervention acts as an amplifier and successfully shifts the behavior from all injection positions.
}
\label{fig:mi_peq_gemma-2-2b-it_cr_L-6_query-first}
\end{figure}

\begin{figure}[!ht]
\centering
\includegraphics[width=0.98\textwidth]{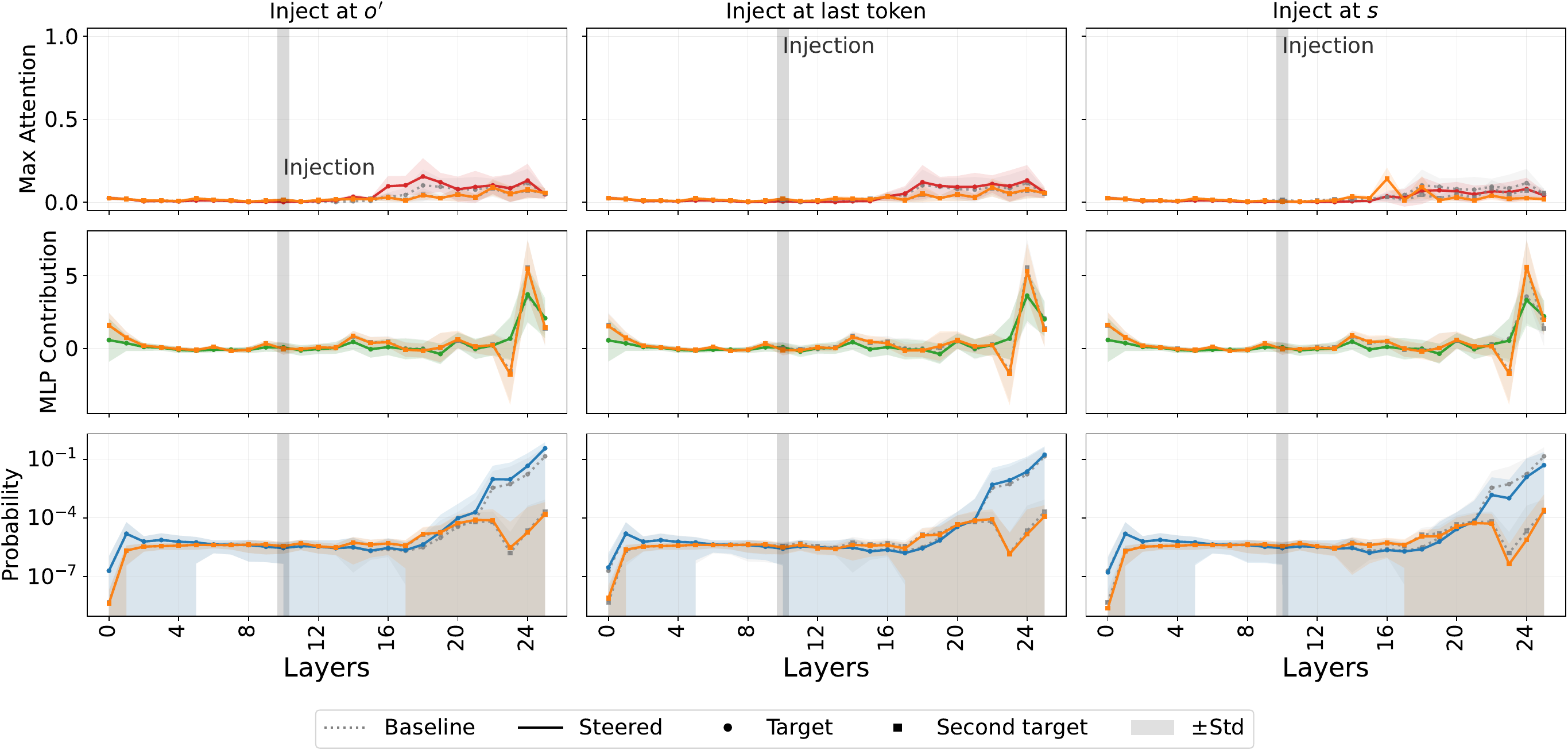}
\caption{
(R$\rightarrow$C, Gemma2 - Context-First: Object vs. last token vs. subject) Showing the mechanistic investigation of parametric steering over 1,026 PEQ samples. Top: Attention (Subject=Orange, Counterfactual=Red). Middle: MLP(Truth=Orange, Counterfactual=Green). Bottom: Probability(Truth=Orange, Counterfactual=Blue). The intervention acts as an amplifier and successfully shifts the behavior from all injection positions.
}
\label{fig:mi_peq_gemma-2-2b-it_rc_L-10_context-first}
\end{figure}

\begin{figure}[!ht]
\centering
\includegraphics[width=0.98\textwidth]{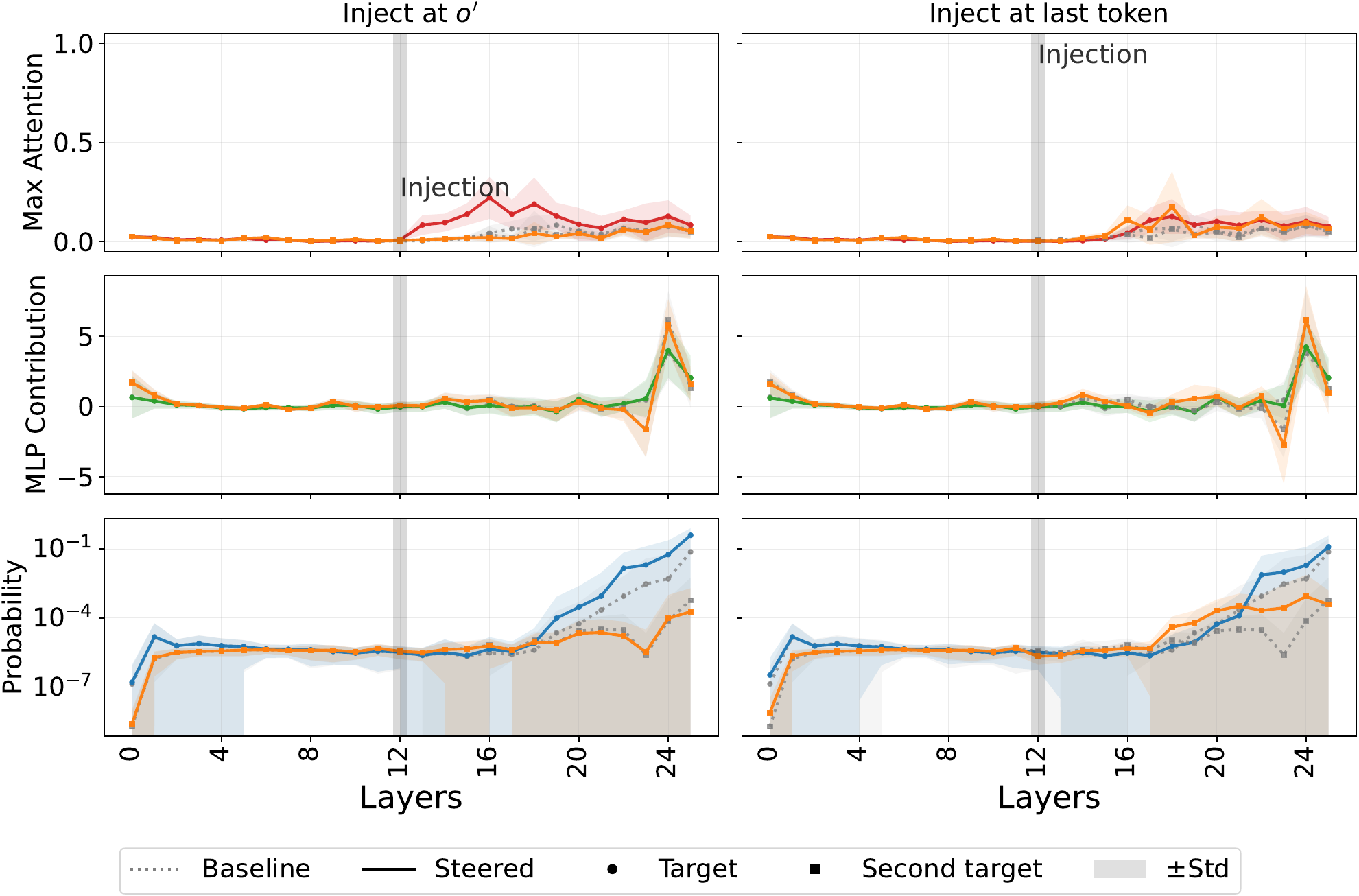}
\caption{
(R$\rightarrow$C, Gemma2 - Query-First: Object vs. last token only) Showing the mechanistic investigation of parametric steering over 690 PEQ samples. Top: Attention (Subject=Orange, Counterfactual=Red). Middle: MLP(Truth=Orange, Counterfactual=Green). Bottom: Probability(Truth=Orange, Counterfactual=Blue). The intervention acts as an amplifier and successfully shifts the behavior from all injection positions.
}
\label{fig:mi_peq_gemma-2-2b-it_rc_L-12_query-first}
\end{figure}

\begin{figure}[!ht]
\centering
\includegraphics[width=0.98\textwidth]{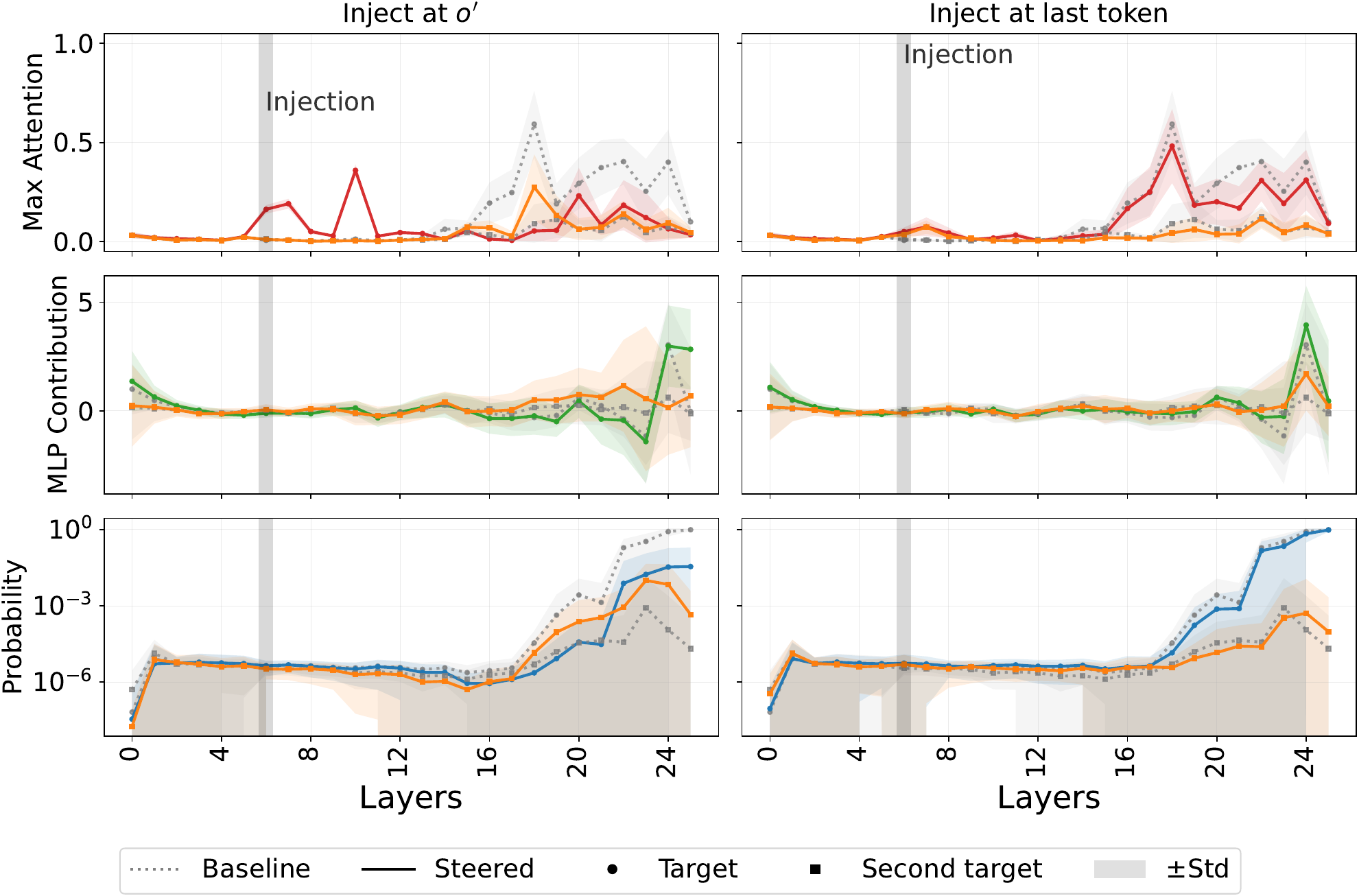}
\caption{
(C$\rightarrow$R, Gemma2 - Query-First: Object vs. last token only) Showing the mechanistic investigation of parametric steering over 1,032 PopQA samples. Top: Attention (Subject=Orange, Counterfactual=Red). Middle: MLP(Truth=Orange, Counterfactual=Green). Bottom: Probability(Truth=Orange, Counterfactual=Blue). The intervention acts as an amplifier and successfully shifts the behavior from all injection positions.
}
\label{fig:mi_popqa_gemma-2-2b-it_cr_L-6_query-first}
\end{figure}

\begin{figure}[!ht]
\centering
\includegraphics[width=0.98\textwidth]{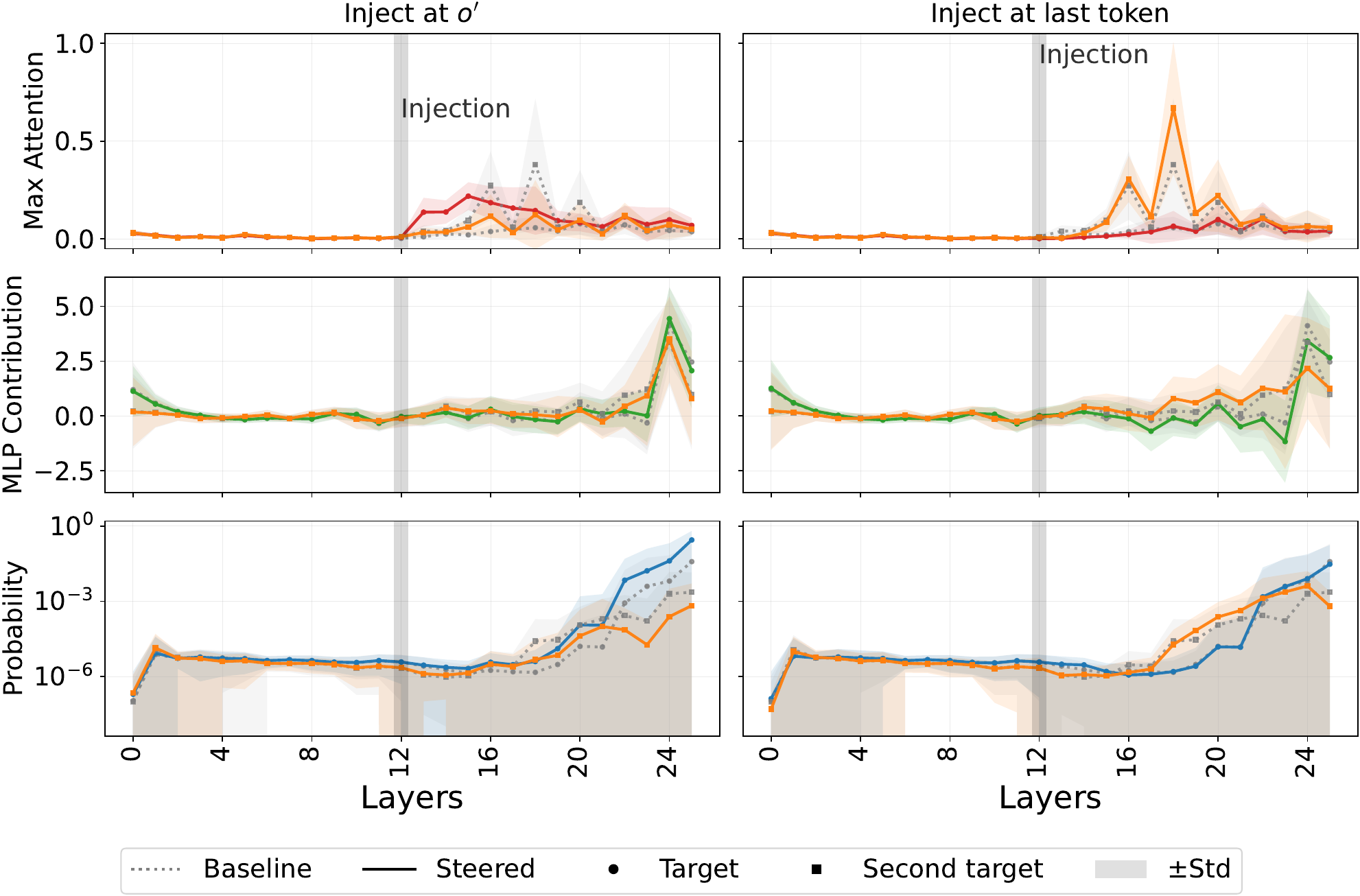}
\caption{
(R$\rightarrow$C, Gemma2 - Query-First: Object vs. last token only) Showing the mechanistic investigation of parametric steering over 1,032 PopQA samples. Top: Attention (Subject=Orange, Counterfactual=Red). Middle: MLP(Truth=Orange, Counterfactual=Green). Bottom: Probability(Truth=Orange, Counterfactual=Blue). The intervention acts as an amplifier and successfully shifts the behavior from all injection positions.
}
\label{fig:mi_popqa_gemma-2-2b-it_rc_L-12_query-first}
\end{figure}

\begin{figure}[!ht]
\centering
\includegraphics[width=0.98\textwidth]{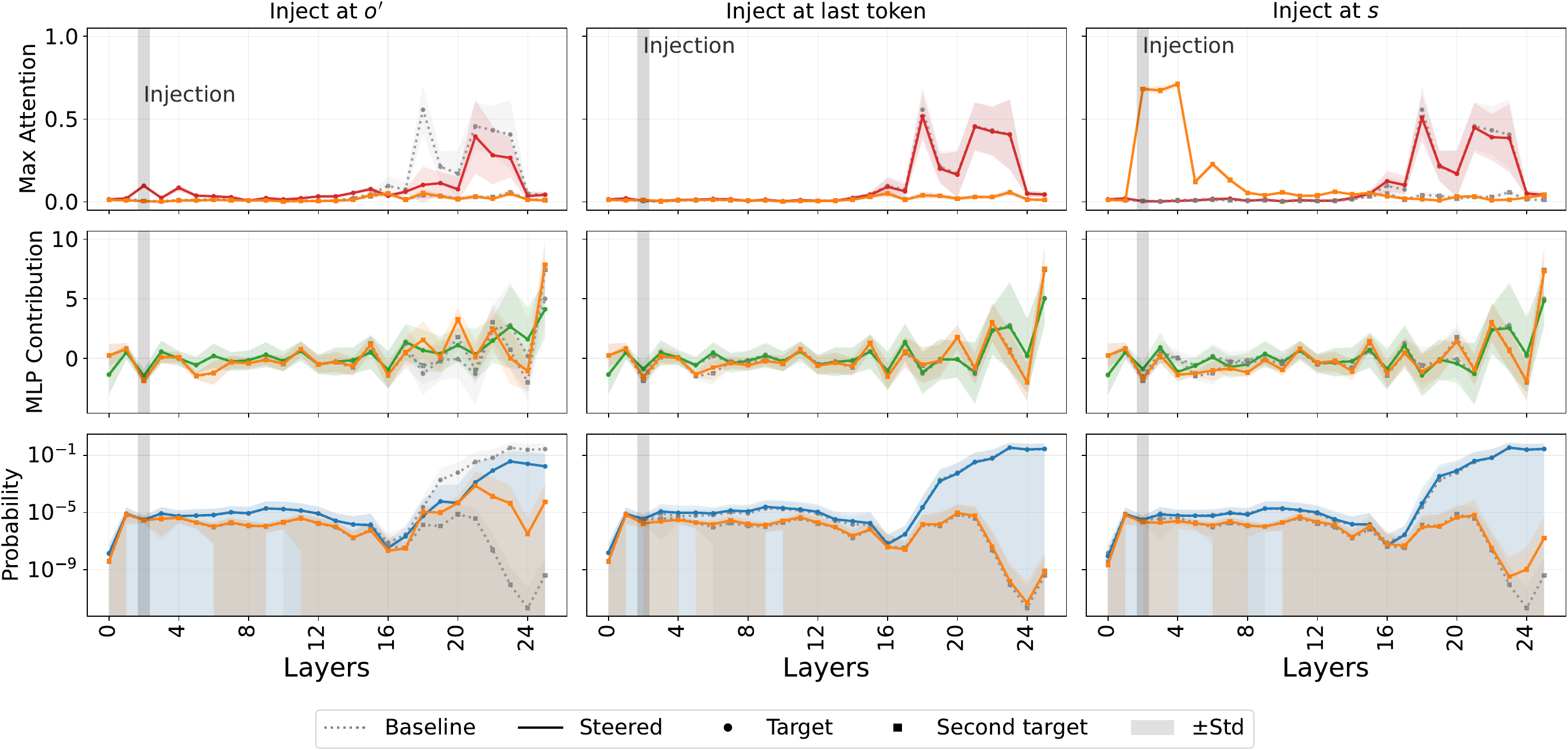}
\caption{
(C$\rightarrow$R, T5Gemma - Context-First: Object vs. last token vs. subject) Showing the mechanistic investigation of parametric steering over 1,221 PEQ samples. Top: Attention (Subject=Orange, Counterfactual=Red). Middle: MLP(Truth=Orange, Counterfactual=Green). Bottom: Probability(Truth=Orange, Counterfactual=Blue). The intervention acts as an amplifier and successfully shifts the behavior from all injection positions.
}
\label{fig:mi_peq_t5gemma-2b-2b-prefixlm-it_cr_L-2_context-first}
\end{figure}

\begin{figure}[!ht]
\centering
\includegraphics[width=0.98\textwidth]{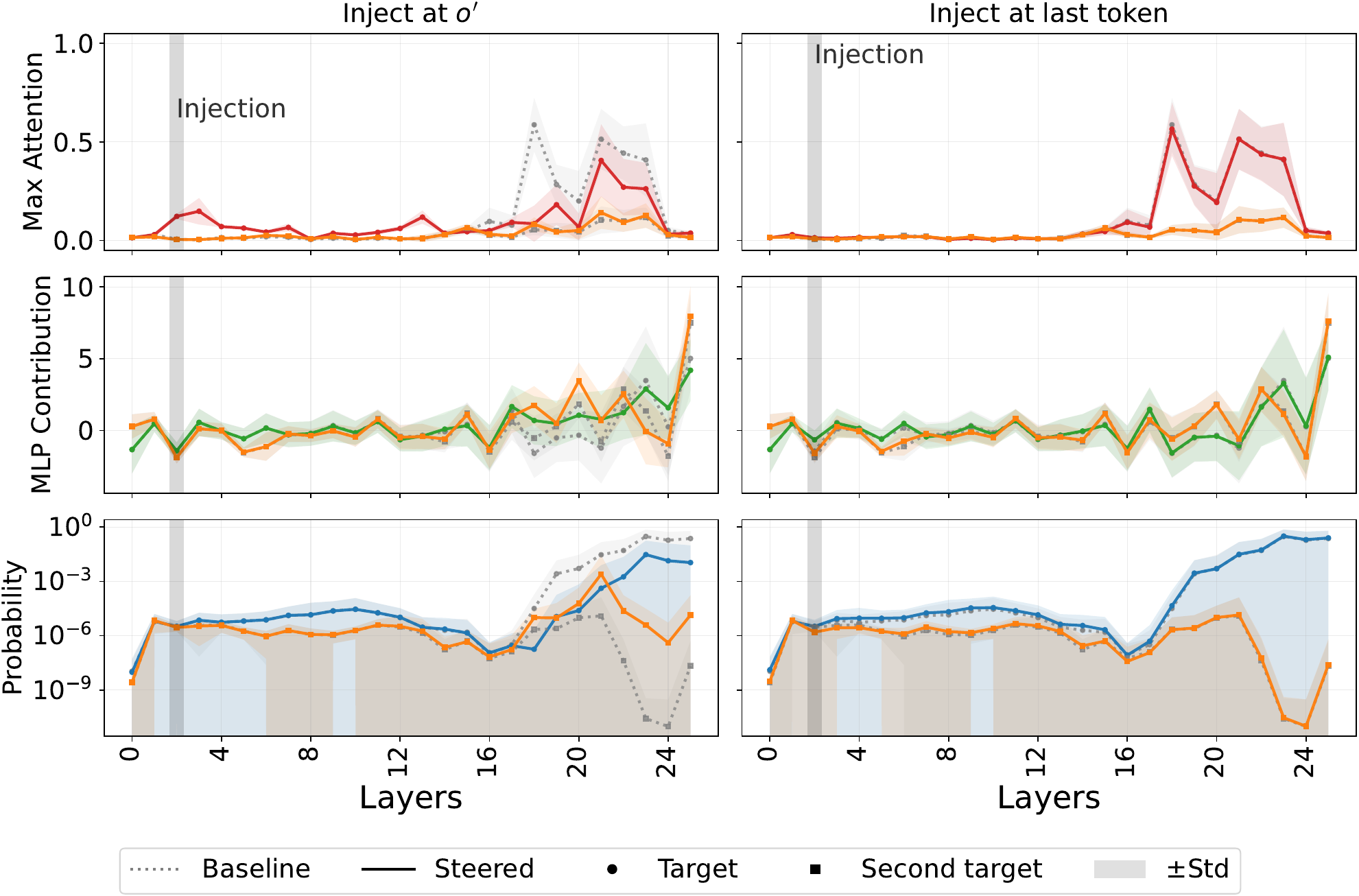}
\caption{
(C$\rightarrow$R, T5Gemma - Query-First: Object vs. last token only) Showing the mechanistic investigation of parametric steering over 1,203 PEQ samples. Top: Attention (Subject=Orange, Counterfactual=Red). Middle: MLP(Truth=Orange, Counterfactual=Green). Bottom: Probability(Truth=Orange, Counterfactual=Blue). The intervention acts as an amplifier and successfully shifts the behavior from all injection positions.
}
\label{fig:mi_peq_t5gemma-2b-2b-prefixlm-it_cr_L-2_query-first}
\end{figure}

\begin{figure}[!ht]
\centering
\includegraphics[width=0.98\textwidth]{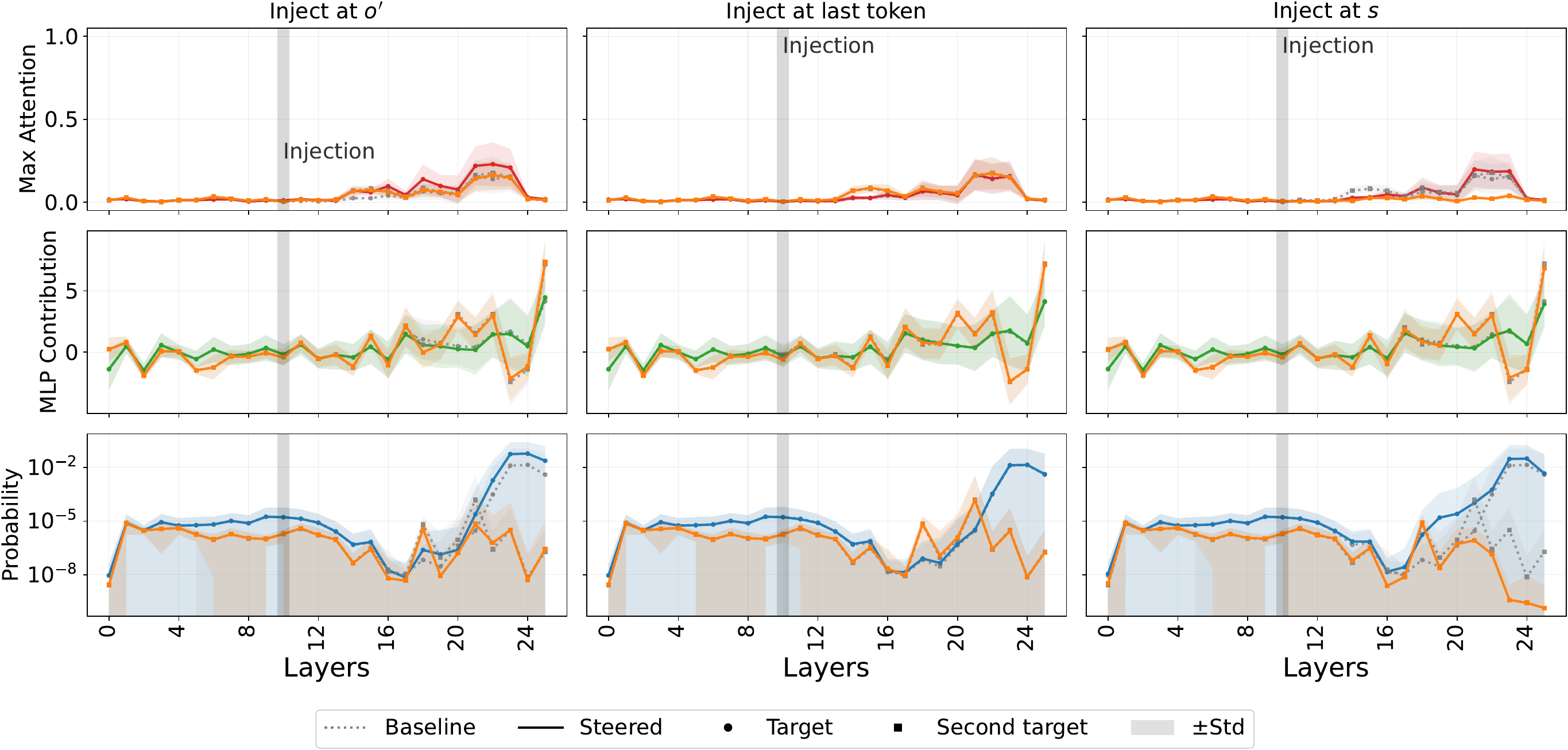}
\caption{
(R$\rightarrow$C, T5Gemma - Context-First: Object vs. last token vs. subject) Showing the mechanistic investigation of parametric steering over 1,221 PEQ samples. Top: Attention (Subject=Orange, Counterfactual=Red). Middle: MLP(Truth=Orange, Counterfactual=Green). Bottom: Probability(Truth=Orange, Counterfactual=Blue). The intervention acts as an amplifier and successfully shifts the behavior from all injection positions.
}
\label{fig:mi_peq_t5gemma-2b-2b-prefixlm-it_rc_L-10_context-first}
\end{figure}

\begin{figure}[!ht]
\centering
\includegraphics[width=0.98\textwidth]{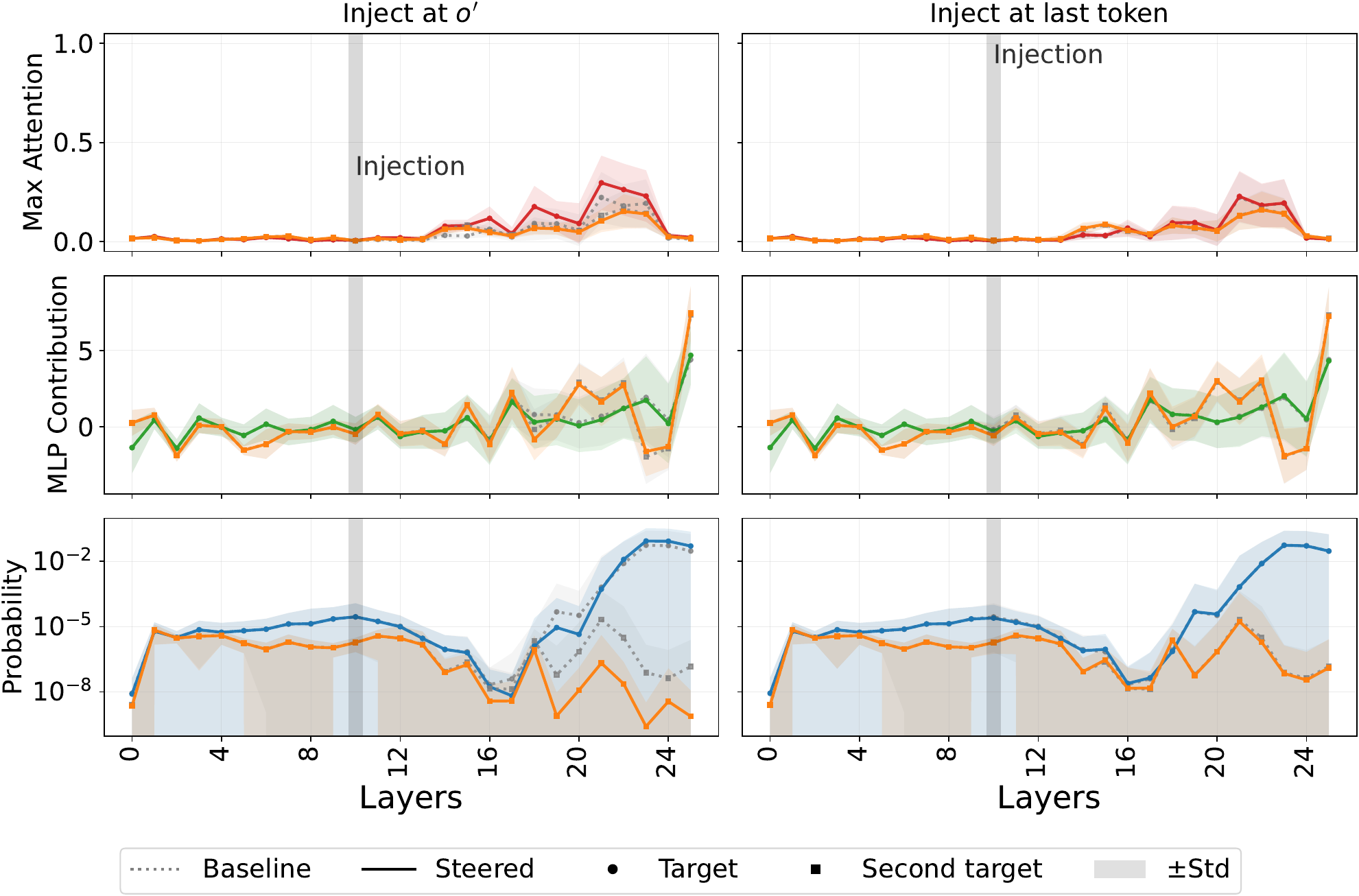}
\caption{
(R$\rightarrow$C, T5Gemma - Query-First: Object vs. last token only) Showing the mechanistic investigation of parametric steering over 1,203 PEQ samples. Top: Attention (Subject=Orange, Counterfactual=Red). Middle: MLP(Truth=Orange, Counterfactual=Green). Bottom: Probability(Truth=Orange, Counterfactual=Blue). The intervention acts as an amplifier and successfully shifts the behavior from all injection positions.
}
\label{fig:mi_peq_t5gemma-2b-2b-prefixlm-it_rc_L-10_query-first}
\end{figure}

\begin{figure}[!ht]
\centering
\includegraphics[width=0.98\textwidth]{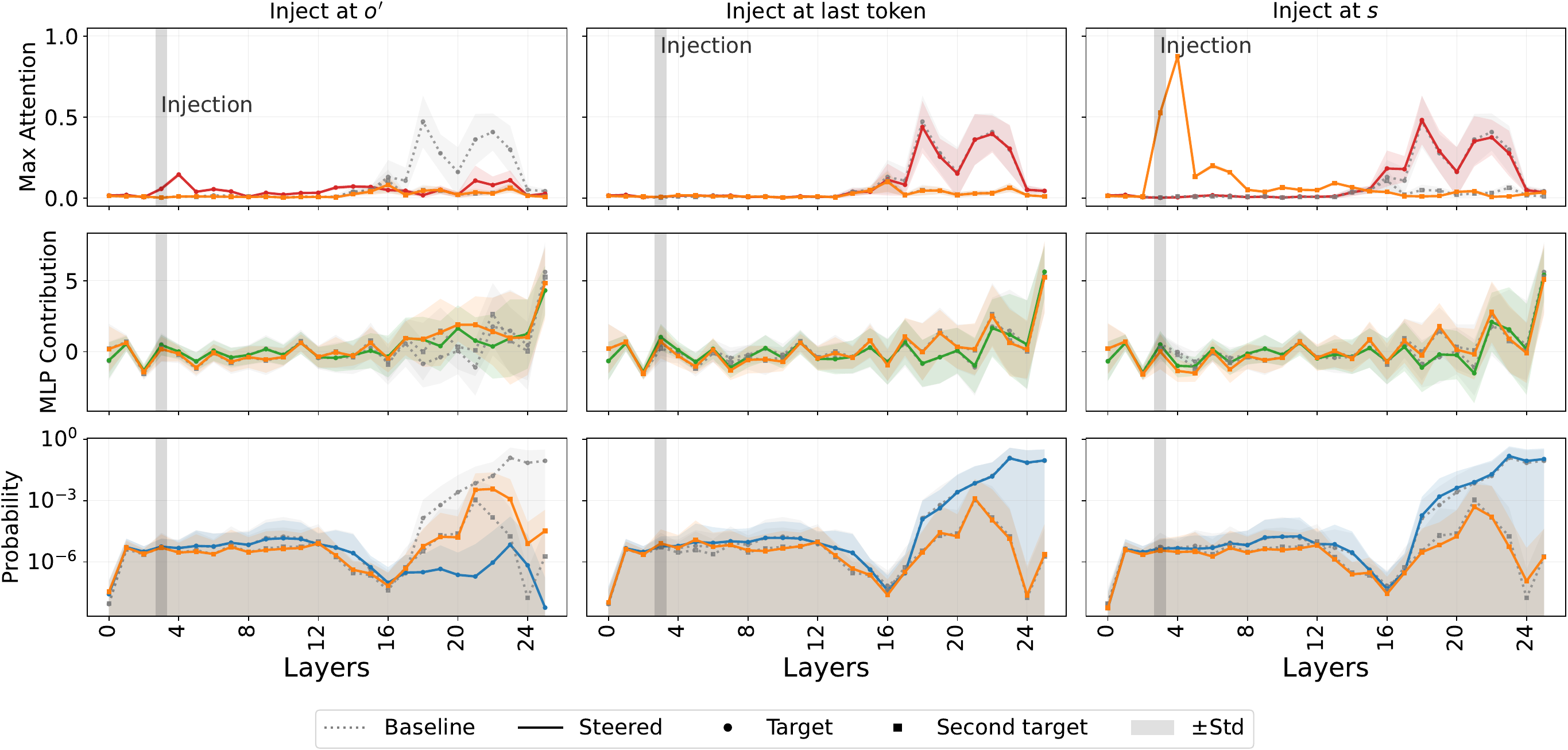}
\caption{
(C$\rightarrow$R, T5Gemma - Context-First: Object vs. last token vs. subject) Showing the mechanistic investigation of parametric steering over 1,776 PopQA samples. Top: Attention (Subject=Orange, Counterfactual=Red). Middle: MLP(Truth=Orange, Counterfactual=Green). Bottom: Probability(Truth=Orange, Counterfactual=Blue). The intervention acts as an amplifier and successfully shifts the behavior from all injection positions.
}
\label{fig:mi_popqa_t5gemma-2b-2b-prefixlm-it_cr_L-3_context-first}
\end{figure}

\begin{figure}[!ht]
\centering
\includegraphics[width=0.98\textwidth]{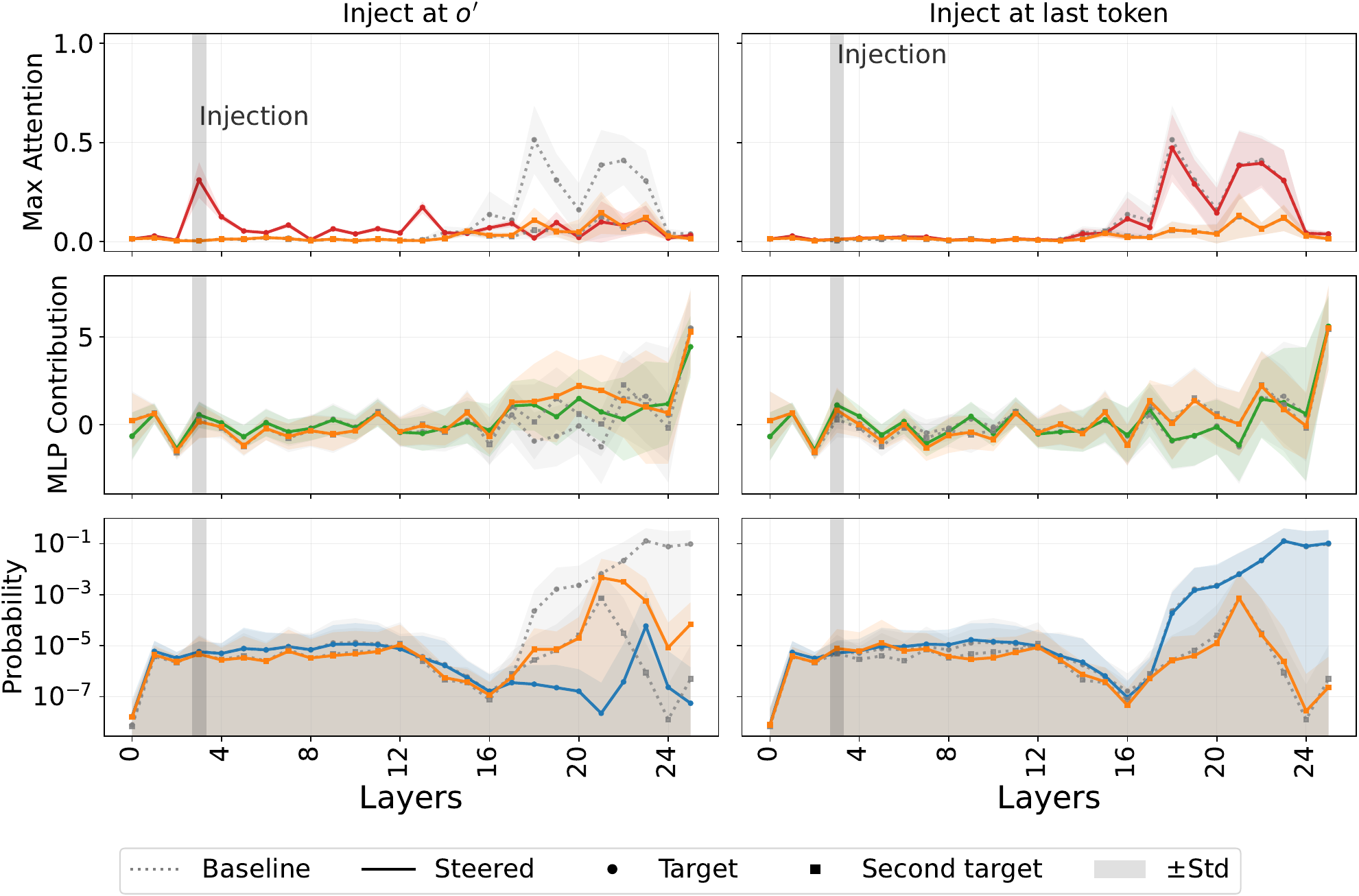}
\caption{
(C$\rightarrow$R, T5Gemma - Query-First: Object vs. last token only) Showing the mechanistic investigation of parametric steering over 1,125 PopQA samples. Top: Attention (Subject=Orange, Counterfactual=Red). Middle: MLP(Truth=Orange, Counterfactual=Green). Bottom: Probability(Truth=Orange, Counterfactual=Blue). The intervention acts as an amplifier and successfully shifts the behavior from all injection positions.
}
\label{fig:mi_popqa_t5gemma-2b-2b-prefixlm-it_cr_L-3_query-first}
\end{figure}

\begin{figure}[!ht]
\centering
\includegraphics[width=0.98\textwidth]{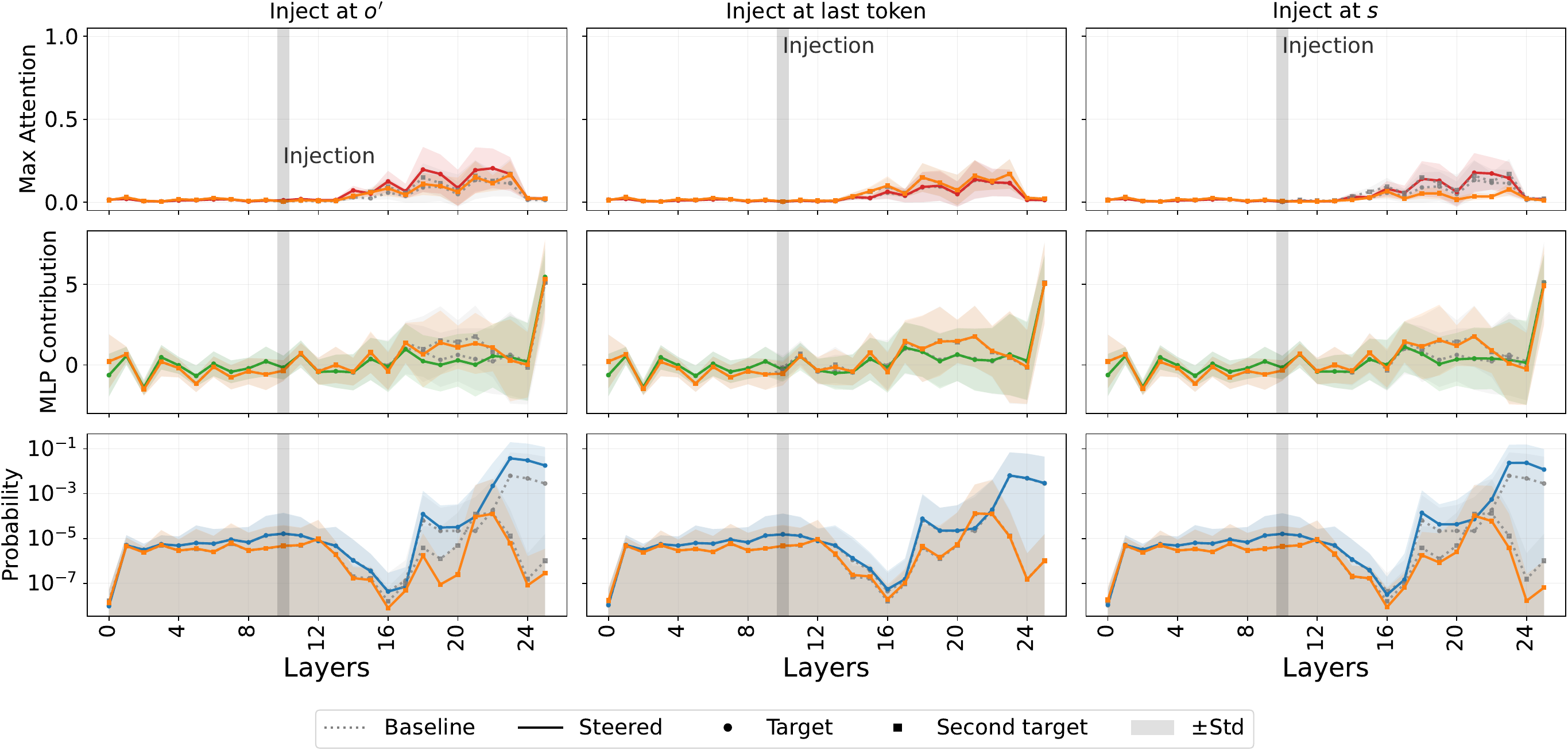}
\caption{
(R$\rightarrow$C, T5Gemma - Context-First: Object vs. last token vs. subject) Showing the mechanistic investigation of parametric steering over 1,776 PopQA samples. Top: Attention (Subject=Orange, Counterfactual=Red). Middle: MLP(Truth=Orange, Counterfactual=Green). Bottom: Probability(Truth=Orange, Counterfactual=Blue). The intervention acts as an amplifier and successfully shifts the behavior from all injection positions.
}
\label{fig:mi_popqa_t5gemma-2b-2b-prefixlm-it_rc_L-10_context-first}
\end{figure}

\begin{figure}[!ht]
\centering
\includegraphics[width=0.98\textwidth]{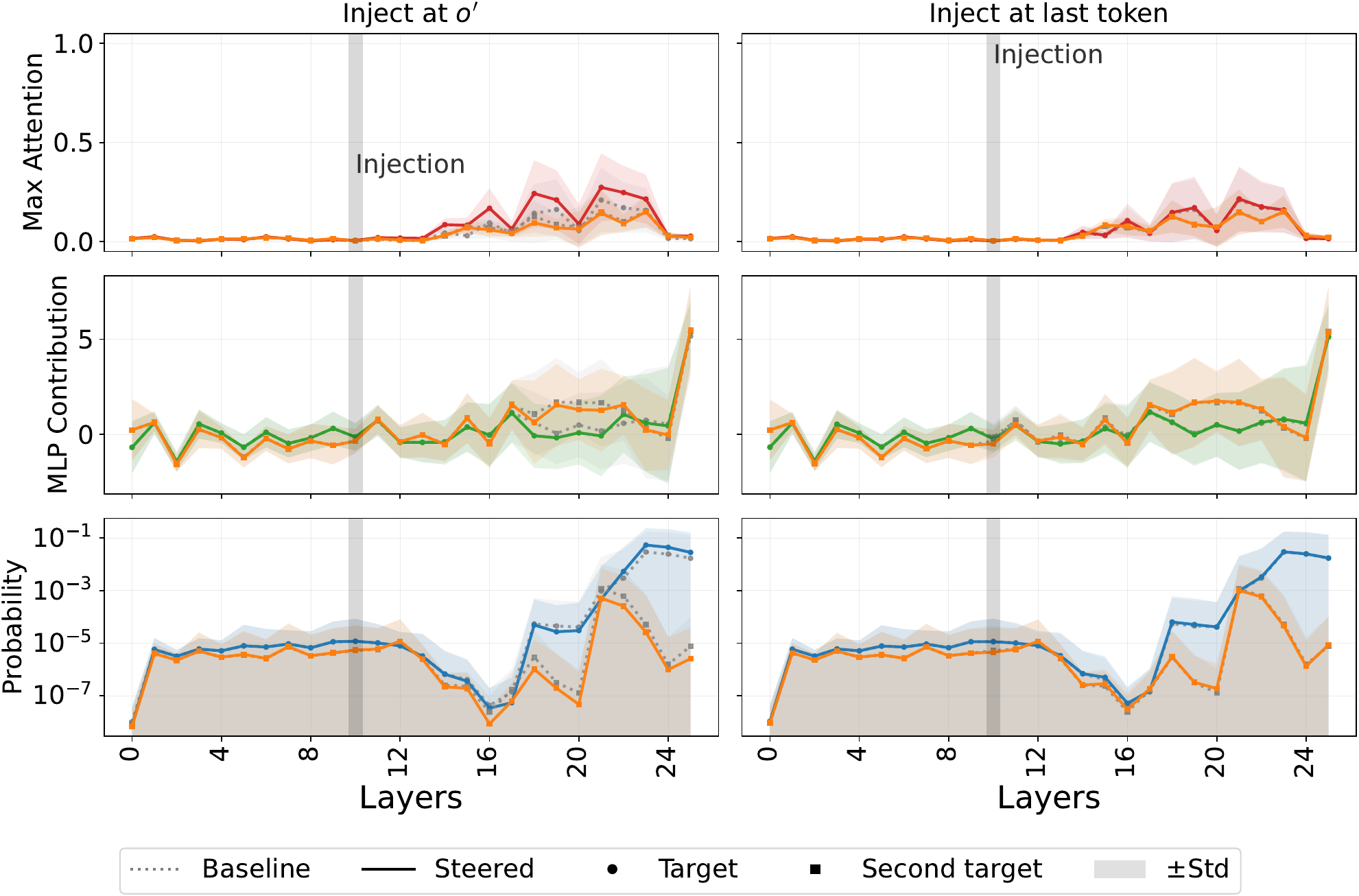}
\caption{
(R$\rightarrow$C, T5Gemma - Query-First: Object vs. last token only) Showing the mechanistic investigation of parametric steering over 1,125 PopQA samples. Top: Attention (Subject=Orange, Counterfactual=Red). Middle: MLP(Truth=Orange, Counterfactual=Green). Bottom: Probability(Truth=Orange, Counterfactual=Blue). The intervention acts as an amplifier and successfully shifts the behavior from all injection positions.
}
\label{fig:mi_popqa_t5gemma-2b-2b-prefixlm-it_rc_L-10_query-first}
\end{figure}

\begin{figure}[!ht]
\centering
\includegraphics[width=0.98\textwidth]{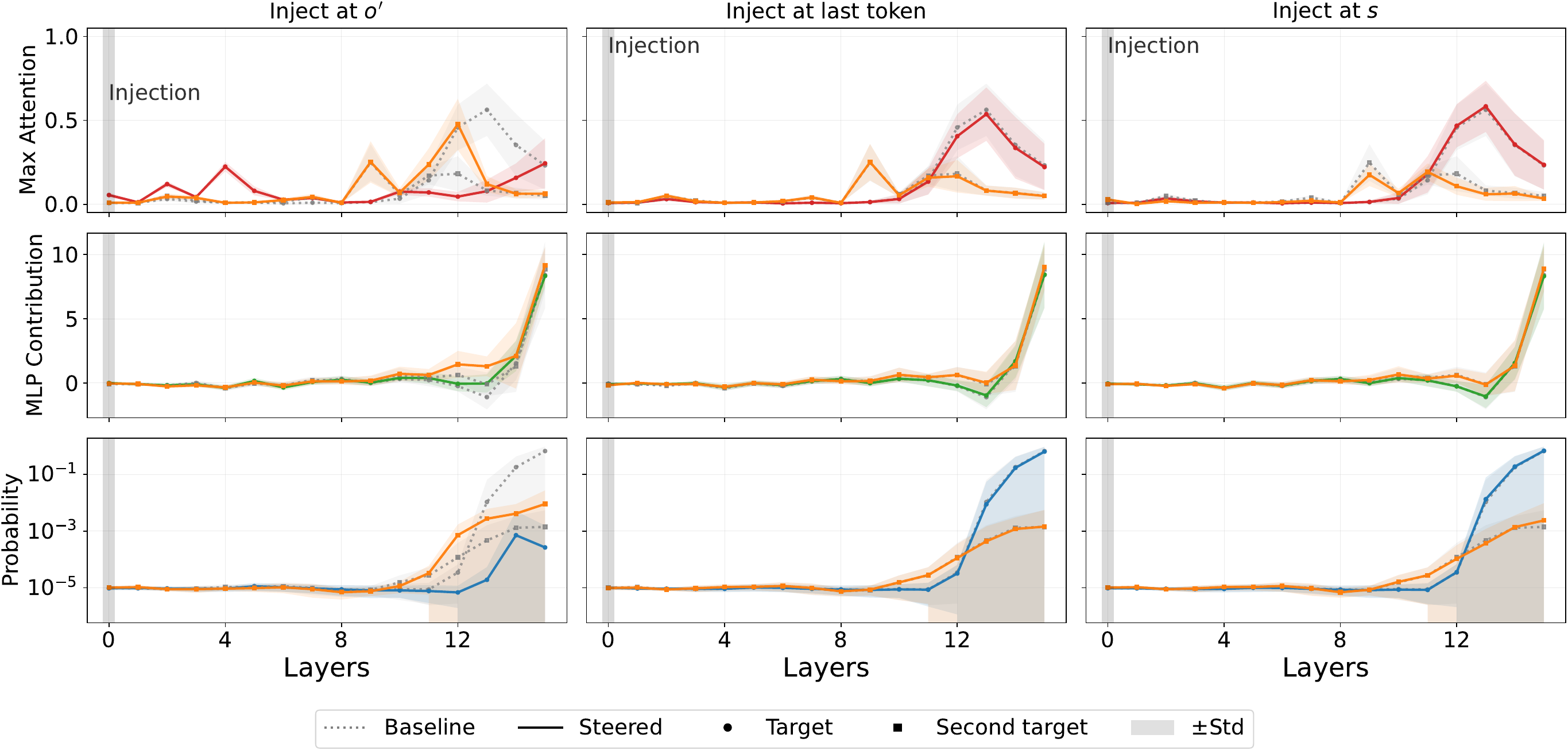}
\caption{
(C$\rightarrow$R, OLMo-2 - Context-First: Object vs. last token vs. subject) Showing the mechanistic investigation of parametric steering over 112 PEQ samples. Top: Attention (Subject=Orange, Counterfactual=Red). Middle: MLP(Truth=Orange, Counterfactual=Green). Bottom: Probability(Truth=Orange, Counterfactual=Blue). The intervention acts as an amplifier and successfully shifts the behavior from all injection positions.
}
\label{fig:mi_peq_OLMo-2-0425-1B-Instruct_cr_L-0_context-first}
\end{figure}

\begin{figure}[!ht]
\centering
\includegraphics[width=0.98\textwidth]{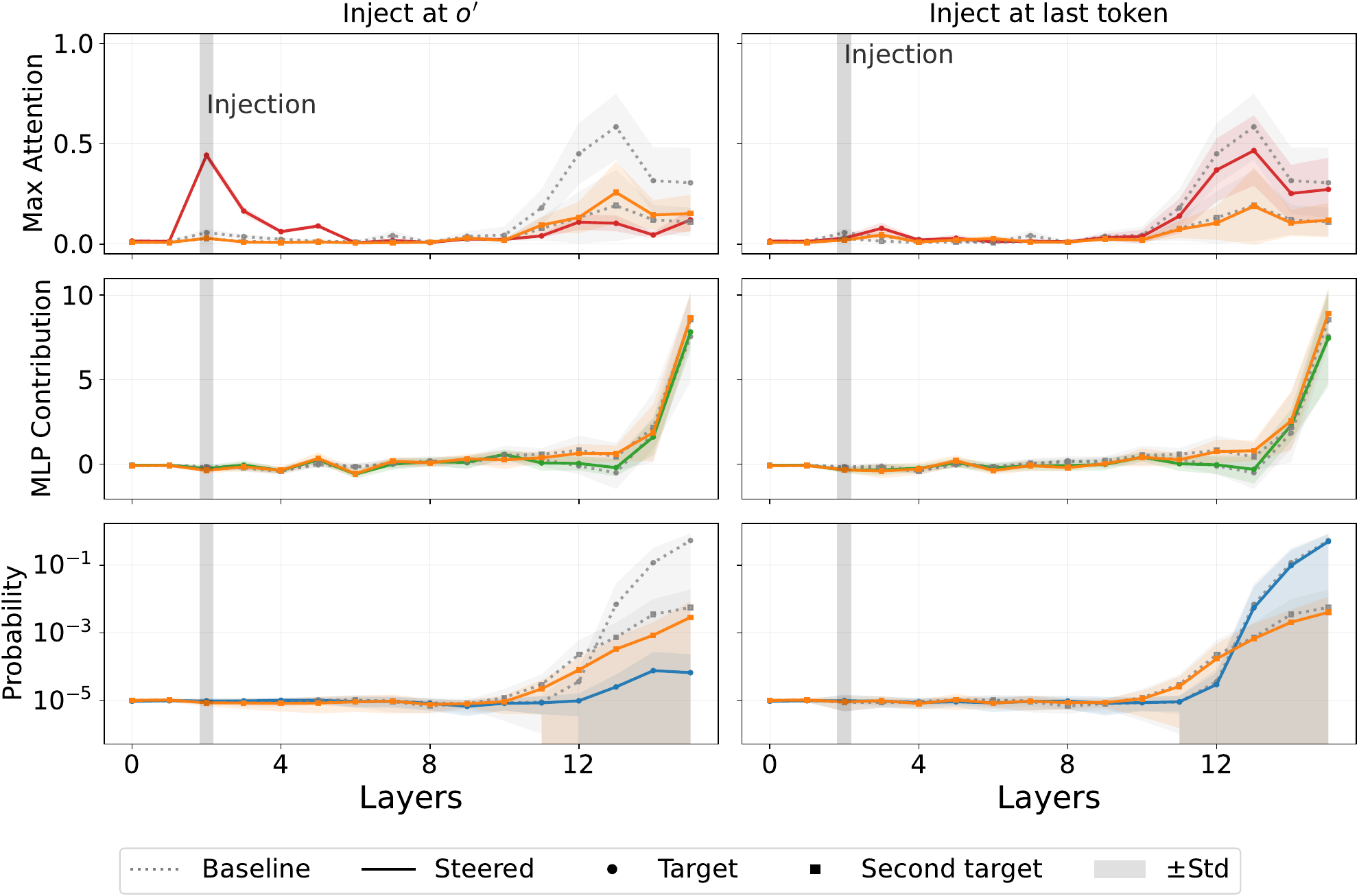}
\caption{
(C$\rightarrow$R, OLMo-2 - Query-First: Object vs. last token only) Showing the mechanistic investigation of parametric steering over 103 PEQ samples. Top: Attention (Subject=Orange, Counterfactual=Red). Middle: MLP(Truth=Orange, Counterfactual=Green). Bottom: Probability(Truth=Orange, Counterfactual=Blue). The intervention acts as an amplifier and successfully shifts the behavior from all injection positions.
}
\label{fig:mi_peq_OLMo-2-0425-1B-Instruct_cr_L-2_query-first}
\end{figure}

\begin{figure}[!ht]
\centering
\includegraphics[width=0.98\textwidth]{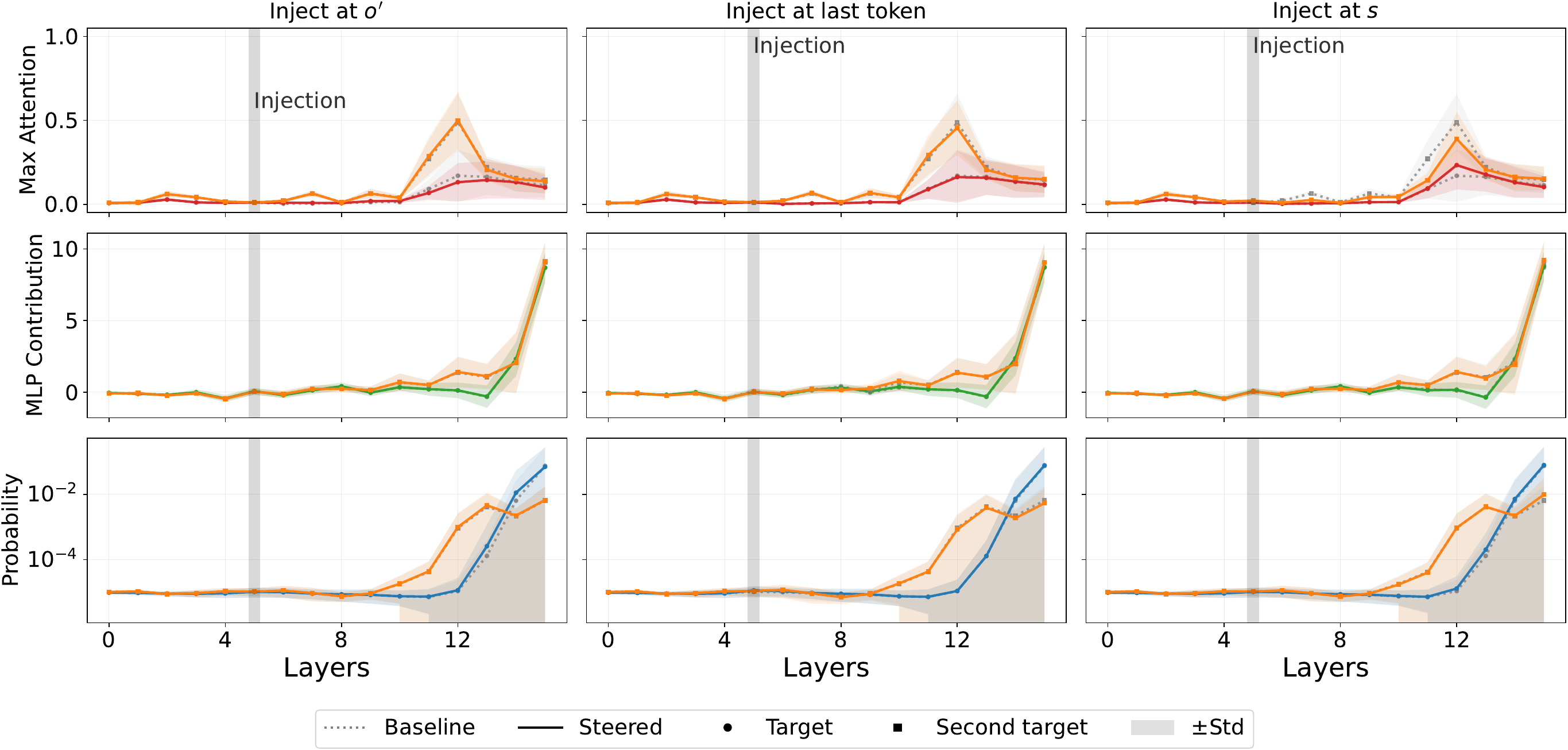}
\caption{
(R$\rightarrow$C, OLMo-2 - Context-First: Object vs. last token vs. subject) Showing the mechanistic investigation of parametric steering over 112 PEQ samples. Top: Attention (Subject=Orange, Counterfactual=Red). Middle: MLP(Truth=Orange, Counterfactual=Green). Bottom: Probability(Truth=Orange, Counterfactual=Blue). The intervention acts as an amplifier and successfully shifts the behavior from all injection positions.
}
\label{fig:mi_peq_OLMo-2-0425-1B-Instruct_rc_L-5_context-first}
\end{figure}

\begin{figure}[!ht]
\centering
\includegraphics[width=0.98\textwidth]{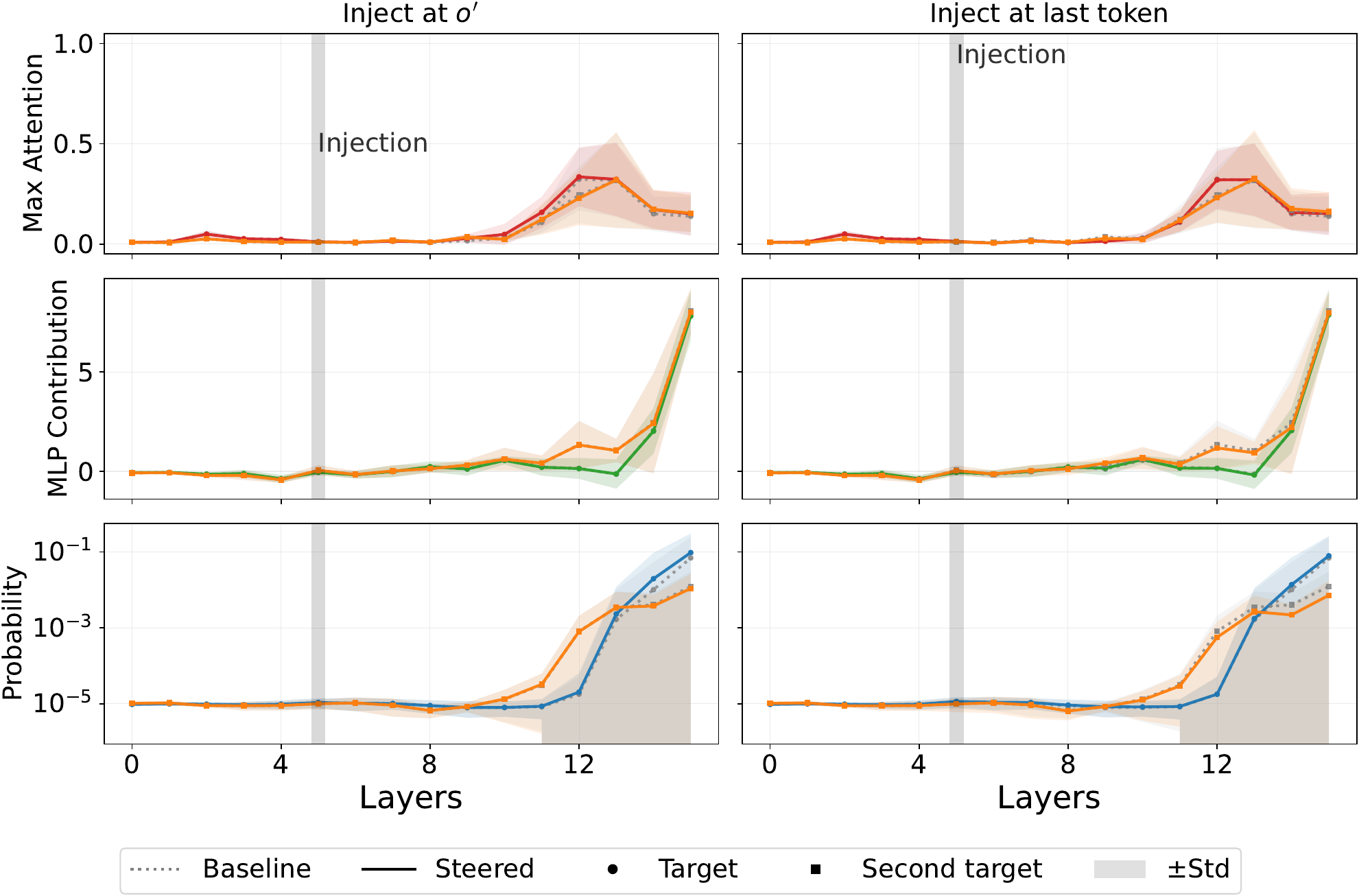}
\caption{
(R$\rightarrow$C, OLMo-2 - Query-First: Object vs. last token only) Showing the mechanistic investigation of parametric steering over 103 PEQ samples. Top: Attention (Subject=Orange, Counterfactual=Red). Middle: MLP(Truth=Orange, Counterfactual=Green). Bottom: Probability(Truth=Orange, Counterfactual=Blue). The intervention acts as an amplifier and successfully shifts the behavior from all injection positions.
}
\label{fig:mi_peq_OLMo-2-0425-1B-Instruct_rc_L-5_query-first}
\end{figure}

\begin{figure}[!ht]
\centering
\includegraphics[width=0.98\textwidth]{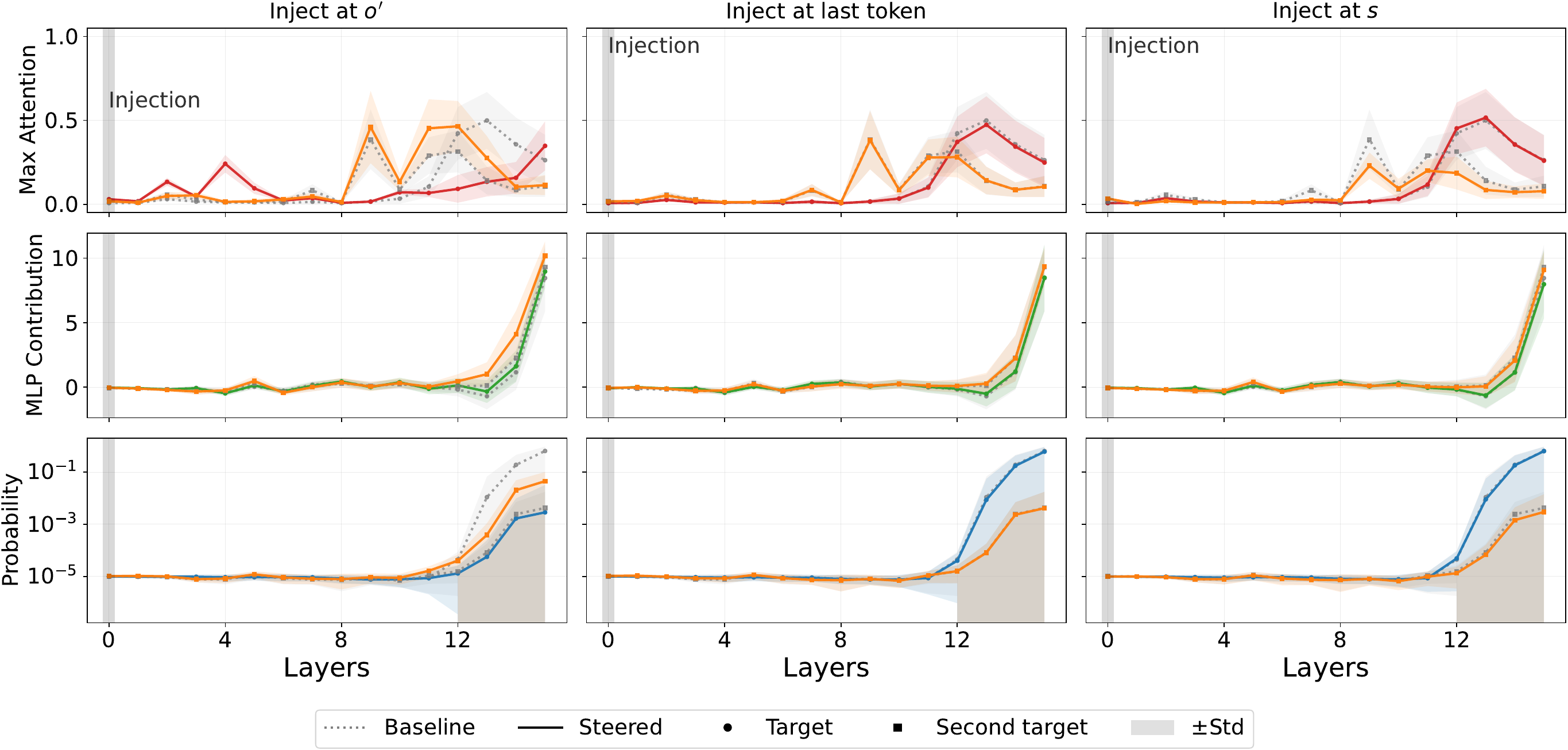}
\caption{
(C$\rightarrow$R, OLMo-2 - Context-First: Object vs. last token vs. subject) Showing the mechanistic investigation of parametric steering over 655 PopQA samples. Top: Attention (Subject=Orange, Counterfactual=Red). Middle: MLP(Truth=Orange, Counterfactual=Green). Bottom: Probability(Truth=Orange, Counterfactual=Blue). The intervention acts as an amplifier and successfully shifts the behavior from all injection positions.
}
\label{fig:mi_popqa_OLMo-2-0425-1B-Instruct_cr_L-0_context-first}
\end{figure}

\begin{figure}[!ht]
\centering
\includegraphics[width=0.98\textwidth]{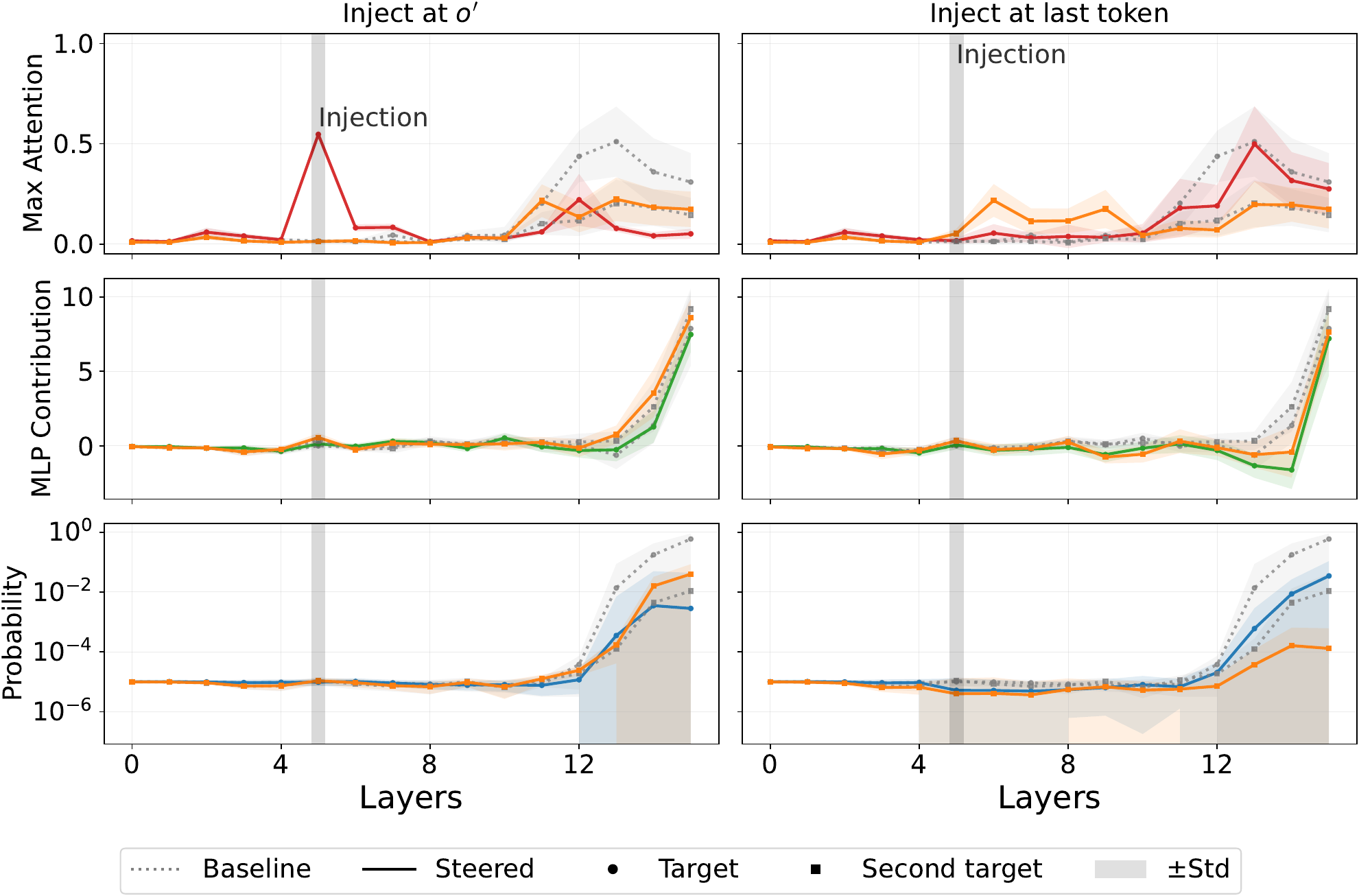}
\caption{
(C$\rightarrow$R, OLMo-2 - Query-First: Object vs. last token only) Showing the mechanistic investigation of parametric steering over 510 PopQA samples. Top: Attention (Subject=Orange, Counterfactual=Red). Middle: MLP(Truth=Orange, Counterfactual=Green). Bottom: Probability(Truth=Orange, Counterfactual=Blue). The intervention acts as an amplifier and successfully shifts the behavior from all injection positions.
}
\label{fig:mi_popqa_OLMo-2-0425-1B-Instruct_cr_L-5_query-first}
\end{figure}

\begin{figure}[!ht]
\centering
\includegraphics[width=0.98\textwidth]{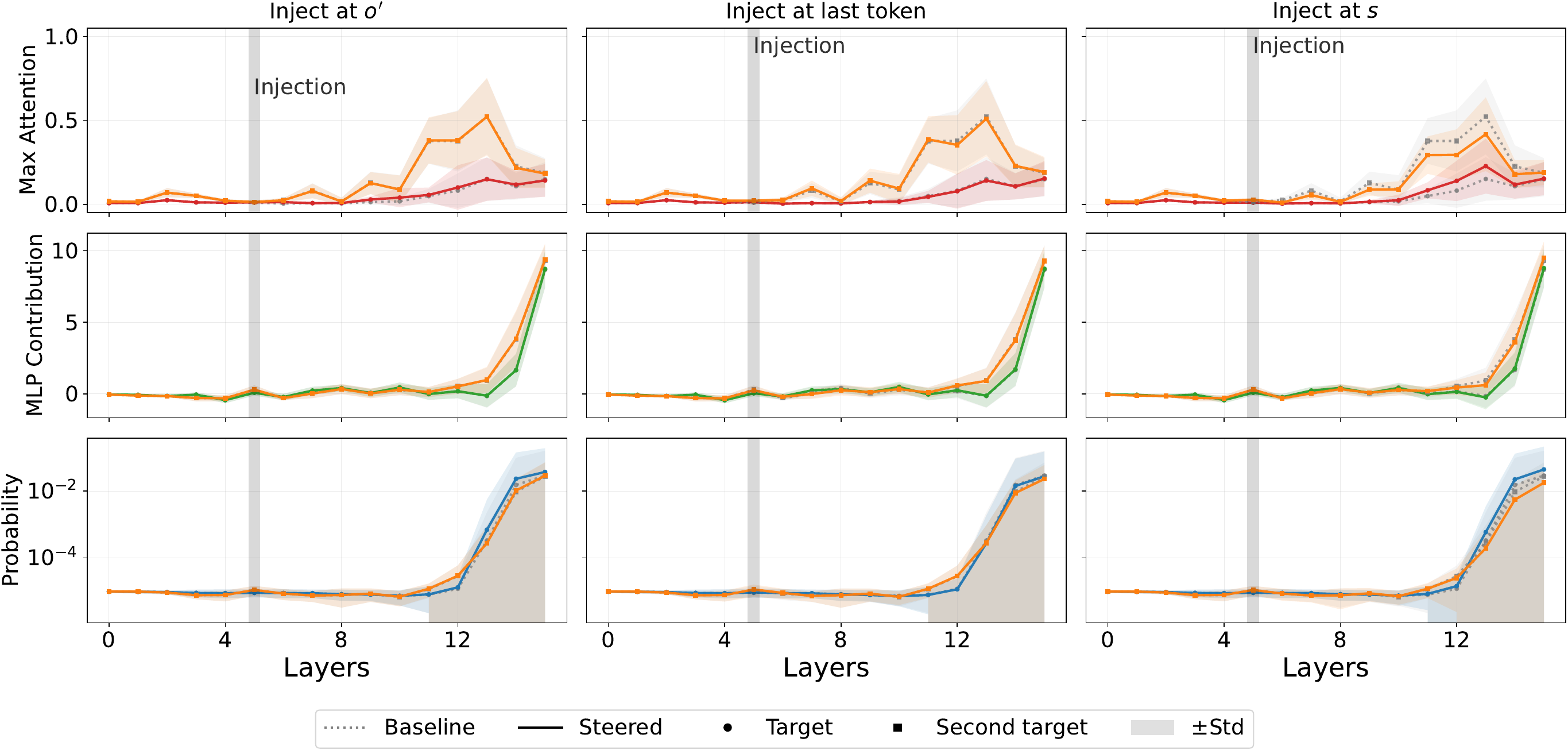}
\caption{
(R$\rightarrow$C, OLMo-2 - Context-First: Object vs. last token vs. subject) Showing the mechanistic investigation of parametric steering over 655 PopQA samples. Top: Attention (Subject=Orange, Counterfactual=Red). Middle: MLP(Truth=Orange, Counterfactual=Green). Bottom: Probability(Truth=Orange, Counterfactual=Blue). The intervention acts as an amplifier and successfully shifts the behavior from all injection positions.
}
\label{fig:mi_popqa_OLMo-2-0425-1B-Instruct_rc_L-5_context-first}
\end{figure}

\begin{figure}[!ht]
\centering
\includegraphics[width=0.98\textwidth]{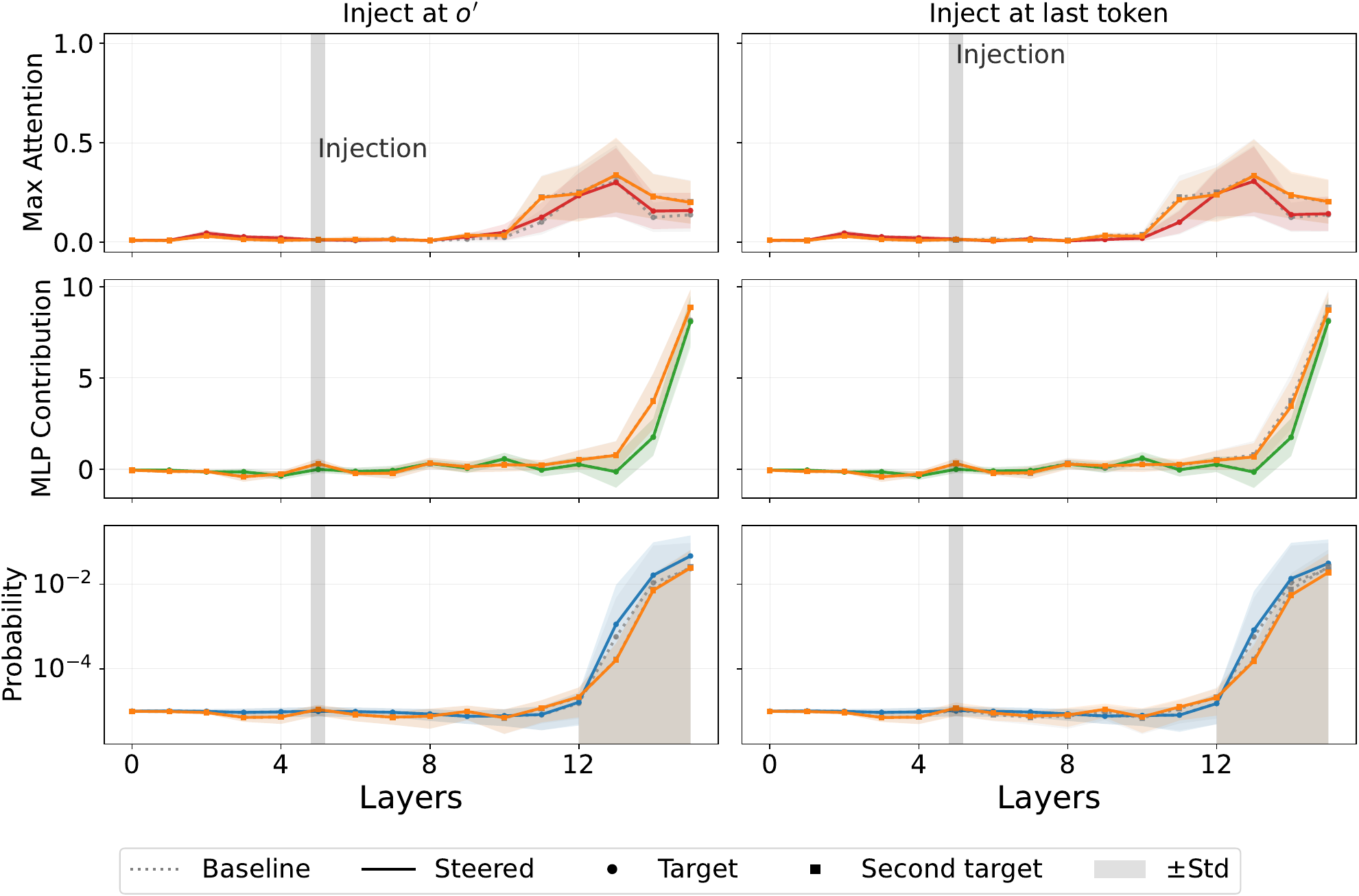}
\caption{
(R$\rightarrow$C, OLMo-2 - Query-First: Object vs. last token only) Showing the mechanistic investigation of parametric steering over 510 PopQA samples. Top: Attention (Subject=Orange, Counterfactual=Red). Middle: MLP(Truth=Orange, Counterfactual=Green). Bottom: Probability(Truth=Orange, Counterfactual=Blue). The intervention acts as an amplifier and successfully shifts the behavior from all injection positions.
}
\label{fig:mi_popqa_OLMo-2-0425-1B-Instruct_rc_L-5_query-first}
\end{figure}

Similarly, for the Gemma-2, T5Gemma, and OLMo-2 models in both directions, see Figures~\ref{fig:mi_popqa_gemma-2-2b-it_cr_L-6_context-first}-\ref{fig:mi_popqa_OLMo-2-0425-1B-Instruct_rc_L-5_query-first} for PopQA. The results also show similar behavior for PEQ dataset as shown in the Figures~\ref{fig:mi_peq_gemma-2-2b-it_cr_L-6_context-first}-\ref{fig:mi_peq_OLMo-2-0425-1B-Instruct_rc_L-5_query-first}.

\begin{figure}[!ht]
\centering
\includegraphics[width=0.98\textwidth]{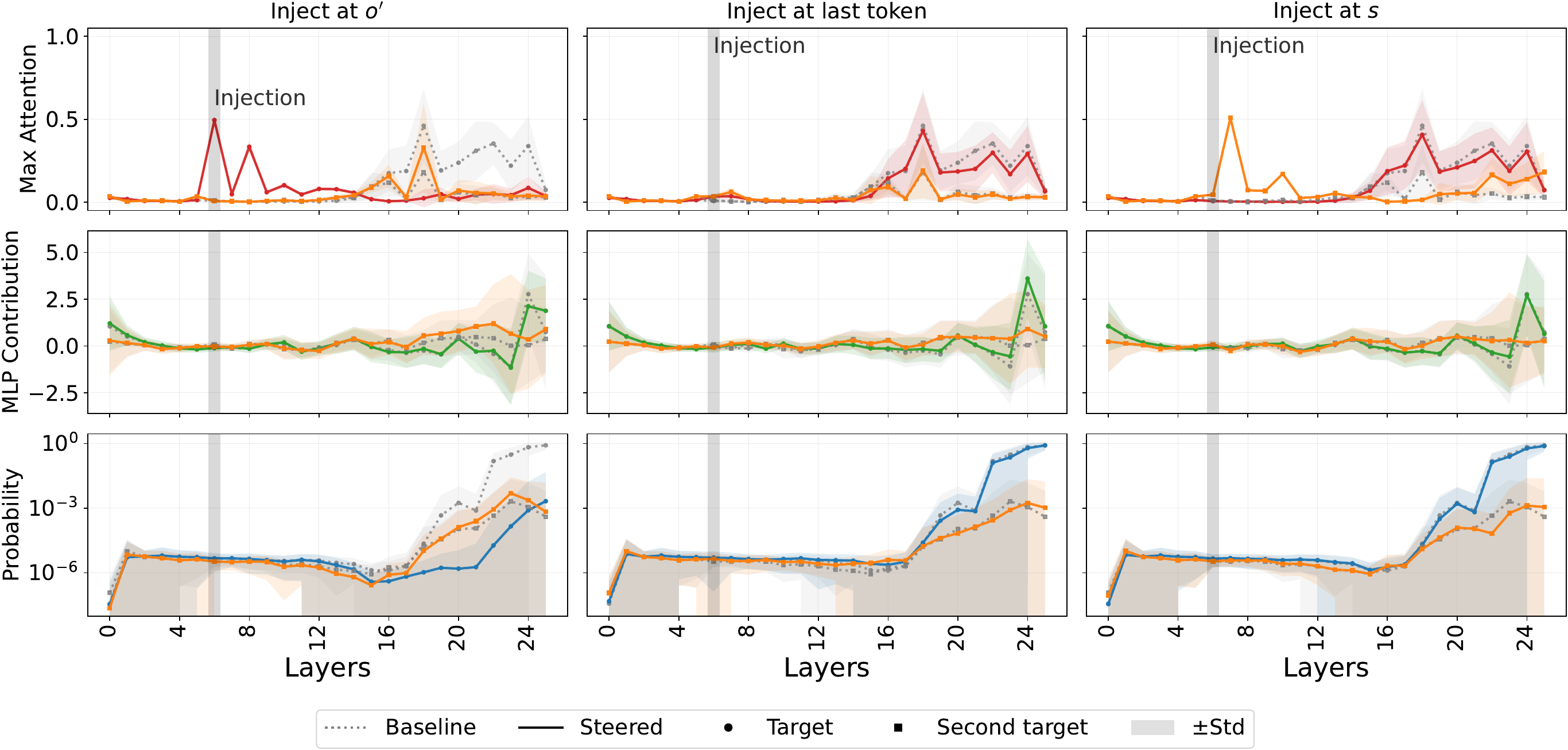}
\caption{
(C$\rightarrow$R, Gemma2 - Context-First: Object vs. last token vs. subject) Showing the mechanistic investigation of parametric steering over 1,434 PopQA samples. Top: Attention (Subject=Orange, Counterfactual=Red). Middle: MLP(Truth=Orange, Counterfactual=Green). Bottom: Probability(Truth=Orange, Counterfactual=Blue). The intervention acts as an amplifier and successfully shifts the behavior from all injection positions.
}
\end{figure}

\newpage

\newpage

\clearpage
\section{Behavioral Metrics (Performance vs. Fluency)}
\label{app:perf-prob-flu}

\subsection{PopQA: Object Token (Query-First)}

\subsubsection{Parametric Steering (Copy$\rightarrow$Recall) - Gemma2-2B}
\begin{figure}[H]
\centering
\includegraphics[width=\textwidth]{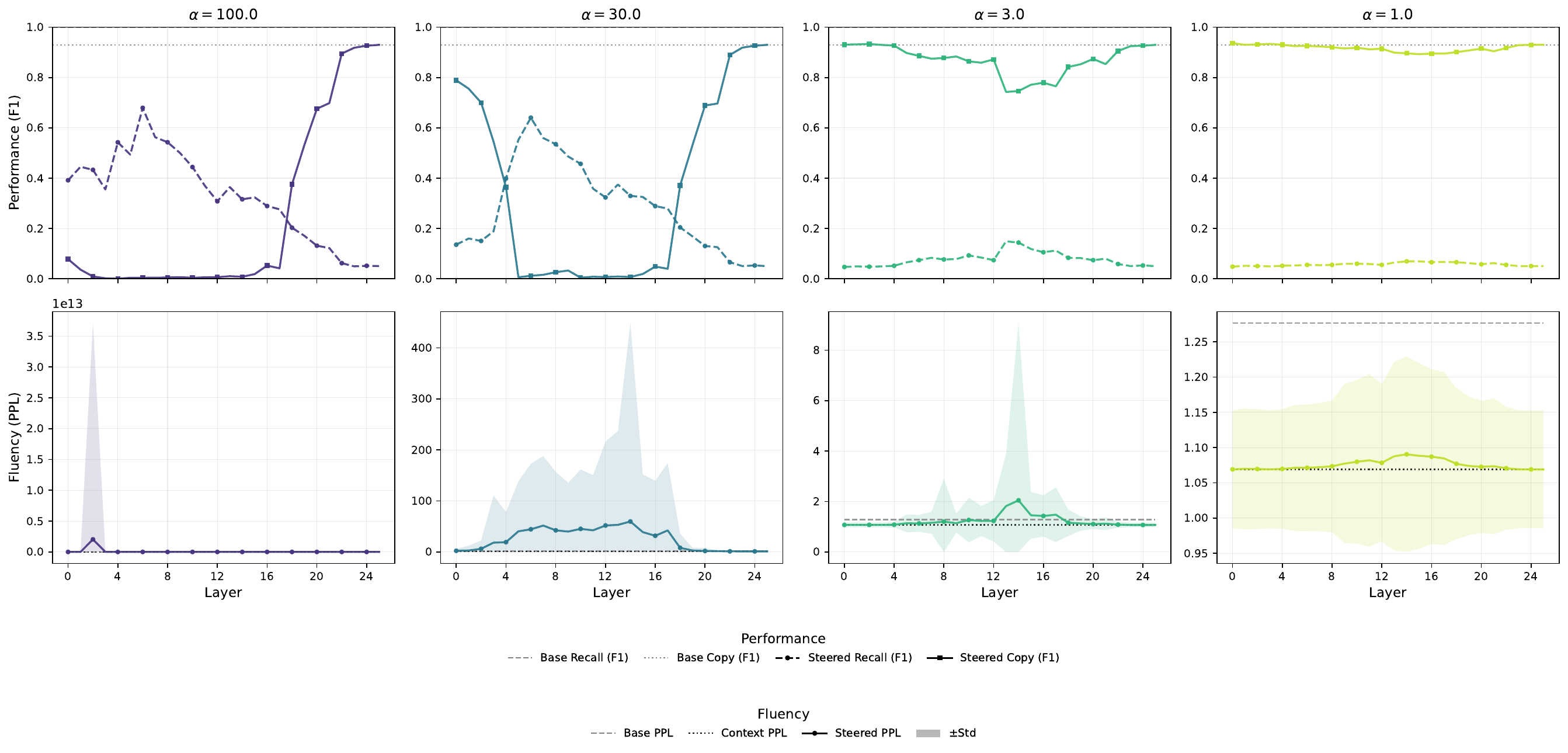}
\caption{Layer-wise effects of steering across four $\alpha$. Rows show performance (F1) and fluency (PPL). Columns correspond to $\alpha$.}
\label{fig:ov-perf-prob-flu-popqa-gemma2-cr-qf-obj}
\end{figure}

\subsubsection{Contextual Steering (Recall$\rightarrow$Copy) - Gemma2-2B}
\begin{figure}[H]
\centering
\includegraphics[width=\textwidth]{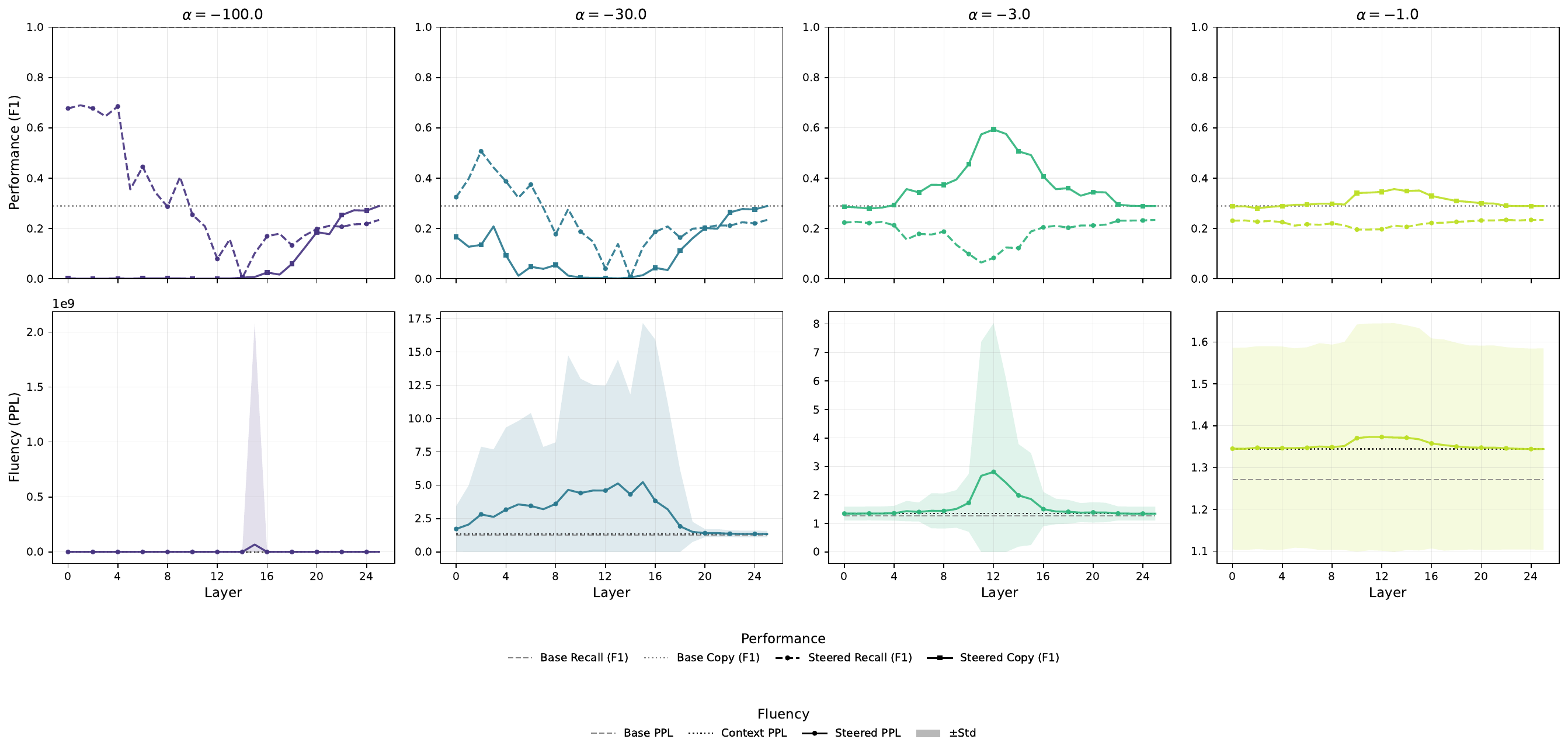}
\caption{Layer-wise effects of steering across four $\alpha$. Rows show performance (F1) and fluency (PPL). Columns correspond to $\alpha$.}
\label{fig:ov-perf-prob-flu-popqa-gemma2-rc-qf-obj}
\end{figure}

\subsubsection{Parametric Steering (Copy$\rightarrow$Recall) - T5Gemma-2B}
\begin{figure}[H]
\centering
\includegraphics[width=\textwidth]{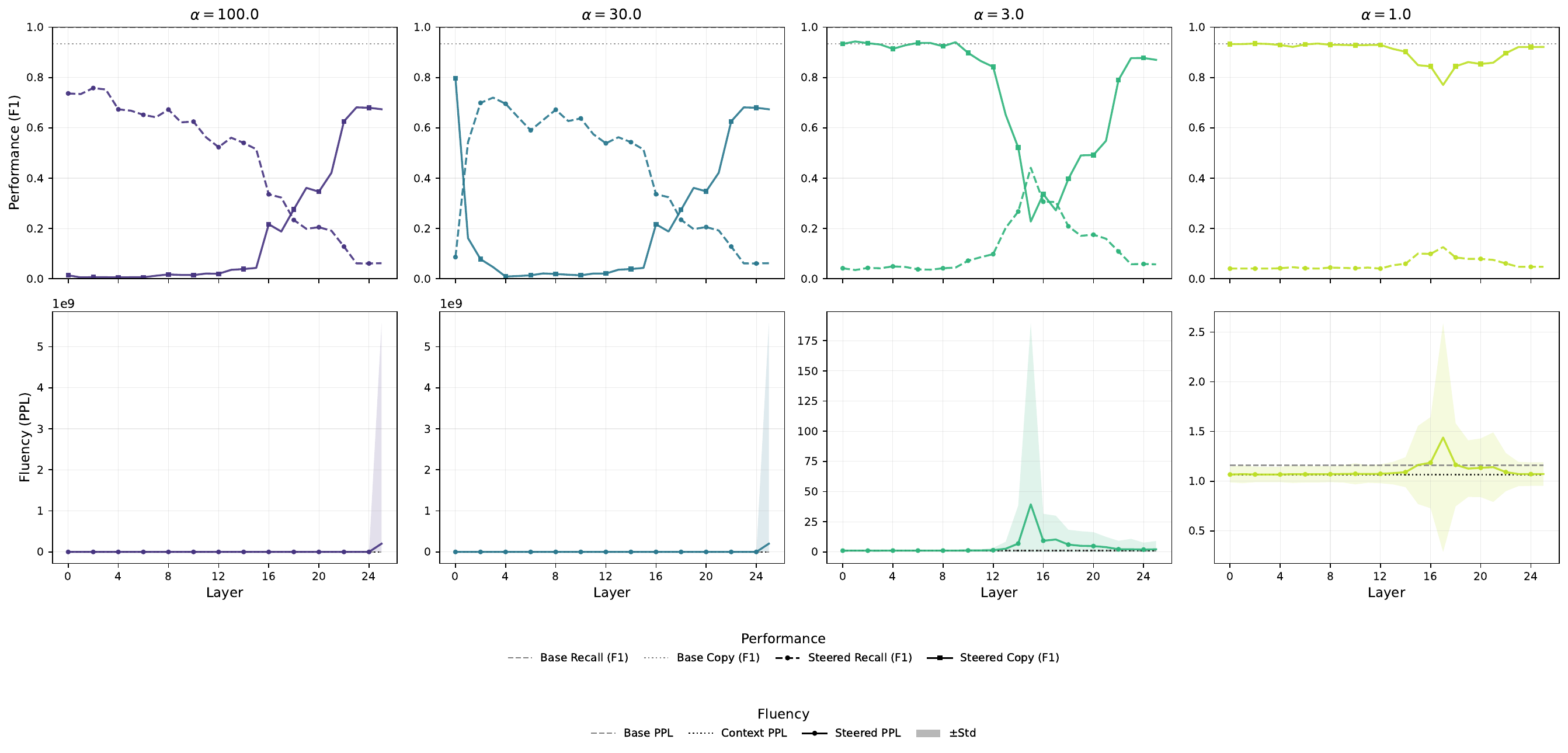}
\caption{Layer-wise effects of steering across four $\alpha$. Rows show performance (F1) and fluency (PPL). Columns correspond to $\alpha$.}
\label{fig:ov-perf-prob-flu-popqa-t5gemma-cr-qf-obj}
\end{figure}

\subsubsection{Contextual Steering (Recall$\rightarrow$Copy) - T5Gemma-2B}
\begin{figure}[H]
\centering
\includegraphics[width=\textwidth]{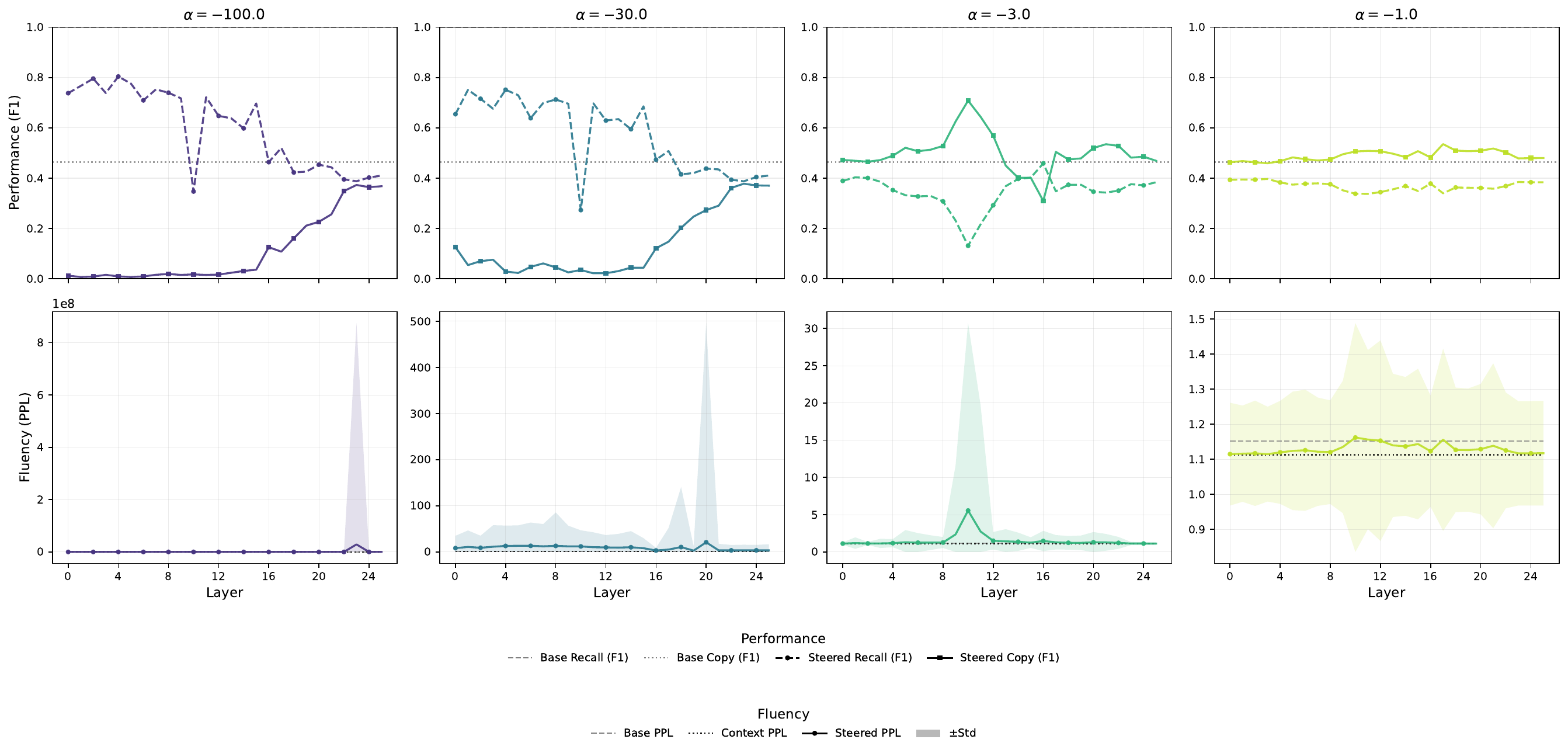}
\caption{Layer-wise effects of steering across four $\alpha$. Rows show performance (F1) and fluency (PPL). Columns correspond to $\alpha$.}
\label{fig:ov-perf-prob-flu-popqa-t5gemma-rc-qf-obj}
\end{figure}

\subsubsection{Parametric Steering (Copy$\rightarrow$Recall) - OLMo-2-0425-1B}
\begin{figure}[H]
\centering
\includegraphics[width=\textwidth]{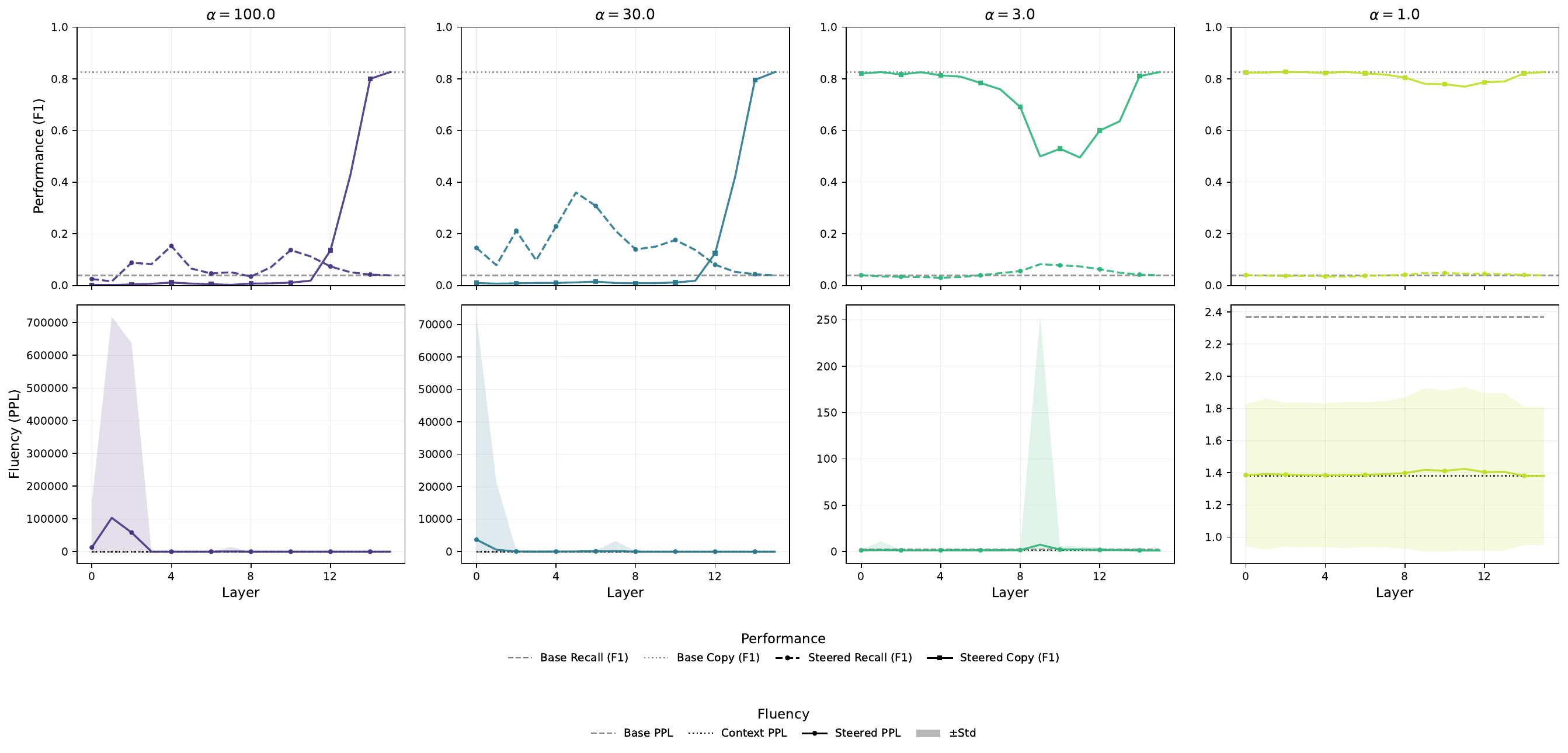}
\caption{Layer-wise effects of steering across four $\alpha$. Rows show performance (F1) and fluency (PPL). Columns correspond to $\alpha$.}
\label{fig:ov-perf-prob-flu-popqa-olmo2-cr-qf-obj}
\end{figure}

\subsubsection{Contextual Steering (Recall$\rightarrow$Copy) - OLMo-2-0425-1B}
\begin{figure}[H]
\centering
\includegraphics[width=\textwidth]{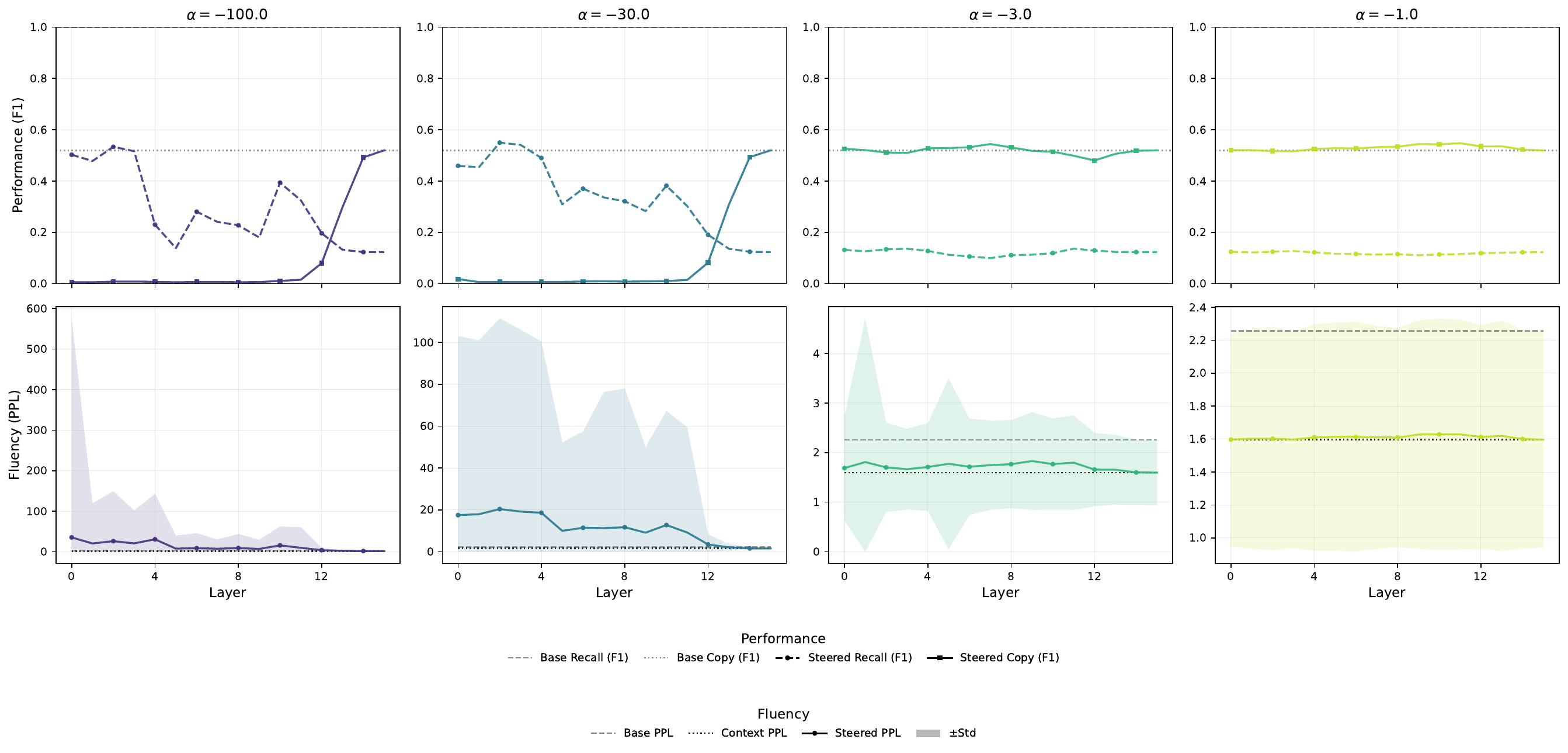}
\caption{Layer-wise effects of steering across four $\alpha$. Rows show performance (F1) and fluency (PPL). Columns correspond to $\alpha$.}
\label{fig:ov-perf-prob-flu-popqa-olmo2-rc-qf-obj}
\end{figure}

\subsection{PopQA: Object Token (Context-First)}

\subsubsection{Parametric Steering (Copy$\rightarrow$Recall) - Gemma2-2B}
\begin{figure}[H]
\centering
\includegraphics[width=\textwidth]{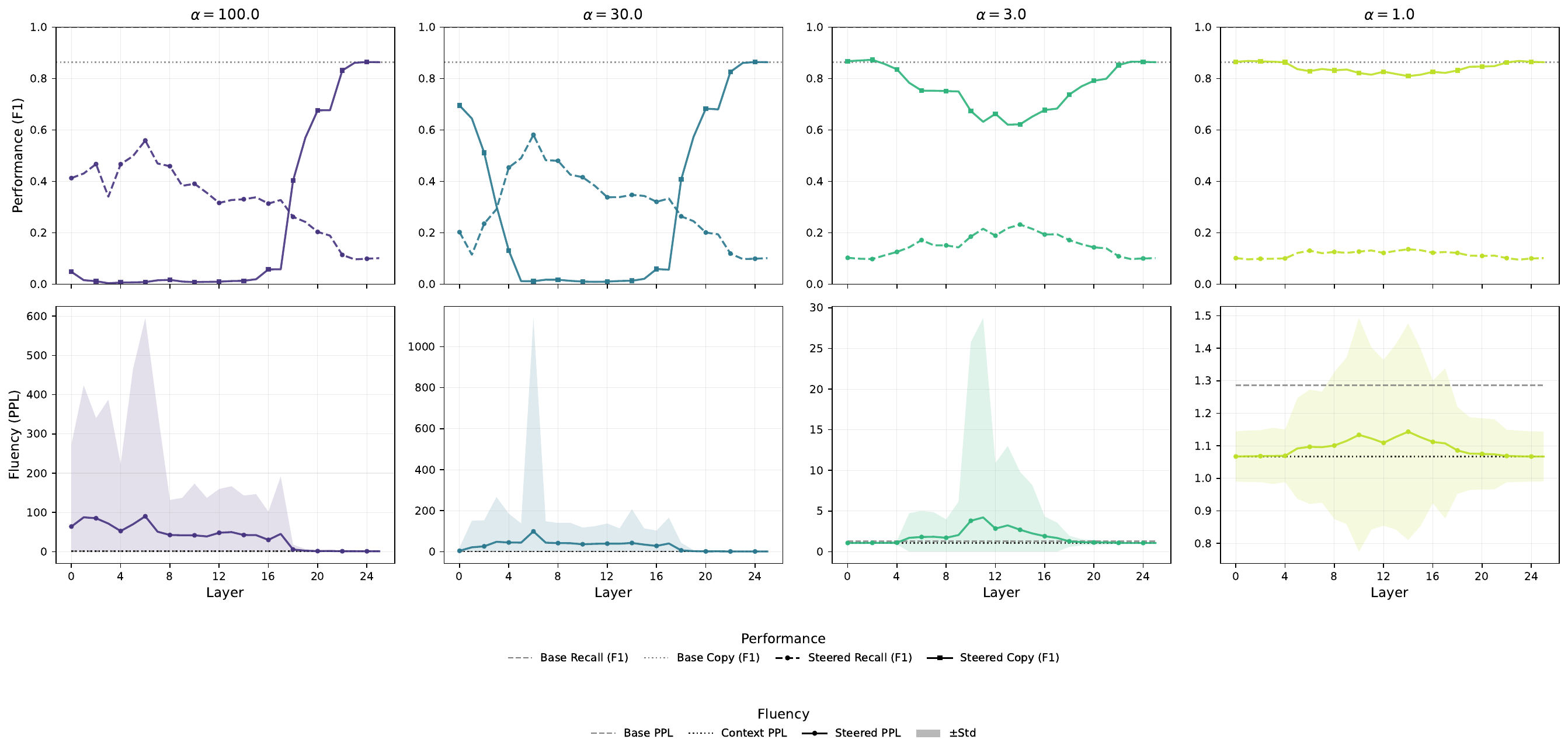}
\caption{Layer-wise effects of steering across four $\alpha$. Rows show performance (F1) and fluency (PPL). Columns correspond to $\alpha$.}
\label{fig:ov-perf-prob-flu-popqa-gemma2-cr-cf-obj}
\end{figure}

\subsubsection{Contextual Steering (Recall$\rightarrow$Copy) - Gemma2-2B}
\begin{figure}[H]
\centering
\includegraphics[width=\textwidth]{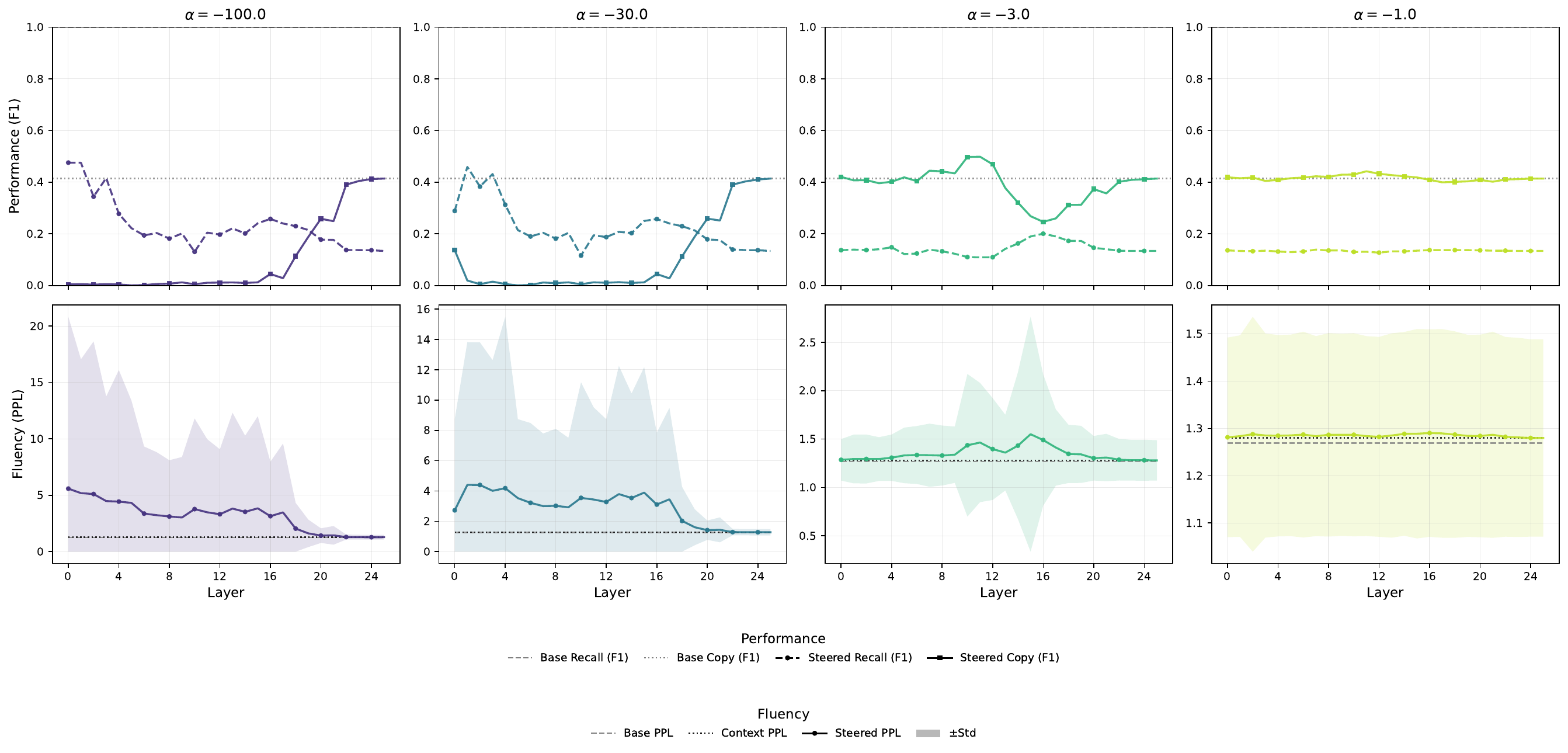}
\caption{Layer-wise effects of steering across four $\alpha$. Rows show performance (F1) and fluency (PPL). Columns correspond to $\alpha$.}
\label{fig:ov-perf-prob-flu-popqa-gemma2-rc-cf-obj}
\end{figure}

\subsubsection{Parametric Steering (Copy$\rightarrow$Recall) - T5Gemma-2B}
\begin{figure}[H]
\centering
\includegraphics[width=\textwidth]{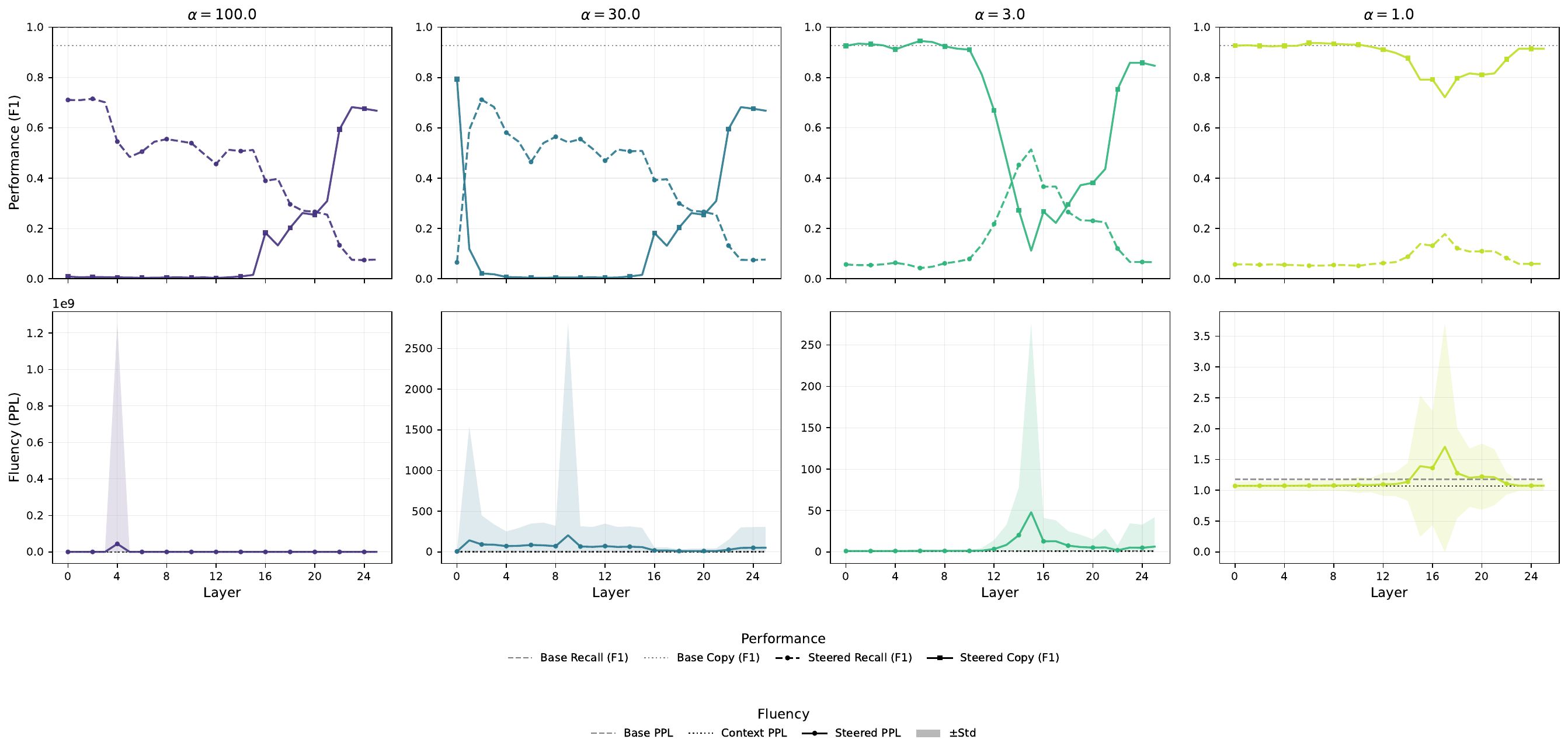}
\caption{Layer-wise effects of steering across four $\alpha$. Rows show performance (F1) and fluency (PPL). Columns correspond to $\alpha$.}
\label{fig:ov-perf-prob-flu-popqa-t5gemma-cr-cf-obj}
\end{figure}

\subsubsection{Contextual Steering (Recall$\rightarrow$Copy) - T5Gemma-2B}
\begin{figure}[H]
\centering
\includegraphics[width=\textwidth]{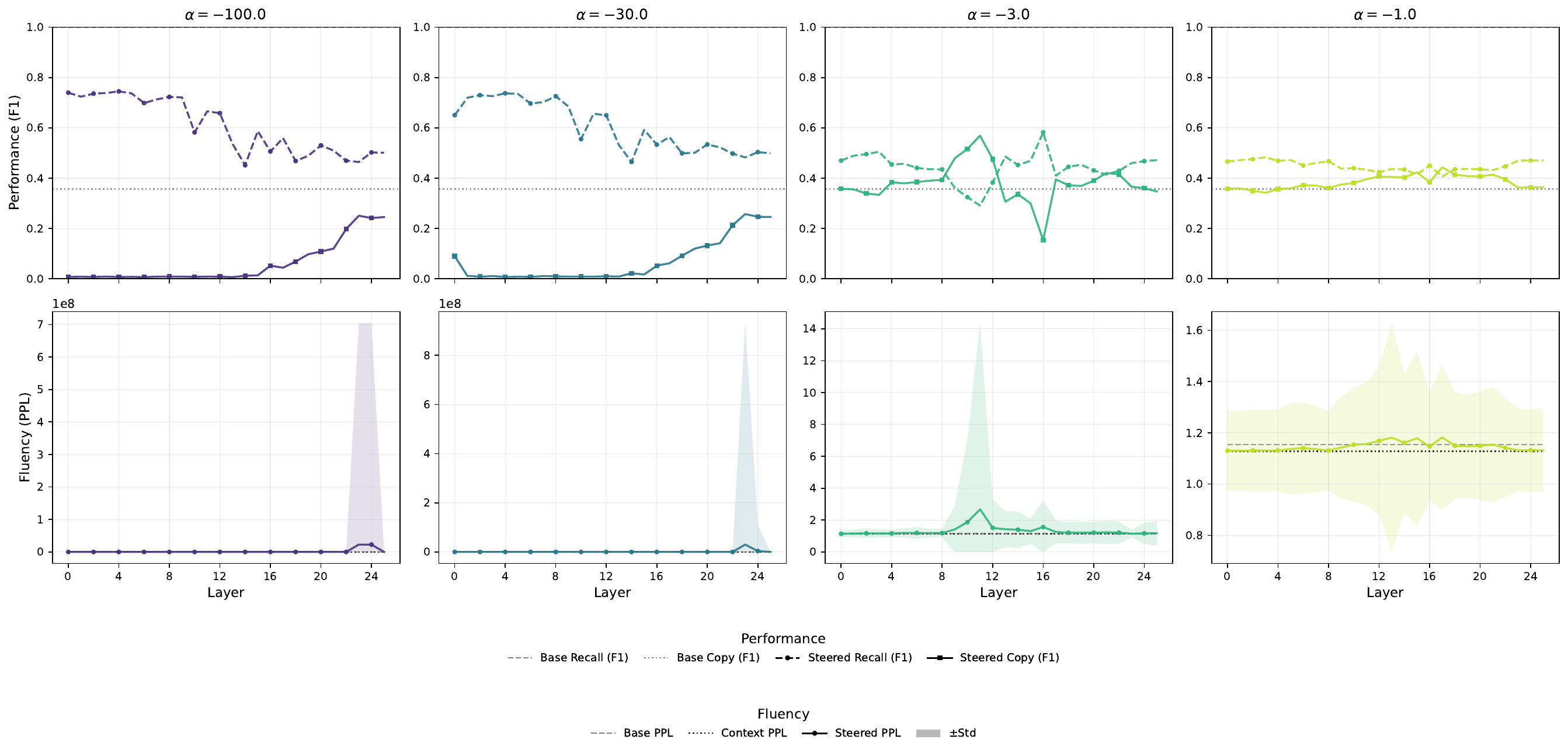}
\caption{Layer-wise effects of steering across four $\alpha$. Rows show performance (F1) and fluency (PPL). Columns correspond to $\alpha$.}
\label{fig:ov-perf-prob-flu-popqa-t5gemma-rc-cf-obj}
\end{figure}

\subsubsection{Parametric Steering (Copy$\rightarrow$Recall) - OLMo-2-0425-1B}
\begin{figure}[H]
\centering
\includegraphics[width=\textwidth]{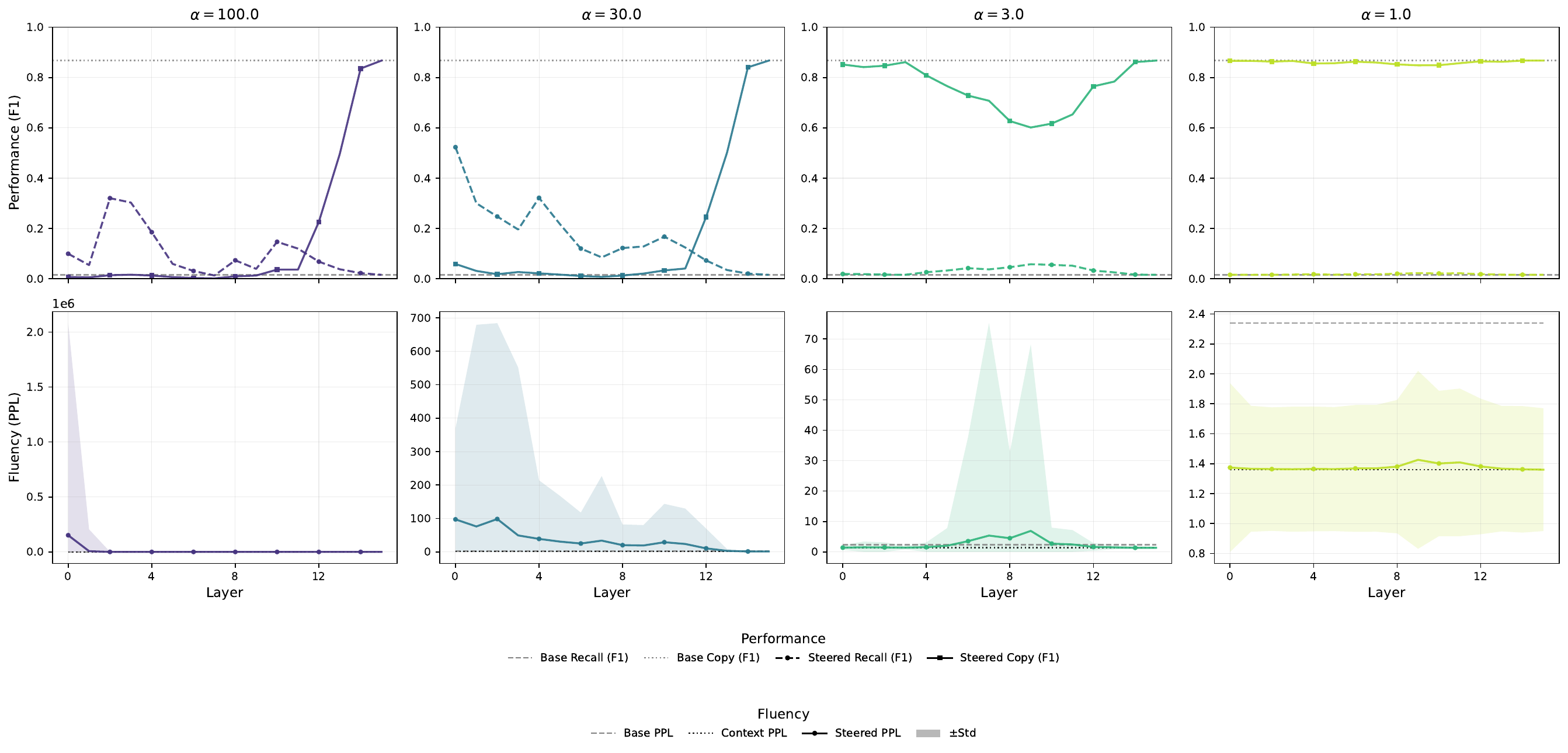}
\caption{Layer-wise effects of steering across four $\alpha$. Rows show performance (F1) and fluency (PPL). Columns correspond to $\alpha$.}
\label{fig:ov-perf-prob-flu-popqa-olmo2-cr-cf-obj}
\end{figure}

\subsubsection{Contextual Steering (Recall$\rightarrow$Copy) - OLMo-2-0425-1B}
\begin{figure}[H]
\centering
\includegraphics[width=\textwidth]{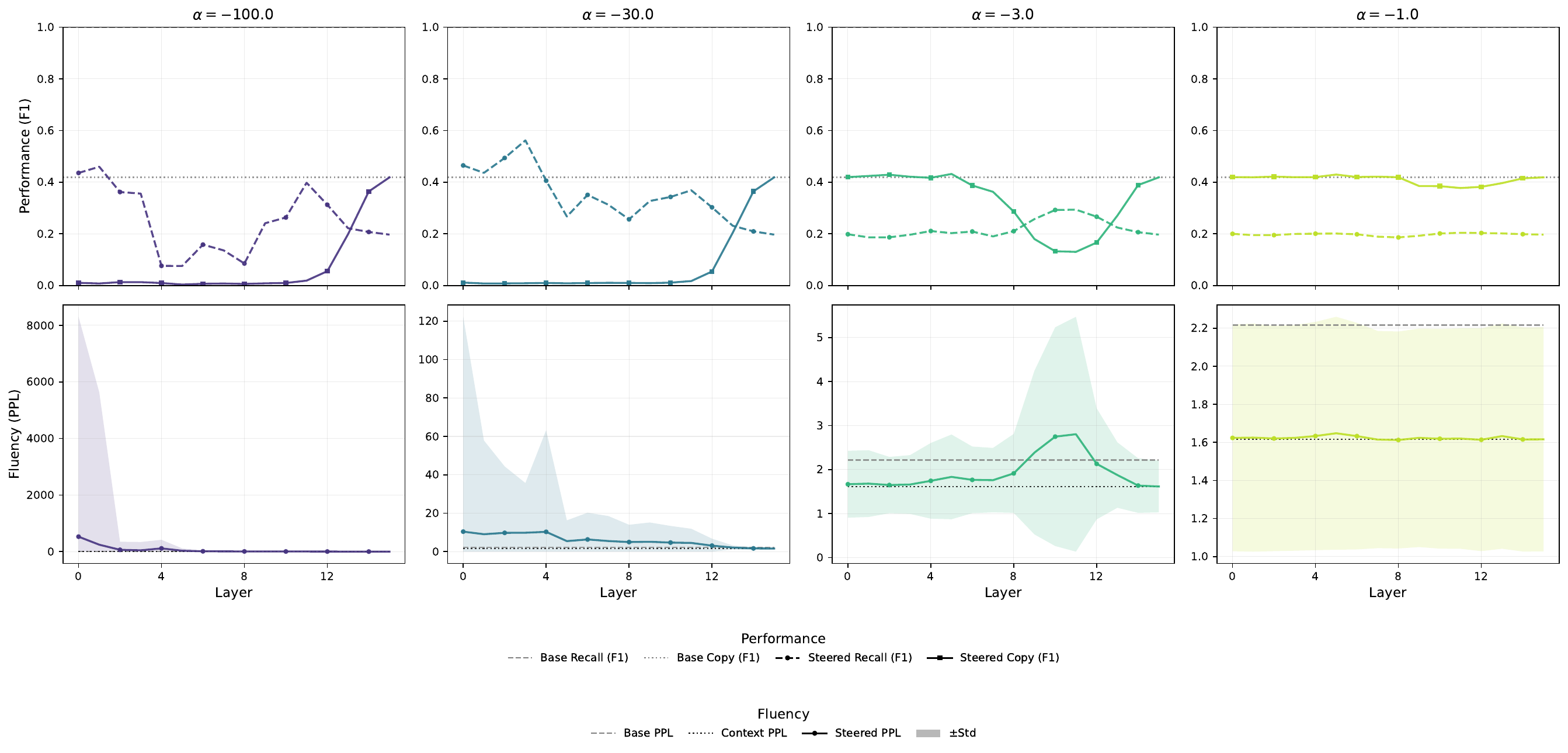}
\caption{Layer-wise effects of steering across four $\alpha$. Rows show performance (F1) and fluency (PPL). Columns correspond to $\alpha$.}
\label{fig:ov-perf-prob-flu-popqa-olmo2-rc-cf-obj}
\end{figure}

\subsection{PopQA: Subject Token (Context-First)}

\subsubsection{Parametric Steering (Copy$\rightarrow$Recall) - Gemma2-2B}
\begin{figure}[H]
\centering
\includegraphics[width=\textwidth]{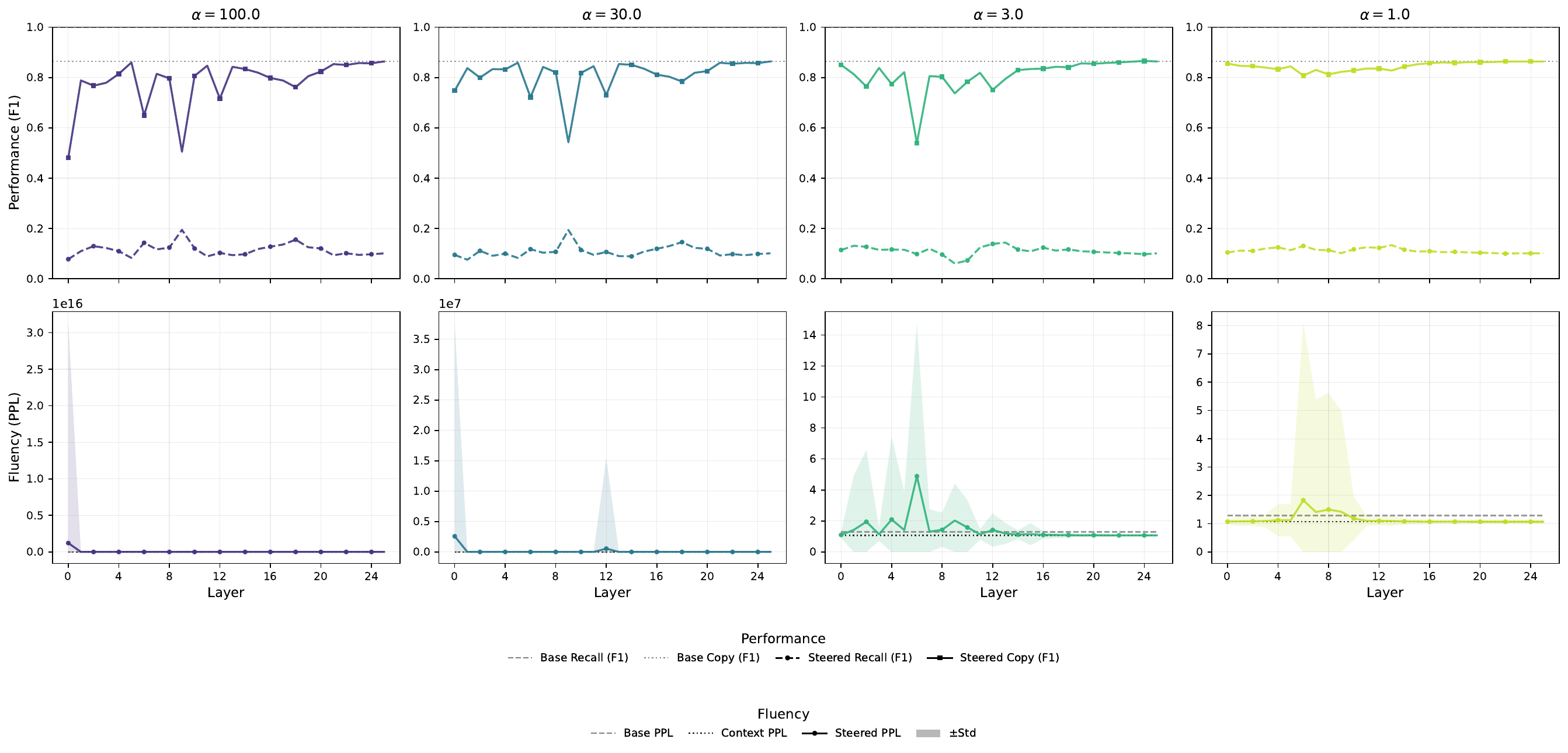}
\caption{Layer-wise effects of steering across four $\alpha$. Rows show performance (F1) and fluency (PPL). Columns correspond to $\alpha$.}
\label{fig:ov-perf-prob-flu-popqa-gemma2-cr-cf-subj}
\end{figure}

\subsubsection{Contextual Steering (Recall$\rightarrow$Copy) - Gemma2-2B}
\begin{figure}[H]
\centering
\includegraphics[width=\textwidth]{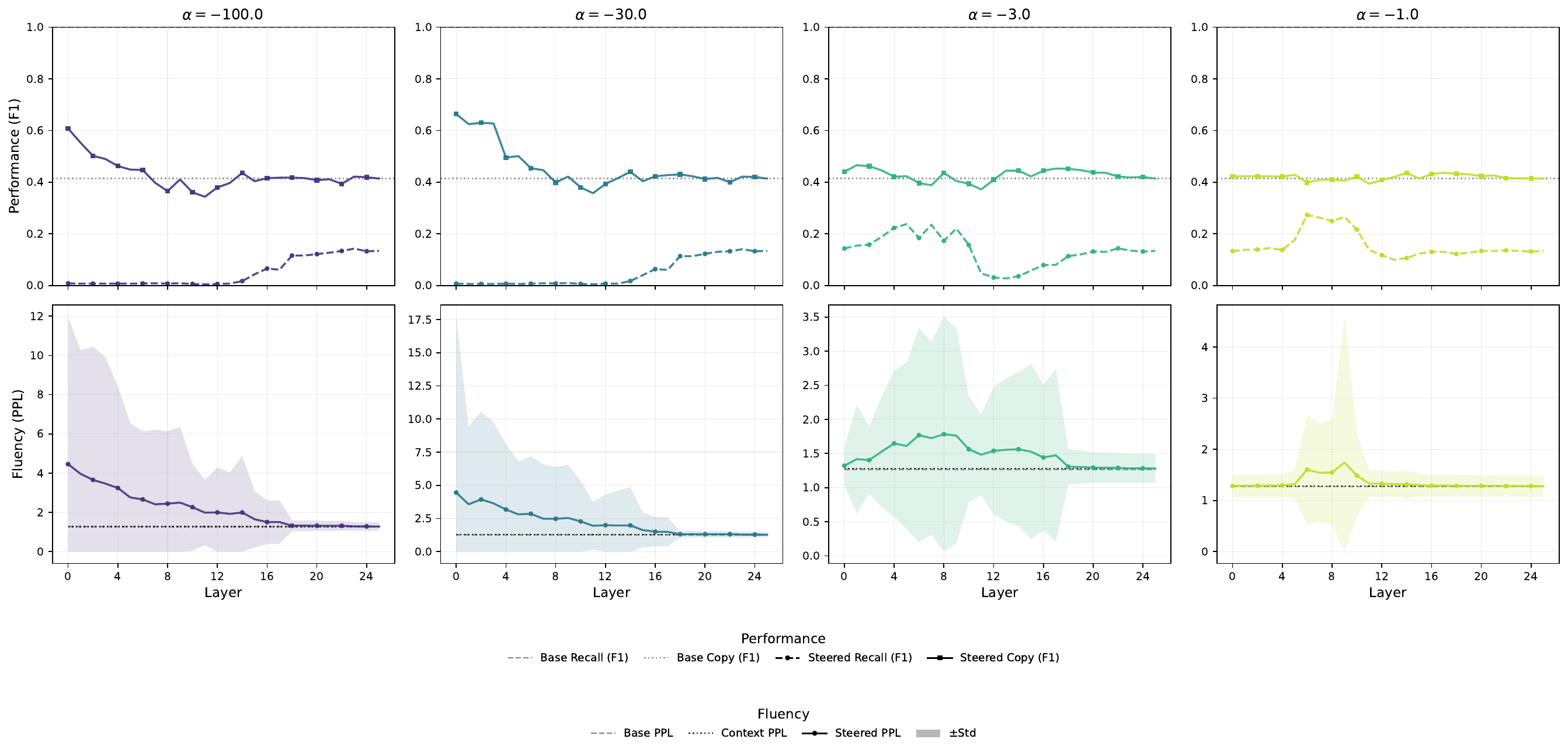}
\caption{Layer-wise effects of steering across four $\alpha$. Rows show performance (F1) and fluency (PPL). Columns correspond to $\alpha$.}
\label{fig:ov-perf-prob-flu-popqa-gemma2-rc-cf-subj}
\end{figure}

\subsubsection{Parametric Steering (Copy$\rightarrow$Recall) - T5Gemma-2B}
\begin{figure}[H]
\centering
\includegraphics[width=\textwidth]{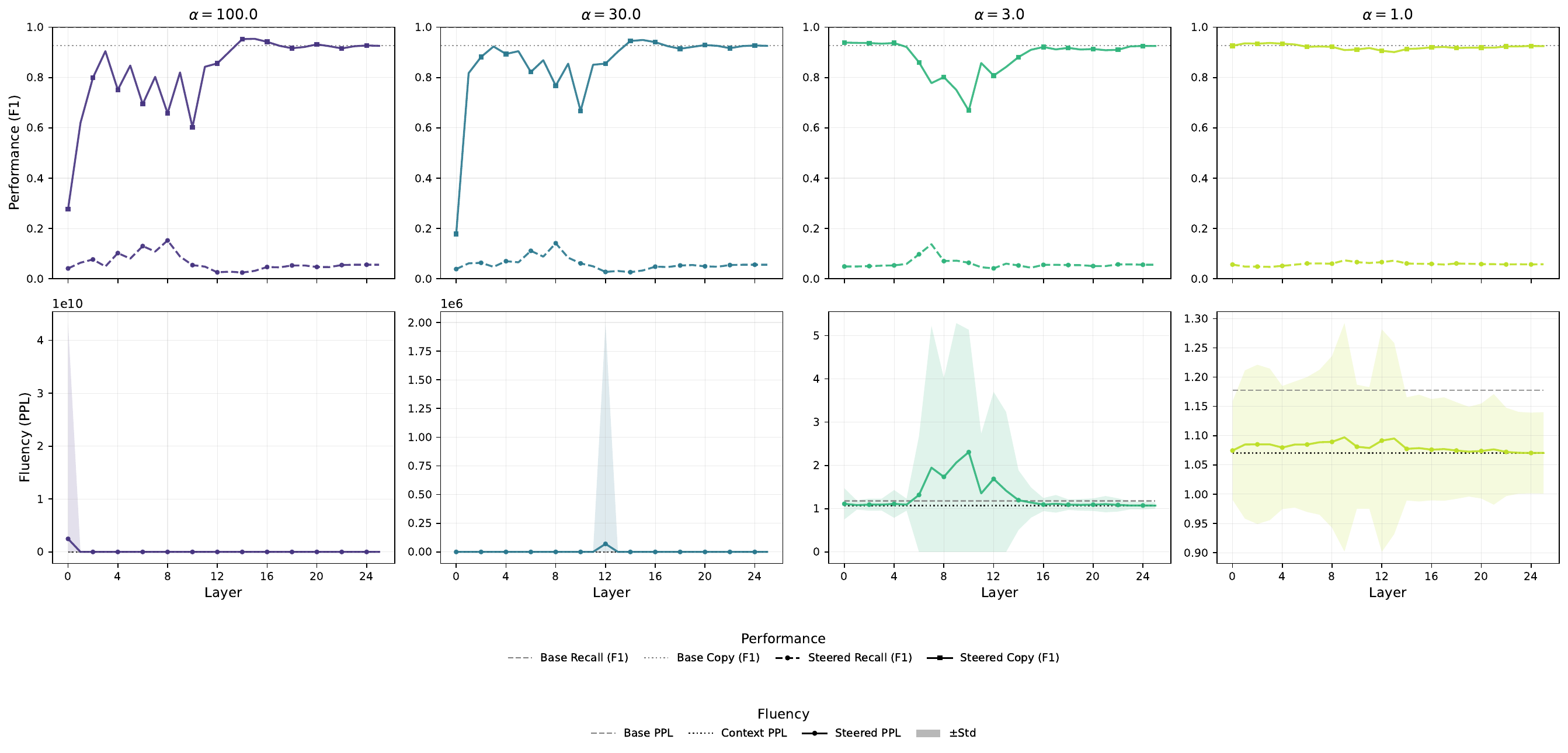}
\caption{Layer-wise effects of steering across four $\alpha$. Rows show performance (F1) and fluency (PPL). Columns correspond to $\alpha$.}
\label{fig:ov-perf-prob-flu-popqa-t5gemma-cr-cf-subj}
\end{figure}

\subsubsection{Contextual Steering (Recall$\rightarrow$Copy) - T5Gemma-2B}
\begin{figure}[H]
\centering
\includegraphics[width=\textwidth]{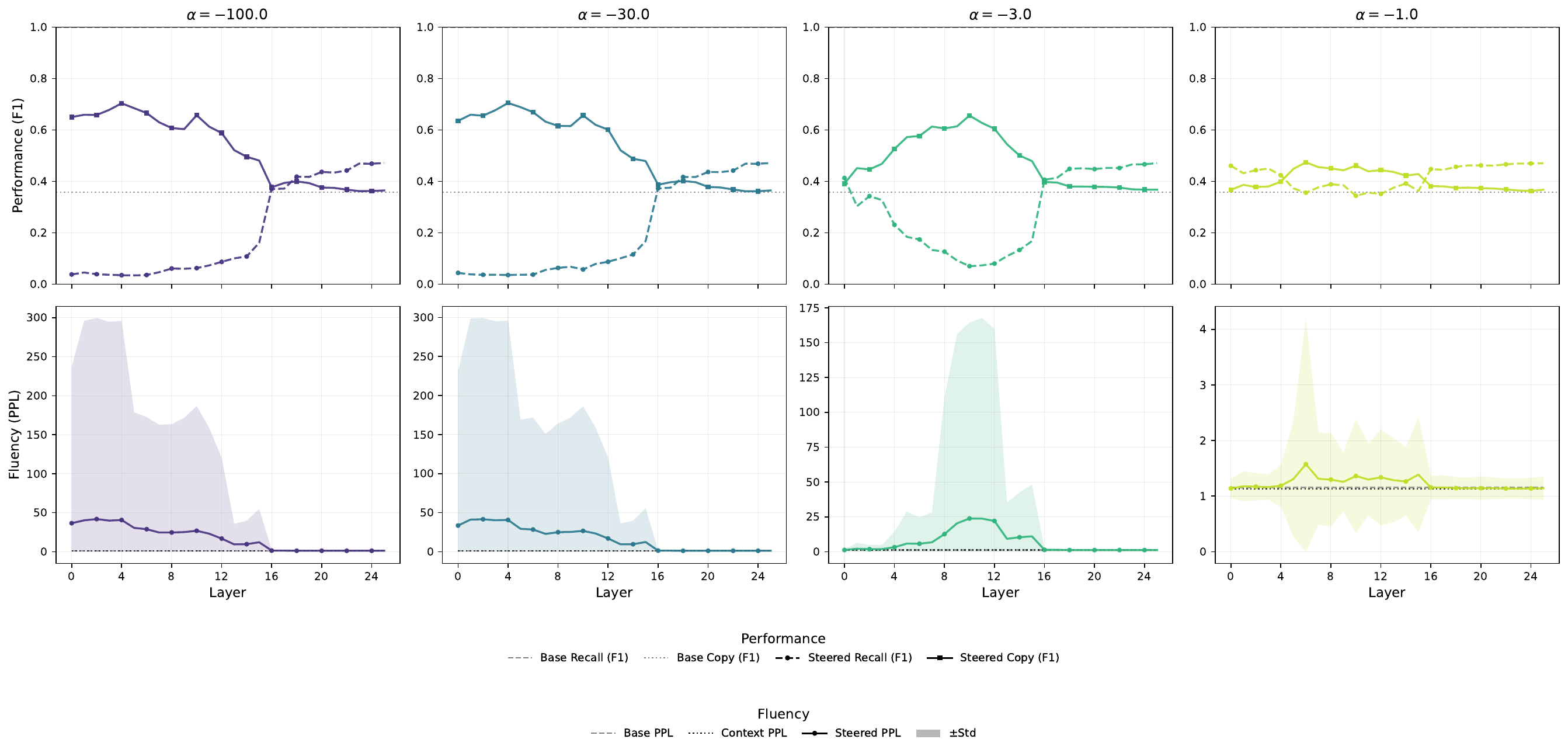}
\caption{Layer-wise effects of steering across four $\alpha$. Rows show performance (F1) and fluency (PPL). Columns correspond to $\alpha$.}
\label{fig:ov-perf-prob-flu-popqa-t5gemma-rc-cf-subj}
\end{figure}

\subsubsection{Parametric Steering (Copy$\rightarrow$Recall) - OLMo-2-0425-1B}
\begin{figure}[H]
\centering
\includegraphics[width=\textwidth]{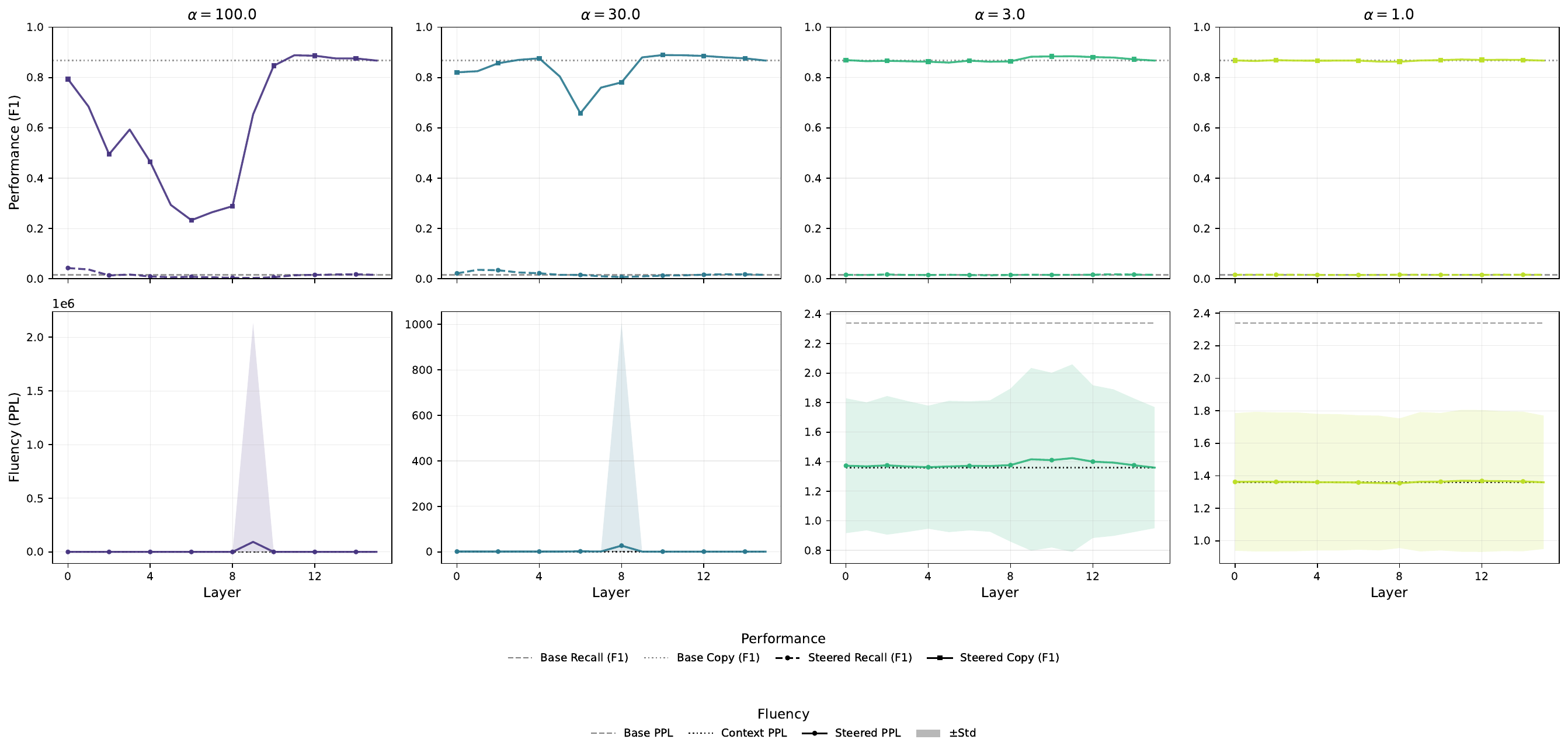}
\caption{Layer-wise effects of steering across four $\alpha$. Rows show performance (F1) and fluency (PPL). Columns correspond to $\alpha$.}
\label{fig:ov-perf-prob-flu-popqa-olmo2-cr-cf-subj}
\end{figure}

\subsubsection{Contextual Steering (Recall$\rightarrow$Copy) - OLMo-2-0425-1B}
\begin{figure}[H]
\centering
\includegraphics[width=\textwidth]{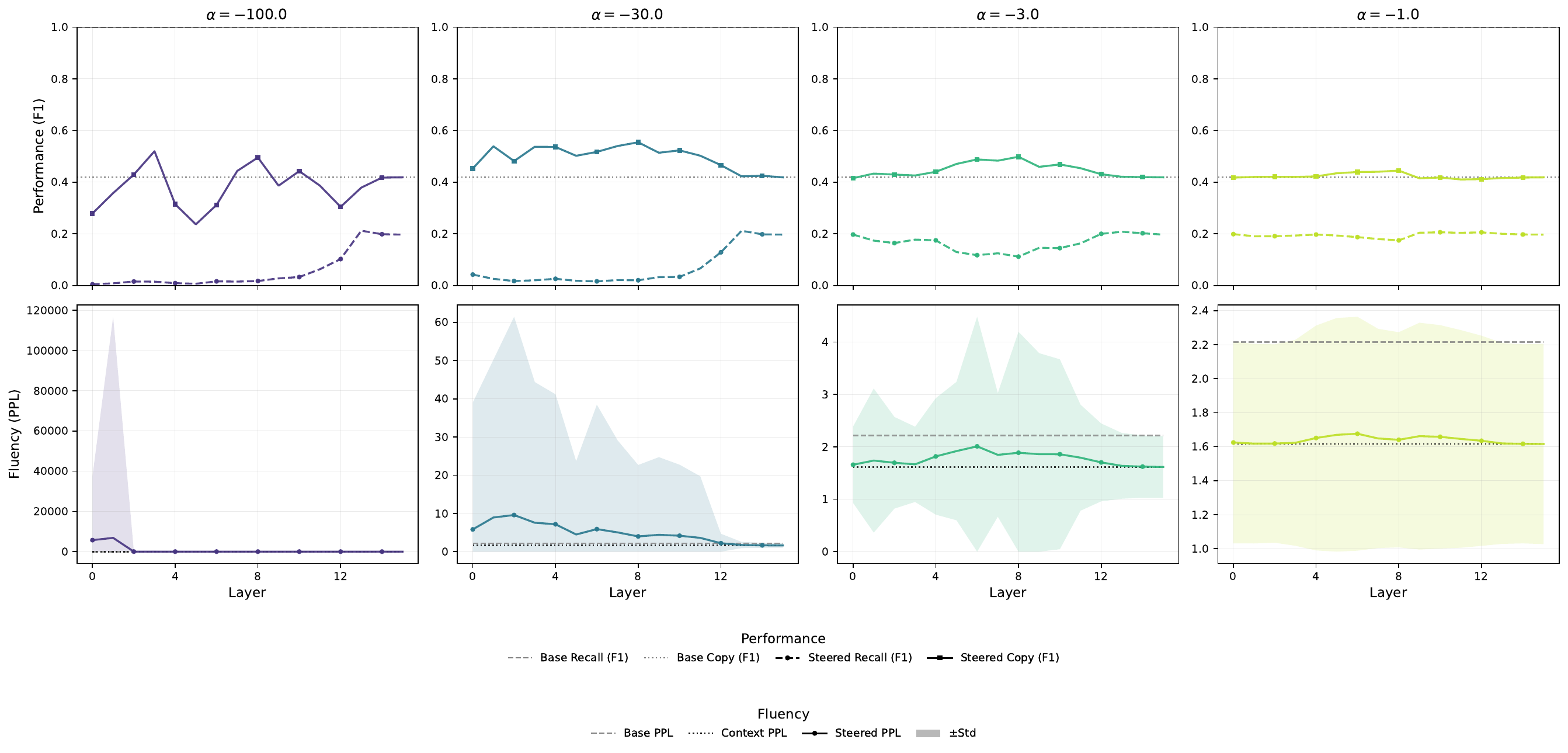}
\caption{Layer-wise effects of steering across four $\alpha$. Rows show performance (F1) and fluency (PPL). Columns correspond to $\alpha$.}
\label{fig:ov-perf-prob-flu-popqa-olmo2-rc-cf-subj}
\end{figure}

\subsection{PopQA: Last Token (Query-First)}

\subsubsection{Parametric Steering (Copy$\rightarrow$Recall) - Gemma2-2B}
\begin{figure}[H]
\centering
\includegraphics[width=\textwidth]{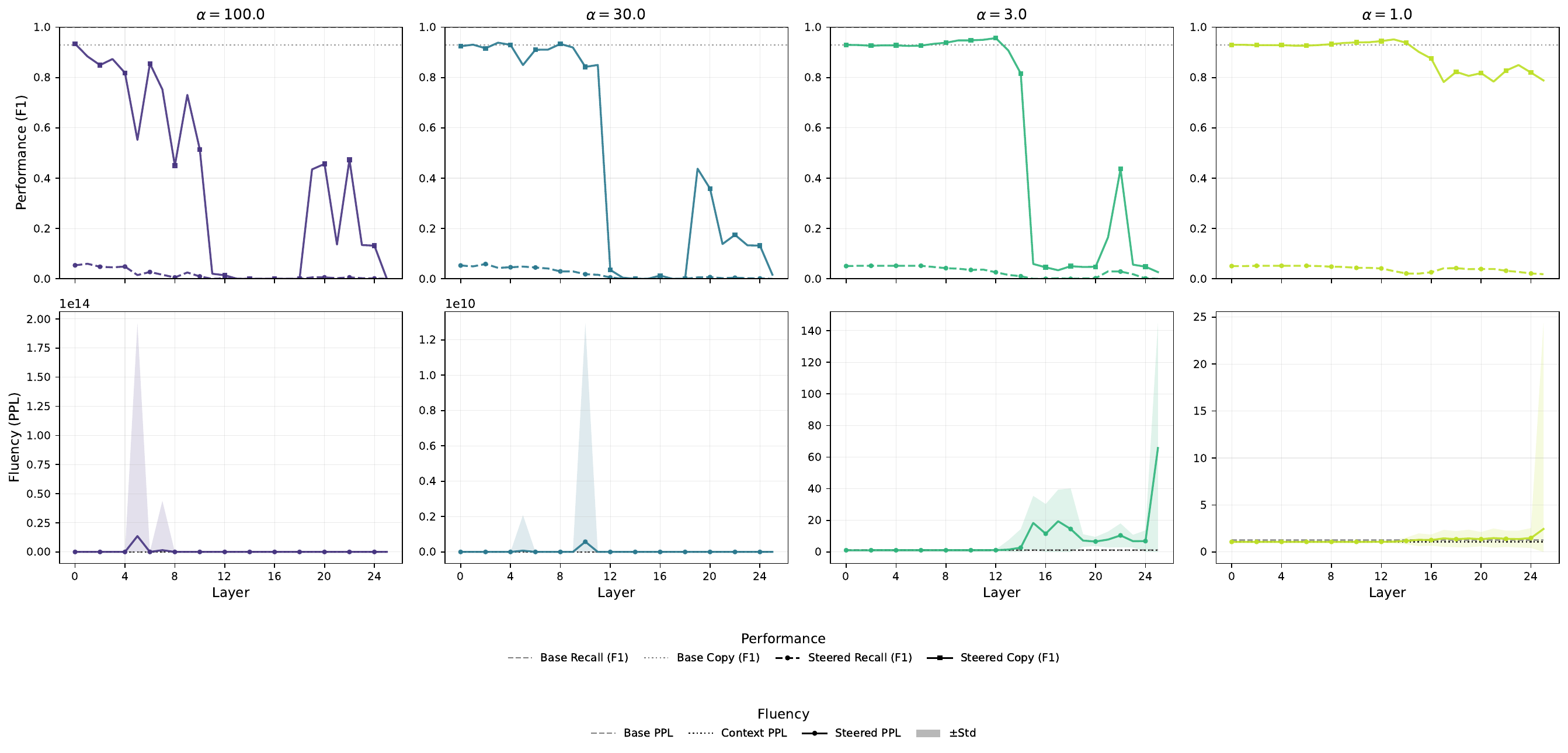}
\caption{Layer-wise effects of steering across four $\alpha$. Rows show performance (F1) and fluency (PPL). Columns correspond to $\alpha$.}
\label{fig:ov-perf-prob-flu-popqa-gemma2-cr-qf-lt}
\end{figure}

\subsubsection{Contextual Steering (Recall$\rightarrow$Copy) - Gemma2-2B}
\begin{figure}[H]
\centering
\includegraphics[width=\textwidth]{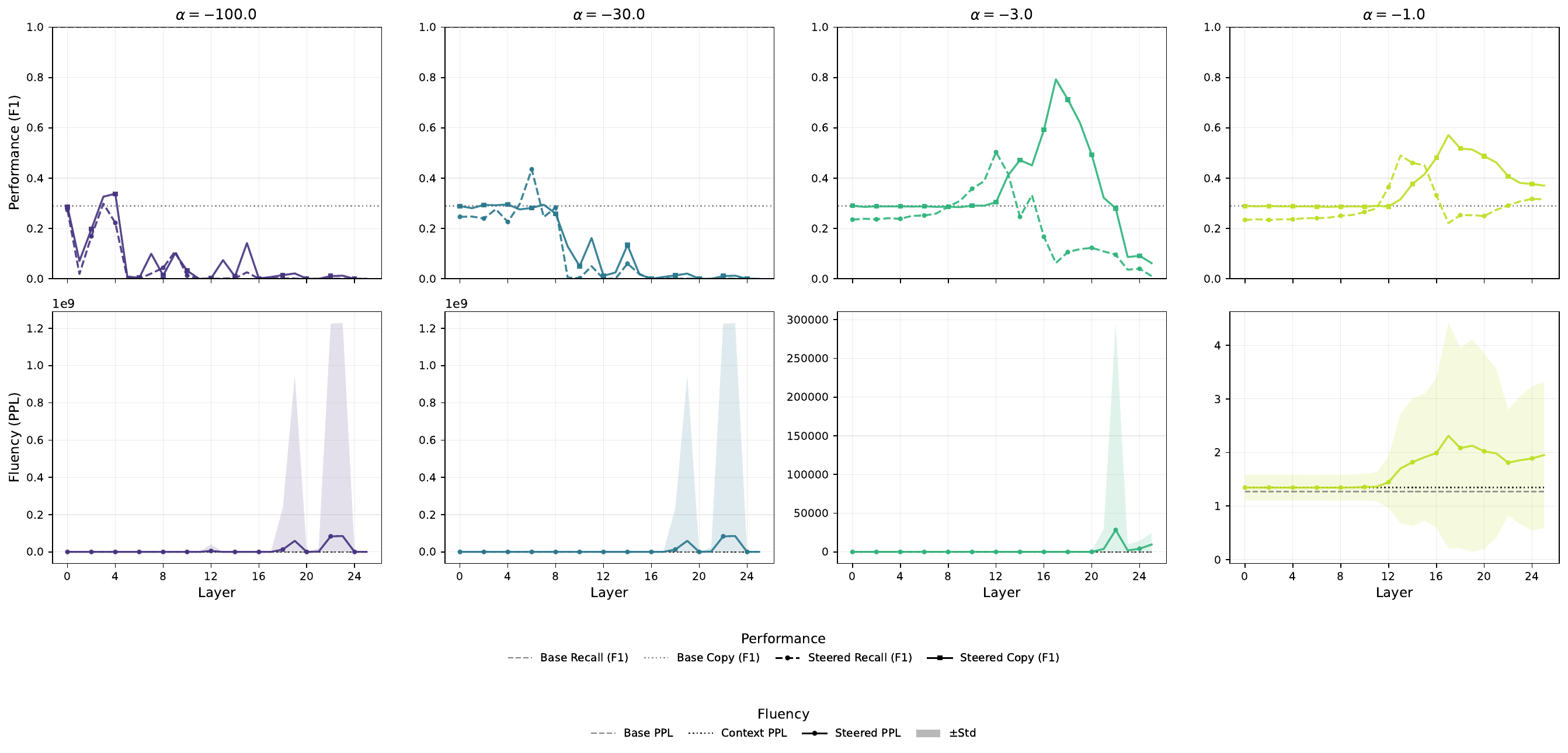}
\caption{Layer-wise effects of steering across four $\alpha$. Rows show performance (F1) and fluency (PPL). Columns correspond to $\alpha$.}
\label{fig:ov-perf-prob-flu-popqa-gemma2-rc-qf-lt}
\end{figure}

\subsubsection{Parametric Steering (Copy$\rightarrow$Recall) - T5Gemma-2B}
\begin{figure}[H]
\centering
\includegraphics[width=\textwidth]{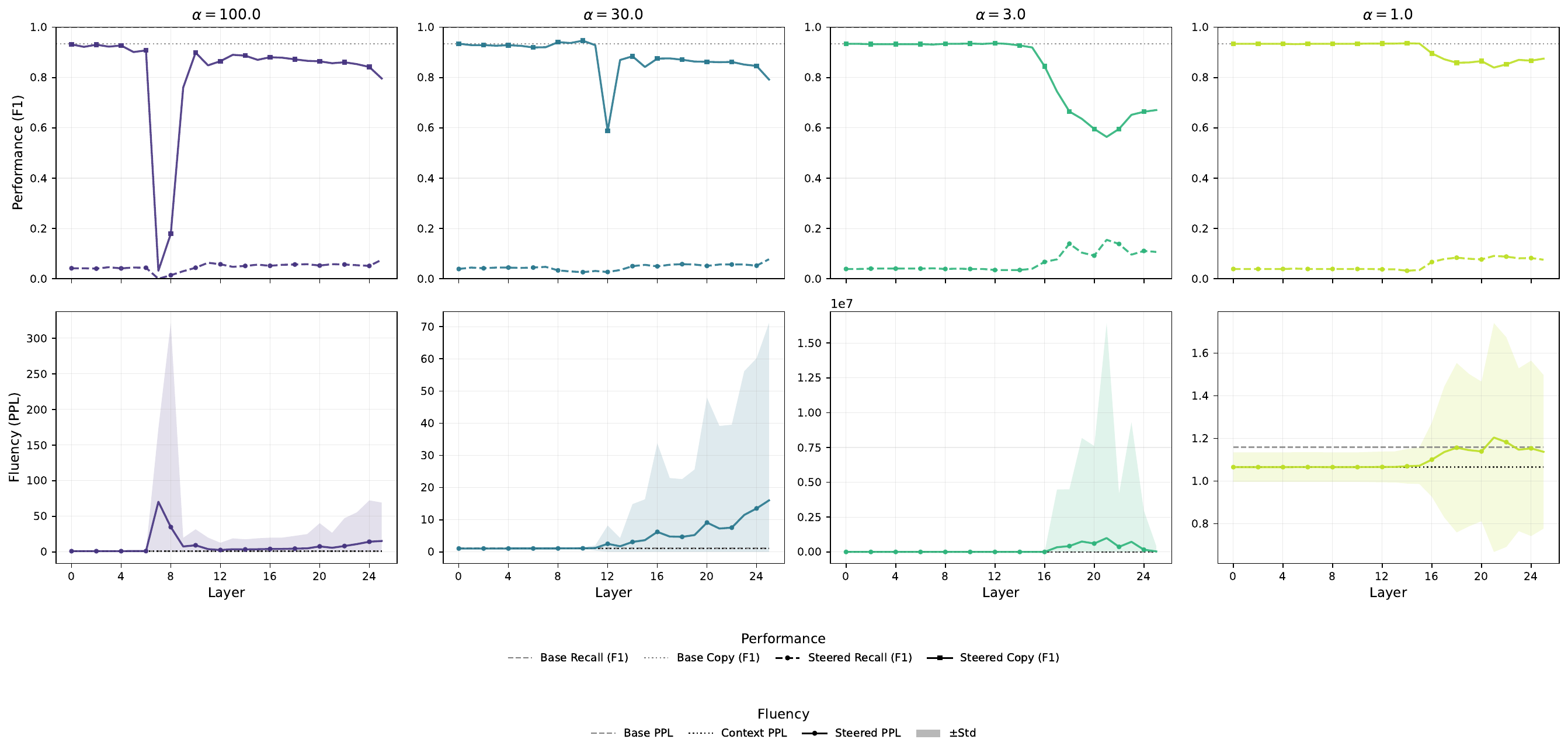}
\caption{Layer-wise effects of steering across four $\alpha$. Rows show performance (F1) and fluency (PPL). Columns correspond to $\alpha$.}
\label{fig:ov-perf-prob-flu-popqa-t5gemma-cr-qf-lt}
\end{figure}

\subsubsection{Contextual Steering (Recall$\rightarrow$Copy) - T5Gemma-2B}
\begin{figure}[H]
\centering
\includegraphics[width=\textwidth]{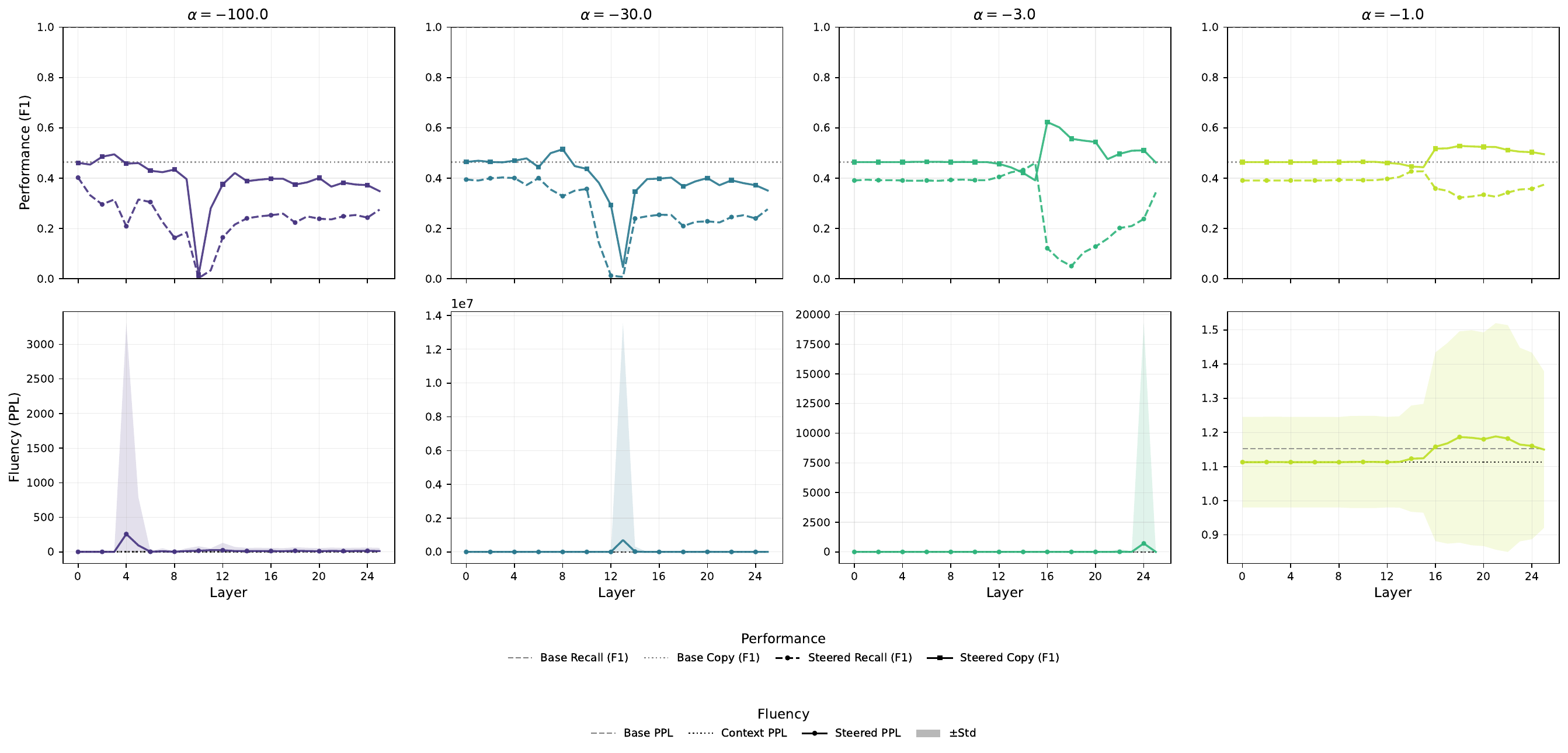}
\caption{Layer-wise effects of steering across four $\alpha$. Rows show performance (F1) and fluency (PPL). Columns correspond to $\alpha$.}
\label{fig:ov-perf-prob-flu-popqa-t5gemma-rc-qf-lt}
\end{figure}

\subsubsection{Parametric Steering (Copy$\rightarrow$Recall) - OLMo-2-0425-1B}
\begin{figure}[H]
\centering
\includegraphics[width=\textwidth]{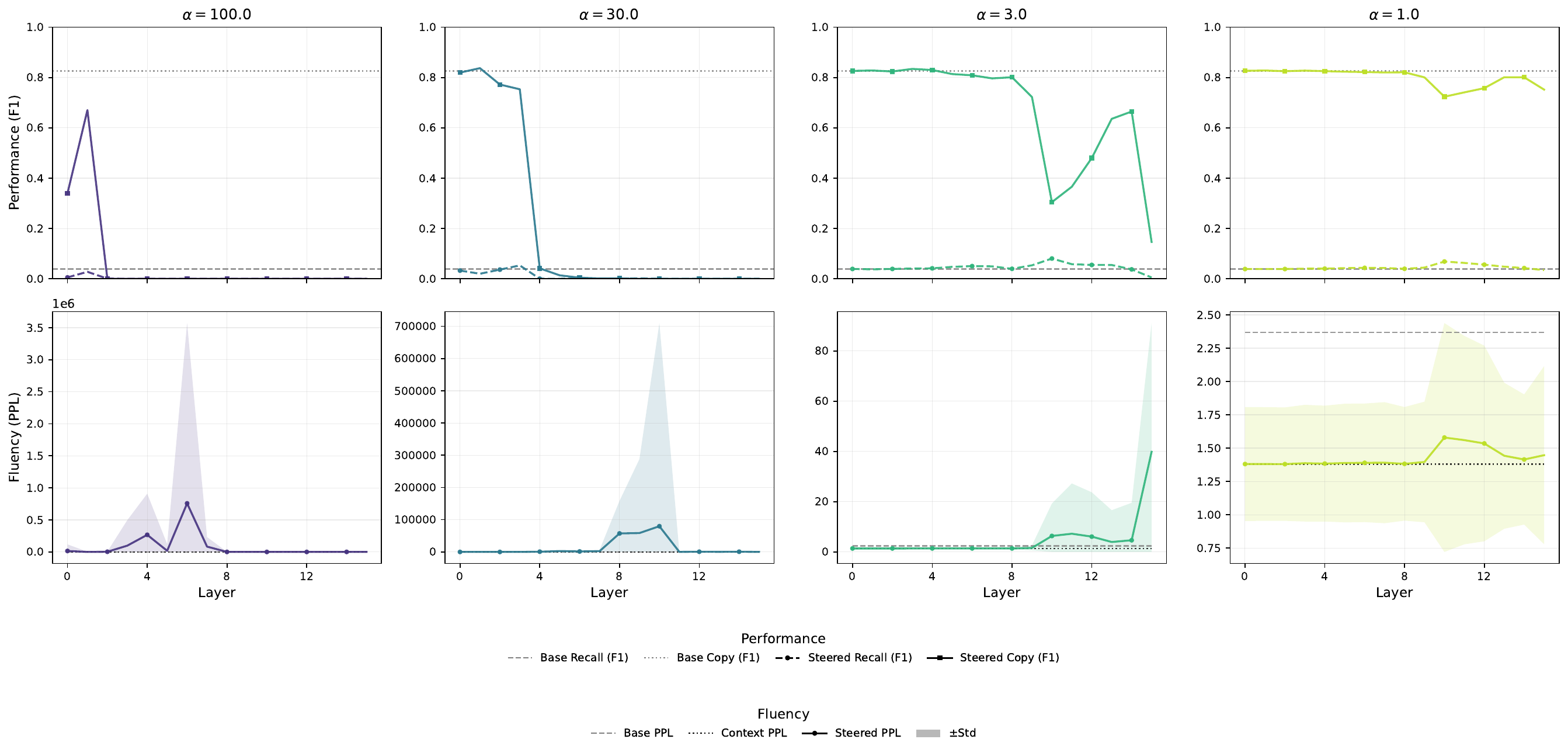}
\caption{Layer-wise effects of steering across four $\alpha$. Rows show performance (F1) and fluency (PPL). Columns correspond to $\alpha$.}
\label{fig:ov-perf-prob-flu-popqa-olmo2-cr-qf-lt}
\end{figure}

\subsubsection{Contextual Steering (Recall$\rightarrow$Copy) - OLMo-2-0425-1B}
\begin{figure}[H]
\centering
\includegraphics[width=\textwidth]{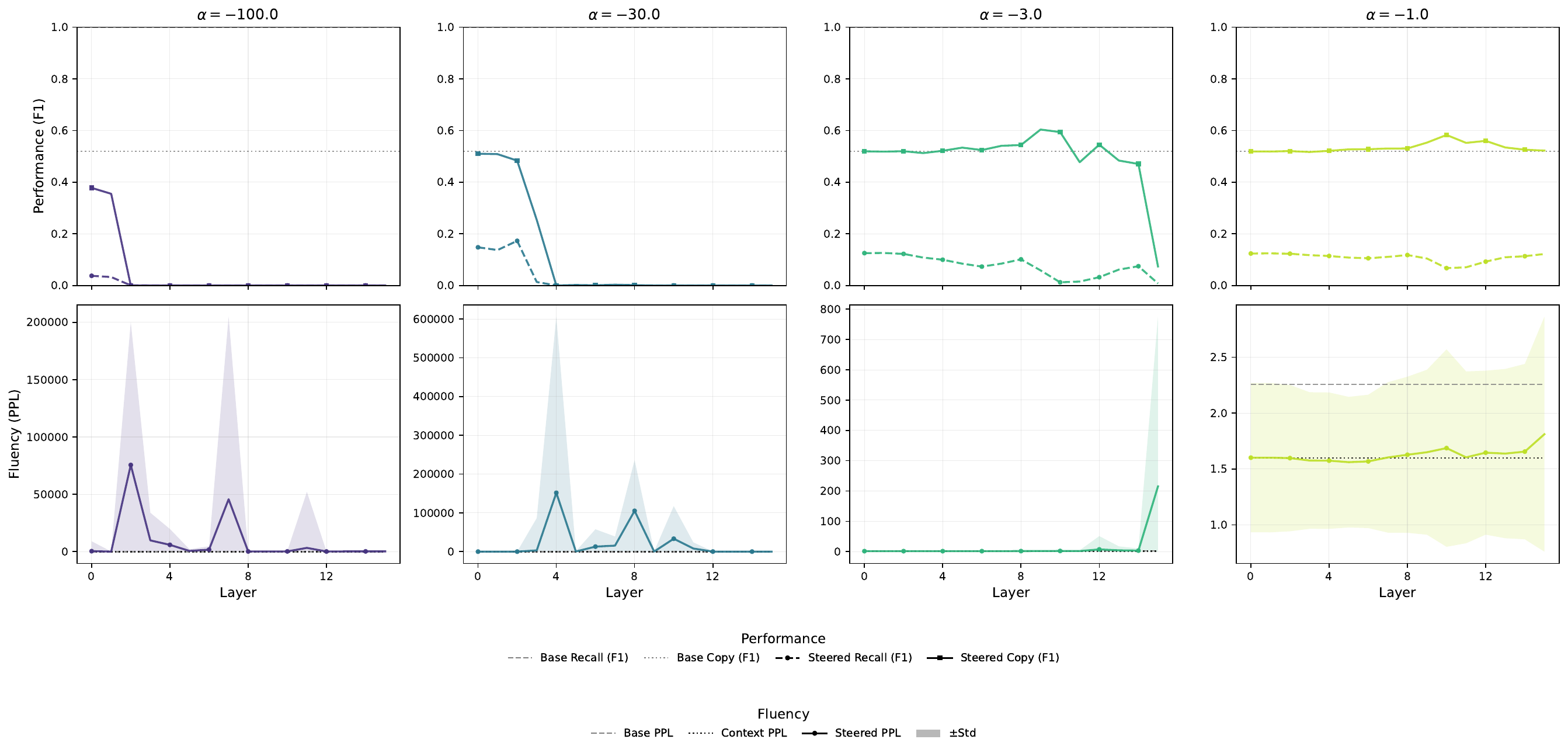}
\caption{Layer-wise effects of steering across four $\alpha$. Rows show performance (F1) and fluency (PPL). Columns correspond to $\alpha$.}
\label{fig:ov-perf-prob-flu-popqa-olmo2-rc-qf-lt}
\end{figure}

\subsection{PopQA: Last Token (Context-First)}

\subsubsection{Parametric Steering (Copy$\rightarrow$Recall) - Gemma2-2B}
\begin{figure}[H]
\centering
\includegraphics[width=\textwidth]{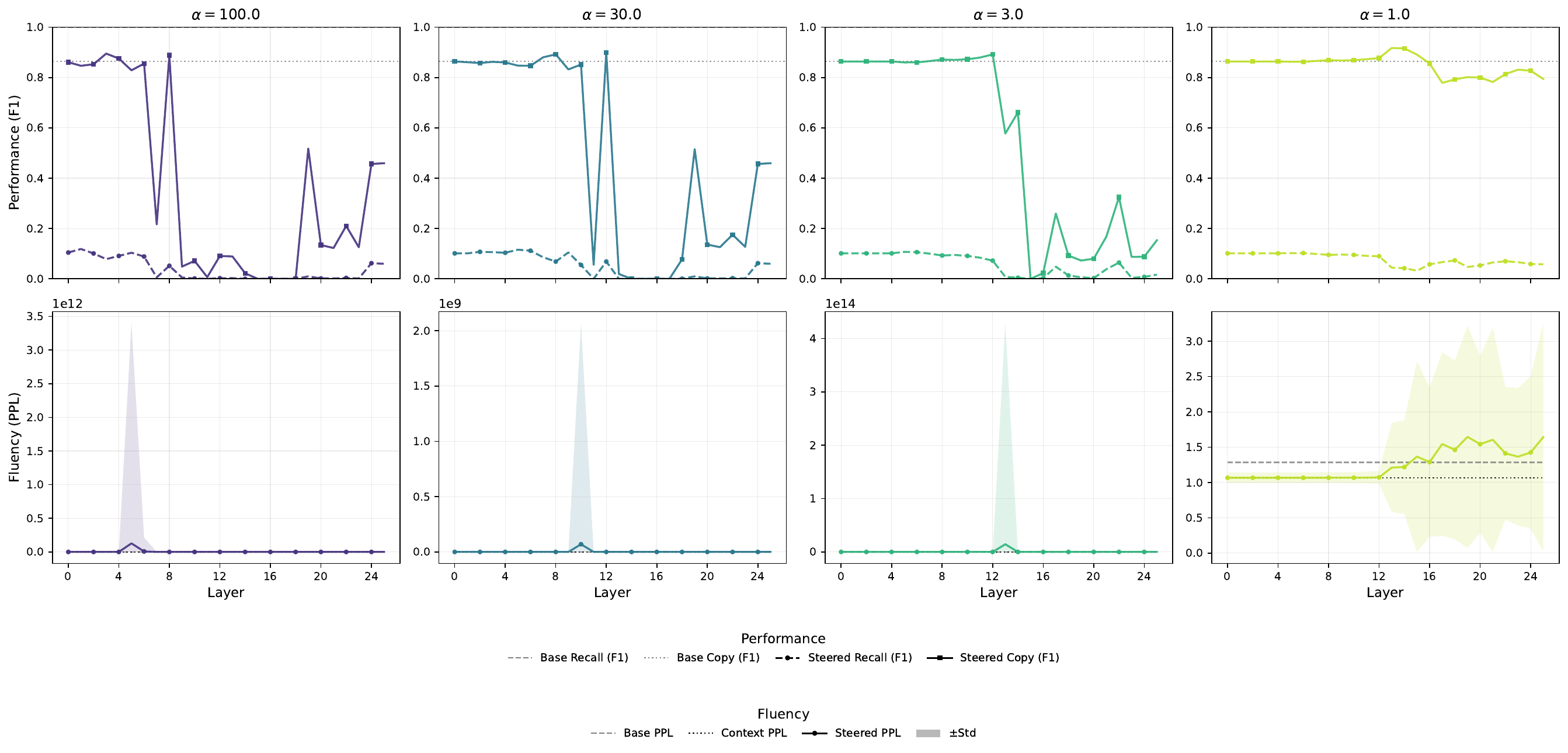}
\caption{Layer-wise effects of steering across four $\alpha$. Rows show performance (F1) and fluency (PPL). Columns correspond to $\alpha$.}
\label{fig:ov-perf-prob-flu-popqa-gemma2-cr-cf-lt}
\end{figure}

\subsubsection{Contextual Steering (Recall$\rightarrow$Copy) - Gemma2-2B}
\begin{figure}[H]
\centering
\includegraphics[width=\textwidth]{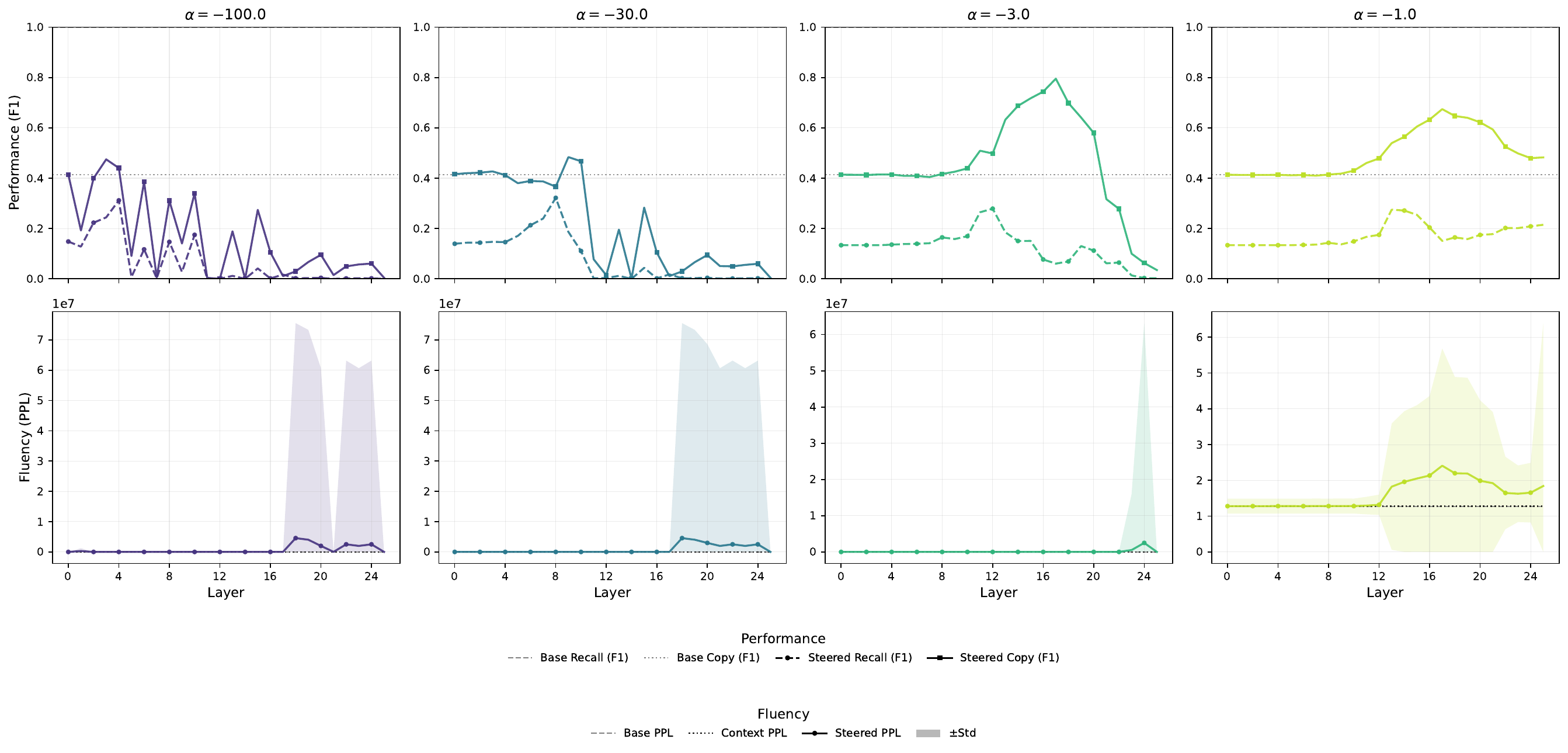}
\caption{Layer-wise effects of steering across four $\alpha$. Rows show performance (F1) and fluency (PPL). Columns correspond to $\alpha$.}
\label{fig:ov-perf-prob-flu-popqa-gemma2-rc-cf-lt}
\end{figure}

\subsubsection{Parametric Steering (Copy$\rightarrow$Recall) - T5Gemma-2B}
\begin{figure}[H]
\centering
\includegraphics[width=\textwidth]{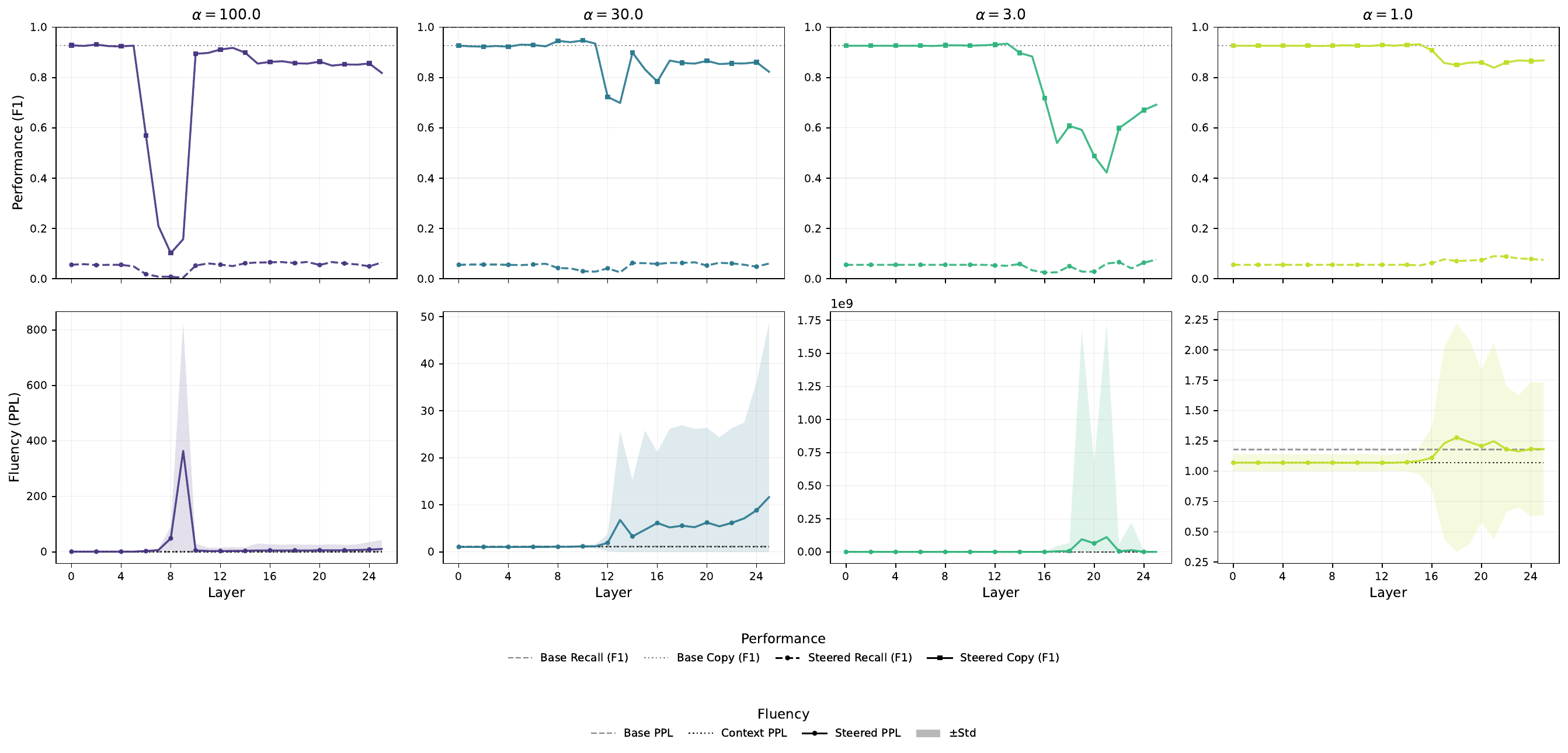}
\caption{Layer-wise effects of steering across four $\alpha$. Rows show performance (F1) and fluency (PPL). Columns correspond to $\alpha$.}
\label{fig:ov-perf-prob-flu-popqa-t5gemma-cr-cf-lt}
\end{figure}

\subsubsection{Contextual Steering (Recall$\rightarrow$Copy) - T5Gemma-2B}
\begin{figure}[H]
\centering
\includegraphics[width=\textwidth]{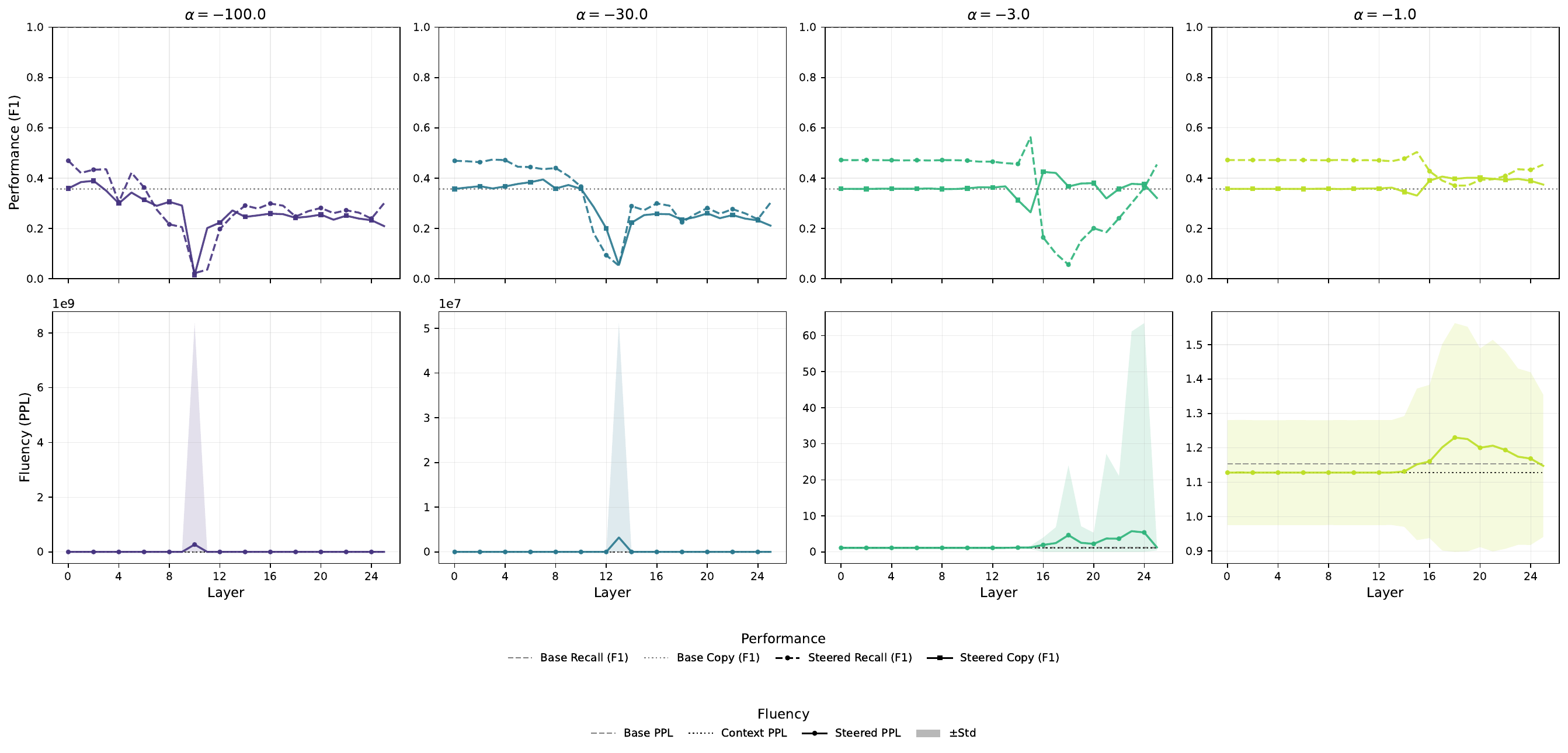}
\caption{Layer-wise effects of steering across four $\alpha$. Rows show performance (F1) and fluency (PPL). Columns correspond to $\alpha$.}
\label{fig:ov-perf-prob-flu-popqa-t5gemma-rc-cf-lt}
\end{figure}

\subsubsection{Parametric Steering (Copy$\rightarrow$Recall) - OLMo-2-0425-1B}
\begin{figure}[H]
\centering
\includegraphics[width=\textwidth]{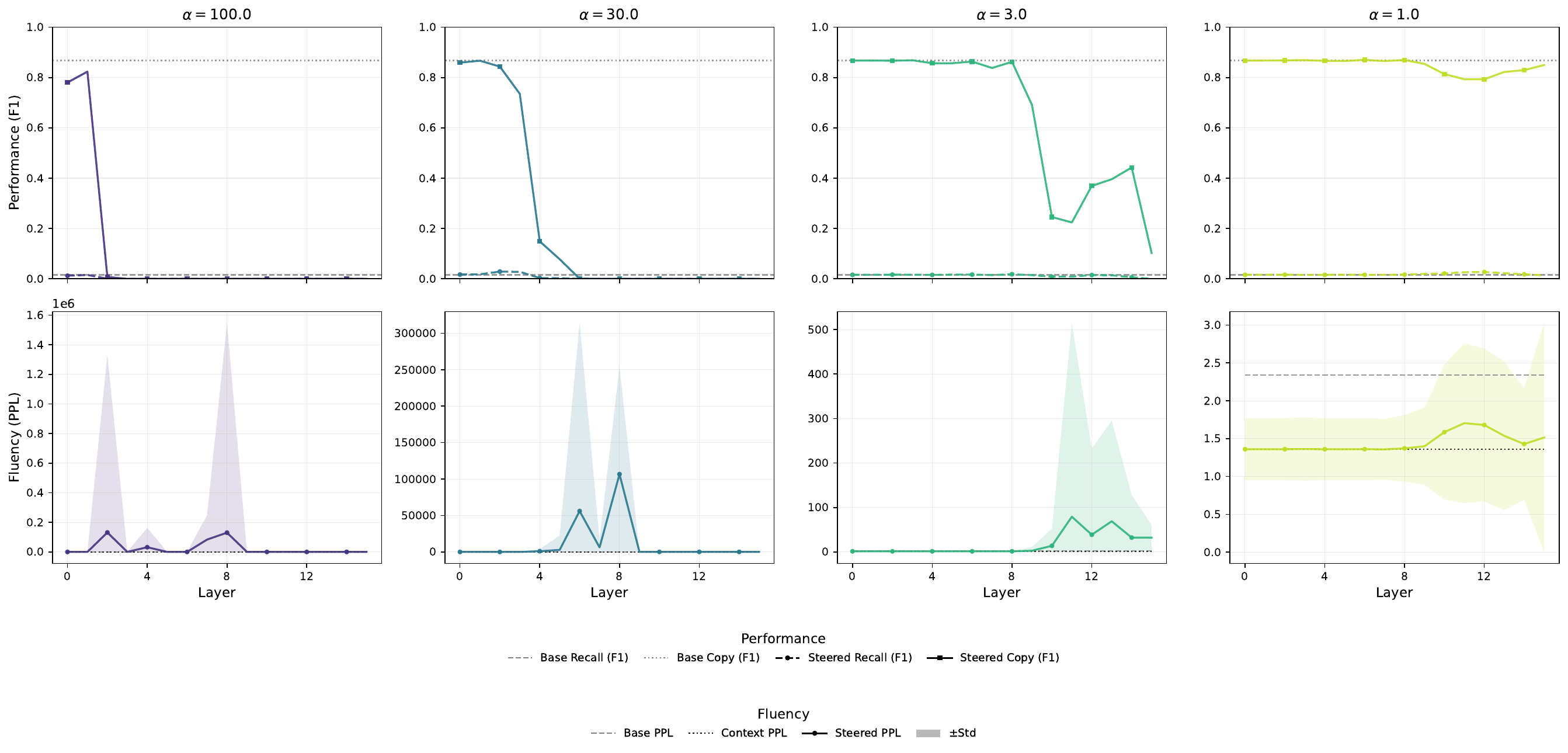}
\caption{Layer-wise effects of steering across four $\alpha$. Rows show performance (F1) and fluency (PPL). Columns correspond to $\alpha$.}
\label{fig:ov-perf-prob-flu-popqa-olmo2-cr-cf-lt}
\end{figure}

\subsubsection{Contextual Steering (Recall$\rightarrow$Copy) - OLMo-2-0425-1B}
\begin{figure}[H]
\centering
\includegraphics[width=\textwidth]{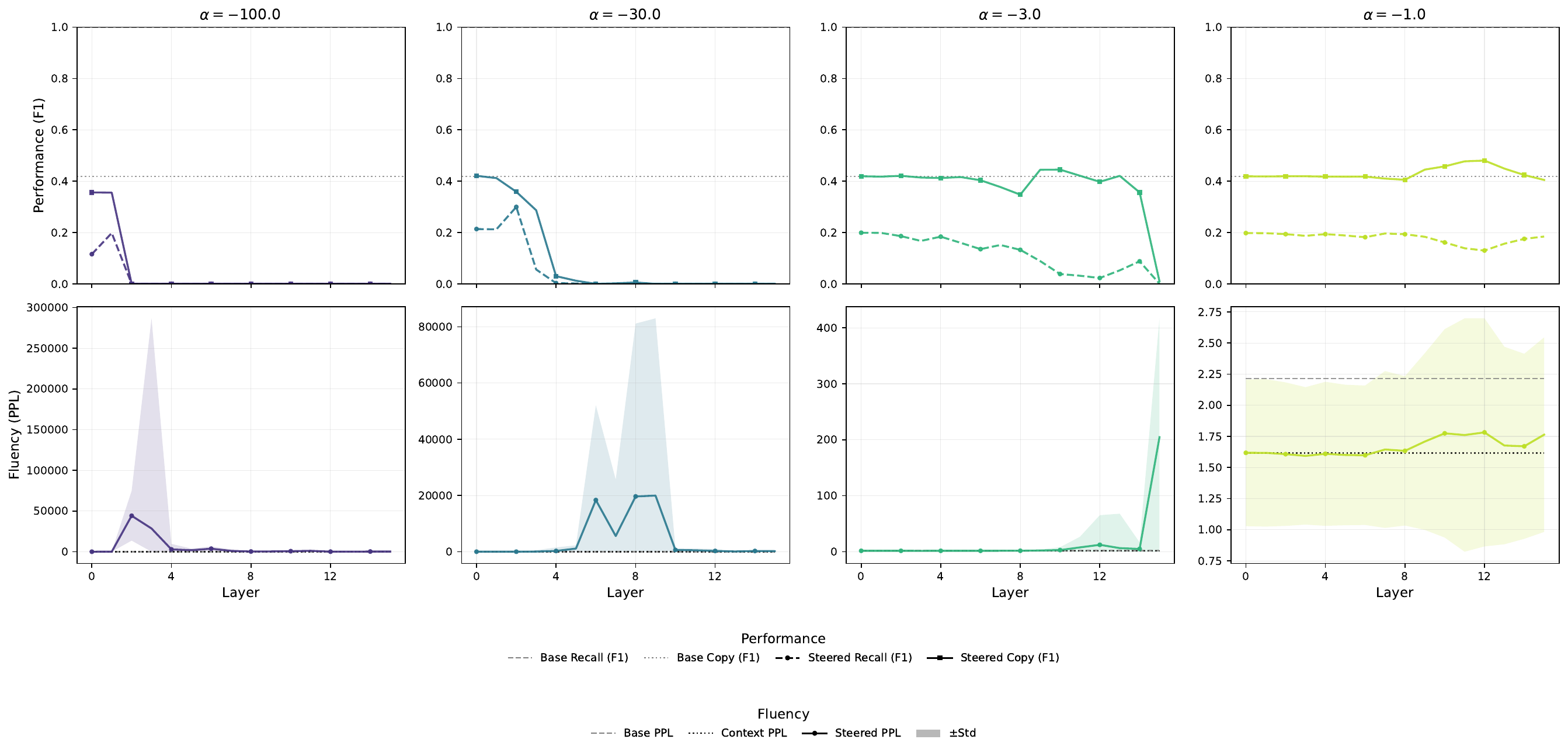}
\caption{Layer-wise effects of steering across four $\alpha$. Rows show performance (F1) and fluency (PPL). Columns correspond to $\alpha$.}
\label{fig:ov-perf-prob-flu-popqa-olmo2-rc-cf-lt}
\end{figure}

\subsection{PEQ: Object Token (Query-First)}

\subsubsection{Parametric Steering (Copy$\rightarrow$Recall) - Gemma2-2B}
\begin{figure}[H]
\centering
\includegraphics[width=\textwidth]{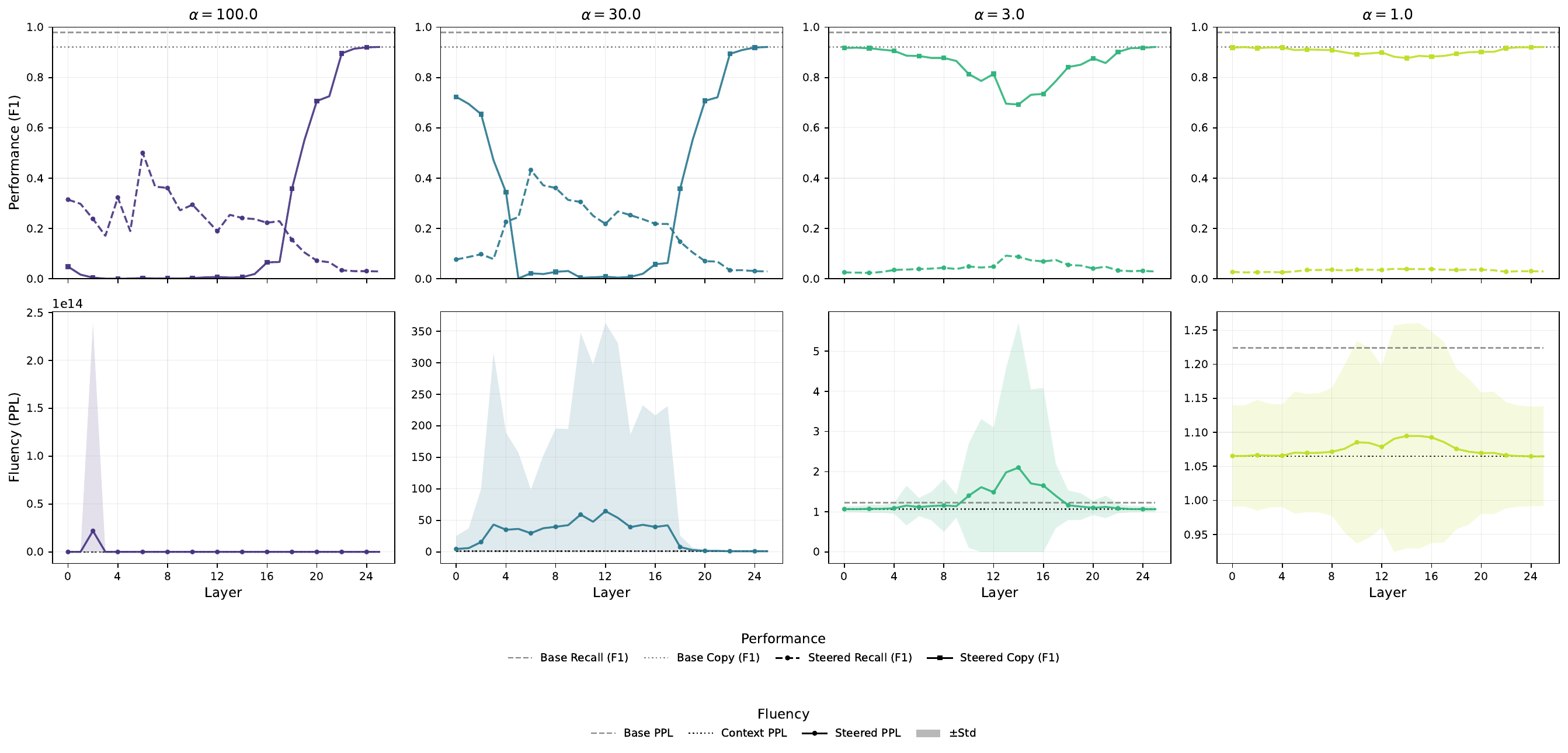}
\caption{Layer-wise effects of steering across four $\alpha$. Rows show performance (F1) and fluency (PPL). Columns correspond to $\alpha$.}
\label{fig:ov-perf-prob-flu-peq-gemma2-cr-qf-obj}
\end{figure}

\subsubsection{Contextual Steering (Recall$\rightarrow$Copy) - Gemma2-2B}
\begin{figure}[H]
\centering
\includegraphics[width=\textwidth]{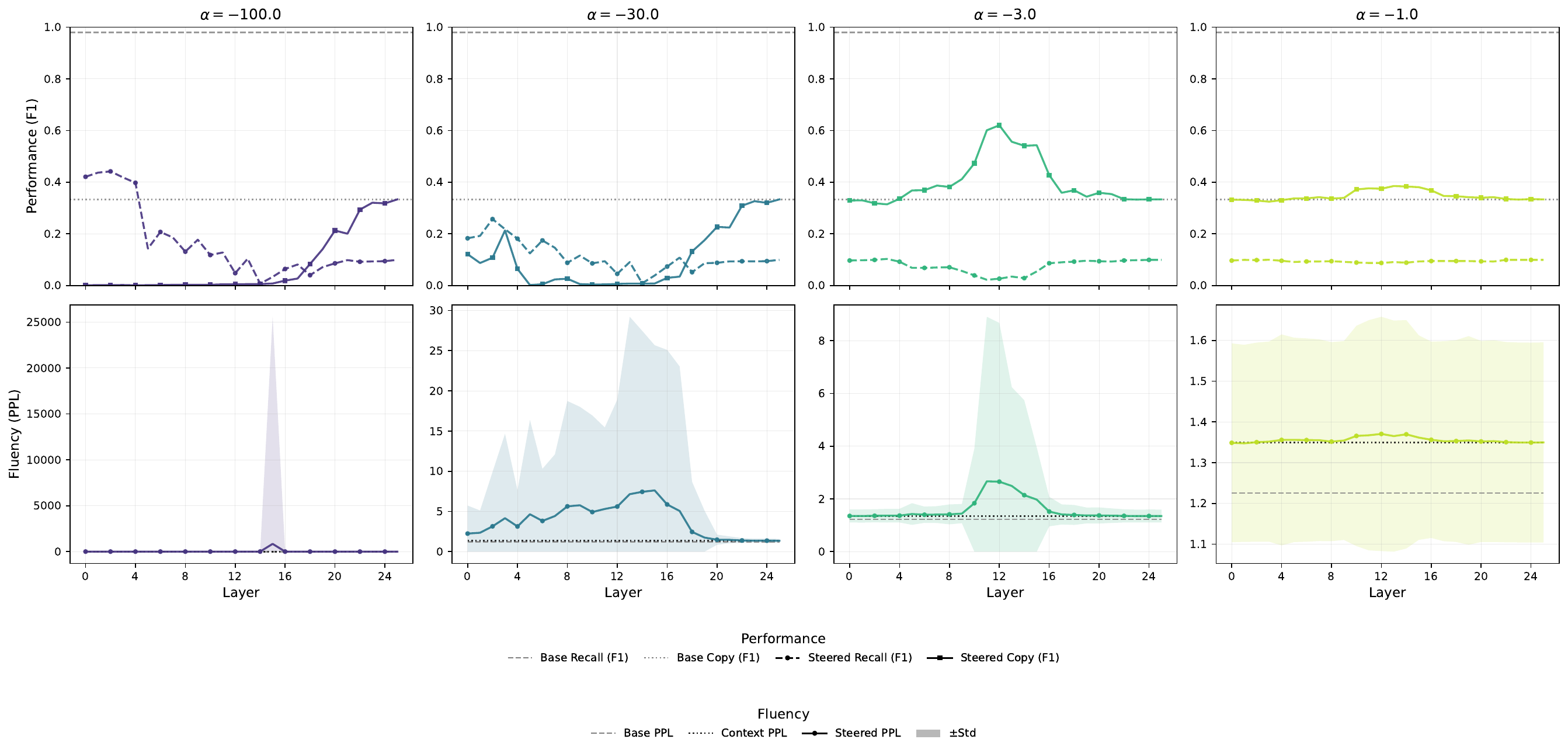}
\caption{Layer-wise effects of steering across four $\alpha$. Rows show performance (F1) and fluency (PPL). Columns correspond to $\alpha$.}
\label{fig:ov-perf-prob-flu-peq-gemma2-rc-qf-obj}
\end{figure}

\subsubsection{Parametric Steering (Copy$\rightarrow$Recall) - T5Gemma-2B}
\begin{figure}[H]
\centering
\includegraphics[width=\textwidth]{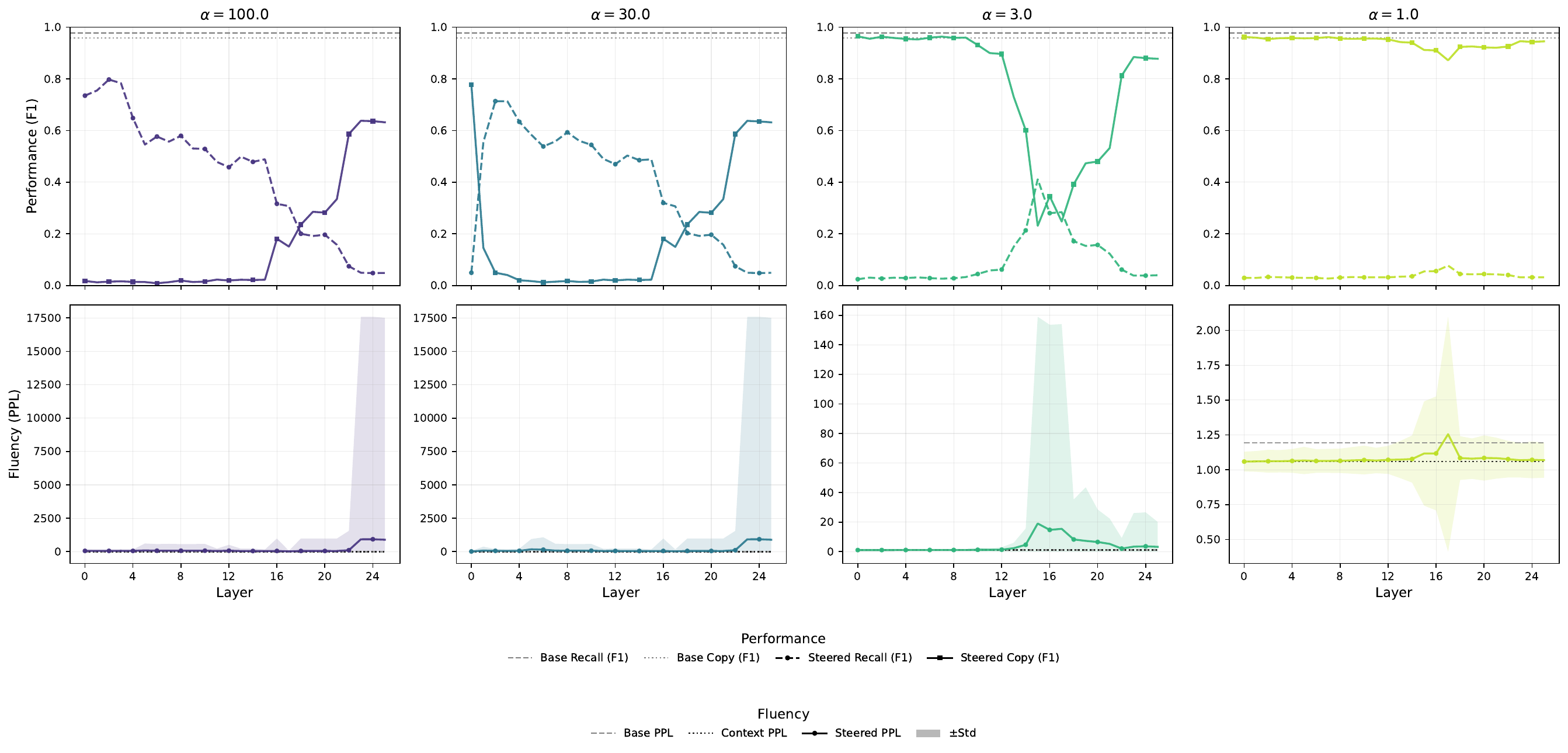}
\caption{Layer-wise effects of steering across four $\alpha$. Rows show performance (F1) and fluency (PPL). Columns correspond to $\alpha$.}
\label{fig:ov-perf-prob-flu-peq-t5gemma-cr-qf-obj}
\end{figure}

\subsubsection{Contextual Steering (Recall$\rightarrow$Copy) - T5Gemma-2B}
\begin{figure}[H]
\centering
\includegraphics[width=\textwidth]{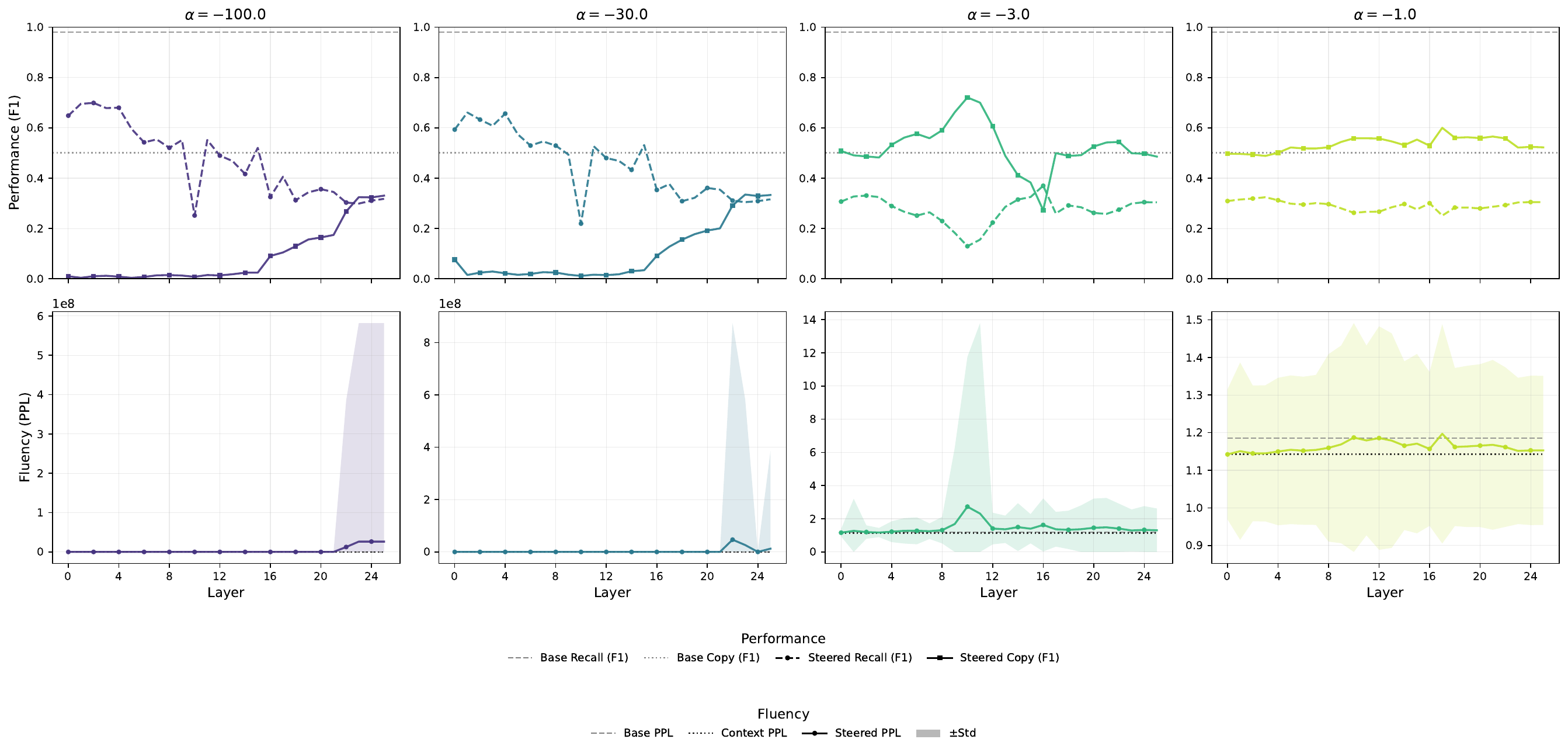}
\caption{Layer-wise effects of steering across four $\alpha$. Rows show performance (F1) and fluency (PPL). Columns correspond to $\alpha$.}
\label{fig:ov-perf-prob-flu-peq-t5gemma-rc-qf-obj}
\end{figure}

\subsubsection{Parametric Steering (Copy$\rightarrow$Recall) - OLMo-2-0425-1B}
\begin{figure}[H]
\centering
\includegraphics[width=\textwidth]{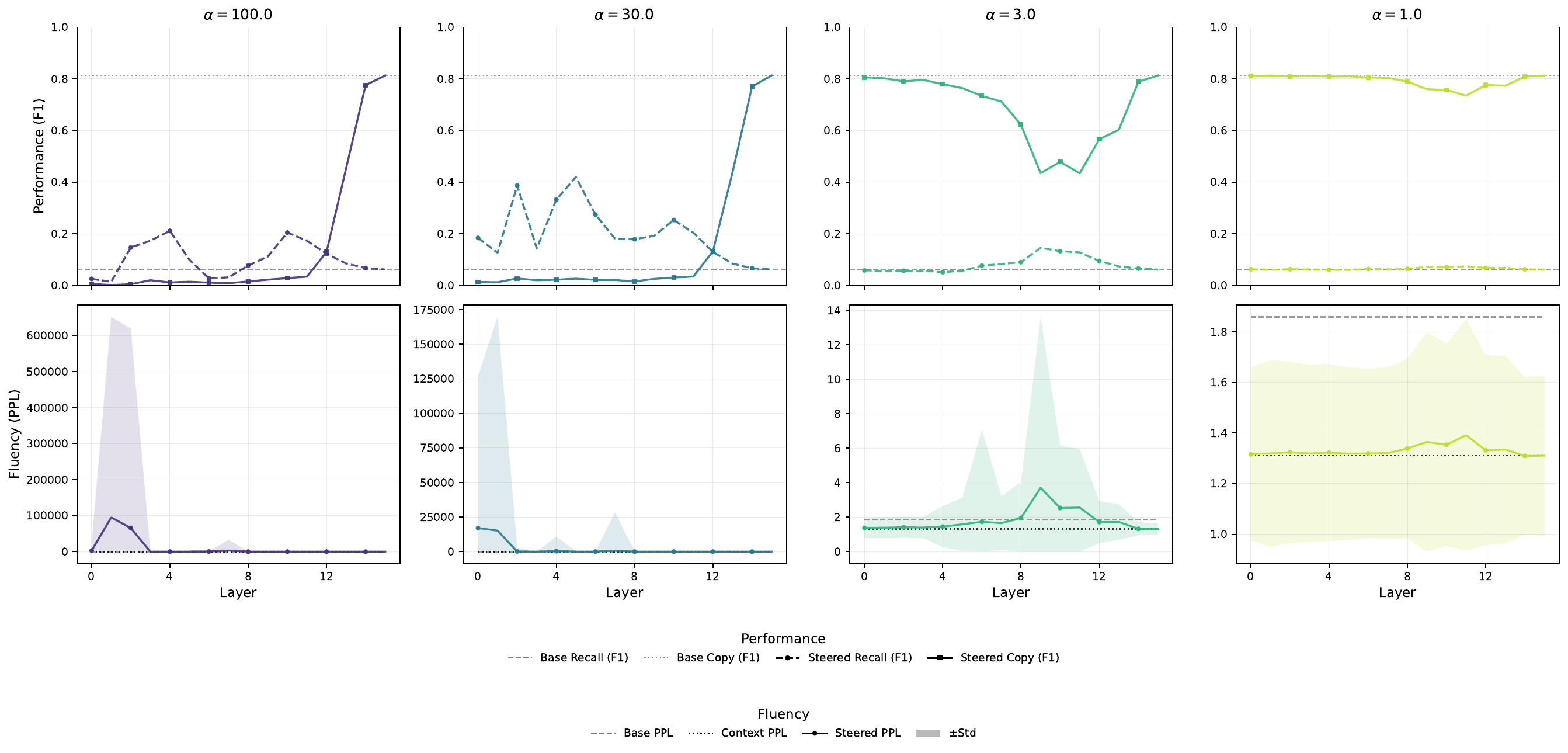}
\caption{Layer-wise effects of steering across four $\alpha$. Rows show performance (F1) and fluency (PPL). Columns correspond to $\alpha$.}
\label{fig:ov-perf-prob-flu-peq-olmo2-cr-qf-obj}
\end{figure}

\subsubsection{Contextual Steering (Recall$\rightarrow$Copy) - OLMo-2-0425-1B}
\begin{figure}[H]
\centering
\includegraphics[width=\textwidth]{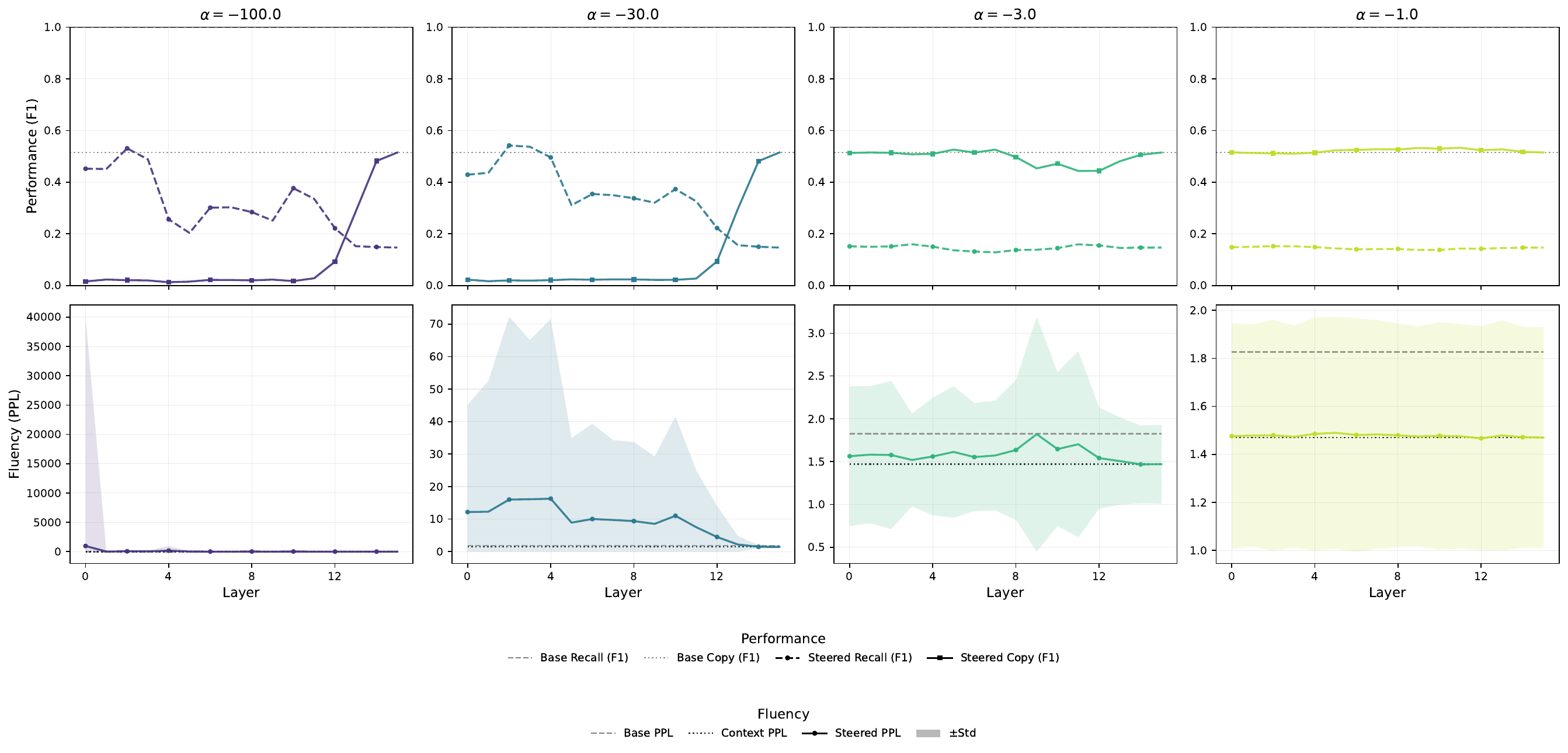}
\caption{Layer-wise effects of steering across four $\alpha$. Rows show performance (F1) and fluency (PPL). Columns correspond to $\alpha$.}
\label{fig:ov-perf-prob-flu-peq-olmo2-rc-qf-obj}
\end{figure}

\subsection{PEQ: Object Token (Context-First)}

\subsubsection{Parametric Steering (Copy$\rightarrow$Recall) - Gemma2-2B}
\begin{figure}[H]
\centering
\includegraphics[width=\textwidth]{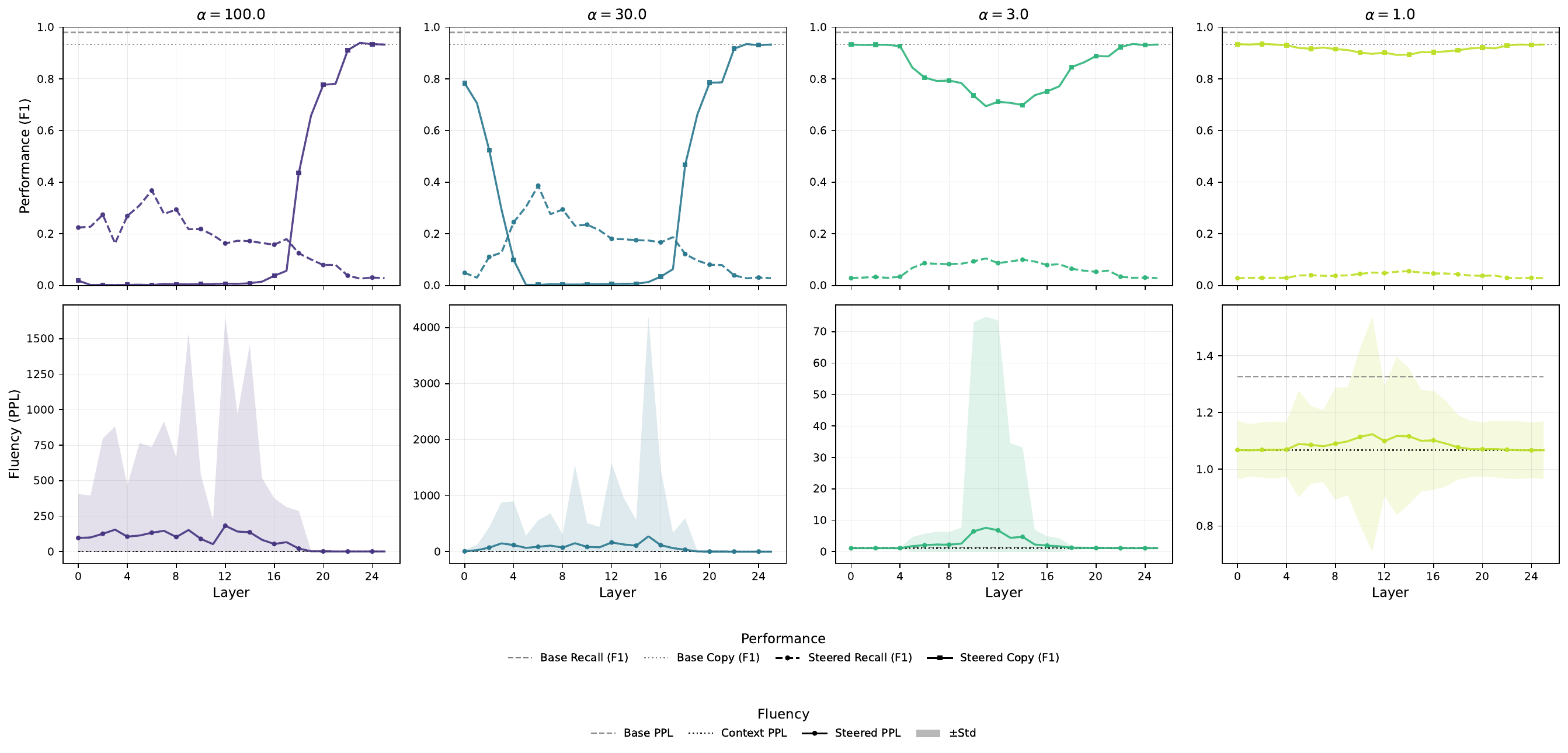}
\caption{Layer-wise effects of steering across four $\alpha$. Rows show performance (F1) and fluency (PPL). Columns correspond to $\alpha$.}
\label{fig:ov-perf-prob-flu-peq-gemma2-cr-cf-obj}
\end{figure}

\subsubsection{Contextual Steering (Recall$\rightarrow$Copy) - Gemma2-2B}
\begin{figure}[H]
\centering
\includegraphics[width=\textwidth]{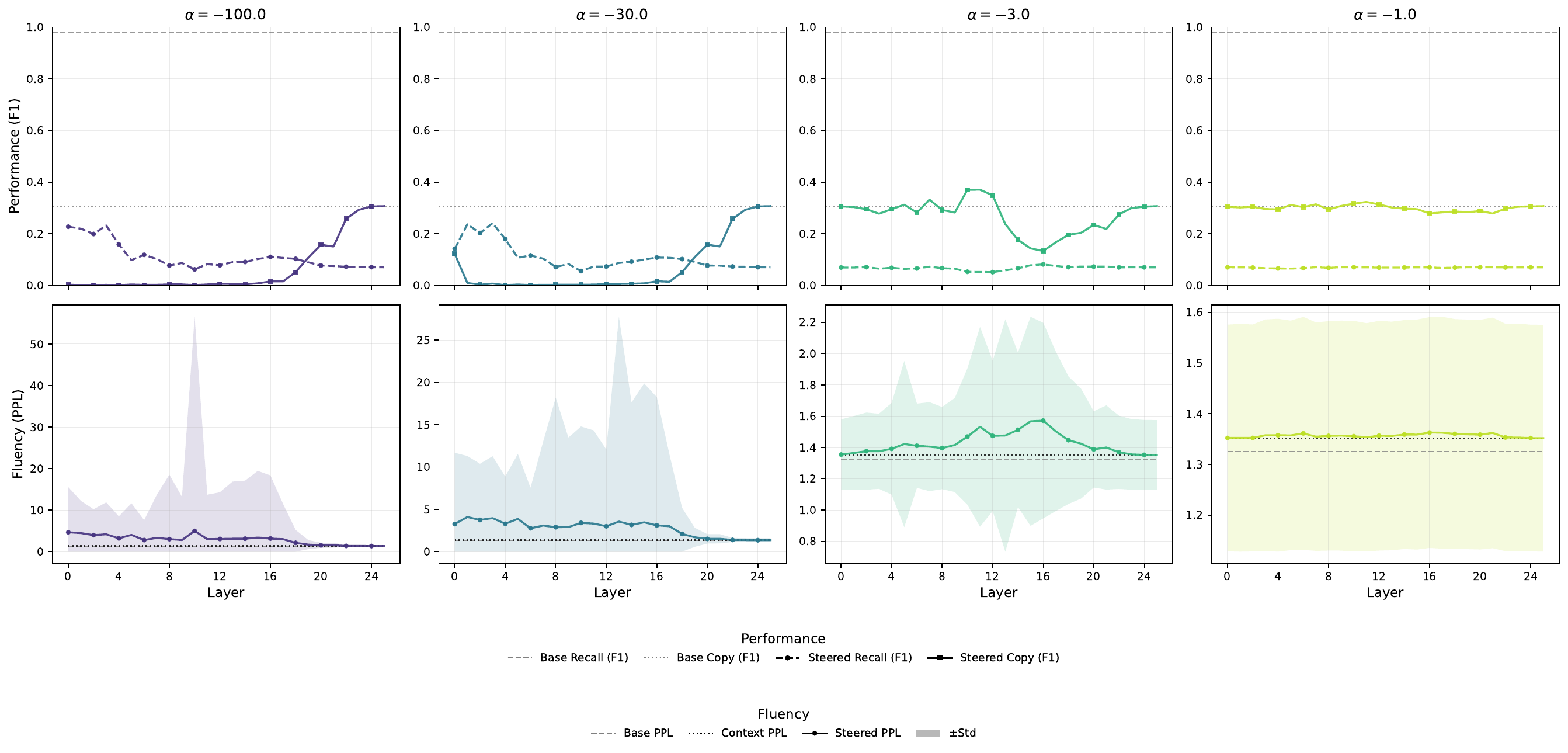}
\caption{Layer-wise effects of steering across four $\alpha$. Rows show performance (F1) and fluency (PPL). Columns correspond to $\alpha$.}
\label{fig:ov-perf-prob-flu-peq-gemma2-rc-cf-obj}
\end{figure}

\subsubsection{Parametric Steering (Copy$\rightarrow$Recall) - T5Gemma-2B}
\begin{figure}[H]
\centering
\includegraphics[width=\textwidth]{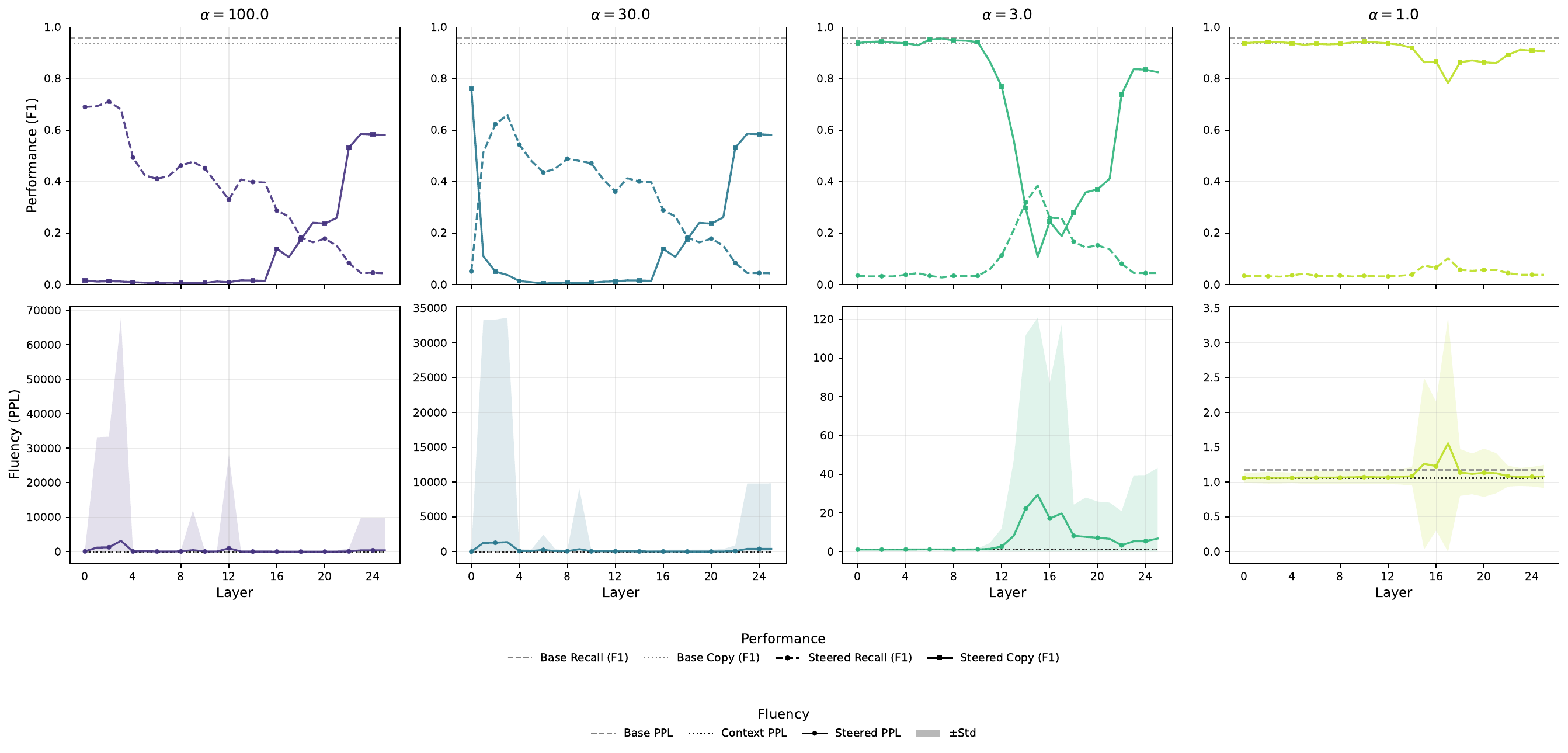}
\caption{Layer-wise effects of steering across four $\alpha$. Rows show performance (F1) and fluency (PPL). Columns correspond to $\alpha$.}
\label{fig:ov-perf-prob-flu-peq-t5gemma-cr-cf-obj}
\end{figure}

\subsubsection{Contextual Steering (Recall$\rightarrow$Copy) - T5Gemma-2B}
\begin{figure}[H]
\centering
\includegraphics[width=\textwidth]{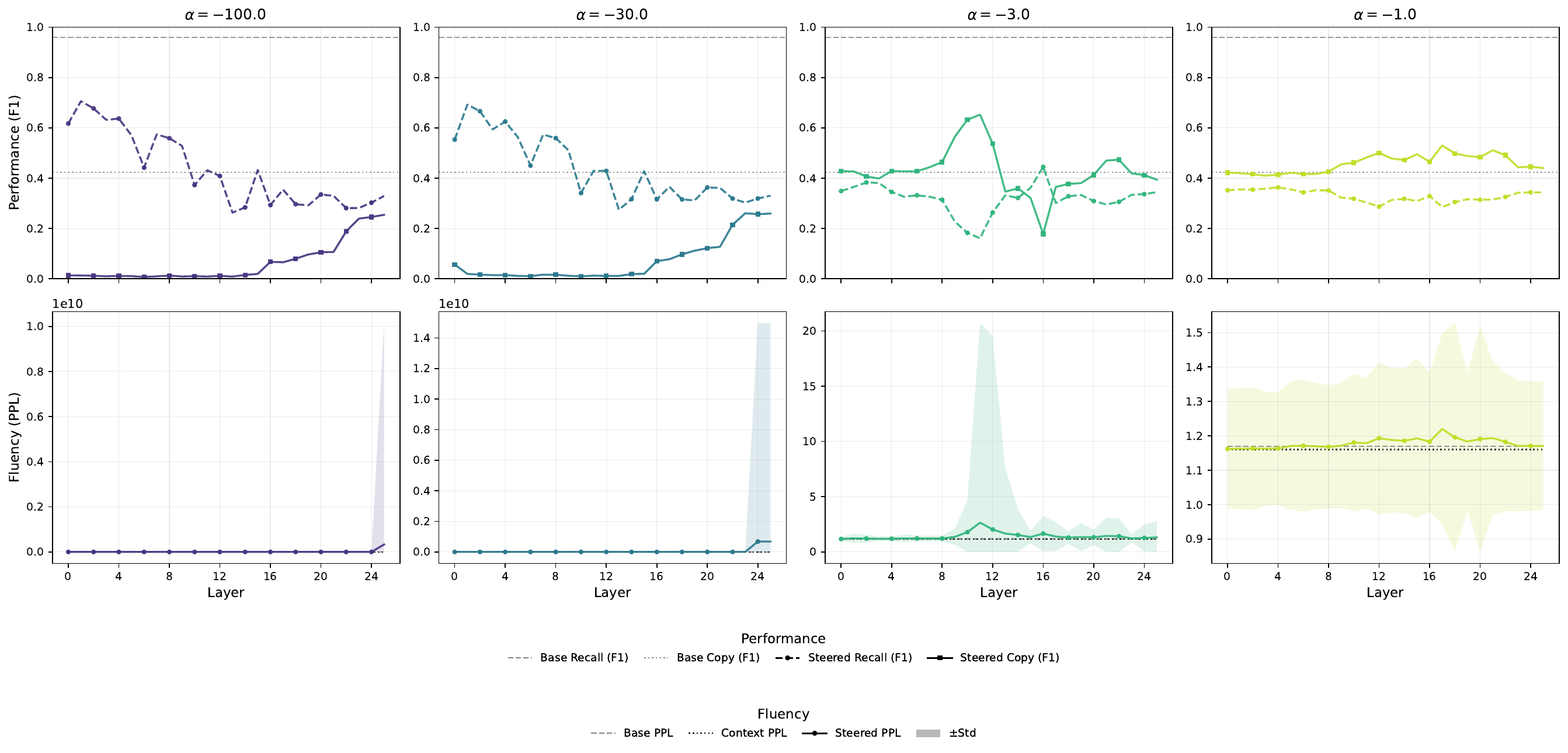}
\caption{Layer-wise effects of steering across four $\alpha$. Rows show performance (F1) and fluency (PPL). Columns correspond to $\alpha$.}
\label{fig:ov-perf-prob-flu-peq-t5gemma-rc-cf-obj}
\end{figure}

\subsubsection{Parametric Steering (Copy$\rightarrow$Recall) - OLMo-2-0425-1B}
\begin{figure}[H]
\centering
\includegraphics[width=\textwidth]{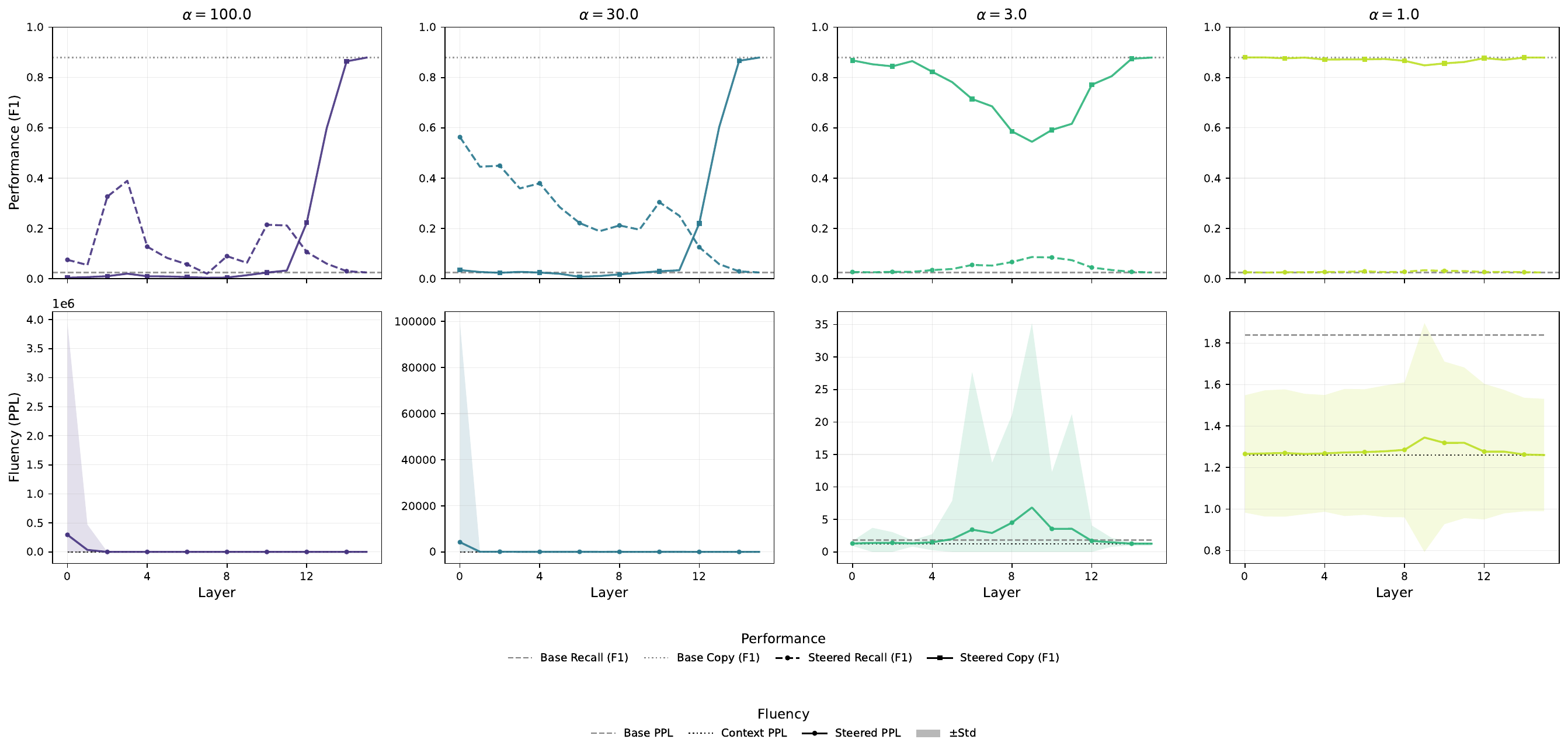}
\caption{Layer-wise effects of steering across four $\alpha$. Rows show performance (F1) and fluency (PPL). Columns correspond to $\alpha$.}
\label{fig:ov-perf-prob-flu-peq-olmo2-cr-cf-obj}
\end{figure}

\subsubsection{Contextual Steering (Recall$\rightarrow$Copy) - OLMo-2-0425-1B}
\begin{figure}[H]
\centering
\includegraphics[width=\textwidth]{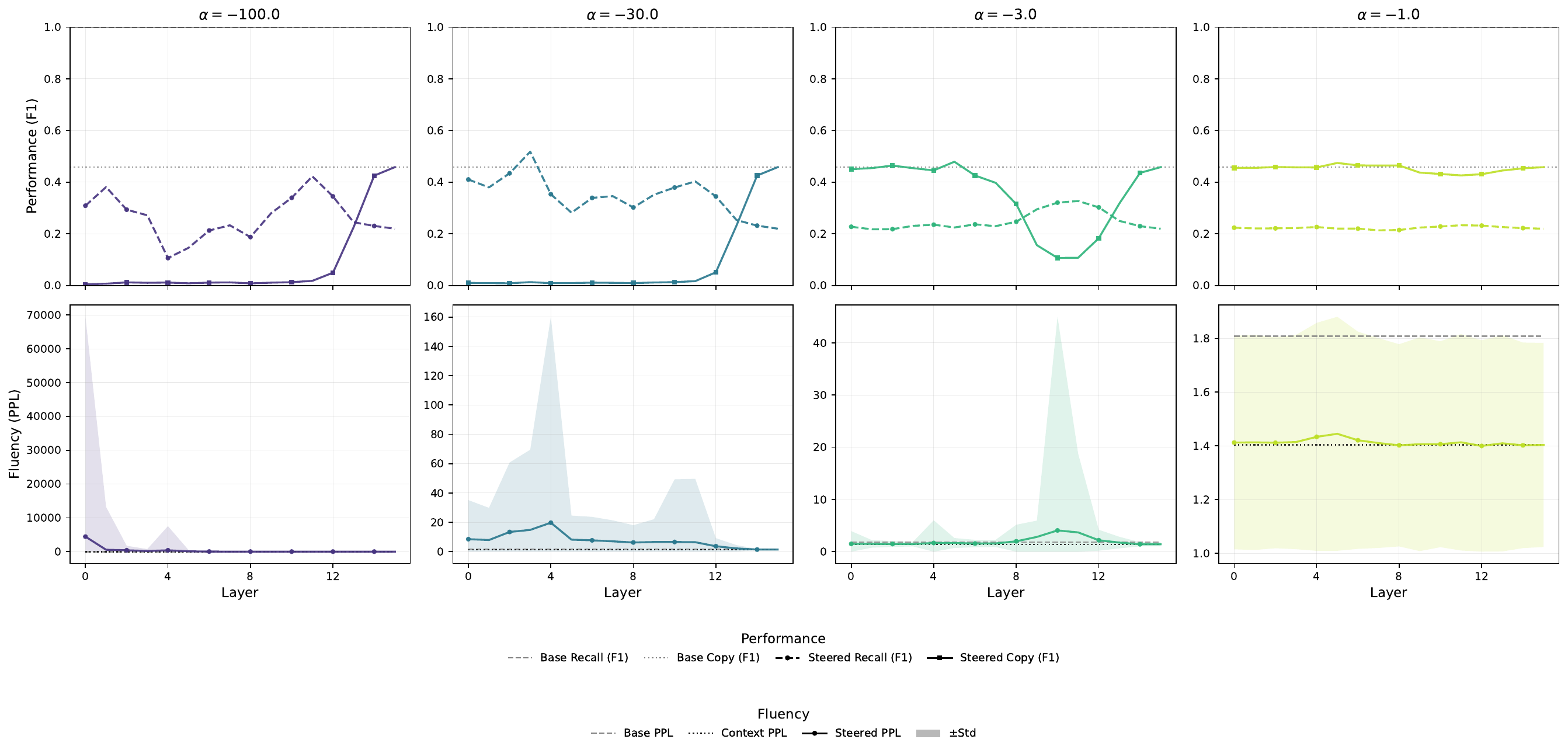}
\caption{Layer-wise effects of steering across four $\alpha$. Rows show performance (F1) and fluency (PPL). Columns correspond to $\alpha$.}
\label{fig:ov-perf-prob-flu-peq-olmo2-rc-cf-obj}
\end{figure}

\subsection{PEQ: Subject Token (Context-First)}

\subsubsection{Parametric Steering (Copy$\rightarrow$Recall) - Gemma2-2B}
\begin{figure}[H]
\centering
\includegraphics[width=\textwidth]{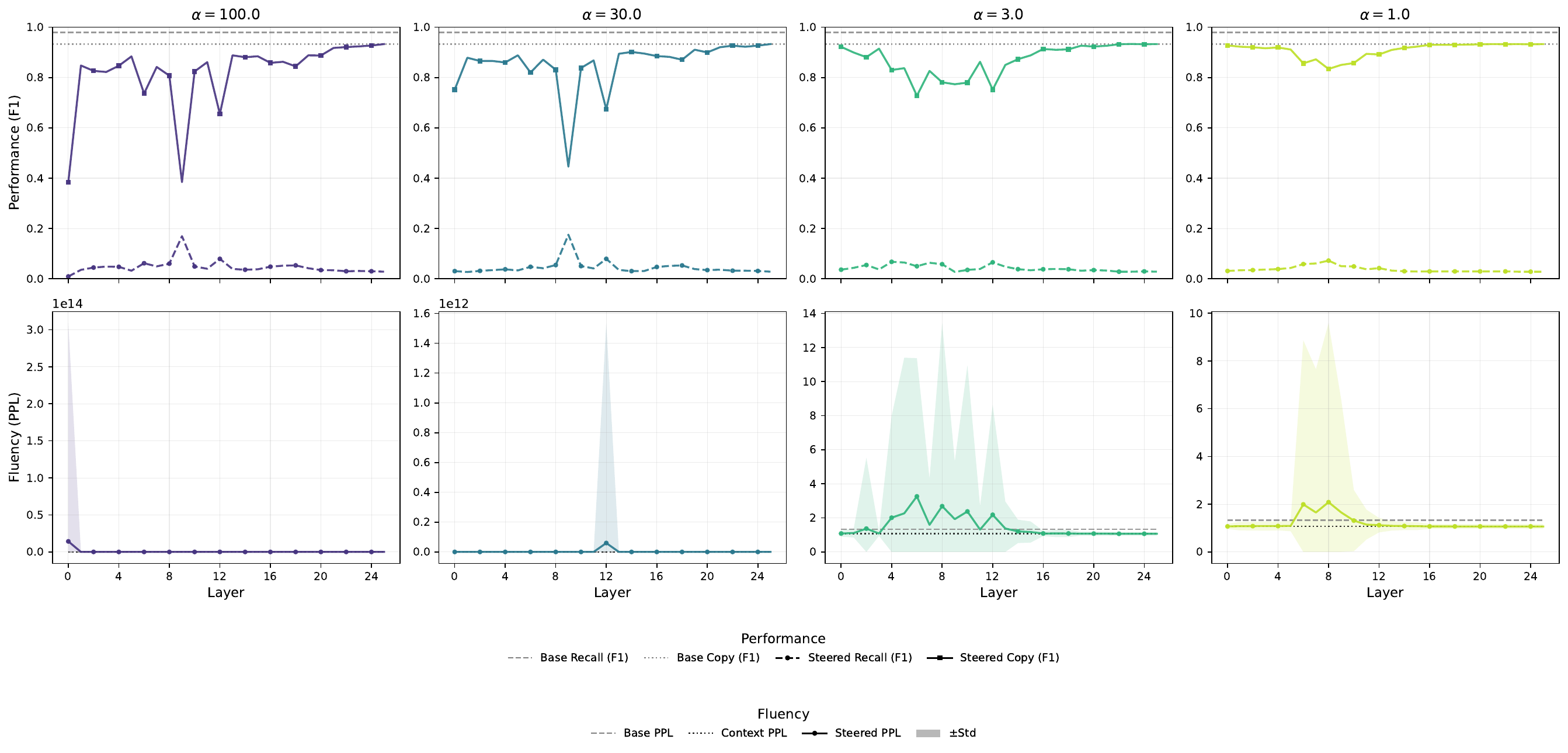}
\caption{Layer-wise effects of steering across four $\alpha$. Rows show performance (F1) and fluency (PPL). Columns correspond to $\alpha$.}
\label{fig:ov-perf-prob-flu-peq-gemma2-cr-cf-subj}
\end{figure}

\subsubsection{Contextual Steering (Recall$\rightarrow$Copy) - Gemma2-2B}
\begin{figure}[H]
\centering
\includegraphics[width=\textwidth]{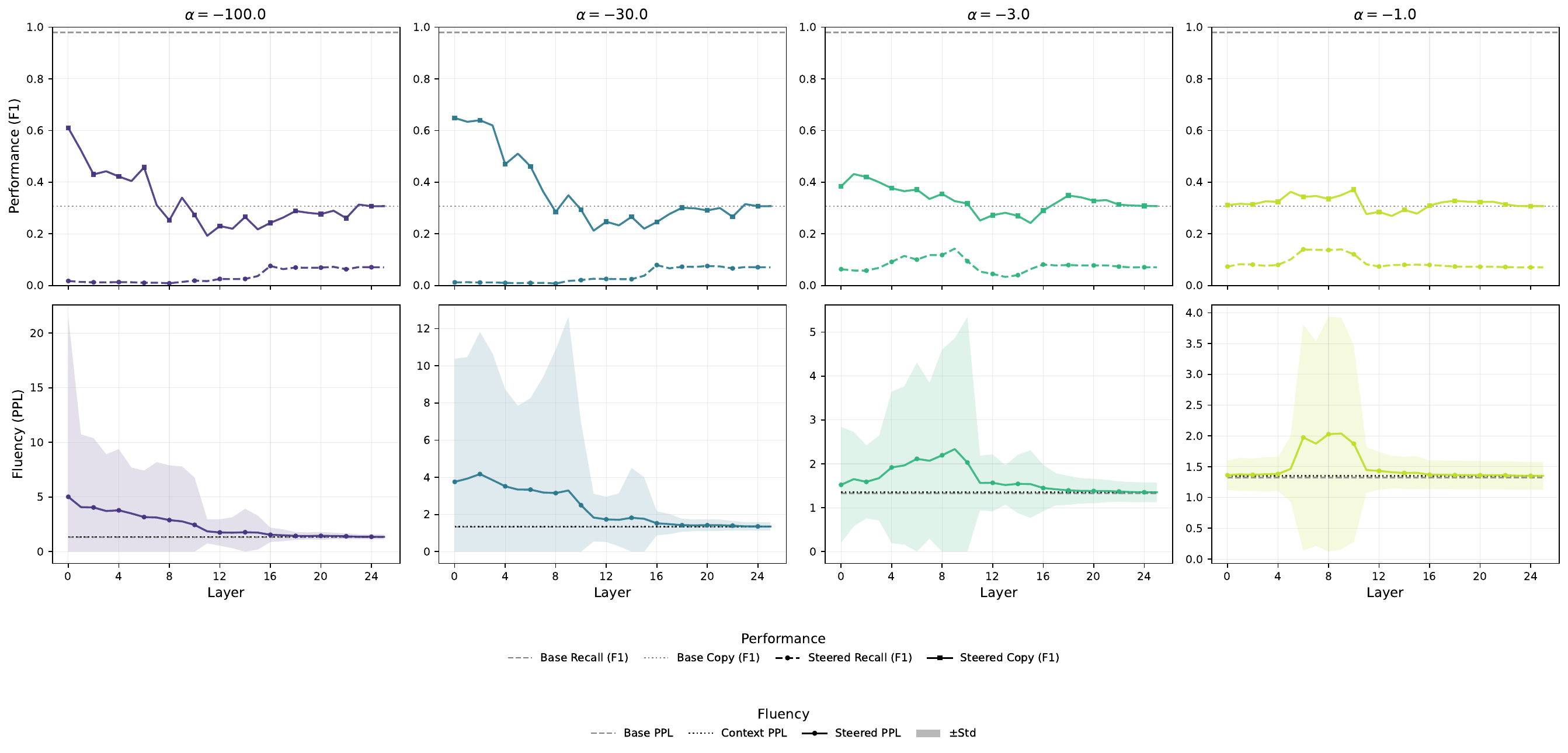}
\caption{Layer-wise effects of steering across four $\alpha$. Rows show performance (F1) and fluency (PPL). Columns correspond to $\alpha$.}
\label{fig:ov-perf-prob-flu-peq-gemma2-rc-cf-subj}
\end{figure}

\subsubsection{Parametric Steering (Copy$\rightarrow$Recall) - T5Gemma-2B}
\begin{figure}[H]
\centering
\includegraphics[width=\textwidth]{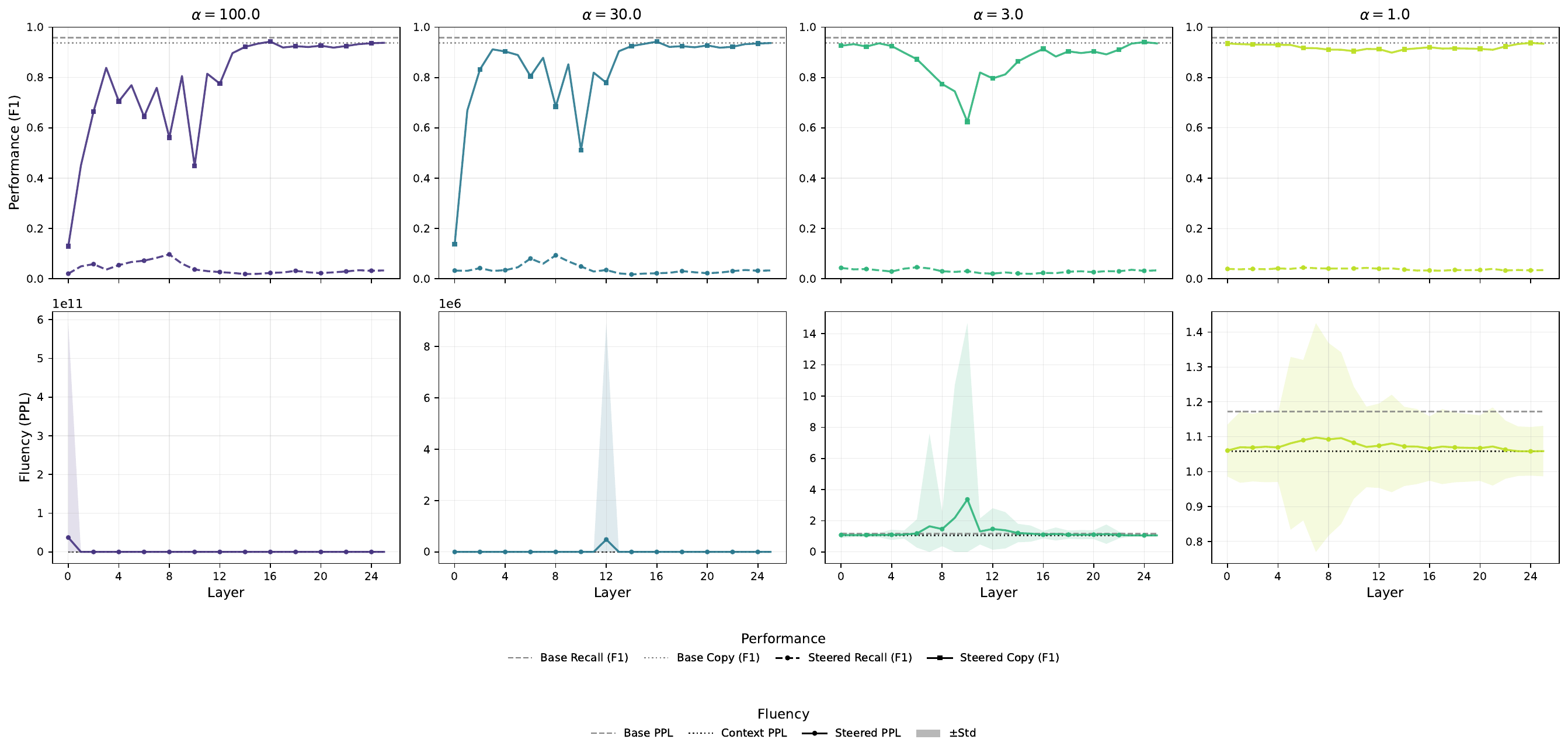}
\caption{Layer-wise effects of steering across four $\alpha$. Rows show performance (F1) and fluency (PPL). Columns correspond to $\alpha$.}
\label{fig:ov-perf-prob-flu-peq-t5gemma-cr-cf-subj}
\end{figure}

\subsubsection{Contextual Steering (Recall$\rightarrow$Copy) - T5Gemma-2B}
\begin{figure}[H]
\centering
\includegraphics[width=\textwidth]{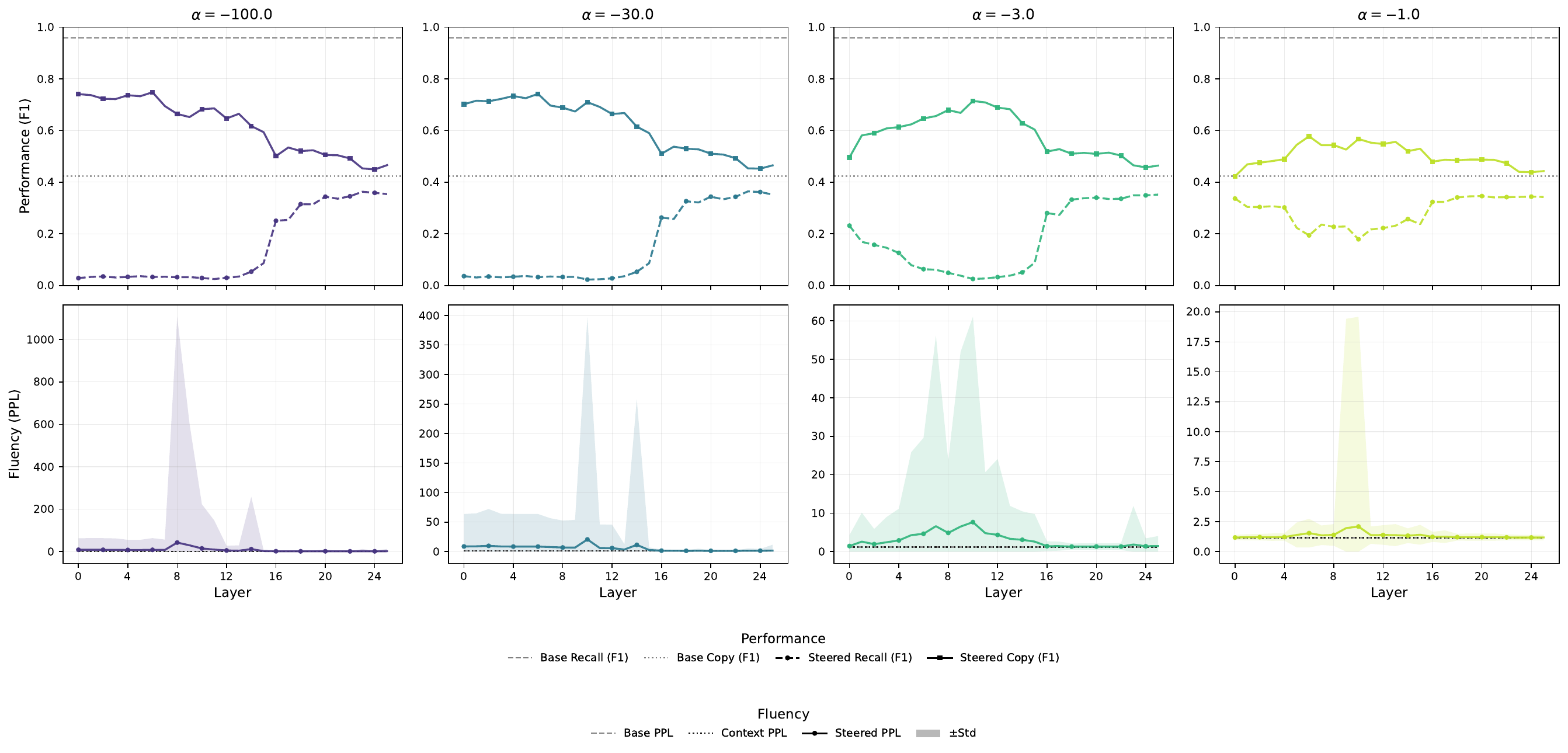}
\caption{Layer-wise effects of steering across four $\alpha$. Rows show performance (F1) and fluency (PPL). Columns correspond to $\alpha$.}
\label{fig:ov-perf-prob-flu-peq-t5gemma-rc-cf-subj}
\end{figure}

\subsubsection{Parametric Steering (Copy$\rightarrow$Recall) - OLMo-2-0425-1B}
\begin{figure}[H]
\centering
\includegraphics[width=\textwidth]{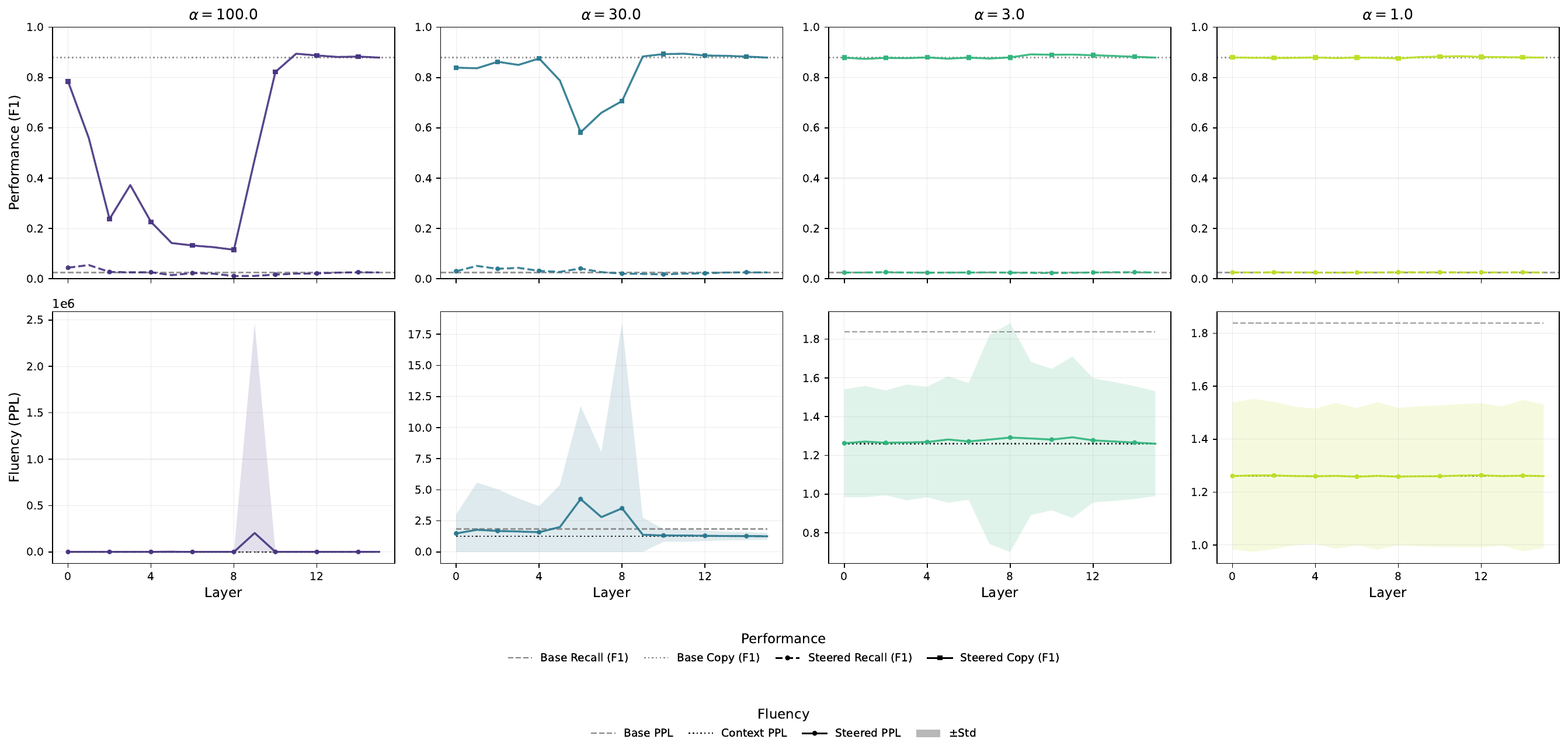}
\caption{Layer-wise effects of steering across four $\alpha$. Rows show performance (F1) and fluency (PPL). Columns correspond to $\alpha$.}
\label{fig:ov-perf-prob-flu-peq-olmo2-cr-cf-subj}
\end{figure}

\subsubsection{Contextual Steering (Recall$\rightarrow$Copy) - OLMo-2-0425-1B}
\begin{figure}[H]
\centering
\includegraphics[width=\textwidth]{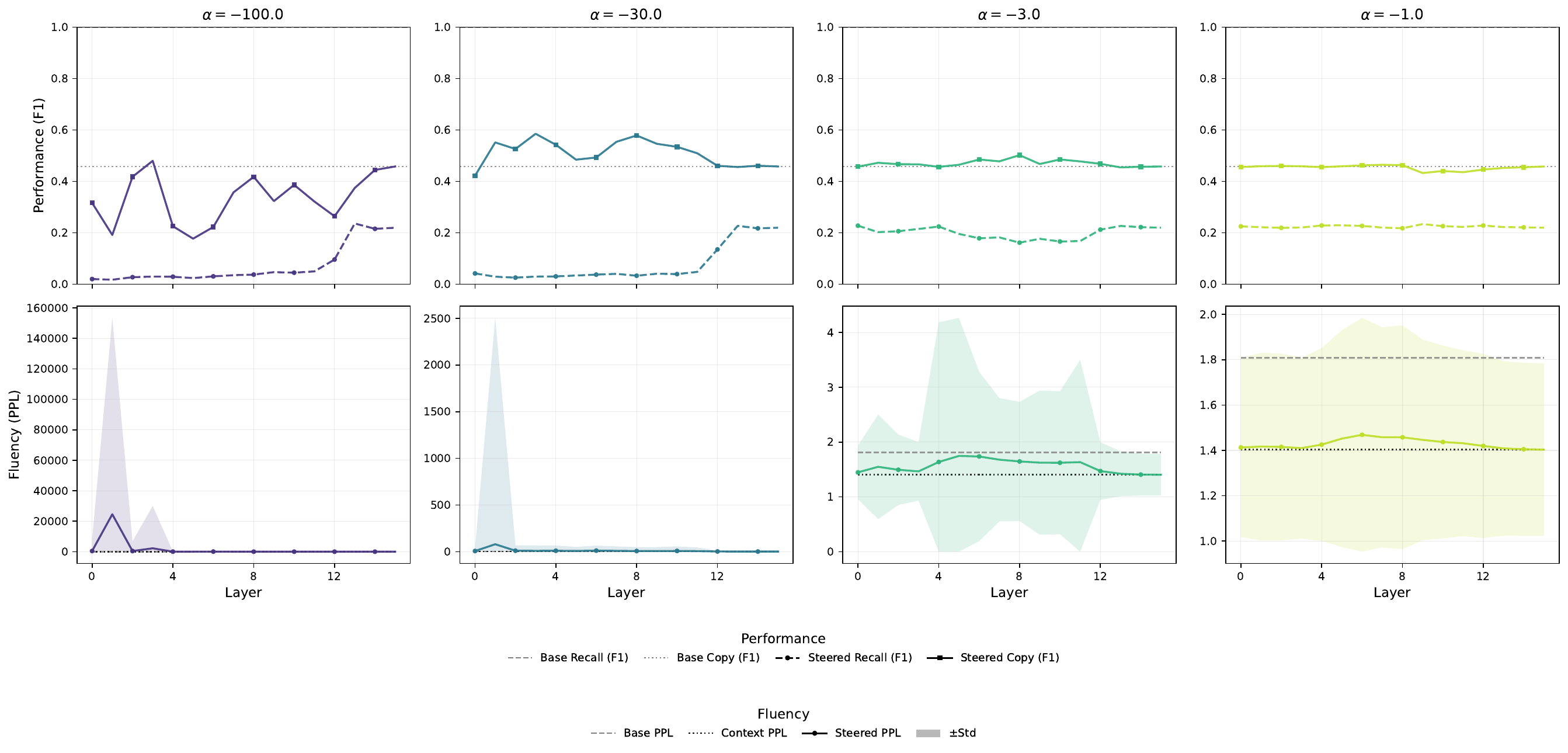}
\caption{Layer-wise effects of steering across four $\alpha$. Rows show performance (F1) and fluency (PPL). Columns correspond to $\alpha$.}
\label{fig:ov-perf-prob-flu-peq-olmo2-rc-cf-subj}
\end{figure}

\subsection{PEQ: Last Token (Query-First)}

\subsubsection{Parametric Steering (Copy$\rightarrow$Recall) - Gemma2-2B}
\begin{figure}[H]
\centering
\includegraphics[width=\textwidth]{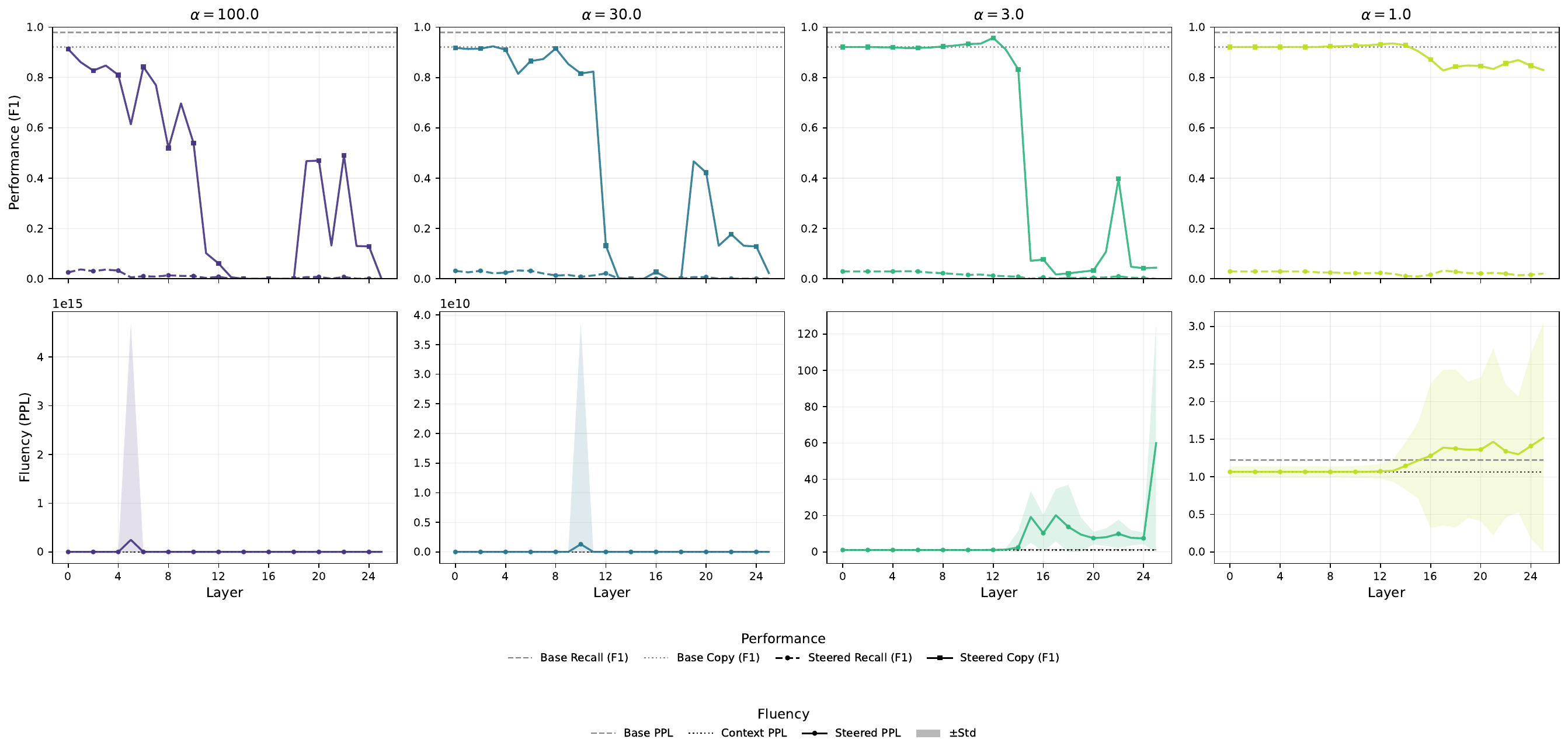}
\caption{Layer-wise effects of steering across four $\alpha$. Rows show performance (F1) and fluency (PPL). Columns correspond to $\alpha$.}
\label{fig:ov-perf-prob-flu-peq-gemma2-cr-qf-lt}
\end{figure}

\subsubsection{Contextual Steering (Recall$\rightarrow$Copy) - Gemma2-2B}
\begin{figure}[H]
\centering
\includegraphics[width=\textwidth]{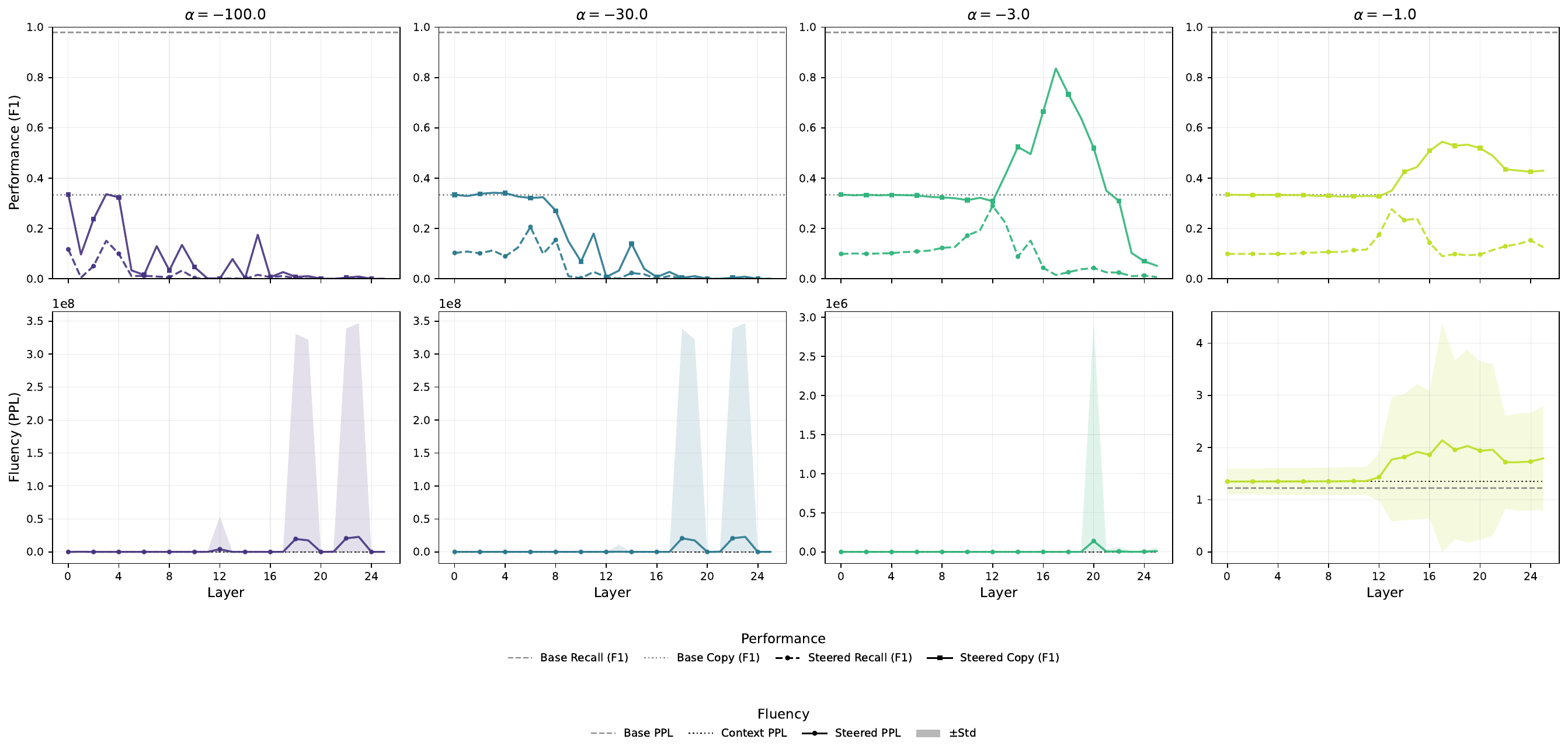}
\caption{Layer-wise effects of steering across four $\alpha$. Rows show performance (F1) and fluency (PPL). Columns correspond to $\alpha$.}
\label{fig:ov-perf-prob-flu-peq-gemma2-rc-qf-lt}
\end{figure}

\subsubsection{Parametric Steering (Copy$\rightarrow$Recall) - T5Gemma-2B}
\begin{figure}[H]
\centering
\includegraphics[width=\textwidth]{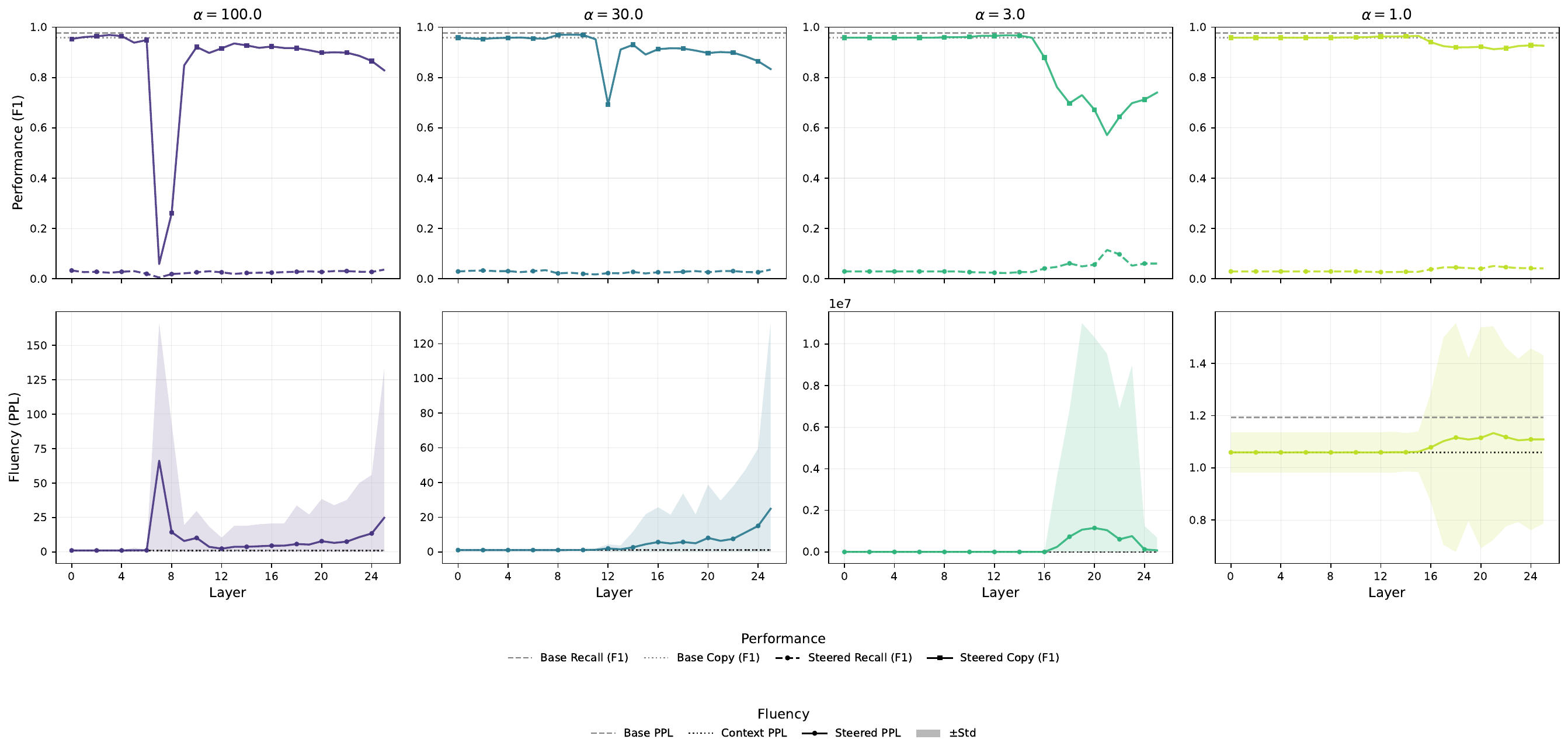}
\caption{Layer-wise effects of steering across four $\alpha$. Rows show performance (F1) and fluency (PPL). Columns correspond to $\alpha$.}
\label{fig:ov-perf-prob-flu-peq-t5gemma-cr-qf-lt}
\end{figure}

\subsubsection{Contextual Steering (Recall$\rightarrow$Copy) - T5Gemma-2B}
\begin{figure}[H]
\centering
\includegraphics[width=\textwidth]{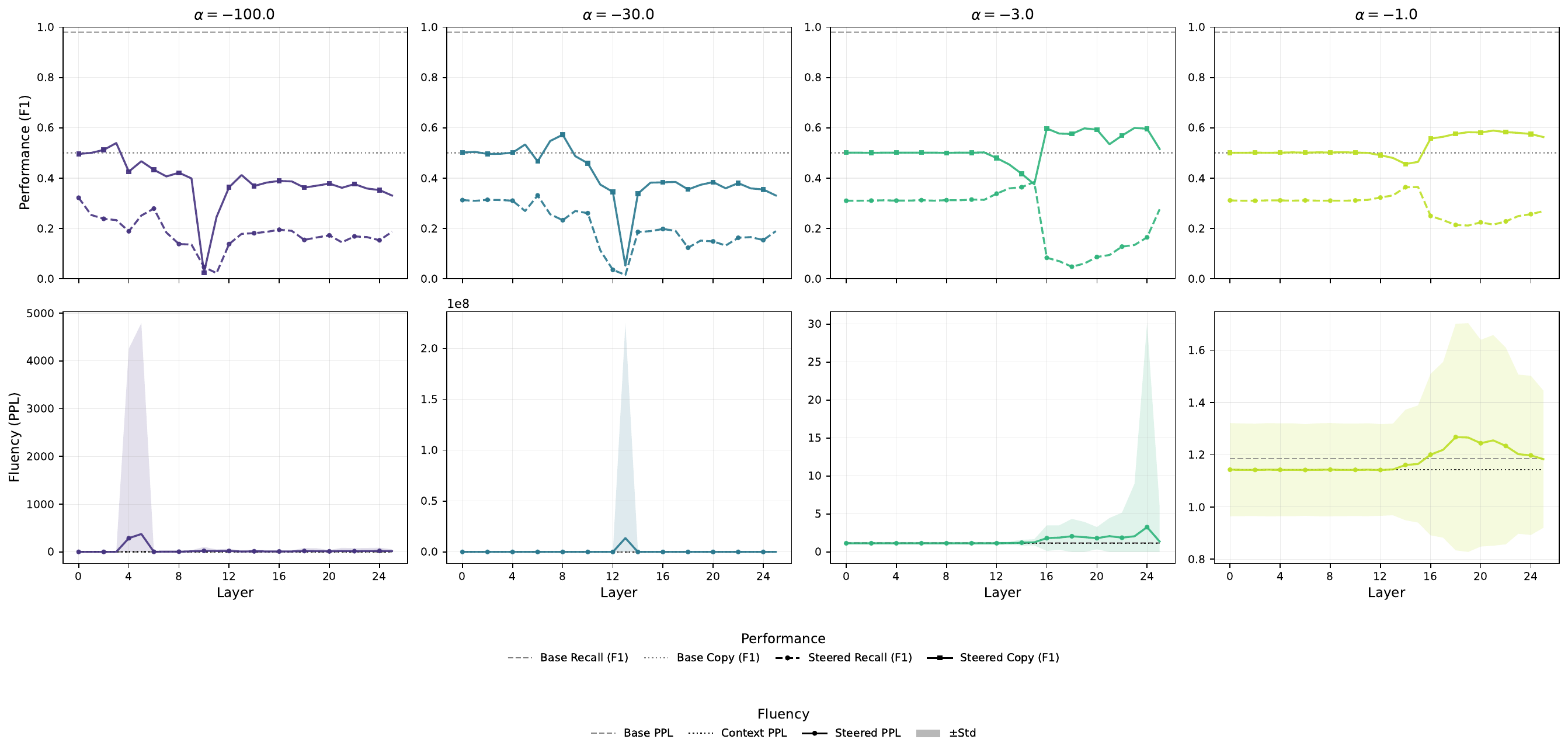}
\caption{Layer-wise effects of steering across four $\alpha$. Rows show performance (F1) and fluency (PPL). Columns correspond to $\alpha$.}
\label{fig:ov-perf-prob-flu-peq-t5gemma-rc-qf-lt}
\end{figure}

\subsubsection{Parametric Steering (Copy$\rightarrow$Recall) - OLMo-2-0425-1B}
\begin{figure}[H]
\centering
\includegraphics[width=\textwidth]{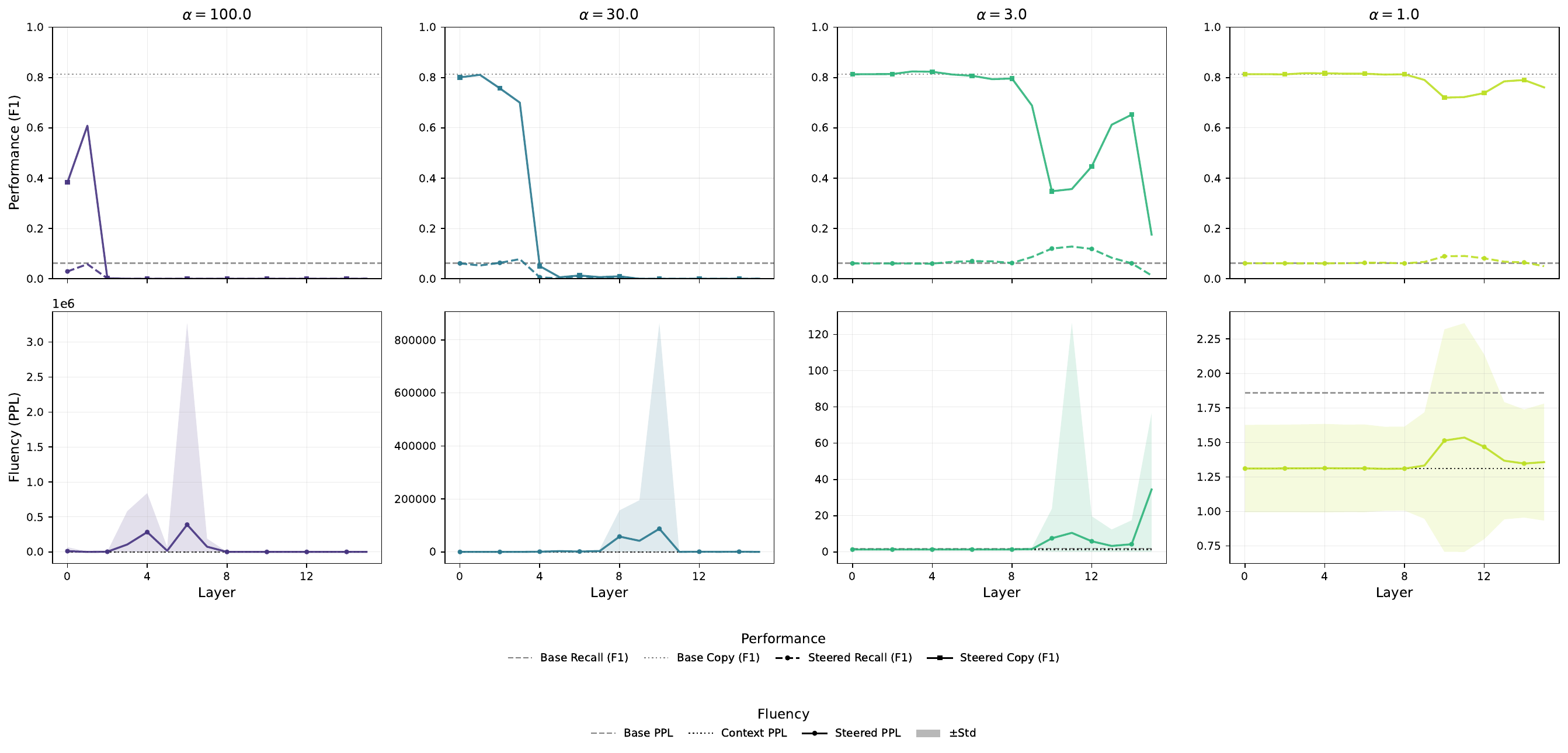}
\caption{Layer-wise effects of steering across four $\alpha$. Rows show performance (F1) and fluency (PPL). Columns correspond to $\alpha$.}
\label{fig:ov-perf-prob-flu-peq-olmo2-cr-qf-lt}
\end{figure}

\subsubsection{Contextual Steering (Recall$\rightarrow$Copy) - OLMo-2-0425-1B}
\begin{figure}[H]
\centering
\includegraphics[width=\textwidth]{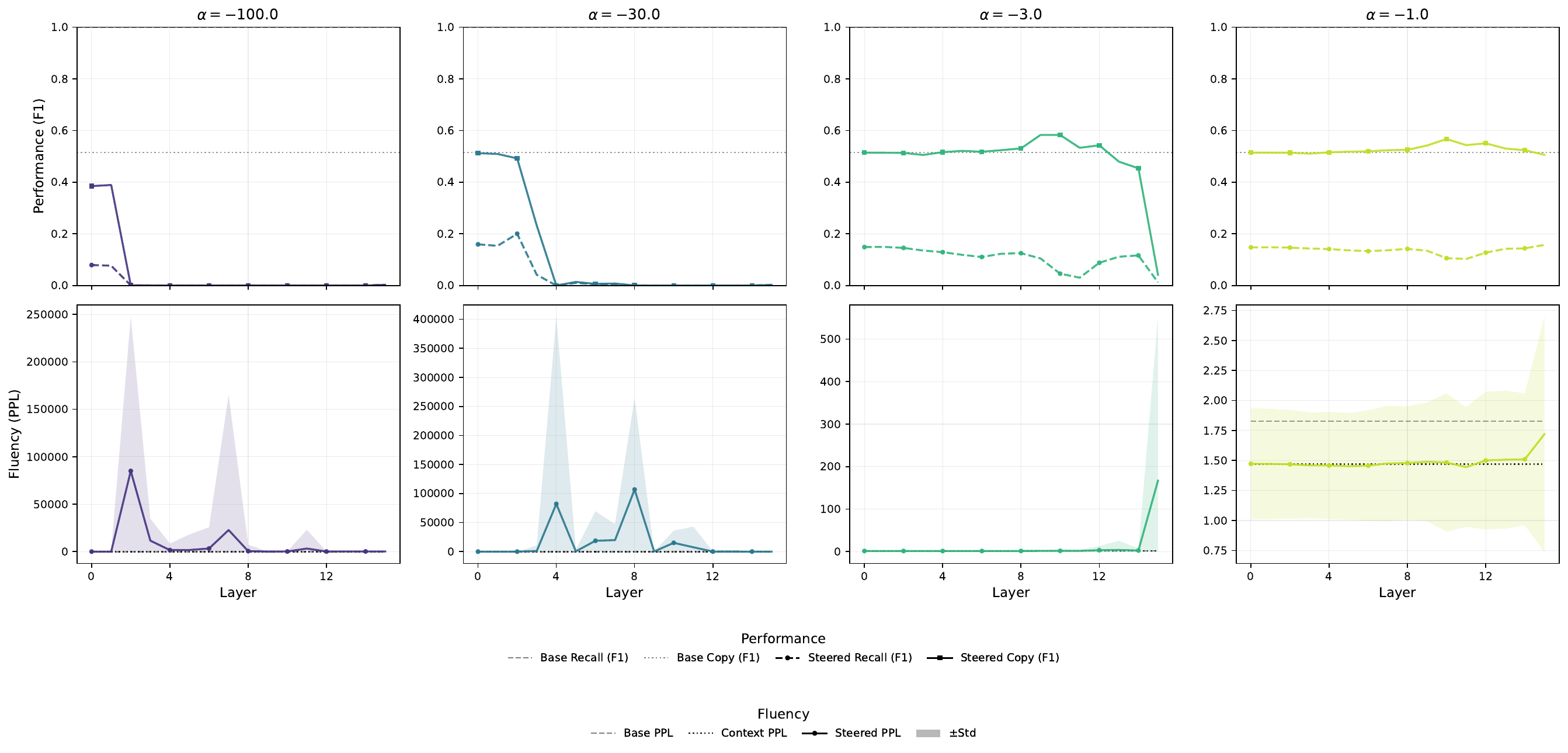}
\caption{Layer-wise effects of steering across four $\alpha$. Rows show performance (F1) and fluency (PPL). Columns correspond to $\alpha$.}
\label{fig:ov-perf-prob-flu-peq-olmo2-rc-qf-lt}
\end{figure}

\subsection{PEQ: Last Token (Context-First)}

\subsubsection{Parametric Steering (Copy$\rightarrow$Recall) - Gemma2-2B}
\begin{figure}[H]
\centering
\includegraphics[width=\textwidth]{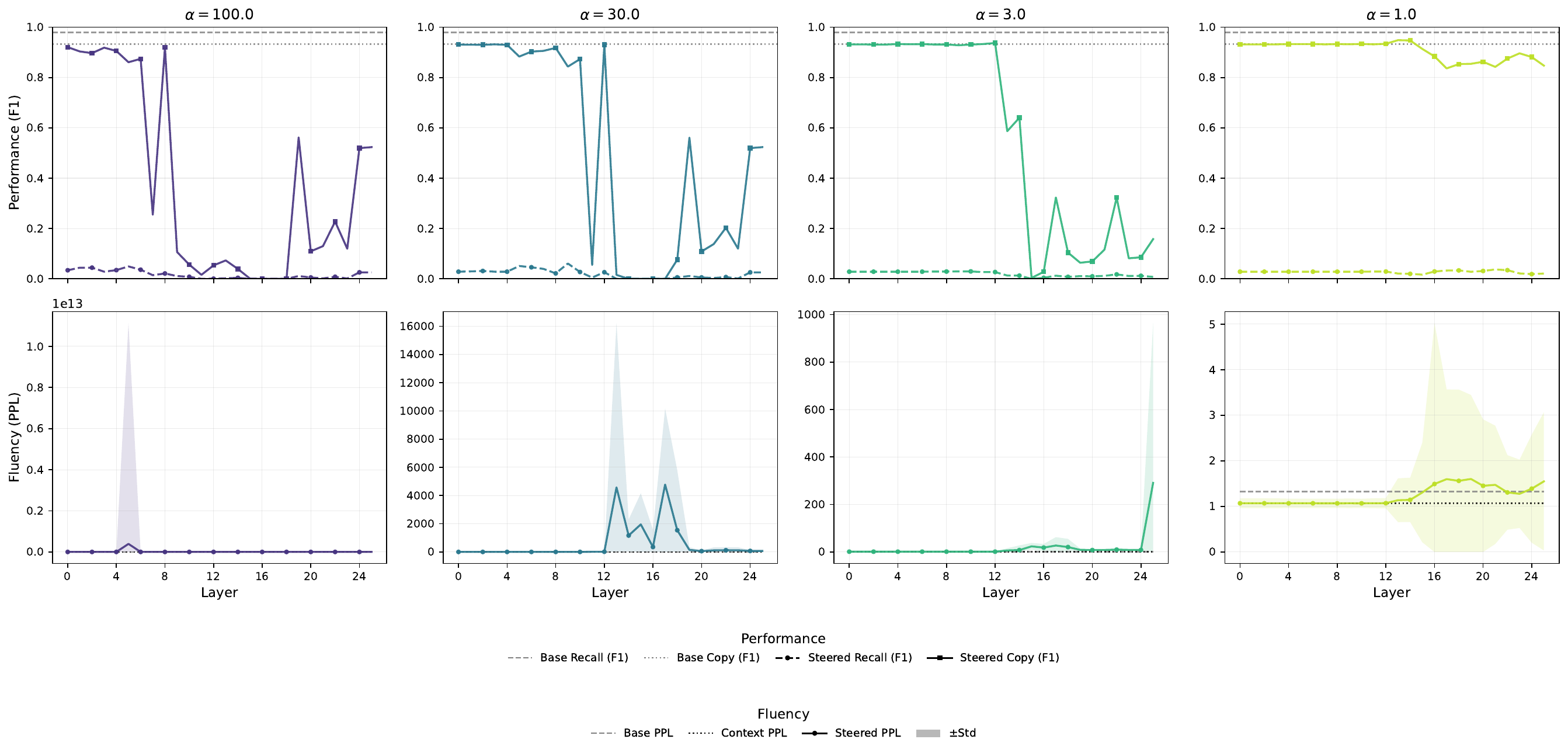}
\caption{Layer-wise effects of steering across four $\alpha$. Rows show performance (F1) and fluency (PPL). Columns correspond to $\alpha$.}
\label{fig:ov-perf-prob-flu-peq-gemma2-cr-cf-lt}
\end{figure}

\subsubsection{Contextual Steering (Recall$\rightarrow$Copy) - Gemma2-2B}
\begin{figure}[H]
\centering
\includegraphics[width=\textwidth]{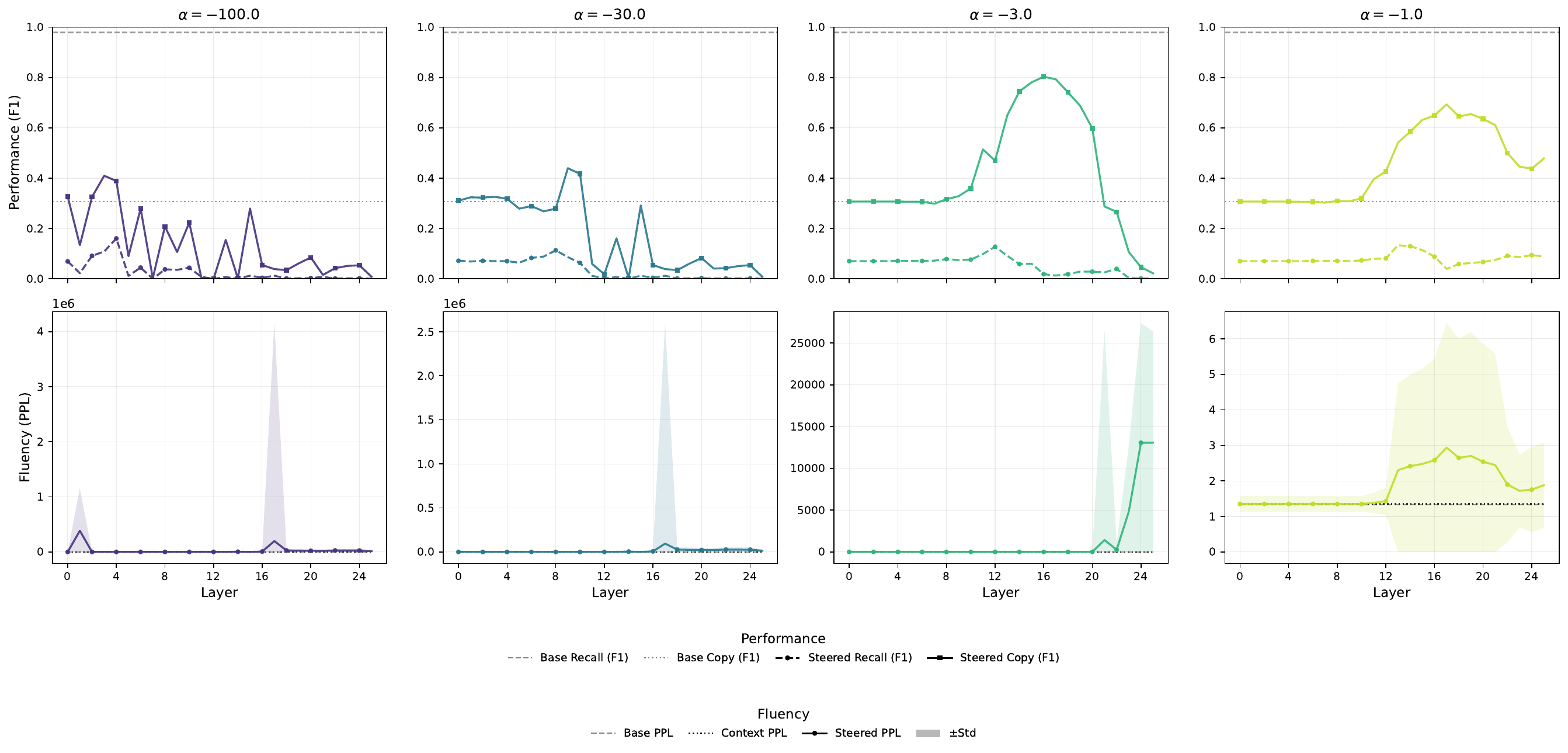}
\caption{Layer-wise effects of steering across four $\alpha$. Rows show performance (F1) and fluency (PPL). Columns correspond to $\alpha$.}
\label{fig:ov-perf-prob-flu-peq-gemma2-rc-cf-lt}
\end{figure}

\subsubsection{Parametric Steering (Copy$\rightarrow$Recall) - T5Gemma-2B}
\begin{figure}[H]
\centering
\includegraphics[width=\textwidth]{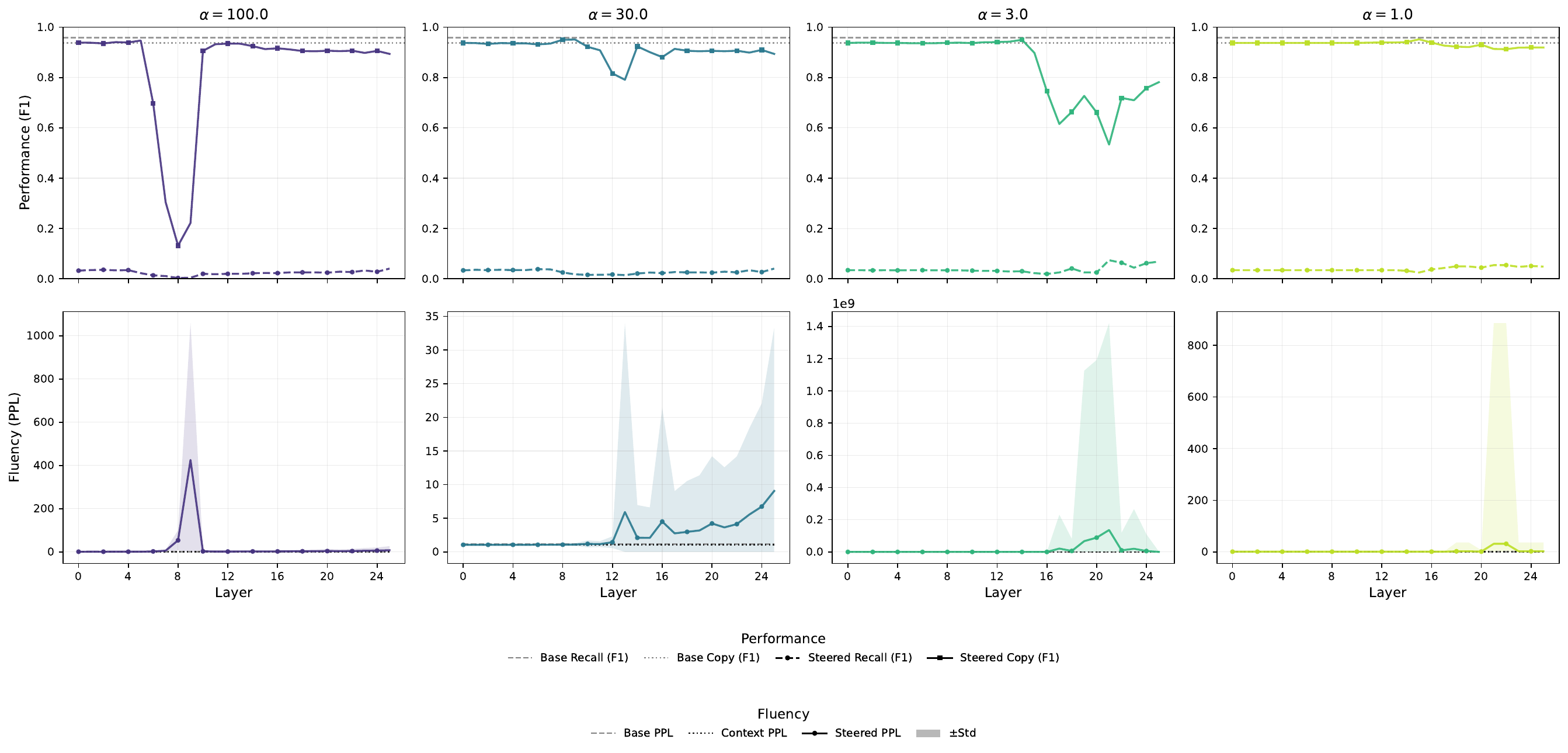}
\caption{Layer-wise effects of steering across four $\alpha$. Rows show performance (F1) and fluency (PPL). Columns correspond to $\alpha$.}
\label{fig:ov-perf-prob-flu-peq-t5gemma-cr-cf-lt}
\end{figure}

\subsubsection{Contextual Steering (Recall$\rightarrow$Copy) - T5Gemma-2B}
\begin{figure}[H]
\centering
\includegraphics[width=\textwidth]{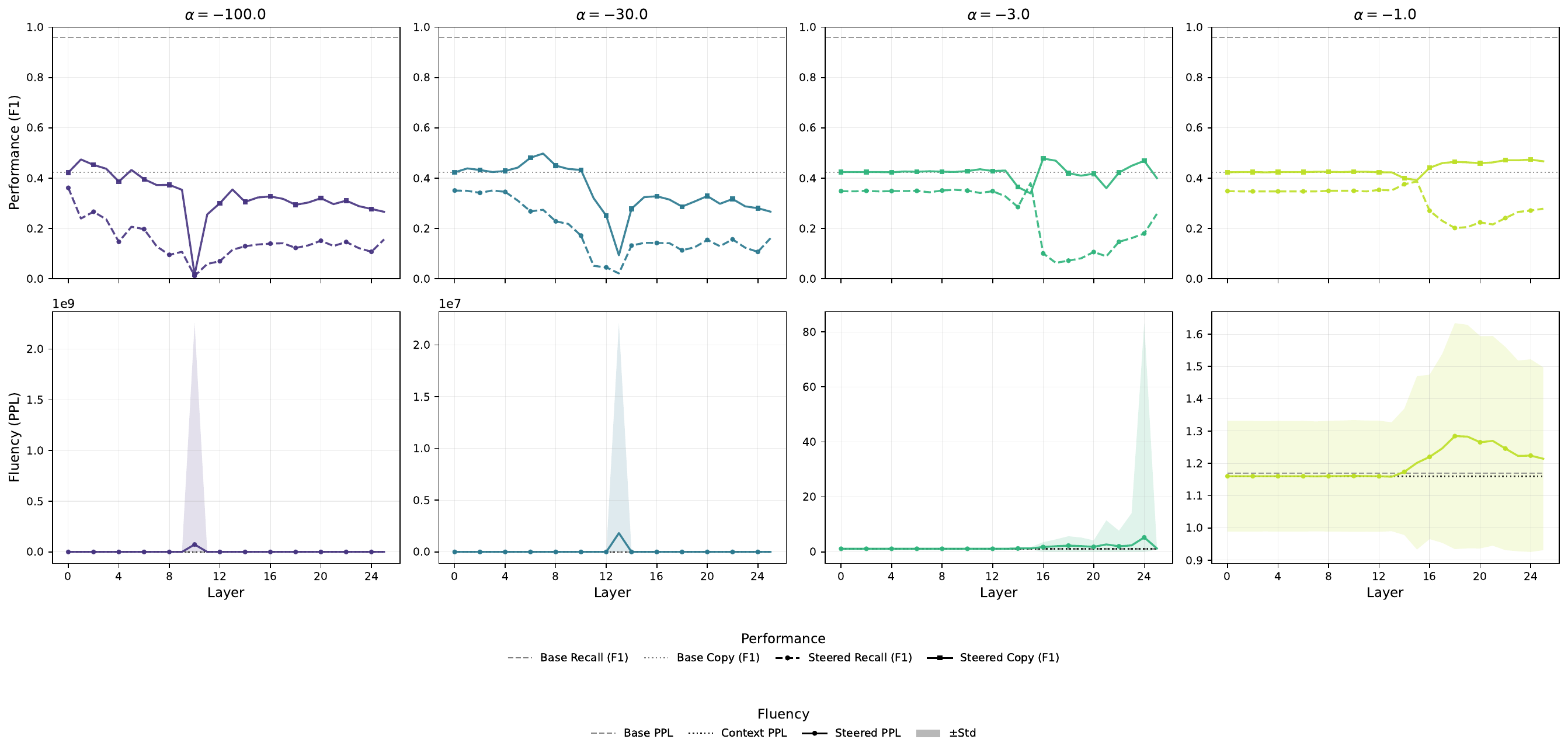}
\caption{Layer-wise effects of steering across four $\alpha$. Rows show performance (F1) and fluency (PPL). Columns correspond to $\alpha$.}
\label{fig:ov-perf-prob-flu-peq-t5gemma-rc-cf-lt}
\end{figure}

\subsubsection{Parametric Steering (Copy$\rightarrow$Recall) - OLMo-2-0425-1B}
\begin{figure}[H]
\centering
\includegraphics[width=\textwidth]{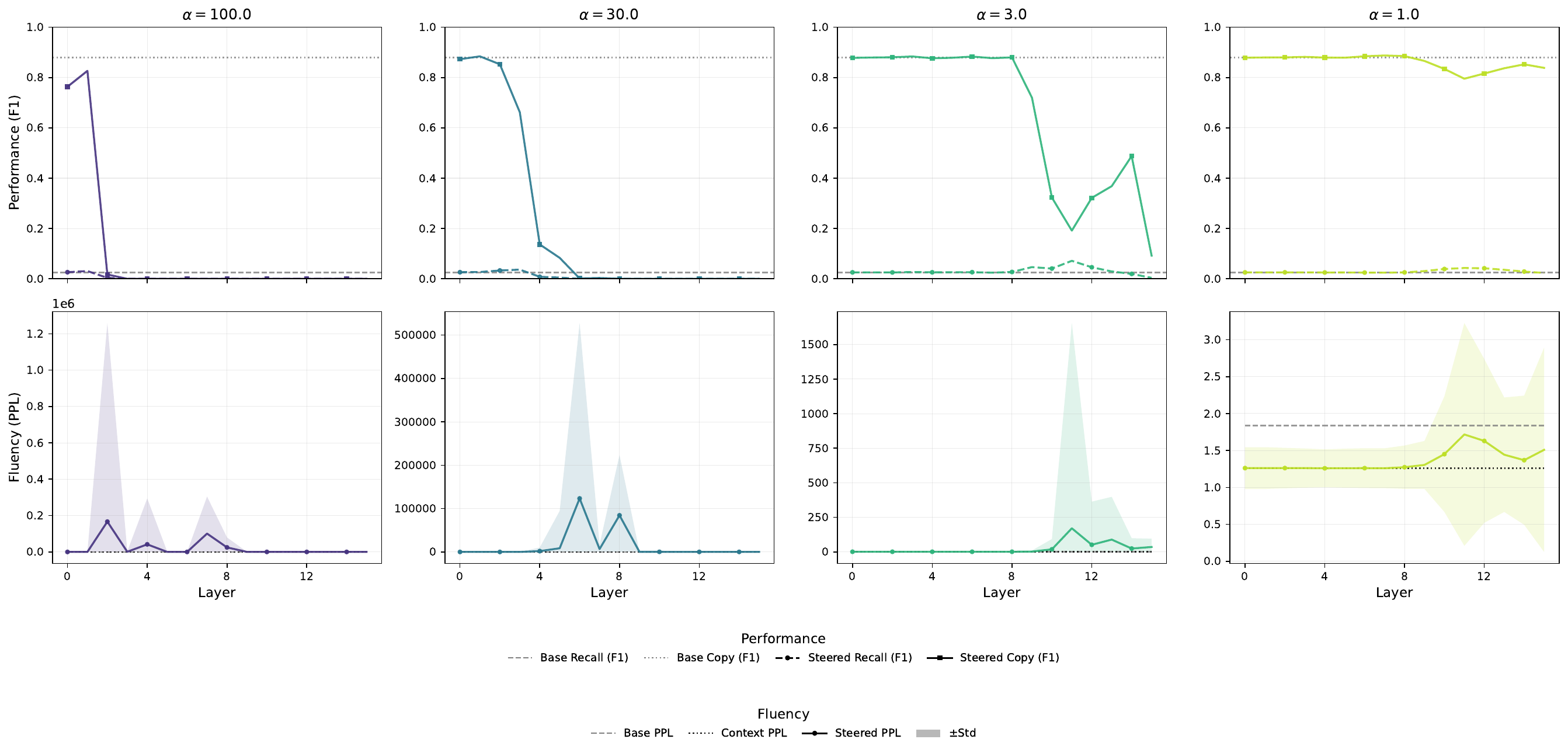}
\caption{Layer-wise effects of steering across four $\alpha$. Rows show performance (F1) and fluency (PPL). Columns correspond to $\alpha$.}
\label{fig:ov-perf-prob-flu-peq-olmo2-cr-cf-lt}
\end{figure}

\subsubsection{Contextual Steering (Recall$\rightarrow$Copy) - OLMo-2-0425-1B}
\begin{figure}[H]
\centering
\includegraphics[width=\textwidth]{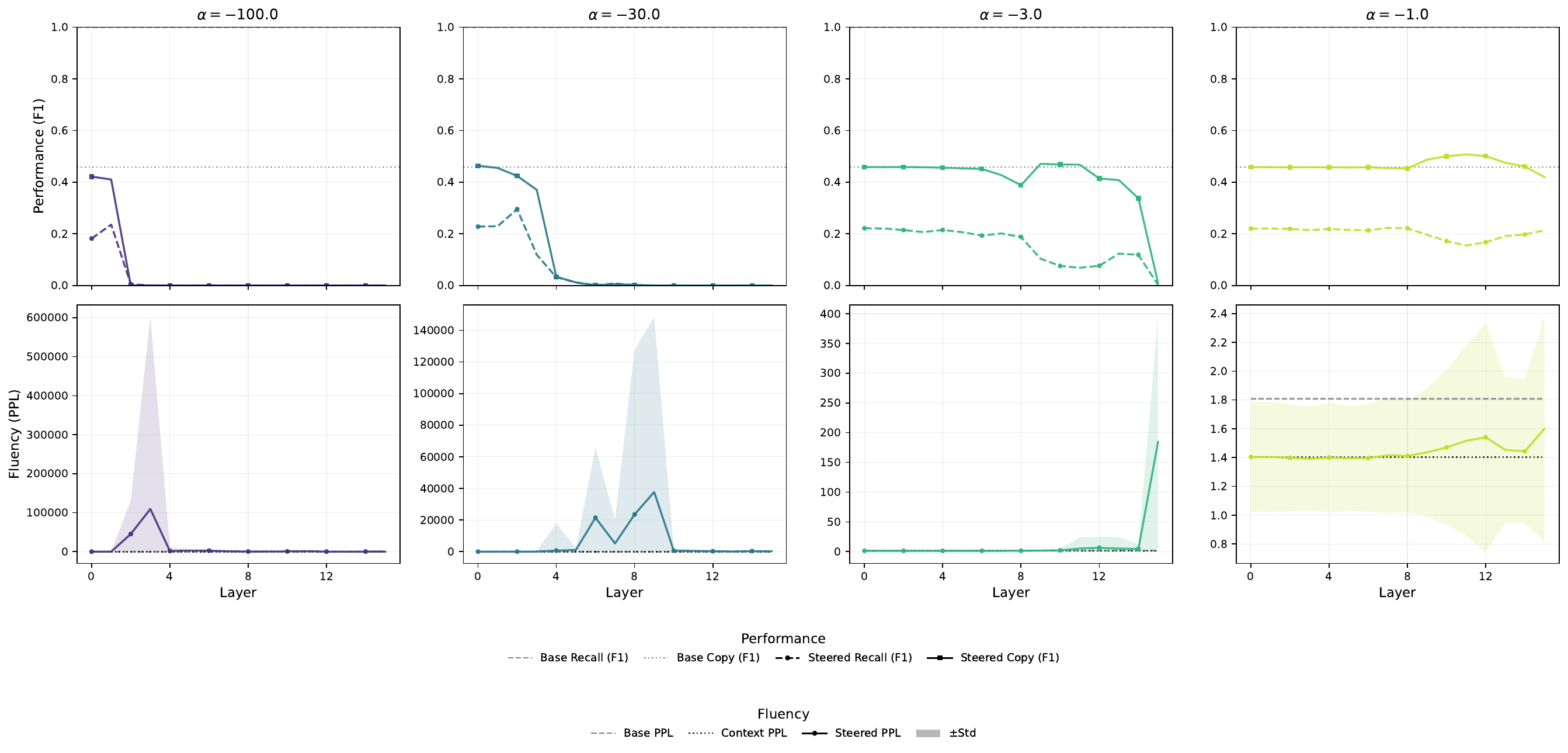}
\caption{Layer-wise effects of steering across four $\alpha$. Rows show performance (F1) and fluency (PPL). Columns correspond to $\alpha$.}
\label{fig:ov-perf-prob-flu-peq-olmo2-rc-cf-lt}
\end{figure}

\end{document}